\documentclass[10pt,twocolumn,letterpaper]{article}

\usepackage[pagenumbers]{iccv} % To force page numbers, e.g. for an arXiv version

\definecolor{iccvblue}{rgb}{0.21,0.49,0.74}
\usepackage[pagebackref,breaklinks,colorlinks,allcolors=iccvblue]{hyperref}

\usepackage{multirow}
\usepackage{float}

\def\paperID{13716} % *** Enter the Paper ID here
\def\confName{ICCV}
\def\confYear{2025}

\title{LBM: Latent Bridge Matching for Fast Image-to-Image Translation}

\author{Clément Chadebec \\ Jasper Research \and Onur Tasar \\ Jasper Research \and Sanjeev Sreetharan \\ Jasper Research \and Benjamin Aubin \\ Jasper Research
}

\begin{document}
\twocolumn[{
\maketitle
\begin{minipage}{\textwidth}
    \begin{figure}[H]
        \centering
    \captionsetup[subfigure]{position=above, labelformat = empty}
    \subfloat[Original]{\includegraphics[width=0.33\linewidth]{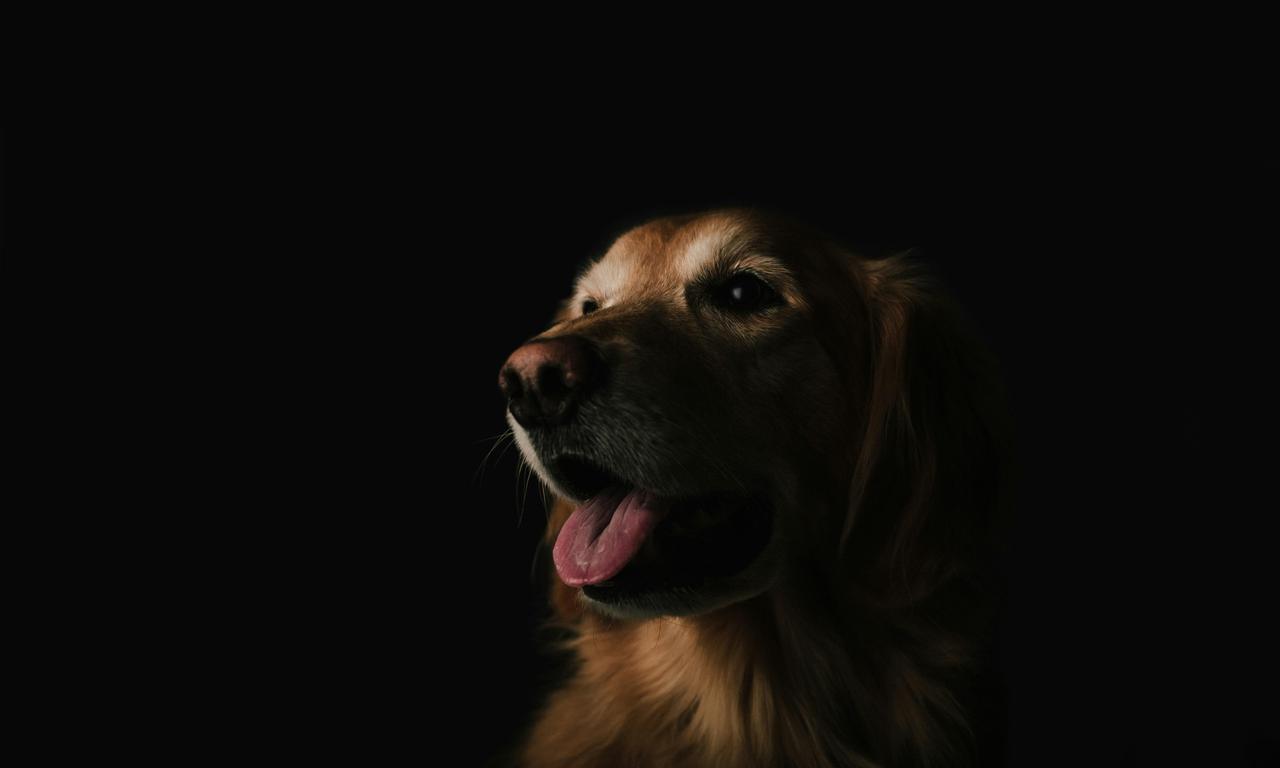}}
    \subfloat{\includegraphics[width=0.33\linewidth]{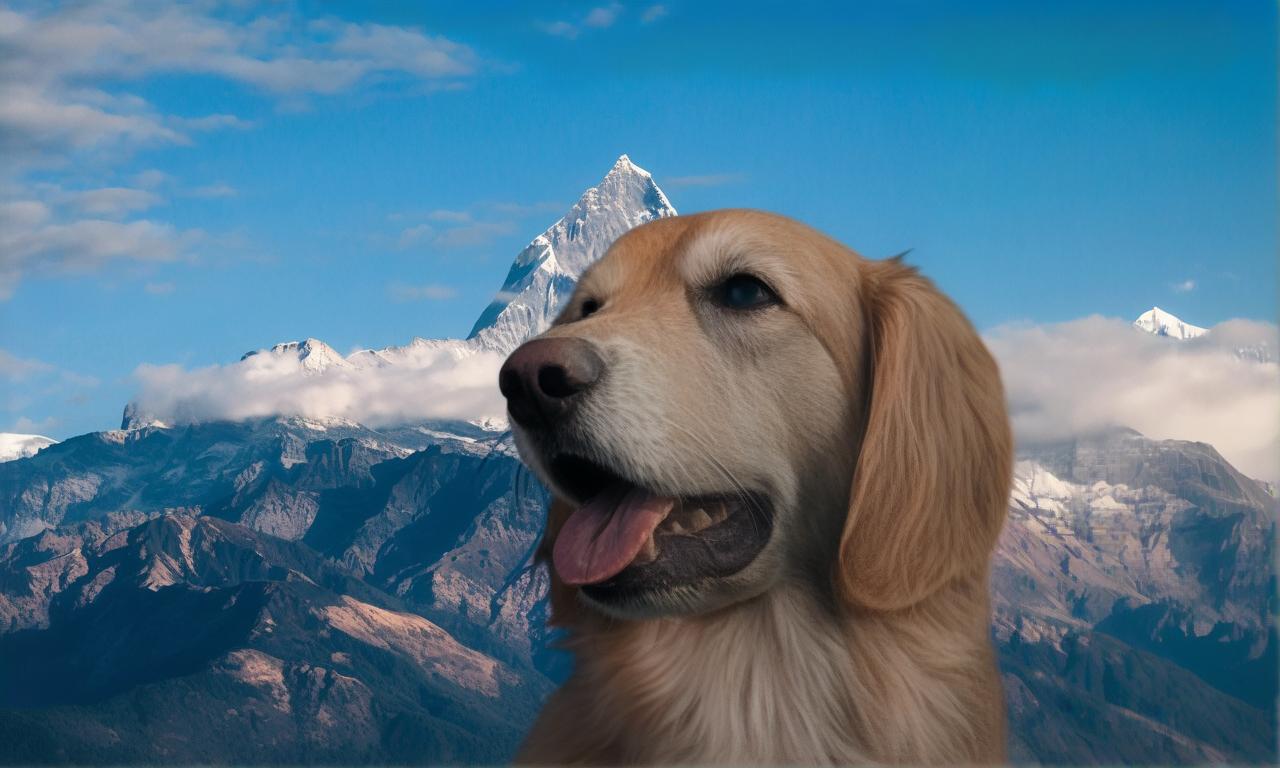}}
    \subfloat{\includegraphics[width=0.33\linewidth]{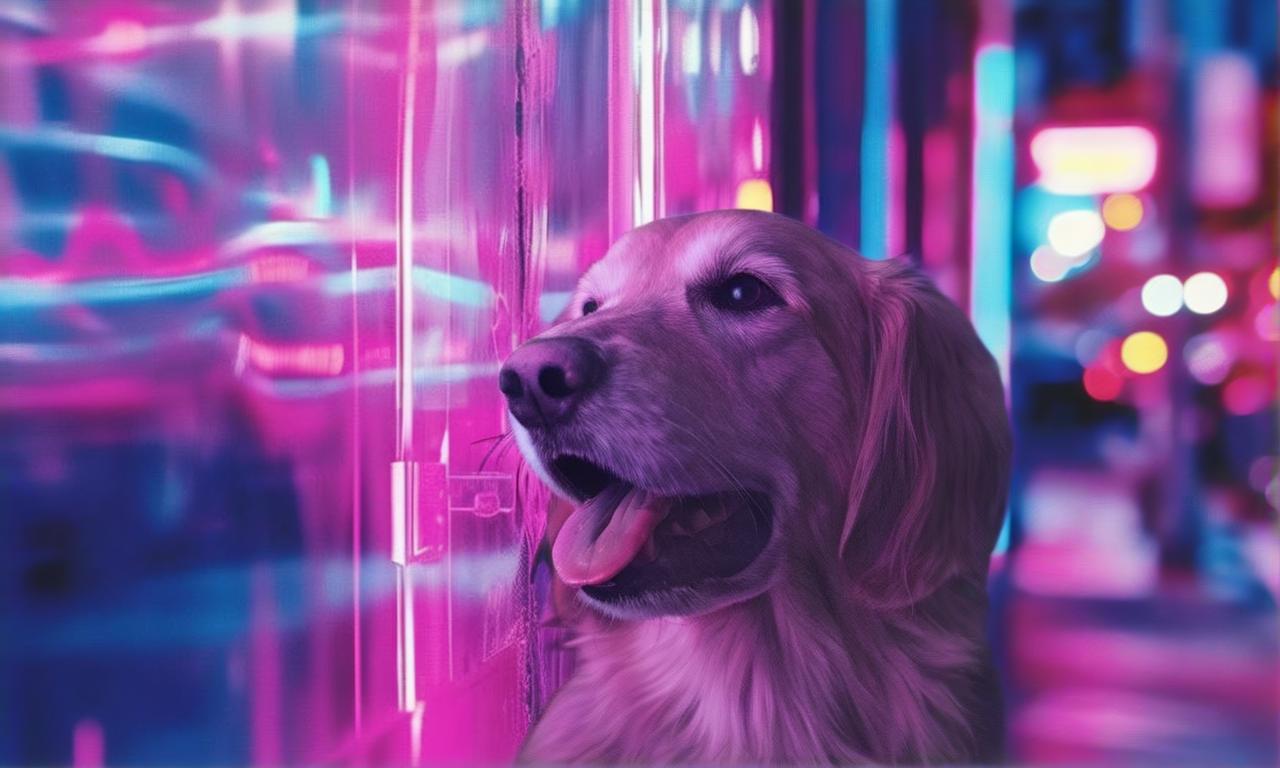}}\\
    \vspace{-0.1em}
    \subfloat{\includegraphics[width=0.33\linewidth]{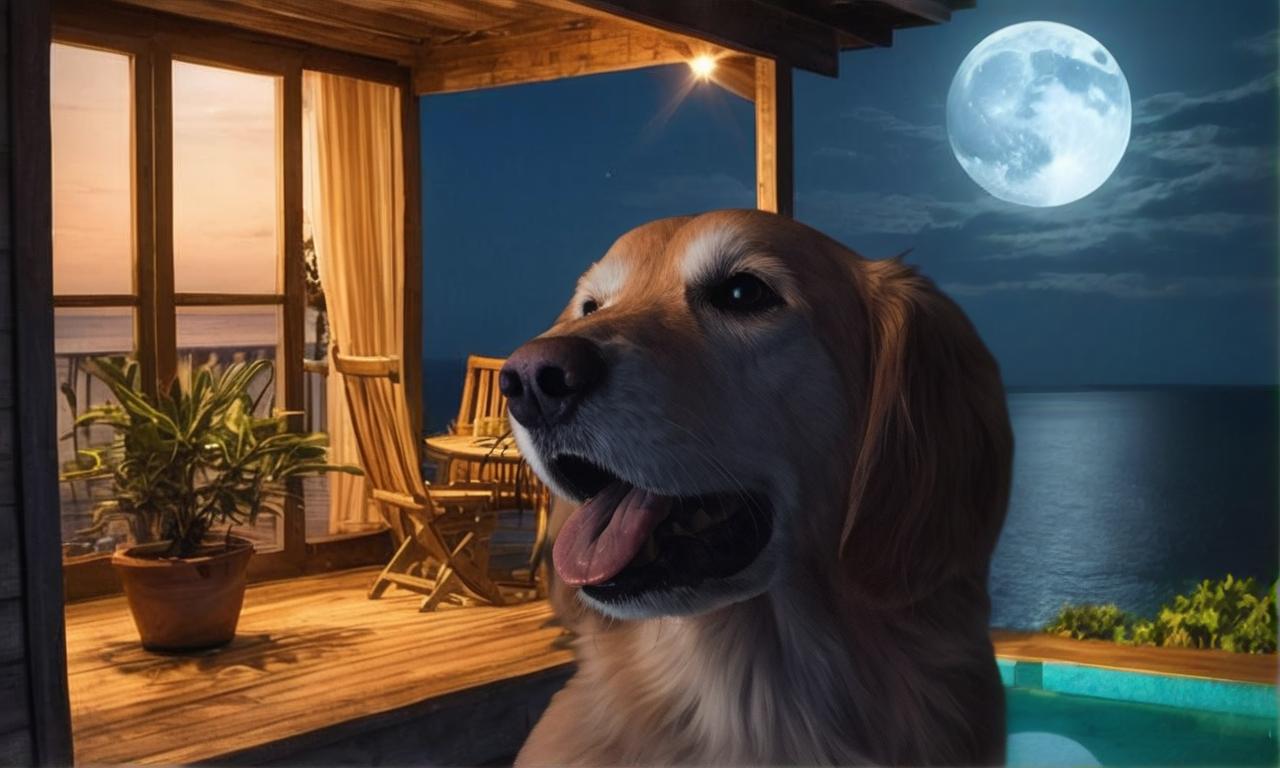}}
    \subfloat{\includegraphics[width=0.33\linewidth]{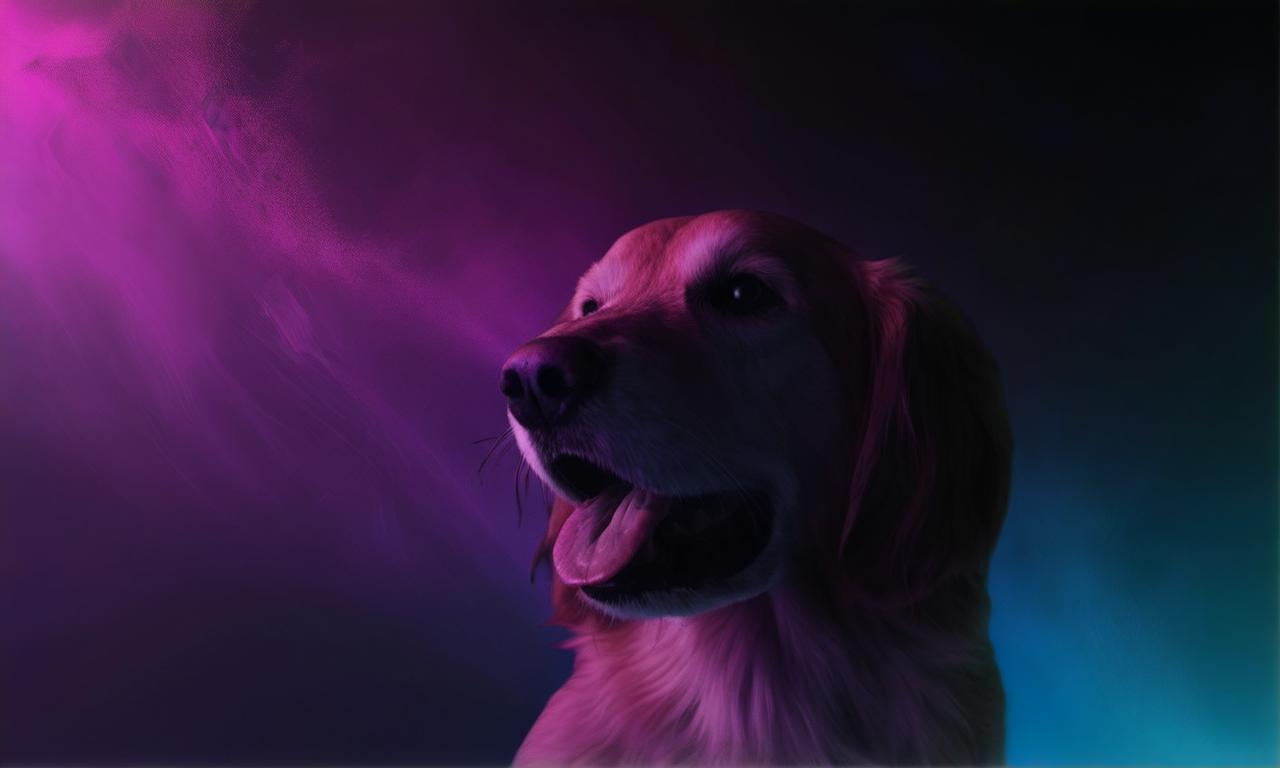}}
    \subfloat{\includegraphics[width=0.33\linewidth]{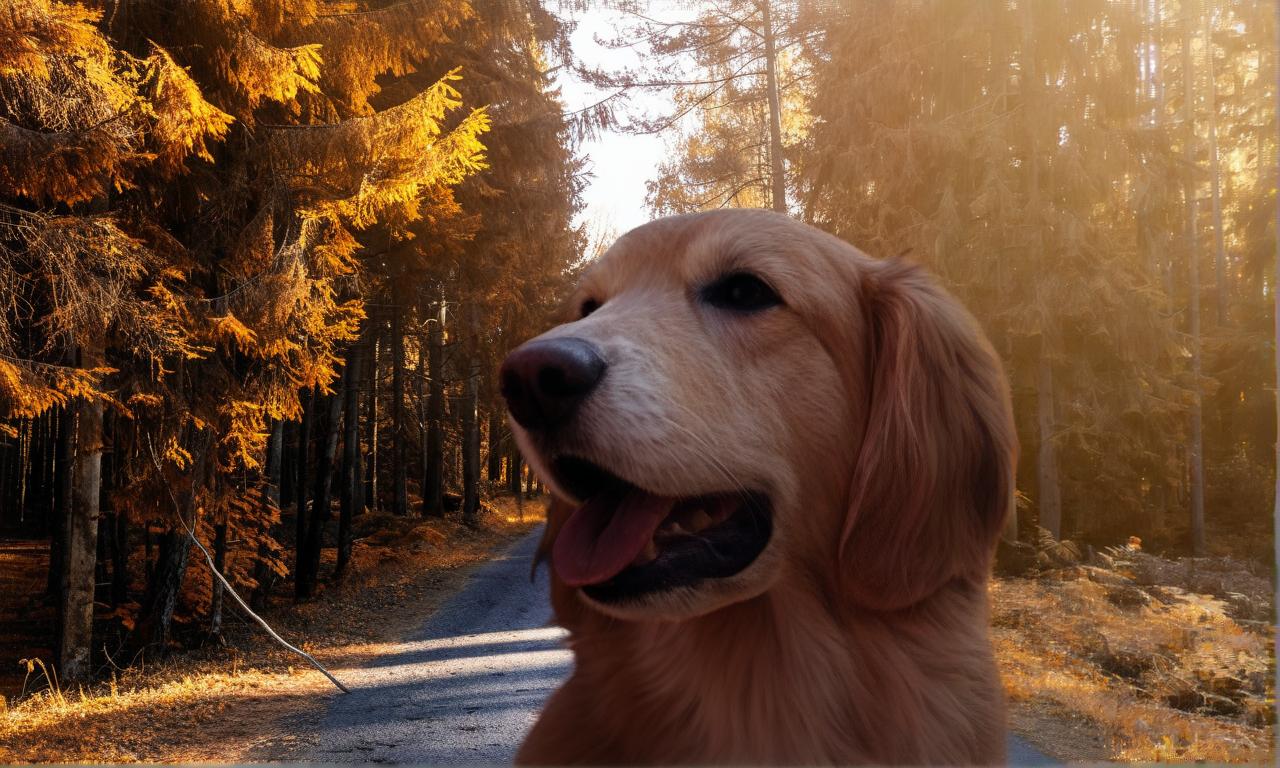}}\\
    % \vspace{-0.1em}
    % \subfloat{\includegraphics[width=0.33\linewidth]{plots/relighting/additional_paper_plots_rf/output_image_1.jpg}}
    % \subfloat{\includegraphics[width=0.33\linewidth]{plots/relighting/additional_paper_plots_rf/output_image_2.jpg}}
    % \subfloat{\includegraphics[width=0.33\linewidth]{plots/relighting/additional_paper_plots_rf/output_image_3.jpg}}\\
    \caption{Relighted images using Latent Bridge Matching (LBM) and 1 neural function evaluation (NFE).}
    \label{fig:example_image}
\end{figure}
\end{minipage}
\vspace{2em}
}]
\begin{abstract}
In this paper, we introduce Latent Bridge Matching (LBM), a new, versatile and scalable method that relies on Bridge Matching in a latent space to achieve fast image-to-image translation. We show that the method can reach state-of-the-art results for various image-to-image tasks using only a single inference step. In addition to its efficiency, we also demonstrate the versatility of the method across different image translation tasks such as object removal, normal and depth estimation, and object relighting. We also derive a conditional framework of LBM and demonstrate its effectiveness by tackling the tasks of controllable image relighting and shadow generation.  We provide an implementation at \url{https://github.com/gojasper/LBM}.
\end{abstract}
\section{Introduction}
\label{sec:intro}
Image translation is a task that consists of mapping an image from a source domain to a target domain. It can be formulated as a transport problem where the goal is to find a mapping that translates samples from the source domain to the target domain \cite{pang2021image}. This field contains a wide range of tasks such as object removal \cite{sun2025attentive}, semantic image synthesis \cite{park2019semantic}, style transfer \cite{zhu2017unpaired,gatys2016image}, image harmonization \cite{chen2023dense,zhang2025scaling} or image segmentation \cite{kirillov2023segment,zheng2024birefnet}.

Diffusion models (DM) \citep{sohl2015deep,ho2020denoising,song2020score} are a type of generative model that learns a denoising mechanism that can be used to generate new samples from a Gaussian noise. These models appear very well suited for image synthesis \cite{dhariwal2021diffusion,ramesh2022hierarchical, rombach2022high,nichol2022glide} and can be conditioned with respect to various types of inputs such as text \cite{dhariwal2021diffusion,ramesh2022hierarchical,rombach2022high,nichol2022glide,esser2024scaling,ho2022imagen,podell2023sdxl} where they demonstrate remarkable performance. In the particular context of image-to-image translation, the conditional setting allows to build a diffusion model conditioned with source images such as low-resolution images \cite{rombach2022high}, depth maps, normal maps or edges \citep{zhang2023adding,mou2024t2i} and generate images belonging to the target domain. While they demonstrate strong results, their intrinsic iterative generation process hinders their usability for real-time applications. Recently, several works have been proposed to accelerate the sampling process of DM through more efficient solvers \citep{lu2022dpm,lu2022dpm+,zhang2022fast,zhao2024unipc} or via distillation \cite{salimans2021progressive,song2023consistency,lin2024sdxl,xu2023ufogen,liu2023instaflow,ren2024hyper,luo2023latent,luo2023lcm,sauer2023adversarial,sauer2024fast,yin2023one,hsiao2024plug}. However, despite revealing promising results, most of these methods are limited to text-to-image or struggle to achieve satisfactory one-step generation.

% Explain briefly the bridge matching method
Drawing inspiration from diffusion models, bridge matching and flow models have been proposed and aim to find transport maps between two distributions using Stochastic Differential Equations (SDEs) or Ordinary Differential Equations (ODEs). The key difference from diffusion models is that they do not involve any noising mechanism and can be applied to any pair of distributions. The main idea behind flow matching \cite{lipman2023flow,albergo2023building,liu2022flow} (resp. bridge matching \cite{peluchetti2023non,shi2024diffusion}) is to define deterministic (resp. stochastic) interpolants between pairs of samples from the source and target distributions and estimate the drift of the associated ODE (resp. SDE) using a denoiser model \cite{albergo2023stochastic}. While there exist some works applying flow matching to image-to-image translation \cite{gui2024depthfm,fischer2023boosting,martin2025pnpflow}, the usability, scalability, and efficiency of their stochastic variant remain an open question. It's worth noting that \cite{liu2023i2sb} previously applied bridge matching to super-resolution and inpainting tasks, though their approach was limited to low-resolution images. 

In this paper, we aim to bridge this gap by introducing Latent Bridge Matching (LBM), a novel and scalable method based on bridge matching able to achieve 1 step inference for various image translation tasks. The main contributions of this paper are as follows:

\begin{itemize}
    \item We propose LBM, a novel, versatile and scalable method based on bridge matching that shows to be very effective for various image-to-image tasks even for high resolution images.
    \item We show that our method can either compete or achieve state-of-the-art performance for object-removal, depth and surface estimation as well as object relighting. In particular, it outperforms both diffusion-based methods requiring multiple sampling steps as well as flow matching models.
    \item We also derive a conditional framework of the method and apply it to controllable object relighting and shadow generation.
    \item Finally, we conduct an extensive ablation study to understand the impact of the different components of our method.
\end{itemize}

\section{Related works}

Diffusion models are generative models that consist in artificially adding noise to samples drawn from a given distribution according to a pre-defined noising mechanism \citep{sohl2015deep,ho2020denoising,song2020score}. This process is such that the final data distribution is roughly equivalent to Gaussian noise. A denoiser model is then trained to denoise the corrupted samples such that at inference time, the model can be used to iteratively generate samples from pure Gaussian noise. These models can also be conditioned with respect to various modalities such as text \cite{dhariwal2021diffusion,ramesh2022hierarchical,rombach2022high,nichol2022glide,esser2024scaling,ho2022imagen,podell2023sdxl}, images \cite{rombach2022high}, depth maps, edges or poses \citep{zhang2023adding,mou2024t2i} to further guide the generation process. Despite their success, these models are limited by their intrinsic iterative generation process that requires multiple evaluations of a potentially very computationally expensive neural network.

Various methods were then proposed in the literature to accelerate the sampling process of diffusion models by reducing the number of denoising steps required to generate new samples at inference time. The research has evolved along two main paths. First, more effective solvers were proposed \citep{lu2022dpm,lu2022dpm+,zhang2022fast,zhao2024unipc} but these methods still require quite a few steps to generate satisfying samples. Second, many works explored distillation methods \citep{hinton2015distilling}, training student networks to approximate the teacher denoiser's generation in fewer steps \citep{luhman2021knowledge,zheng2023fast,kohler2024imagine,yin2023one,salimans2021progressive,liu2023instaflow,hsiao2024plug,luo2023latent,luo2023lcm}. These approaches were further enriched with adversarial training \citep{xu2023ufogen,sauer2023adversarial,sauer2024fast,lin2024sdxl,xu2023ufogen,ren2024hyper}, distribution matching \citep{yin2023one} or both \citep{chadebec2024flash,yin2024improved}. While these approaches show very promising results for text-to-image, they remain specifically tailored to this task or fail to achieve single-step generation.

% \cite{salimans2021progressive,song2023consistency,lin2024sdxl,xu2023ufogen,liu2023instaflow,ren2024hyper,luo2023latent,luo2023lcm,sauer2023adversarial,sauer2024fast,yin2023one,hsiao2024plug} that aim at distilling the knowledge of a pre-trained diffusion model into a smaller model that can generate samples in a single step. 

Driven by the impressive performance of flow-based models for text-to-image \cite{esser2024scaling}, several works have started to extend the applicability of flow matching to other tasks. These models were for instance adapted to the context of super-resolution \cite{fischer2023boosting,martin2025pnpflow}, depth estimation \cite{gui2024depthfm}, video generation \cite{davtyan2023efficient}, audio generation \cite{le2024voicebox}, image editing \cite{hu2024latent} as well as model distillation \cite{liu2023instaflow}. However, its stochastic variant (bridge matching) has seen far less traction and has mainly been used in \citep{liu2023i2sb} for image restoration and image inpainting on low resolution images.  Hence, it remains unclear if such an approach would scale to high resolution images or transfer efficiently to other tasks since it bridges distributions in the pixel space. 

\section{Proposed method}
In this section, we detail Latent Bridge Matching (LBM), the proposed method that is based on the bridge matching framework.

\subsection{Bridge matching}
Let $\pi_0$ and $\pi_1$ be a pair of distributions such that we have access to samples from both distributions $(x_0, x_1) \sim \pi_0\times\pi_1$. The main idea behind bridge matching is to find a transport map from $\pi_0$ to $\pi_1$ \cite{liu2022let,albergo2023stochastic,peluchetti2023non, shi2024diffusion} so one may ultimately sample from $\pi_1$ using samples from $\pi_0$. To do so, given $(x_0, x_1) \sim \pi_0\times\pi_1$, we build a stochastic interpolant $x_t$ such that the conditional distribution of $x_t$ given $(x_0, x_1)$ ($\pi(x_t|x_0, x_1)$) is essentially a Brownian motion (also known as \emph{Brownian bridge}). 
\begin{equation}\label{eq:bridge}
    x_t = (1 - t) x_0 + t x_1 + \sigma \sqrt{t(1-t)} \epsilon\,,
\end{equation} 
where $\epsilon \sim \mathcal{N}(0, I)$, $ \sigma \geq 0$ and $t \in [0, 1]$. Notably, if one further sets $\sigma = 0$, one may retrieve the flow matching formulation \cite{albergo2023stochastic,lipman2023flow,liu2022flow} which can be considered as the \emph{zero-noise} limit of bridge matching. Hence, the evolution in time of $x_t$ is given by the following Stochastic Differential Equation (SDE):
\begin{equation}\label{eq:bridge_sde}
    d x_t = \frac{(x_1 - x_t)}{1-t} \mathrm{d}t + \sigma \mathrm{d}B_t\,,
\end{equation}
where $v(x_t, t) = (x_1 - x_t)/(1-t)$ is called the drift of the SDE. In order to use Eq.~\eqref{eq:bridge_sde} to sample from $\pi_1$ using $\pi_0$, one needs to ensure that the distribution of $x_t$ ($\pi_t$) is Markov and so does not depend on $x_1$. In practice, a Markovian projection is performed and typically consists of regressing over the drift of the SDE using a neural network:
\begin{equation}\label{eq:bridge_loss_function}
    \mathbb{E}_{t, x_0, x_1} \left[ \left\|(x_1 - x_t)/(1-t) - v_{\theta}(x_t, t) \right\|^2 \right]\,.
\end{equation}
Finally, the estimated drift function $v_{\theta}$ can be integrated into standard SDE solvers to solve Eq.~\eqref{eq:bridge_sde} to generate samples that follow $\pi_1$ from initial samples drawn from $\pi_0$. 

% Noteworthy is the fact that if we further set $\sigma = 0$, one may retrieve the Flow Matching formulation \cite{albergo2023stochastic,lipman2023flow,liu2022flow} which can be considered as the "zero-noise" limit of bridge matching.

\begin{figure*}[ht]
    \centering
    \includegraphics[width=\linewidth]{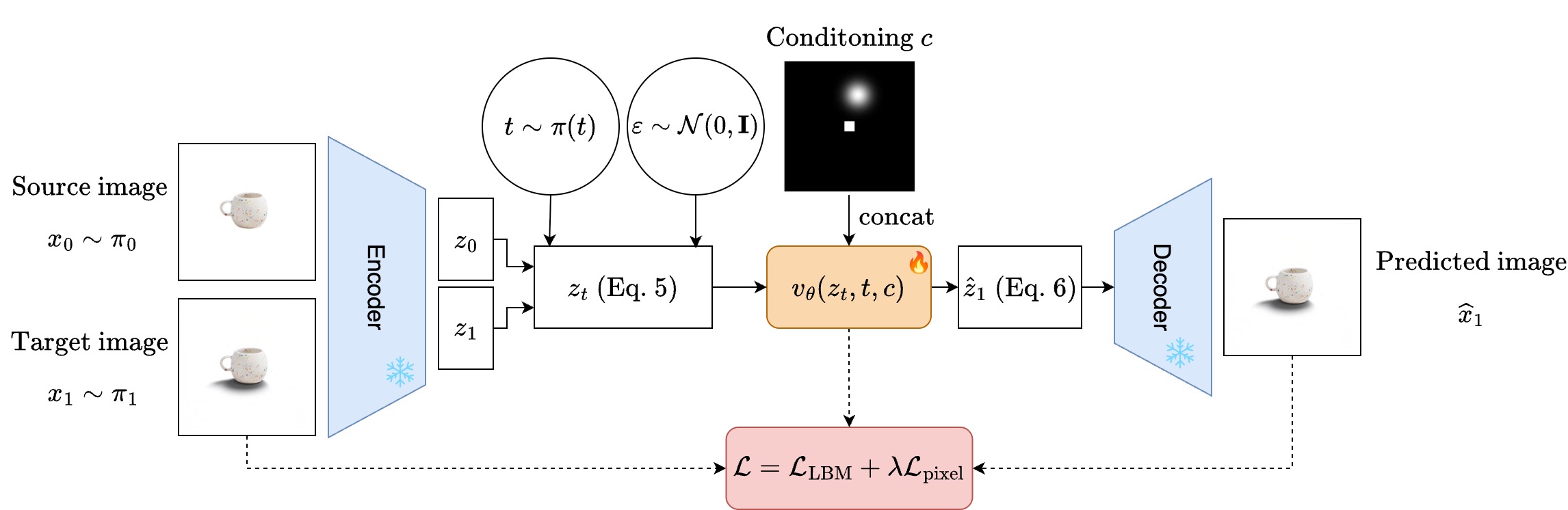}
    \caption{Training procedure for a conditional latent bridge matching model in the context of controllable shadow generation.}
    \label{fig:training_procedure}
\end{figure*}

\subsection{Latent bridge matching}

In our case, since we want the model to handle high resolution images and to have a scalable method, we propose to rely on a \emph{latent} bridge matching approach. In such a case, the samples $(x_0, x_1) \sim \pi_0\times\pi_1$ are first embedded into a latent space using a pre-trained model such as a Variational Autoencoder (VAE) \cite{kingma_auto-encoding_2014} in a similar fashion to \cite{rombach2022high}. Let us denote $z_0$ and $z_1 $ as the latents associated with the samples $x_0$ and $x_1$. Using the same formulation as in the previous section, this leads to the following objective function:
\begin{equation}\label{eq:latent_loss_function}
    \mathcal{L}_{\mathrm{LBM}} = \mathbb{E} \left[ \left\|(\mathcal{E}(x_1) - \mathcal{E}(x_t))/(1-t) - v_{\theta}(z_t, t) \right\|^2 \right]\,,
\end{equation}
where $\mathcal{E}$ is the encoder of the VAE and $z_t$ is given by
\begin{equation}\label{eq:latent_bridge}
    z_t = z_0 (1-t) + z_1 t + \sigma \sqrt{t(1-t)} \epsilon\,,
\end{equation}
where $z_0 = \mathcal{E}(x_0)$,  $z_1 = \mathcal{E}(x_1)$, $\epsilon \sim \mathcal{N}(0, I)$ and $\sigma \geq 0$. At inference time, one may sample from the distribution $\pi_1$ using samples from $\pi_0$ by first drawing a sample from $\pi_0$, mapping it to the latent space, solving the SDE in Eq.~\eqref{eq:bridge_sde} using a standard SDE solver and then mapping the latent back to the image space using the decoder of the VAE. This approach has the benefit of drastically reducing the computational cost of the method by reducing the dimensionality of the data and so allows the training of models that can scale to high dimensional data such as high resolution images. Note that computing the latents associated with any samples from $\pi_0$ or $\pi_1$ can be done before training. In a similar fashion to what was proposed for diffusion models, one may derive a conditional setting of LBM. In such a case, in addition to the pairing $(x_0, x_1)$, an additional conditioning variable $c$ is introduced and will further guide the generation process. Hence, the drift function approximator $v_{\theta}$ is conditioned with respect to $c$ so that $v_{\theta}(z_t, t, c)$ depends on the conditioning variable $c$ as well.

\subsection{Training}
Let us assume we have access to two distributions of images $\pi_0$ and $\pi_1$ and we want to transport samples from $\pi_0$ to $\pi_1$. The training procedure is as follows. First, we draw a pair of samples $(x_0, x_1) \sim \pi_0 \times \pi_1$. Those samples are then encoded into the latent space using a pre-trained VAE giving the corresponding latents $z_0$ and $z_1$. A timestep $t$ is drawn from $\pi(t)$, the timestep distribution and a \emph{noisy} sample $z_t$ is created using Eq.~\eqref{eq:latent_bridge}. This sample is then passed to the denoiser $v_{\theta}(z_t, t$) which is additionally conditioned with respect to the timestep $t$ and predicts the \emph{drift}. Notably, one may easily retreive the corresponding predicted latent $\widehat{z}_1$ for the predicted \emph{drift} using 
\begin{equation}\label{eq:predicted_latent}
\widehat{z}_1 = (1-t) \cdot v_{\theta}(z_t, t) + z_t. 
\end{equation}
During training, we also introduce a pixel loss $\mathcal{L}_{\mathrm{pixel}}$ the influence of which is discussed in \cref{sec:ablations}. The loss consists of decoding the estimated target latent $\widehat{x}_1 = \mathcal{D}(\widehat{z}_1)$ where $\mathcal{D}$ is the decoder of the VAE and comparing it to the real target image $x_1$. Several choices of loss functions are possible such as L1, L2 or LPIPS \cite{zhang2018unreasonable}. We found that LPIPS works well in practice and speeds up domain shift. In order to scale with the image size, we put in place a random cropping strategy and only compute the loss on a patch if the image size is larger than a certain threshold. This limits the memory footprint of the model so it does not become a burden to the training efficiency. The final objective can be summarized as follows:
\begin{equation}\label{eq:full_loss}
  \mathcal{L} = \mathcal{L}_{\mathrm{LBM}}(\mathcal{E}(x_0), \mathcal{E}(x_1)) + \lambda\cdot\mathcal{L}_{\mathrm{pixel}}(\widehat{x}_1, x_1)\,.
\end{equation}
We provide in \cref{fig:training_procedure} a scheme of the training procedure of the proposed method in the conditional setting. For illustration purposes, we elect the context of controllable shadow generation where the generation is further conditioned with respect to a light map $c$ indicating the position of a light source. In this setting, $\pi_0$ corresponds to the distribution of latents associated with images without shadows while $\pi_1$ is the distribution of latents associated with images with shadows. In practice, the conditioning variable $c$ can be injected into the denoiser $v_{\theta}$ by concatenating the latent $z_t$ along the channel dimension.

\subsection{Timestep sampling}\label{sec:timestep_sampling}
One key aspect of the proposed method also relies on the choice of the timestep distribution $\pi(t)$. In several works focusing on accelerating the sampling of diffusion models, it was noted that only selecting a few timesteps during training may be beneficial at inference time \citep{sauer2024fast,chadebec2024flash,luo2023latent}. In particular, training the model to denoise inputs at the same timesteps used during inference proved to be highly effective for model distillation \cite{salimans2021progressive,chadebec2024flash,sauer2024fast,lin2024sdxl}. We follow this approach and propose to only use 4 equally spaced timesteps during training and ensure that these timesteps are the ones used at inference. Notably, this choice limits the maximum number of inference steps to only 4. This is discussed in depth in \cref{sec:ablations}. Note that the proposed framework would also apply to other distributions such as the uniform or logit-normal distribution.

\begin{figure*}[t]
    \captionsetup[subfigure]{position=above, labelformat = empty}
    \centering
    \subfloat[Input]{\includegraphics[width=0.14\linewidth]{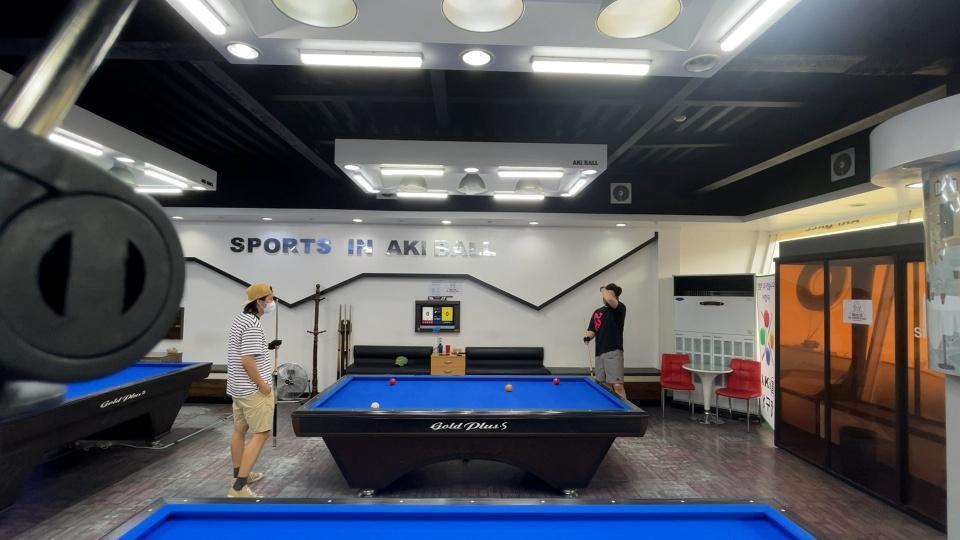}}
    \subfloat[Mask]{\includegraphics[width=0.14\linewidth]{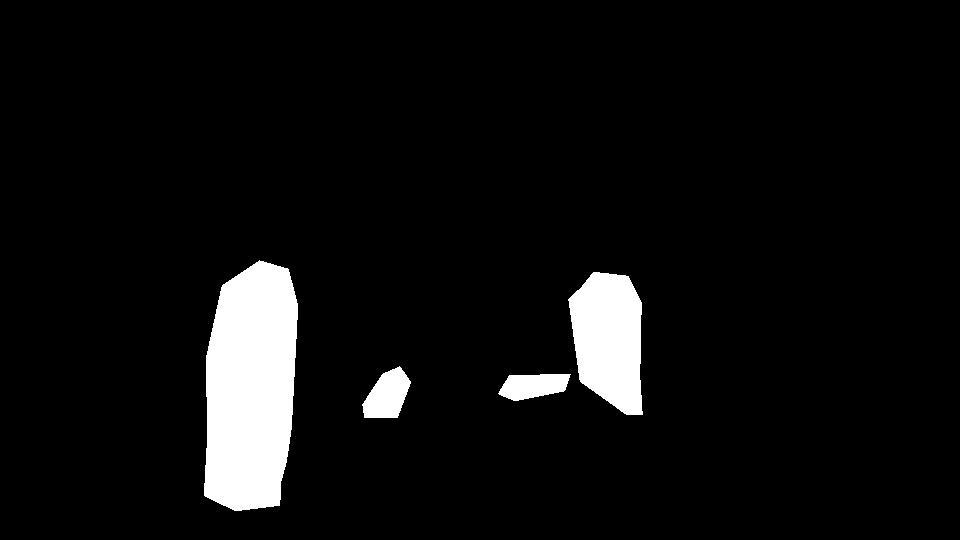}}
    \subfloat[LAMA]{\includegraphics[width=0.14\linewidth]{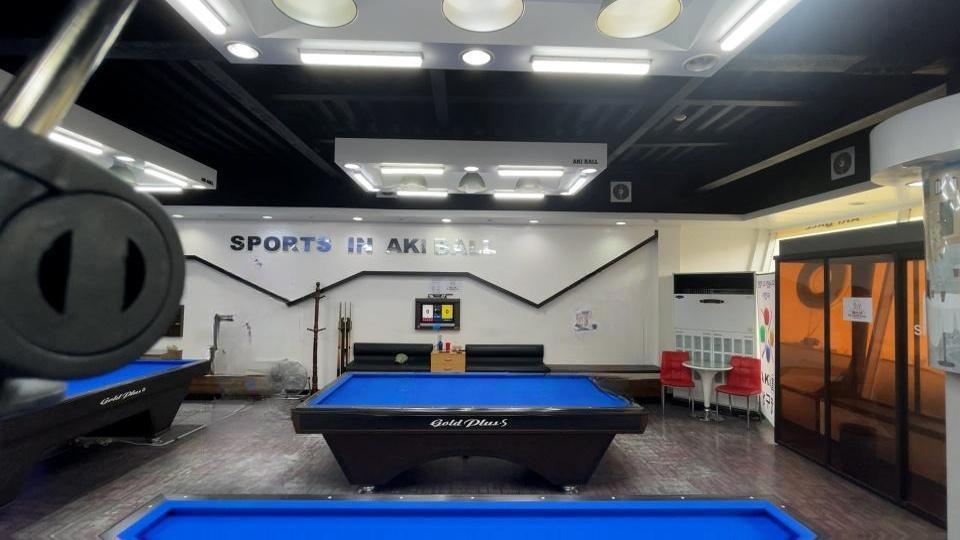}}
    \subfloat[PowerPaint]{\includegraphics[width=0.14\linewidth]{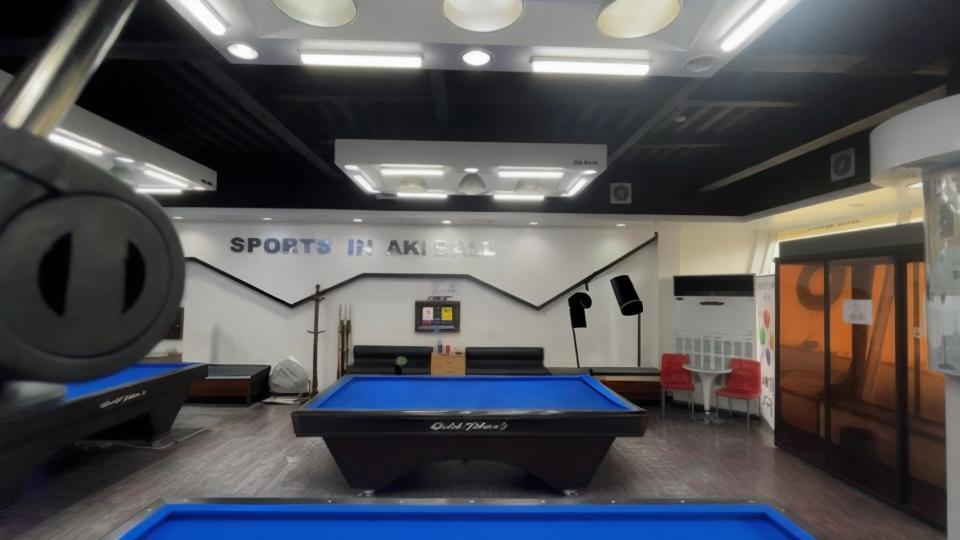}}
    \subfloat[SDXL-Inpainting]{\includegraphics[width=0.14\linewidth]{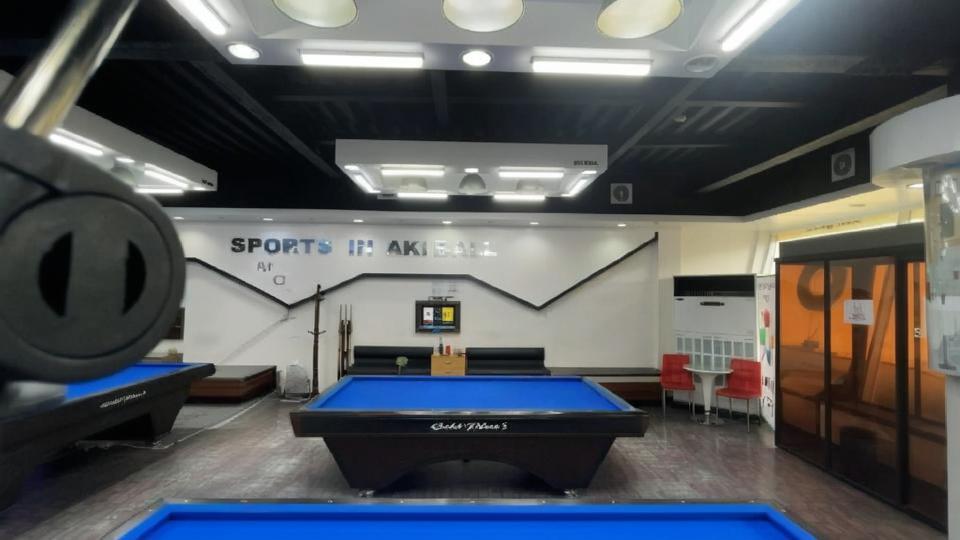}}
    \subfloat[Attentive Eraser]{\includegraphics[width=0.14\linewidth]{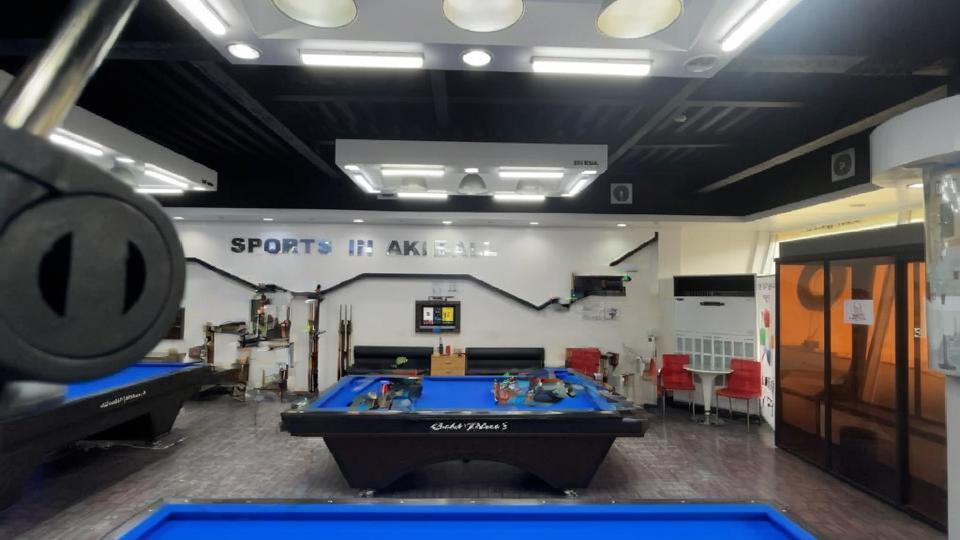}}
    \subfloat[Ours]{\includegraphics[width=0.14\linewidth]{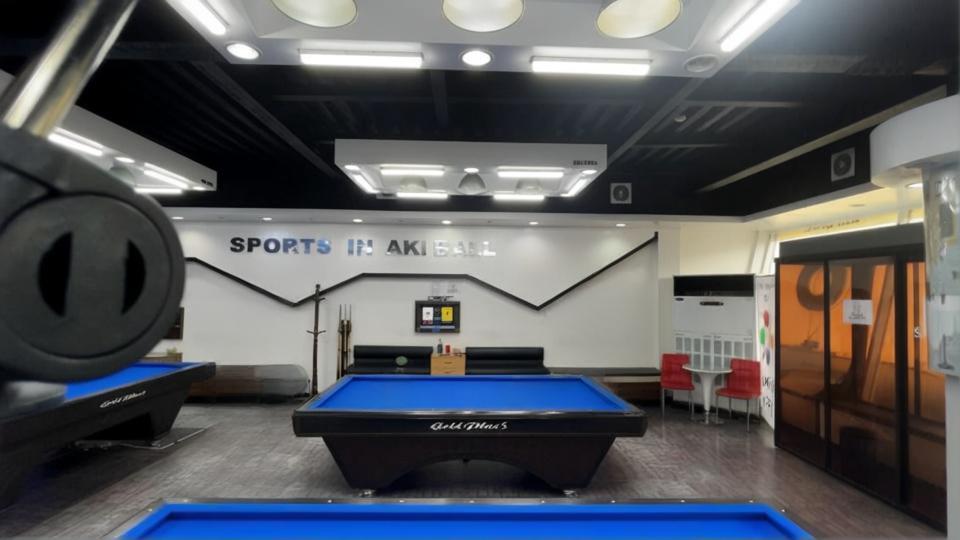}}\\
    \vspace{-0.1em}
    \subfloat{\includegraphics[width=0.14\linewidth]{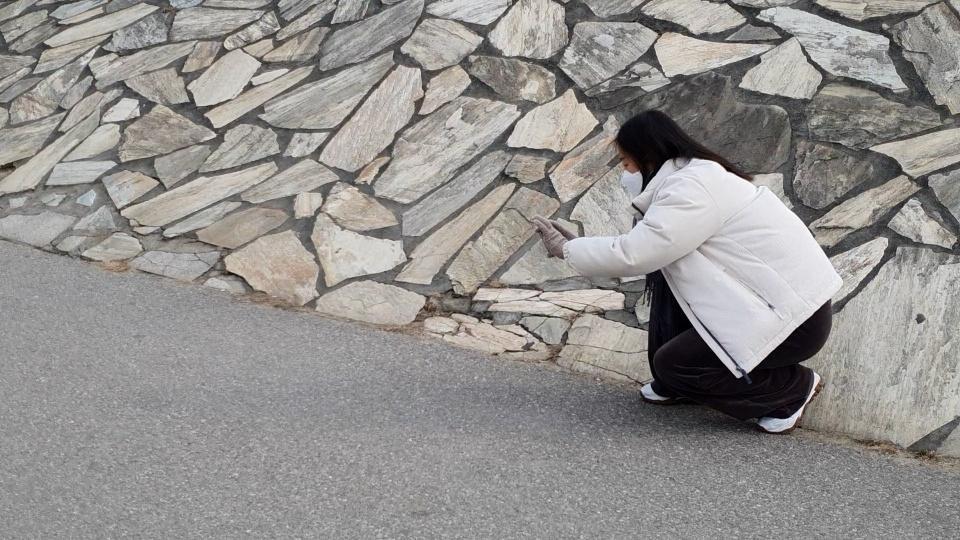}}
    \subfloat{\includegraphics[width=0.14\linewidth]{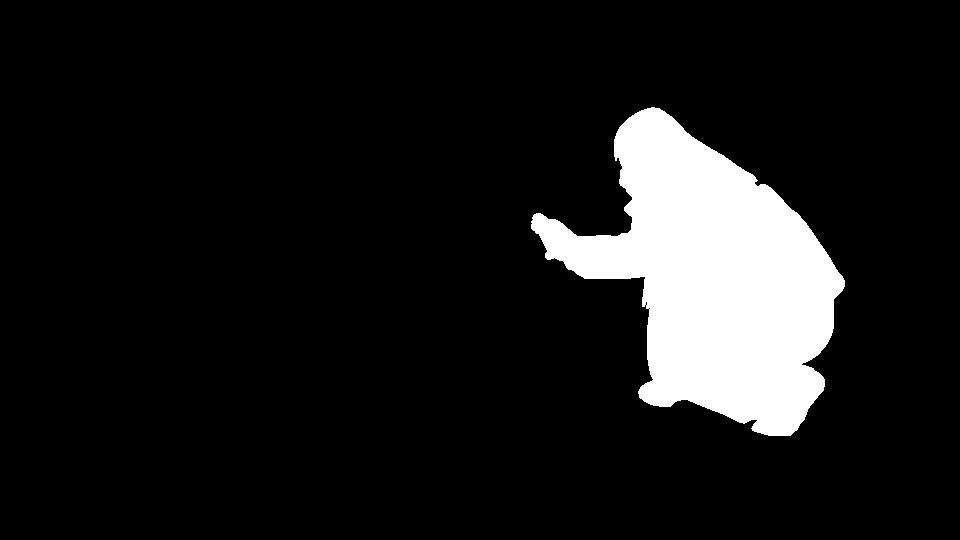}}
    \subfloat{\includegraphics[width=0.14\linewidth]{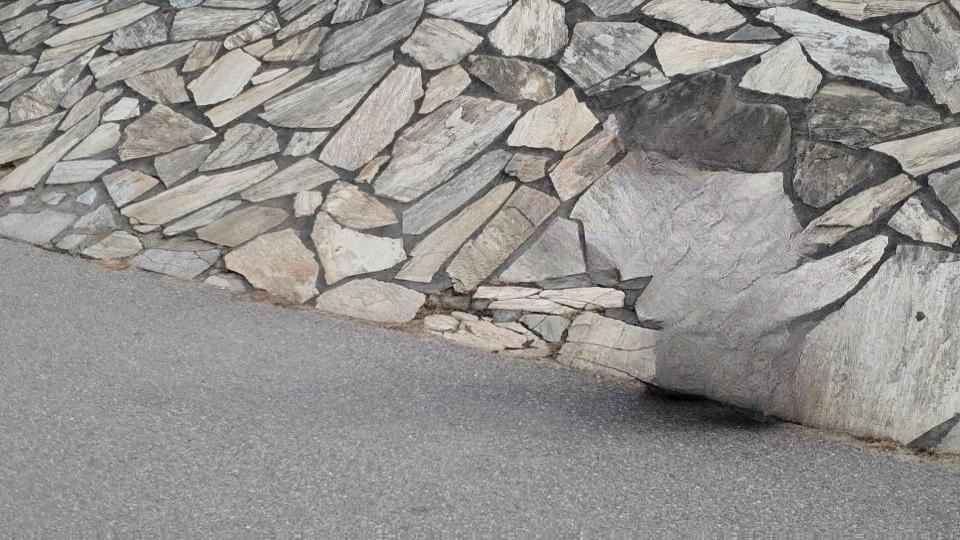}}
    \subfloat{\includegraphics[width=0.14\linewidth]{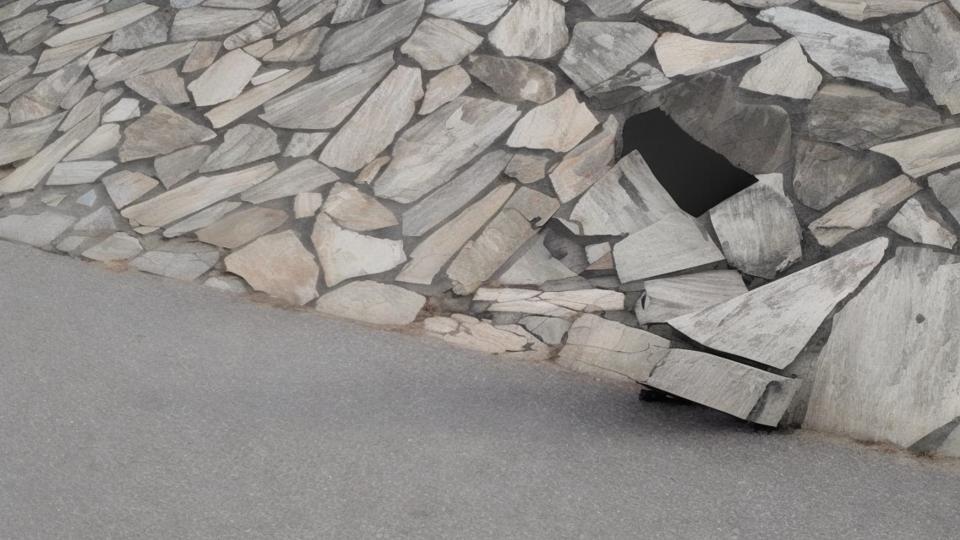}}
    \subfloat{\includegraphics[width=0.14\linewidth]{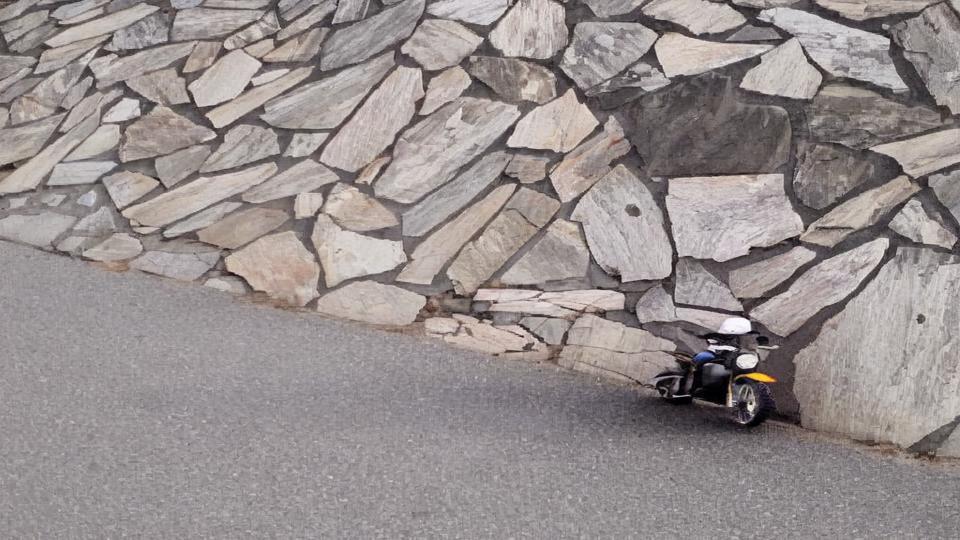}}
    \subfloat{\includegraphics[width=0.14\linewidth]{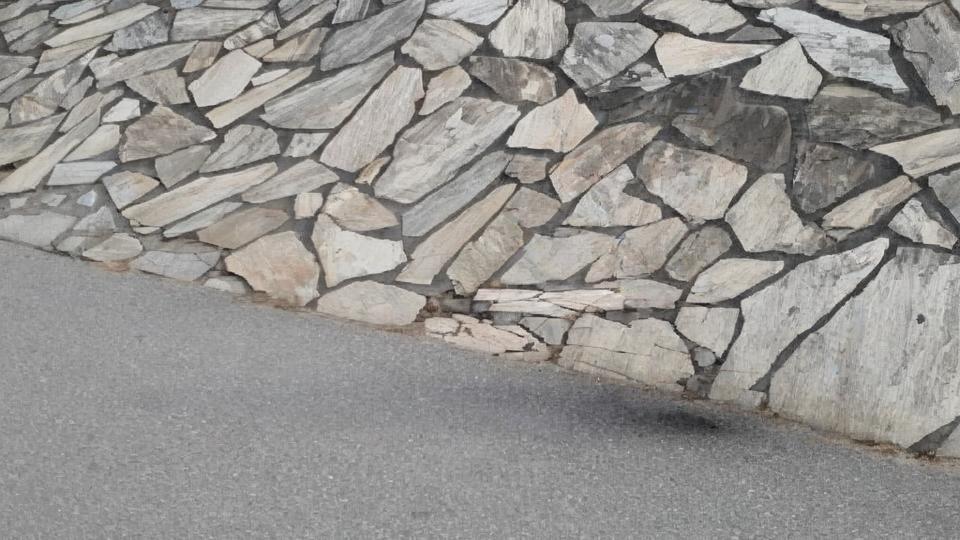}}
    \subfloat{\includegraphics[width=0.14\linewidth]{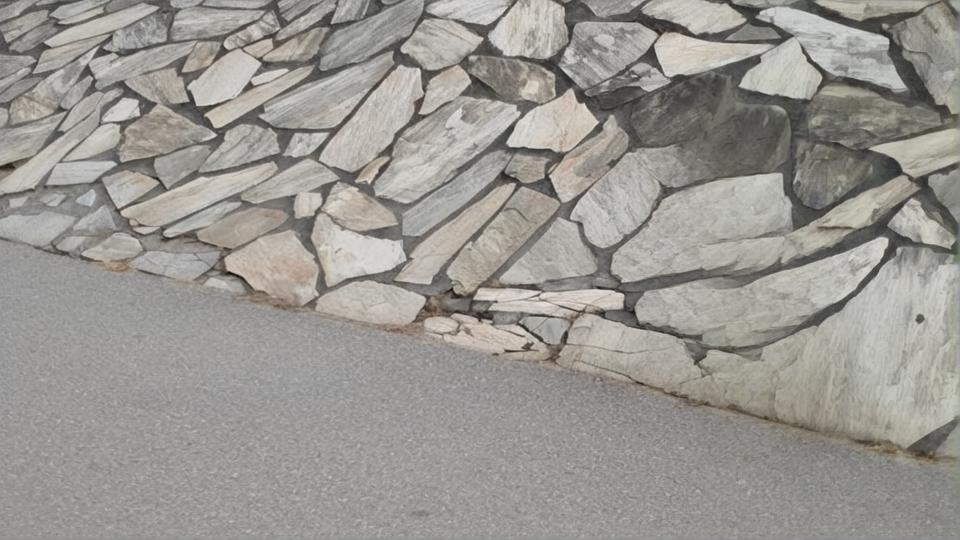}}\\
    \vspace{-0.1em}
    \subfloat{\includegraphics[width=0.14\linewidth]{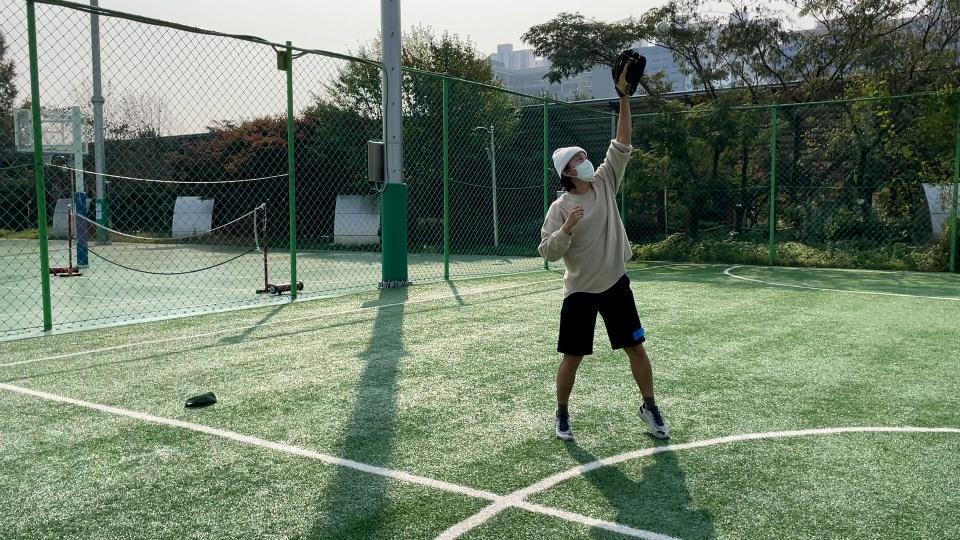}}
    \subfloat{\includegraphics[width=0.14\linewidth]{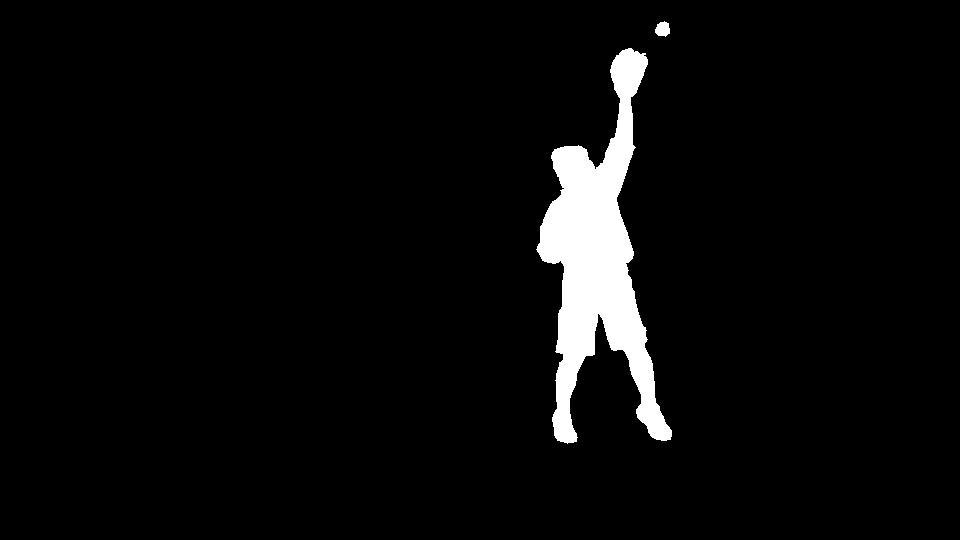}}
    \subfloat{\includegraphics[width=0.14\linewidth]{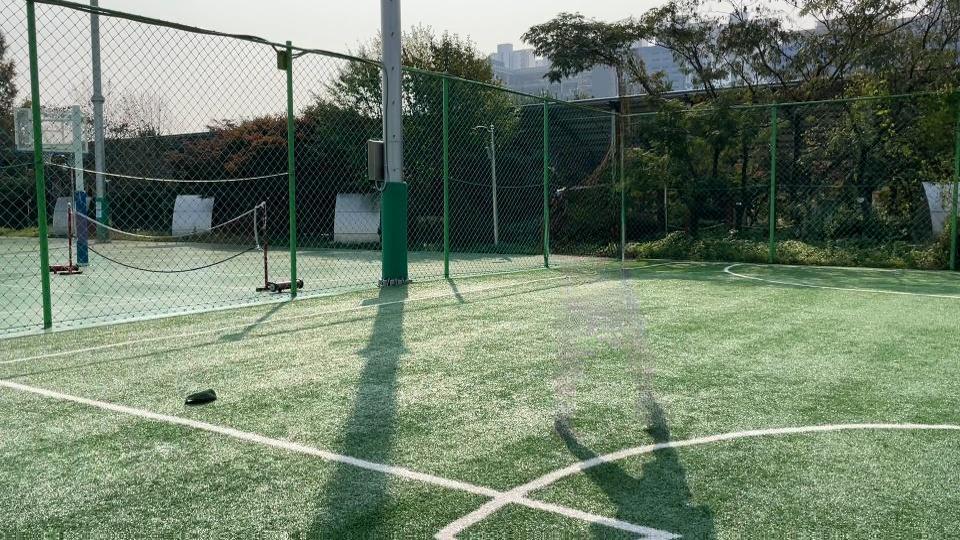}}
    \subfloat{\includegraphics[width=0.14\linewidth]{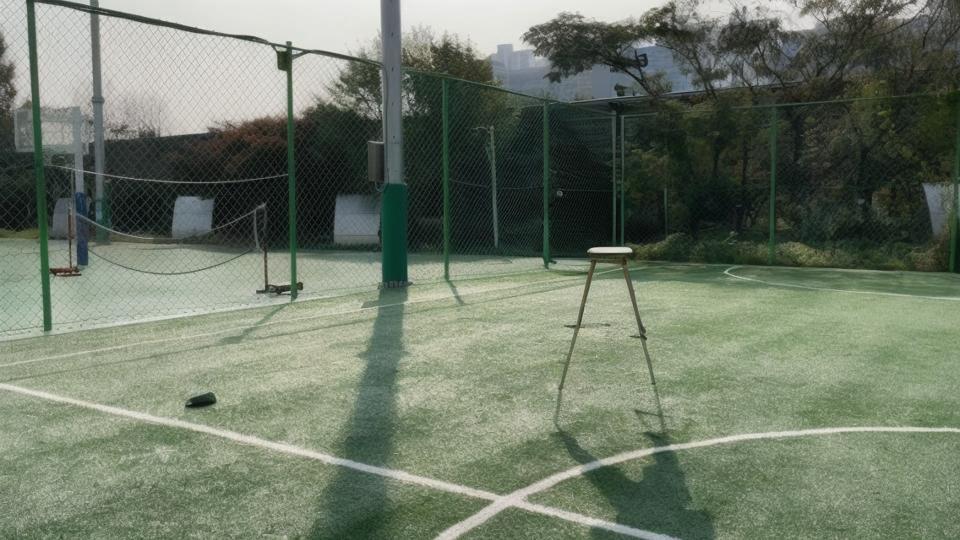}}
    \subfloat{\includegraphics[width=0.14\linewidth]{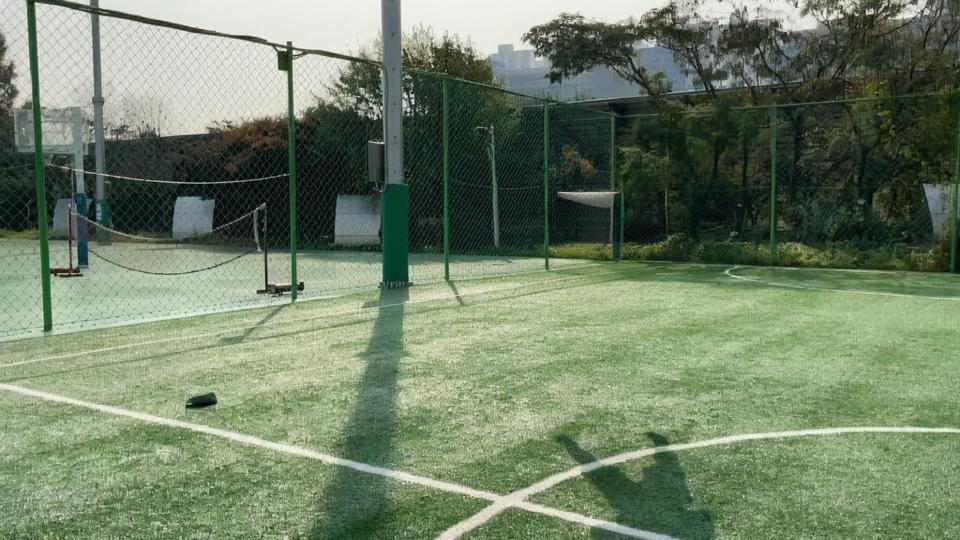}}
    \subfloat{\includegraphics[width=0.14\linewidth]{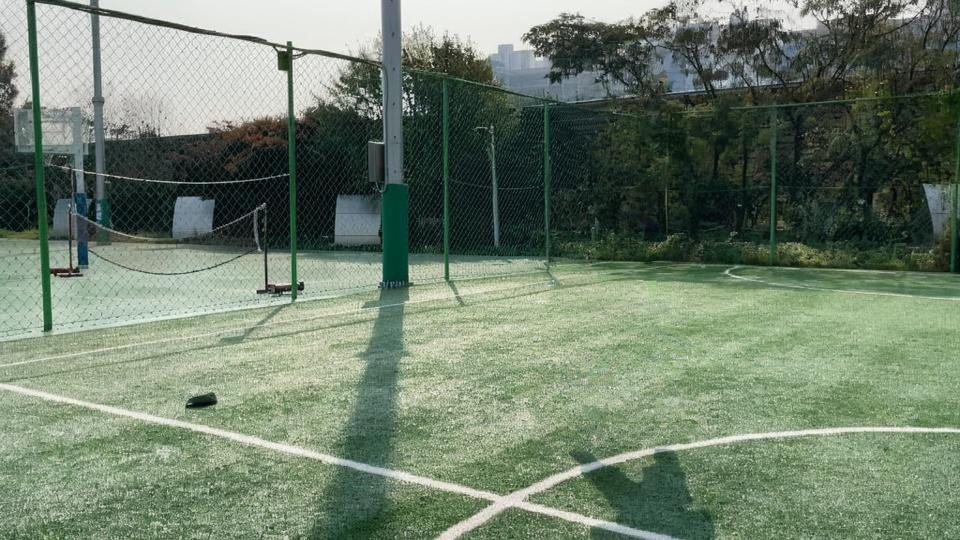}}
    \subfloat{\includegraphics[width=0.14\linewidth]{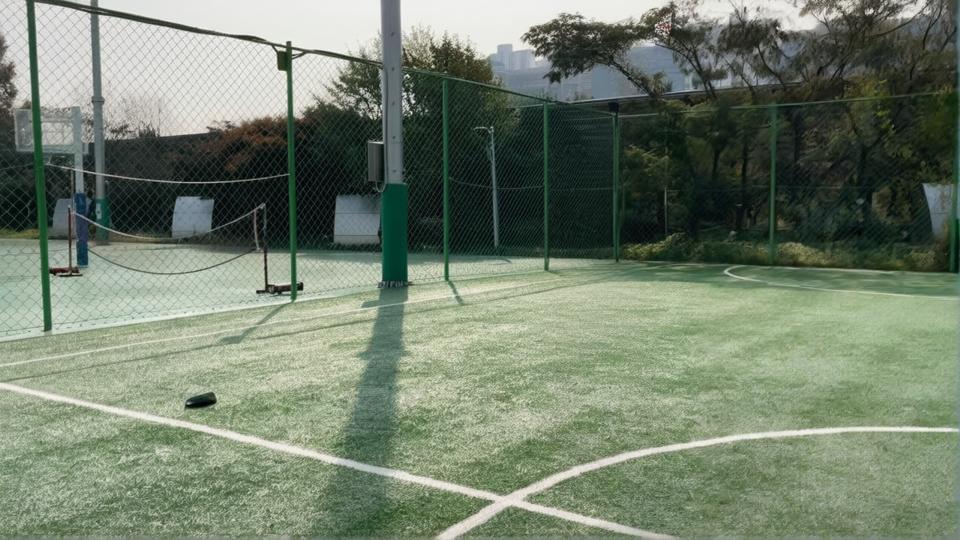}}\\
    \caption{Qualitative results for object-removal on RORD validation dataset \cite{sagong2022rord}. Best viewed zoomed in. Our model uses a single NFE and is able to successfully remove not only the object but also its shadow. Additional results are provided in the appendices.}
    \label{fig:object-removal}
\end{figure*}

% \begin{table*}[ht]
%     \centering
%     \scriptsize
%     \begin{tabular}{ccccccccc}
%         \toprule
%         Method & NFE & LPIPS $\downarrow$ & FID $\downarrow$ & Local FID $\downarrow$ & MSE $\downarrow$ & fMSE $\downarrow$ & PSNR $\uparrow$ & SSIM $\uparrow$ \\
%         \midrule
%         LAMA &1	& \textbf{22.05}	& 30.03	& 35.42	& 906.41	& 1596.71	& 19.65	& 54.49\\
%         SDXL inpainting &50	& 24.56	& 39.30	& 34.47	& 1150.26	& 2976.42	& 19.04	& 65.31\\
%         PowerPaint&50	& 26.31	& 29.83	& 26.04	& 926.31	& 2307.49	& 20.12	& 63.08\\
%         Attentive Eraser&50	& 24.34	& 29.70	& 33.15	& 798.84	& 2029.04	& 20.93	& 65.69\\
%         \midrule
%         Ours & 1 & \underline{23.17}	& \textbf{26.29}	& \textbf{27.91}	& \textbf{552.88}	& \textbf{1314.58}	& \textbf{22.38}	& \textbf{69.06}\\
%         \bottomrule
%     \end{tabular}
%     \caption{Metrics for object-removal task computed on RORD validation set (52k images) using the corse semantic masks. Our method uses a single NFE. The same results using the fine semantic masks are provided in Appendices.}
%     \label{tab:object-removal}
% \end{table*}

\section{Experiments}
In this section, we validate our method on 6 different image-to-image tasks: object-removal, depth and surface estimation, object relighting with respect to a given background image or light conditions as well as shadow generation. Additionally, we also provide a qualitative overview of how our method performs for image restoration in the appendices. Finally, we also ablate the main components of the proposed method such as the choice of the timestep distribution, the loss function, the number of inference steps and the choice of $\sigma$. In the following, unless stated otherwise, we use a latent approach and so embed the source and target images in a latent space using a VAE. The parametrized drift function $v_{\theta}$ is a U-Net \cite{ronneberger2015u} initialized with the weights of the pre-trained text-to-image model SDXL \cite{podell2023sdxl} and we train the full U-Net using 2 H100 GPUs.

\subsection{Object-removal}\label{sec:object-removal}
The first task we consider consists of removing objects from an image the position of which are specified with a mask. For this setting, $\pi_0$ corresponds to the distribution of latents associated with the masked images while $\pi_1$ is the distribution of latents associated with the images without the objects. We create the masked images by replacing the pixels in the masked region with uniformly sampled random pixels. The model is then trained to find a transport map from $\pi_0$ to $\pi_1$ \emph{i.e.} a mapping that transports the masked images to the images without the objects. We train our model for 20k iterations on a combination of: 1) the RORD train dataset \cite{sagong2022rord} (composed of paired images with and without objects and associated masks), 2) a synthetic dataset where we created pairs of images with and without objects using the rendering engine Blender\footnote{https://www.blender.org} and 3) in-the-wild images where we randomly masked an area of the image in a similar fashion as \cite{suvorov2022resolution}. In the latter case, since the mask is created randomly, there may not be any object in the masked region and so the task consists in simply reconstructing the original image. This allows the model to handle cases where the mask does not contain any object at inference time without compromising the quality of the generation. See the appendices for all relevant training parameters.

We compare our approach with LAMA \cite{suvorov2022resolution}, SDXL-Inpainting \cite{rombach2022high}, PowerPaint \cite{zhuang2024task} and Attentive Eraser \cite{sun2025attentive} and evaluate all the methods on the validation set of RORD dataset \cite{sagong2022rord} composed of approximately 52k pairs of images with and without objects. For each image, both fine semantic masks and coarse masks indicating the location of the objects to be removed are provided. We compute the FID score \cite{heusel2017gans}, Local FID (computed only on the masked region) \cite{xie2023smartbrush}, foreground MSE (fMSE), SSIM and PSNR metrics. As illustrated in \cref{tab:object-removal}, our model is able to outperform other approaches for most metrics even when using a single inference step. The evolution of the performance with respect to the number of inference steps is further discussed in \cref{sec:ablations}. We also provide a qualitative comparison of all the methods considered in \cref{fig:object-removal}. As illustrated, our model can remove not only the object but also its shadow as shown in the figure. See the appendices for additional samples as well as a discussion of the failure cases.

\begin{table}[ht]
    \centering
    \scriptsize
    \begin{tabular}{cccccc}
        \toprule
        Method (NFE) &FID $\downarrow$ & Local FID $\downarrow$ &fMSE $\downarrow$ & PSNR $\uparrow$ & SSIM $\uparrow$ \\
        \midrule
        LAMA (1) 	& 30.03	& 35.42	&  \underline{1596.71}	& 19.65	& 54.49\\
        SDXL inp. (50) 	& 39.30	& 34.47		& 2976.42	& 19.04	& 65.31\\
        PowerPaint (50)	& 29.83	& \textbf{26.04}	&  2307.49	& 20.12	& 63.08\\
        AE (50) & \underline{29.70}	& 33.15	&  2029.04	& \underline{20.93}	& \underline{65.69}\\
        \midrule
        Ours (1) &  \textbf{26.29}	& \underline{27.91}	&  \textbf{1314.58}	& \textbf{22.38}	& \textbf{69.06}\\
        \bottomrule
    \end{tabular}
    \caption{Metrics for object-removal task computed on RORD validation set (52k images) using the coarse semantic masks. Our method uses a single neural function evaluation (NFE). Best results are in bold, second best are underlined. The same results using the fine semantic masks are provided in appendices.}
    \label{tab:object-removal}
\end{table}

\begin{table*}[t]
    \centering
    \scriptsize
    \begin{tabular}{cccc|ccc|ccc|ccc|ccc|c}
        \toprule
        \multirow{2}{*}{Method} & \multicolumn{3}{c|}{NYUv2} & \multicolumn{3}{c|}{ScanNet} & \multicolumn{3}{c|}{iBims} & \multicolumn{3}{c|}{Sintel} & Avg.\\
        & m.$\downarrow$ & $11.25^{\circ} \uparrow$ & $30.0^{\circ} \uparrow$ & m.$\downarrow$ & $11.25^{\circ}\uparrow$ & $30.0^{\circ} \uparrow$ & m.$\downarrow$ & $11.25^{\circ}\uparrow$ & $30.0^{\circ} \uparrow$ & m.$\downarrow$ & $11.25^{\circ}\uparrow$ & $30.0^{\circ} \uparrow$ & Rank\\
        \midrule
        OASIS             & 29.2             & 23.8             & 60.7             & 32.8 & 15.4 & 52.6 & 32.6 & 23.5 & 57.4 & 43.1 & 7.0  & 35.7 & 12.3\\
        Omnidata          & 23.1             & 45.8             & 73.6             & 22.9 & 47.4 & 73.2 & 19.0 & 62.1 & 80.1 & 41.5 & 11.4 & 42.0 & 10.7\\
        EESNU             & \underline{16.2} & 58.6             & 83.5             & -    & -    & -    & 20.0 & 58.5 & 78.2 & 42.1 & 11.5 & 41.2 & 8.7\\
        GenPercept        & 18.2             & 56.3             & 81.4             & 17.7 & 58.3 & 82.7 & 18.2 & 64.0 & 82.0 & 37.6 & 16.2 & 51.0 & 7.5\\
        Omnidata V2       & 17.2             & 55.5             & 83.0             & 16.2 & 60.2 & 84.7 & 18.2 & 63.9 & 81.1 & 40.5 & 14.7 & 43.5 & 7.1\\
        DSINE             & 16.4             & 59.6             & 83.5             & 16.2 & 61.0 & 84.4 & 17.1 & 67.4 & 82.3 & 34.9 & 21.5 & 52.7 & 4.4 \\
        Marigold          & 20.9             & 50.5             & -                & 21.3 & 45.6 & -    & 18.5 & 64.7 & -    & -    & -    & - &  9.8 \\
        GeoWizard         & 18.9             & 50.7             & 81.5             & 17.4 & 53.8 & 83.5 & 19.3 & 63.0 & 80.3 & 40.3 & 12.3 & 43.5 & 9.1\\
        StableNormal      & 18.6             & 53.5             & 81.7             & 17.1 & 57.4 & 84.1 & 18.2 & 65.0 & 82.4 & 36.7 & 14.1 & 50.7 & 7.2\\
        Lotus-D           & \underline{16.2} & 59.8             & \underline{83.9} & \underline{14.7} & 64.0 & \underline{86.1} & 17.1 & 66.4 & \underline{83.0} & \underline{32.3} & \underline{22.4} & \underline{57.0} & \underline{2.4}\\
        Lotus-G           & 16.5             & 59.4             & 83.5             & 15.1 & 63.9 & 85.3 & 17.2 & 66.2 & 82.7 & 33.6 & 21.0 & 53.8 & 4.1\\
        Diff.-E2E-FT  & 16.5             & \underline{60.4} & 83.1             & \underline{14.7} & \textbf{66.1} & 85.1 & \textbf{16.1} & \textbf{69.7} & \textbf{83.9} & 33.5 & 22.3 & 53.5 & 2.8\\
        \midrule
        Ours   & \textbf{15.5} & \textbf{62.5} & \textbf{84.9} & \textbf{14.1} & \underline{65.8} & \textbf{87.0} & \underline{16.9} & \underline{68.3} & 82.7 & \textbf{32.2} & \textbf{24.0} & \textbf{58.6} & \textbf{1.4}\\
        \bottomrule
    \end{tabular}
    \caption{Quantitative results for normal estimation. Our method uses a single NFE. Competitors results are taken from \cite{he2024lotus}. Best results are in bold, second best are underlined. We provide the same table for depth estimation in appendices.}
    \label{tab:normal}
\end{table*}

\subsection{Surface and depth estimation}
Monocular depth and normal estimation are typically challenging image translation tasks that require the model to build an understanding of the geometry of a given scene with a single image. There exist a large number of methods trying to tackle either monocular depth estimation \cite{fu2018deep, lee2019big,yuan2022neural,yin2021virtual,ranftl2020towards,ranftl2021vision,eftekhar2021omnidata,yin2021learning,zhang2022hierarchical,yang2024depth,yang2025depth,yin2023metric3d,hu2024metric3d,kar20223d,he2024lotus,gui2024depthfm,bochkovskii2024depth,garcia2024fine,ke2024repurposing,xu2024diffusion} or normal estimation from a single image \cite{yin2021virtual,chen2020oasis,bae2021estimating,eftekhar2021omnidata,bae2024rethinking,he2024lotus,garcia2024fine,xu2024diffusion,fu2024geowizard,ye2024stablenormal}. In this setting, $\pi_0$ is the latent distribution of the images while $\pi_1$ is the distribution of the latents of the depth maps (resp. normal maps). We train a model using our framework for both tasks and provide in appendices any relevant training parameter.

We perform a zero-shot evaluation on commonly used evaluation datasets such as NYUv2 \cite{silberman2012indoor}, KITTI \cite{geiger2013vision}, ETH3D \cite{schops2017multi}, ScanNet \cite{dai2017scannet} and DIODE \cite{vasiljevic2019diode} for depth estimation and NYUv2, ScanNet, i-Bims \cite{koch2018evaluation} and Sintel \cite{butler2012naturalistic} for normal estimation. For the latter, we report in \cref{tab:normal} the mean angular error and the percentage of pixels with an angular error below 11.25 and 30 degrees. As highlighted on \cref{tab:normal}, the method is either able to outperform competitors or be competitive since it ranks amongst the top three models for each metric on all datasets. Moreover, our approach ranks 1.4 on average. We provide in appendices the same table for depth estimation showing the same tendency since the model achieves again the best average ranking.

\subsection{Image relighting}

\paragraph{Setting}
Another task we decide to tackle is object relighting which consists of manipulating the appearance of an image by changing the illumination of the scene. This can be performed by relighting the foreground of an image using a target background \cite{ke2022harmonizer,chen2023dense,guerreiro2023pct,wang2023semi,ren2024relightful,kim2024switchlight,zhang2025scaling}, or modifying an object or the full scene appearance using target lightings (\emph{e.g.} high dynamic range (HDR) maps) \cite{zeng2024dilightnet,deng2024flashtex,kocsis2024lightit,jin2024neural}. In particular, portrait relighting is a special case of image relighting that has driven strong interest in the past few years \cite{zhou2019deep,wang2020single,hou2021towards,nestmeyer2020learning,pandey2021total,sun2019single,yeh2022learning,zhang2021neural,zhang2020portrait}.

In this section, we focus on the task aiming at relighting a foreground object according to a given background, also known as image harmonization. In this case, we set $\pi_0$ to the encoded source images created by pasting the foreground onto the target background image and $\pi_1$ is the desired target relighted image. 

\paragraph{Dataset creation}
This task is quite challenging since most of the time there exist no such pairs of images \emph{i.e.} images with the exact same foreground but on different backgrounds and so under different light conditions. Since we do not have access to such data we rely on the following data creation strategy. 

We collect a set of various publicly available and free-to-use images with saliant foreground and compute the foreground mask for each of them using \cite{zheng2024birefnet} leading to a set of images $\mathcal{X}$. Then, given a pair of images $x_1, x_2 \in \mathcal{X}$, we use the foreground of $x_1$ (resp. $x_2$) and the IC-light model \cite{zhang2025scaling} to produce a relighted foreground $x_1^{\mathrm{fg}}$ (resp. $x_2^{\mathrm{fg}}$) according to the background of $x_2$ (resp. $x_1$). Finally, $x_1^{\mathrm{fg}}$ and $x_2^{\mathrm{fg}}$ are pasted back onto the original images $x_1$ and $x_2$ to produce the source images $y_1$ and $y_2$ while $x_1$ and $x_2$ are used as target images.

\begin{figure}[ht]
    \centering
    \scriptsize
    \captionsetup[subfigure]{position=above, labelformat = empty}
    \subfloat[BG 1]{\includegraphics[width=0.20\linewidth]{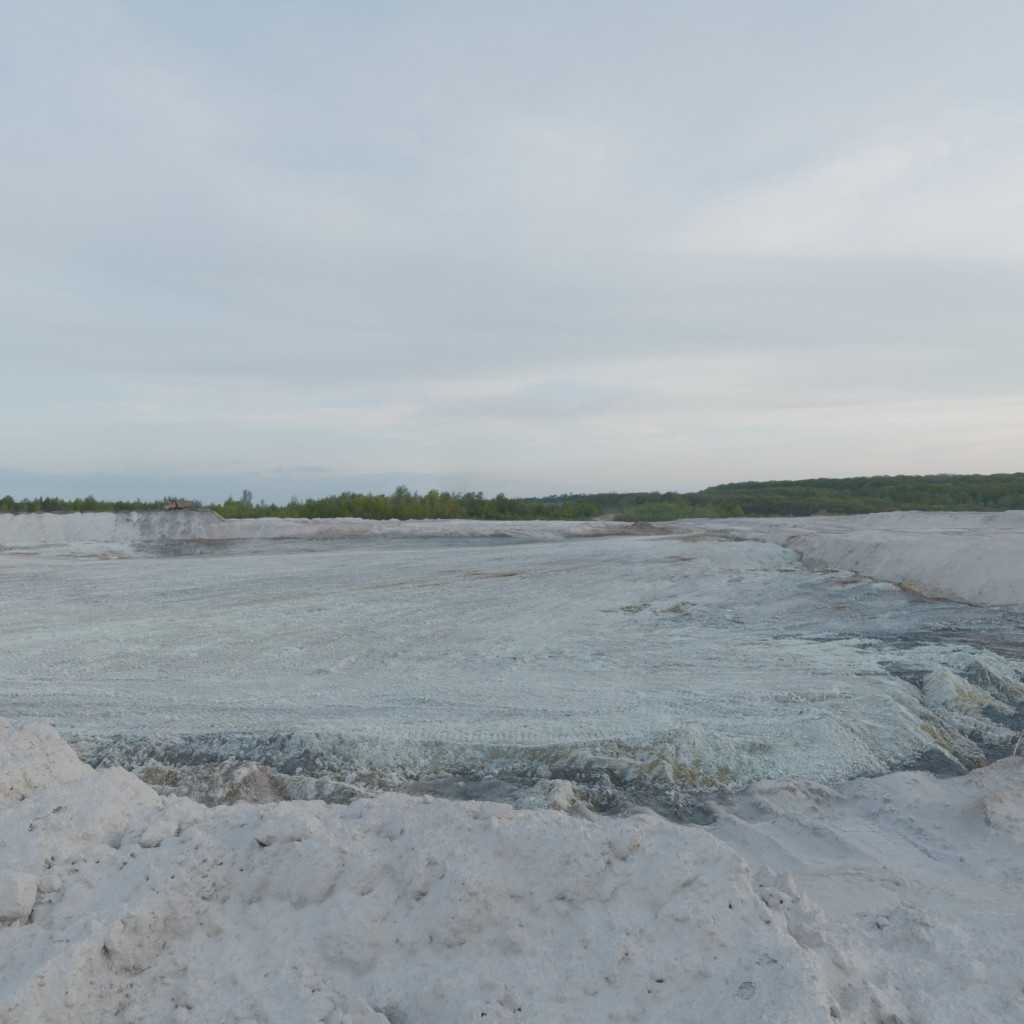}}
    \subfloat[$x_1$]{\includegraphics[width=0.20\linewidth]{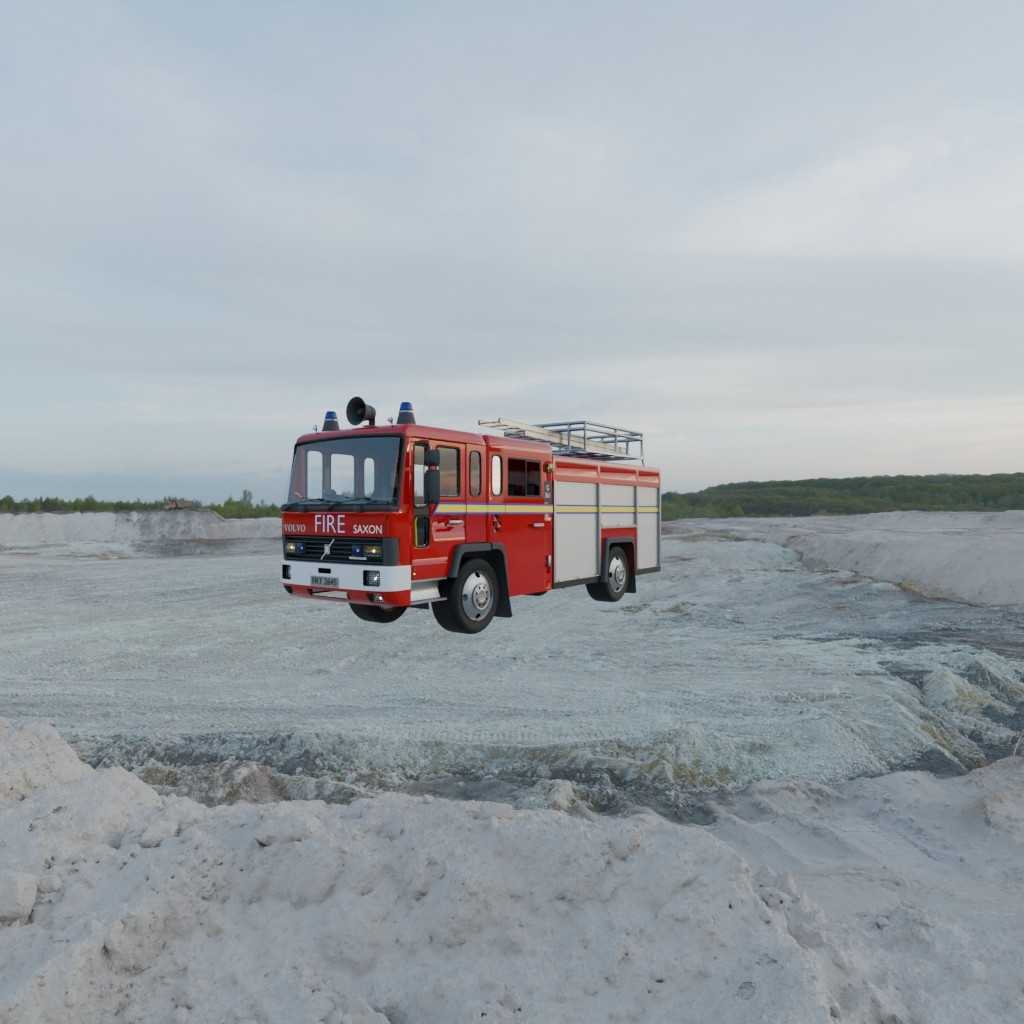}}
    \subfloat[BG 2]{\includegraphics[width=0.20\linewidth]{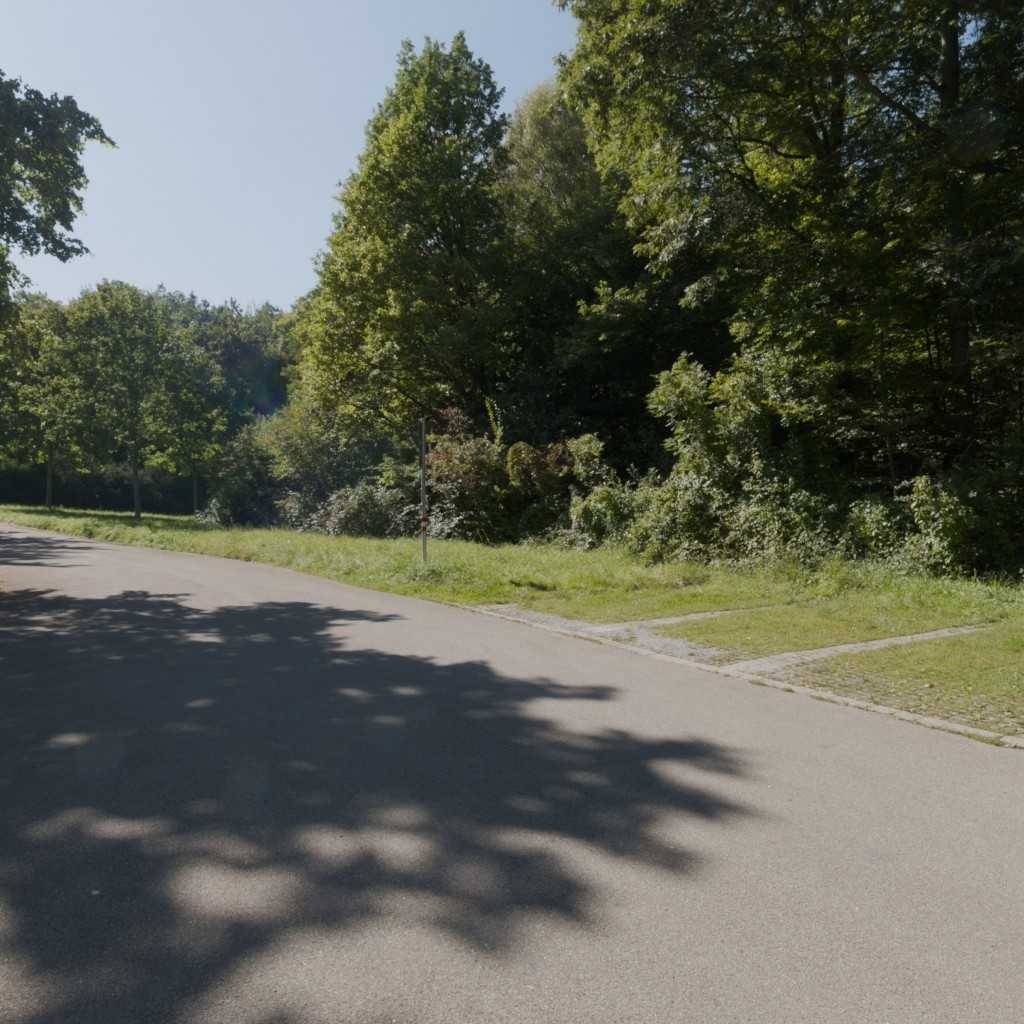}}
    \subfloat[$x_2$ (target)]{\includegraphics[width=0.20\linewidth]{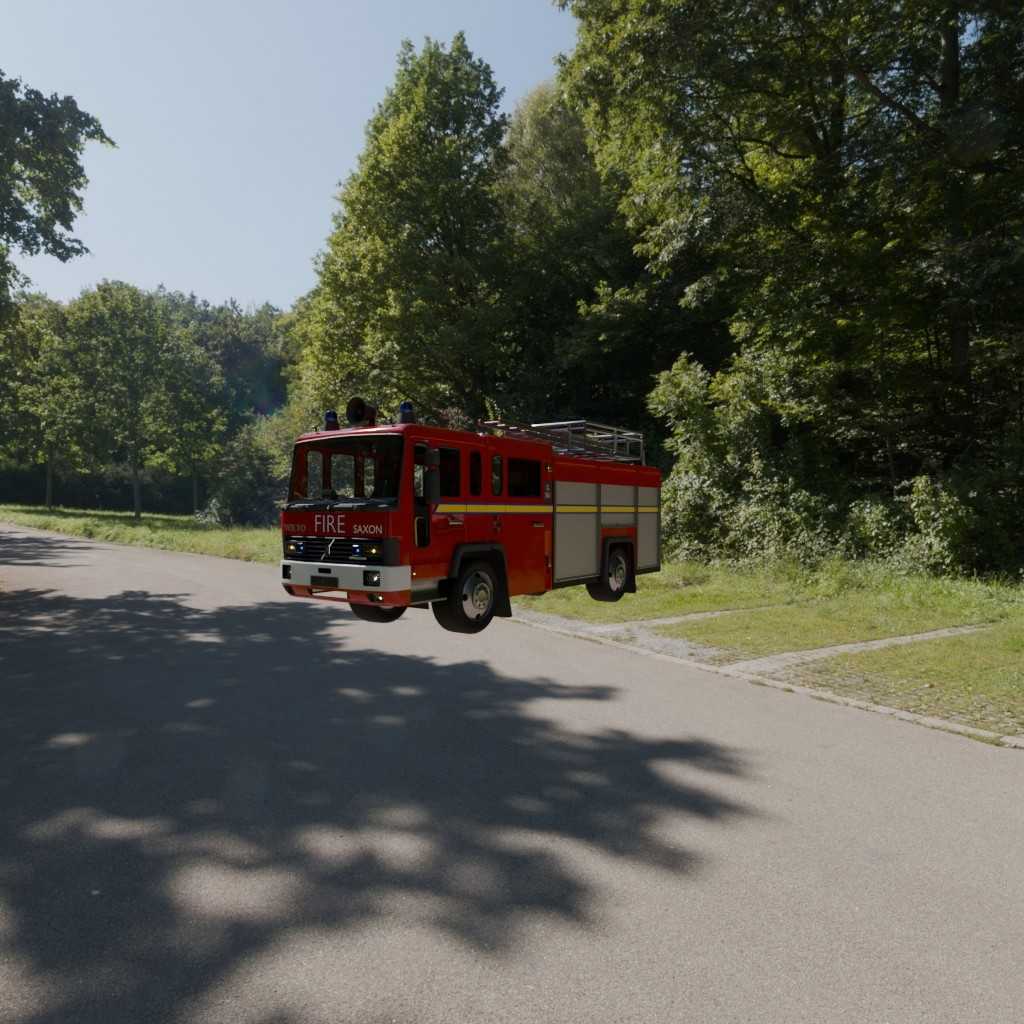}}
    \subfloat[$y_2$ (source)]{\includegraphics[width=0.20\linewidth]{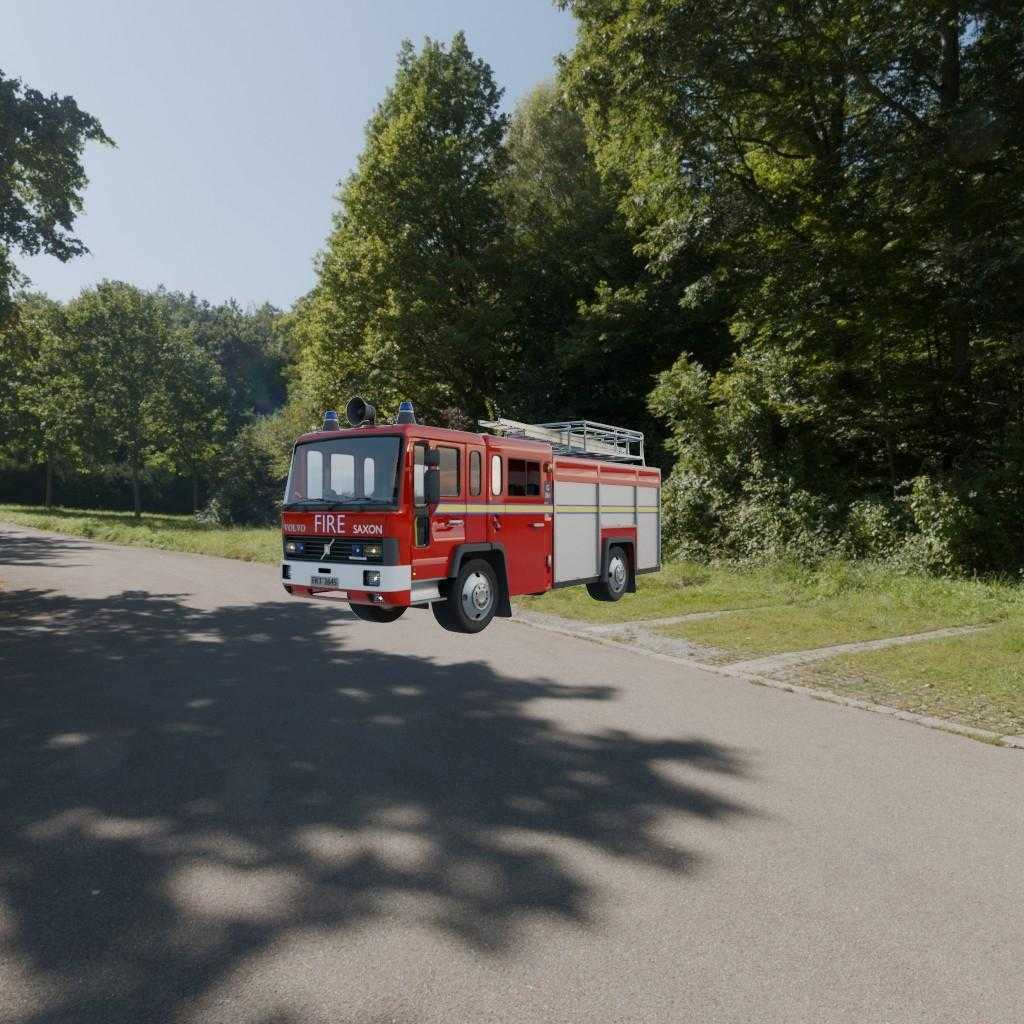}}
    \caption{Sample renders from our synthetic dataset. BG 1 is used to relight the 3D model and create $x_1$. BG 2 is used to produce another image with the same 3D model $x_2$ which is used as target.  Finally, the source image $y_2$ is created by pasting the foreground of $x_1$ on BG 2.}
    \label{fig:example_harmonization_renders}
\end{figure}

Additionally, we also rely on synthetic data created with the rendering engine Blender. Our synthetic dataset creation process begins by assembling a diverse collection of 3D object and human models, along with HDR images. These elements are then used to render high-quality images. For the objects, we collect an extensive selection of high-quality 3D models from BlenderKit\footnote{https://www.blenderkit.com}, a platform featuring professionally crafted assets available under a free-to-use license. For humans, we use a Blender addon\footnote{https://www.humgen3d.com} to generate unique 3D human models by randomly customizing facial features, body shapes, poses, hair, and clothing options. In each iteration of the dataset creation process, we begin by randomly selecting a 3D model. We then randomly select HDR images to illuminate the foreground object. We render the scene, and save the image and associated foreground mask giving $x_1$. We perform the same using another HDR map but with the same 3D object giving $x_2$. Finally, we can paste the foreground of $x_1$ on $x_2$ and vice-versa to create the source images $y_2$ and $y_1$ and use again $x_1$ and $x_2$ as target. Example renders from our dataset are shown in \cref{fig:example_harmonization_renders}.

% We then position between one and three light sources at random locations, each assigned a different color and energy level. Using this setup, we render an image, $y_1$, along with its corresponding object mask, $y_1^m$. Afterward, we remove the light sources, and load a randomly selected HDRI to illuminate the foreground object. We render the scene, producing the target image $y_2$. We then remove the foreground 3D model, and render the scene once more to create the target background, $y_2^{bg}$. Using $y_1^m$, we extract the foreground object from $y_1$ and paste it into $y_1^{bg}$, creating the source image $x_1$ that is designed to be transformed into $y_2$. Example renders from our dataset are shown in Fig.~\cref{fig:example_harmonization_renders}.

% \begin{figure}[t]
%     \centering
%     \begin{subfigure}[b]{0.24\linewidth}
%         \centering
%         \includegraphics[width=\linewidth]{plots/synthetic/source.jpg}
%     \end{subfigure}
%     \hfill
%     \begin{subfigure}[b]{0.24\linewidth}
%         \centering
%         \includegraphics[width=\linewidth]{plots/synthetic/target_bg.jpg}
%     \end{subfigure}
%     \hfill
%     \begin{subfigure}[b]{0.24\linewidth}
%         \centering
%         \includegraphics[width=\linewidth]{plots/synthetic/source_pasted_on_target.jpg}
%     \end{subfigure}
%     \hfill
%     \begin{subfigure}[b]{0.24\linewidth}
%         \centering
%         \includegraphics[width=\linewidth]{plots/synthetic/target.jpg}
%     \end{subfigure}
%     \caption{Sample renders from our synthetic dataset. From left to right: an object illuminated with light sources, a target background, the object from the first image placed onto the target background, and the target image itself.}
%     \label{fig:example_harmonization_renders}
% \end{figure}

\begin{figure*}[ht]
    \captionsetup[subfigure]{position=above, labelformat = empty}
    \centering
    % \subfloat{\includegraphics[width=\linewidth]{plots/relighting/plot_2.jpg}}\\
    % \vspace{-0.2em}
    \subfloat[Background]{\includegraphics[width=0.124\linewidth]{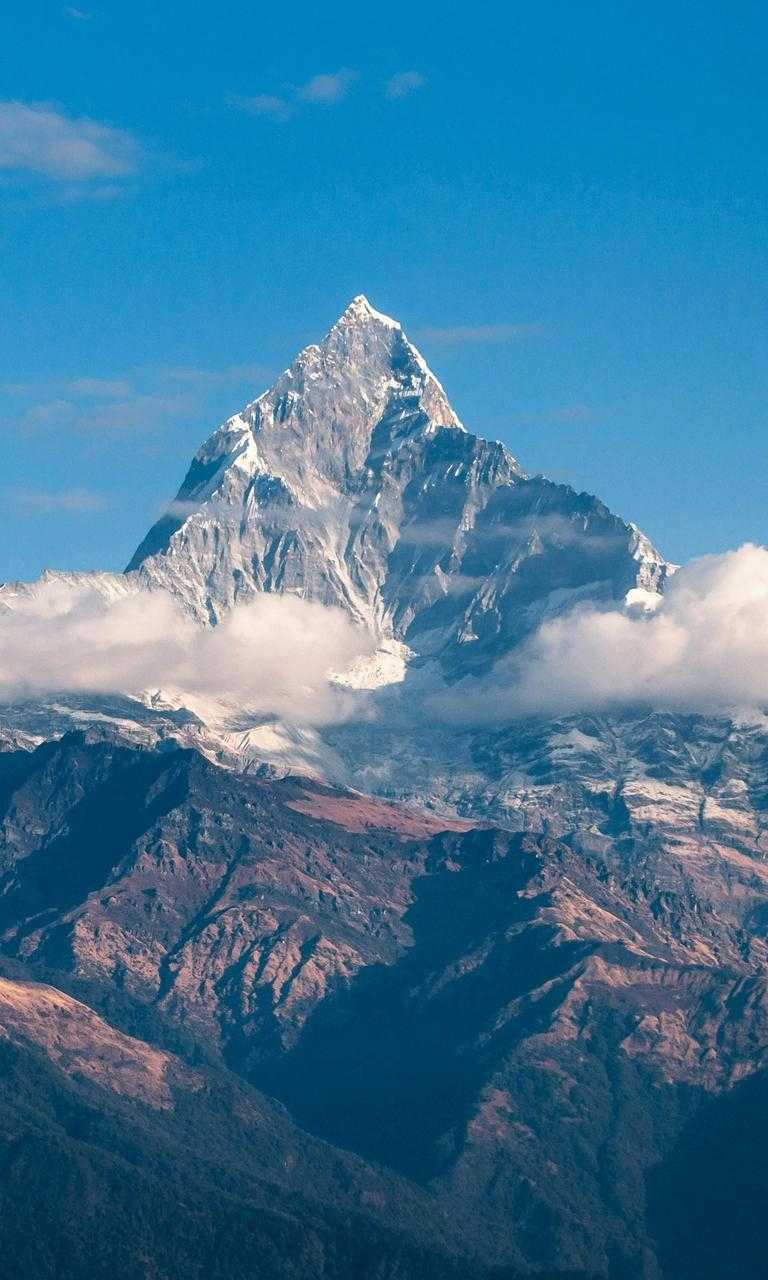}}
    \subfloat[Composite]{\includegraphics[width=0.124\linewidth]{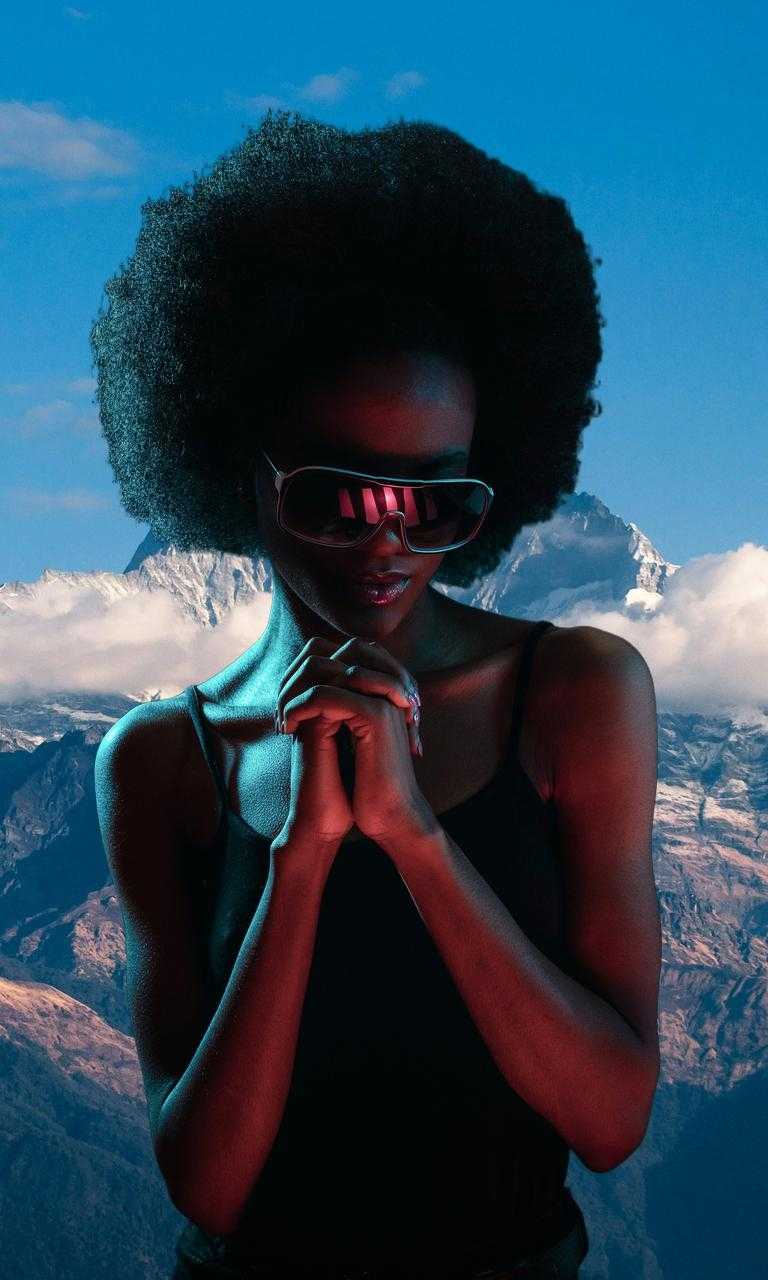}}
    \subfloat[Harmonizer]{\includegraphics[width=0.124\linewidth]{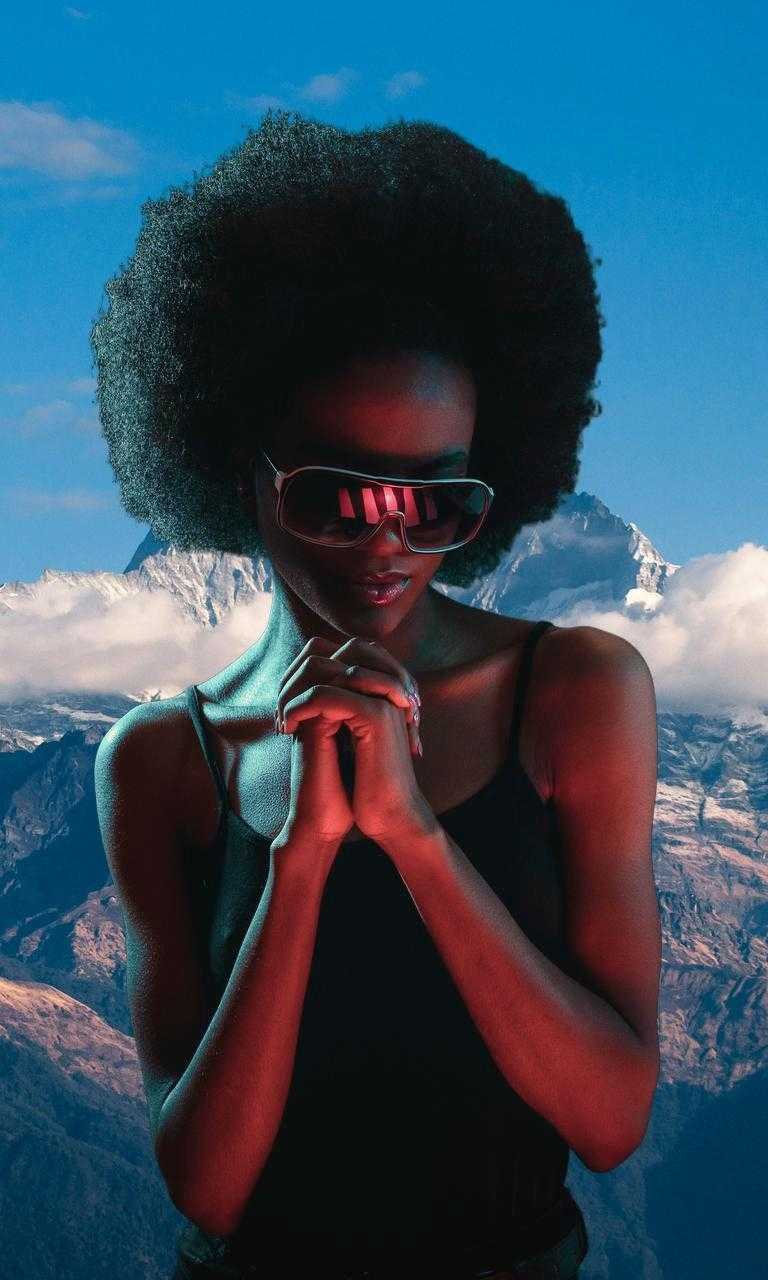}}
    \subfloat[PIH]{\includegraphics[width=0.124\linewidth]{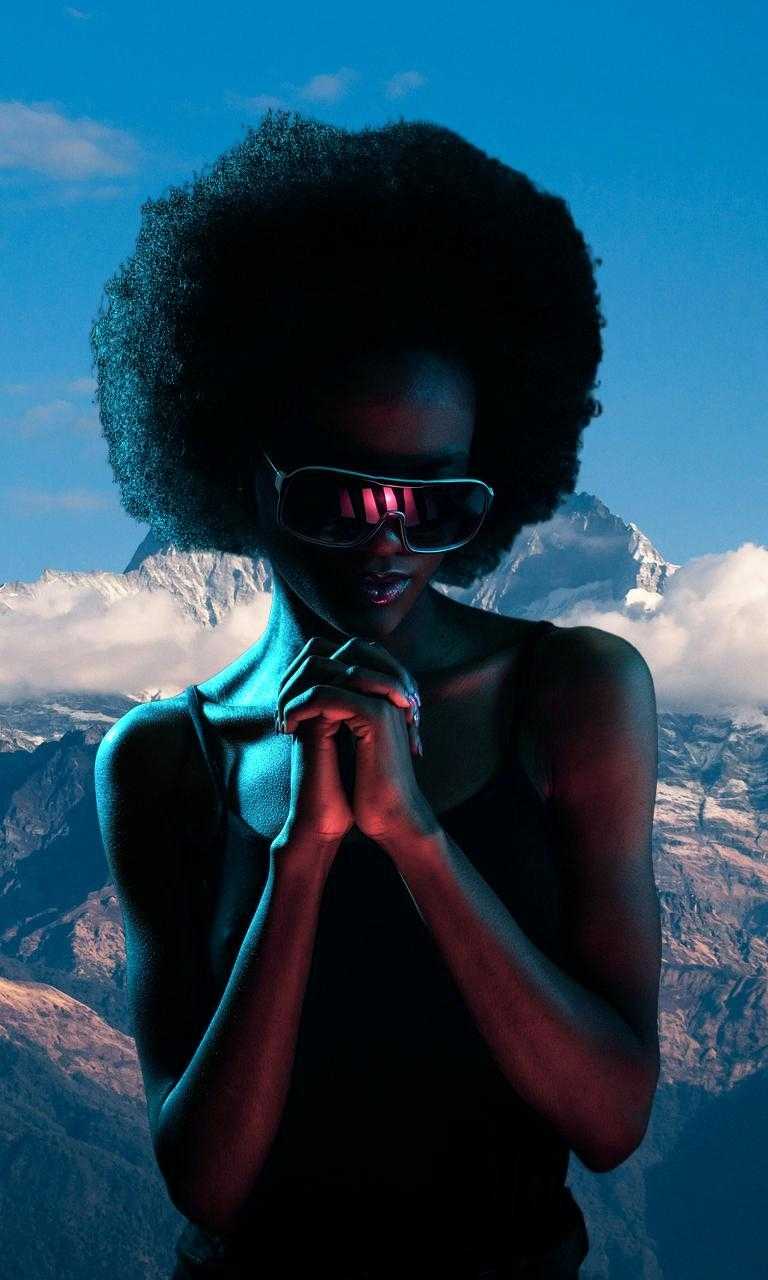}}
    \subfloat[PCT-Net]{\includegraphics[width=0.124\linewidth]{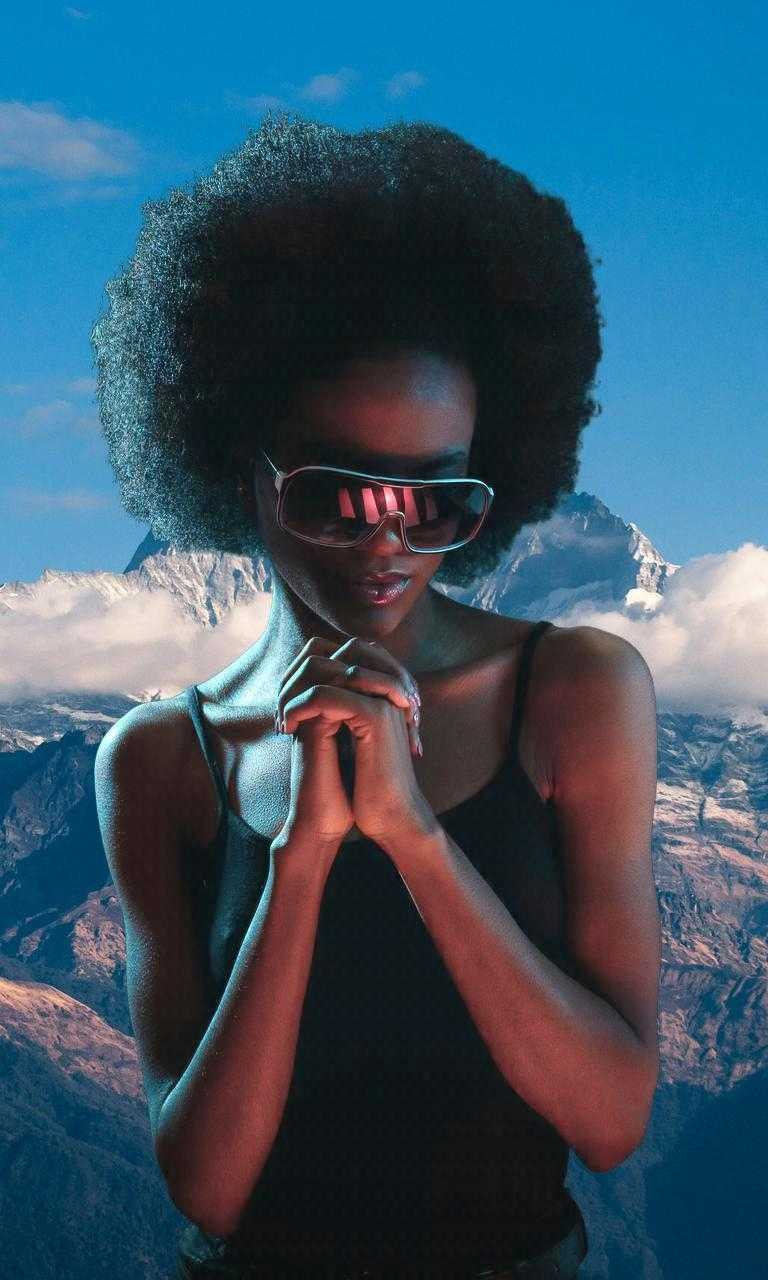}}
    \subfloat[INR]{\includegraphics[width=0.124\linewidth]{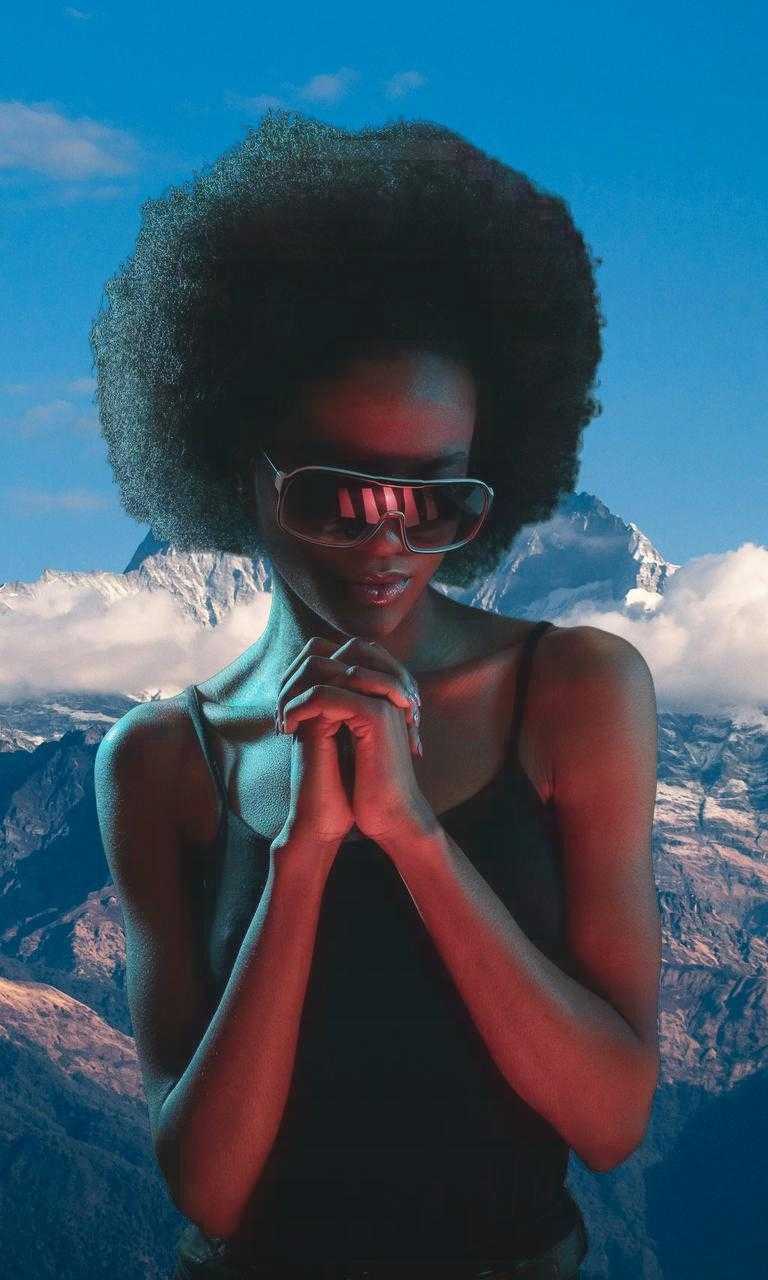}}
    \subfloat[IC-Light]{\includegraphics[width=0.124\linewidth]{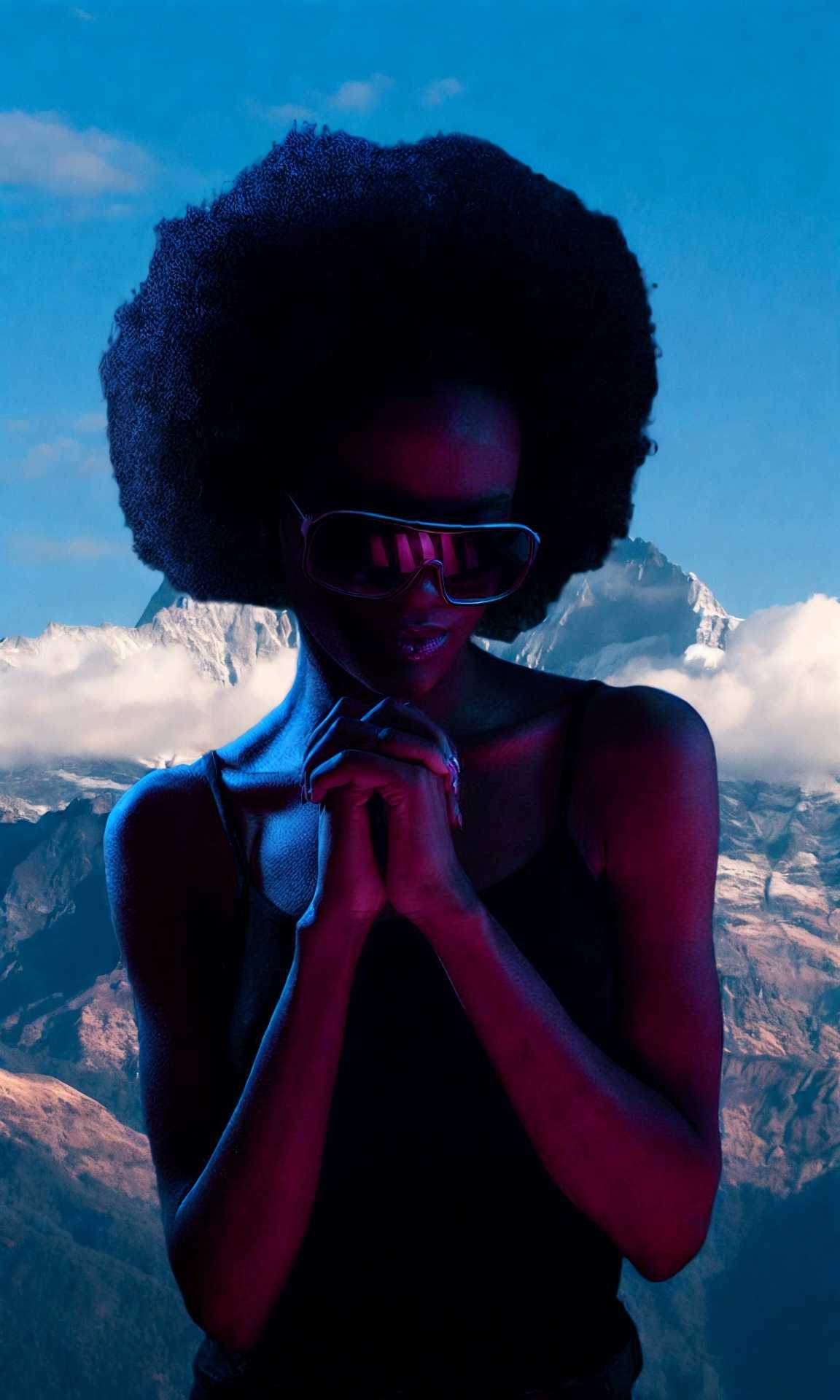}}
    \subfloat[Ours]{\includegraphics[width=0.124\linewidth]{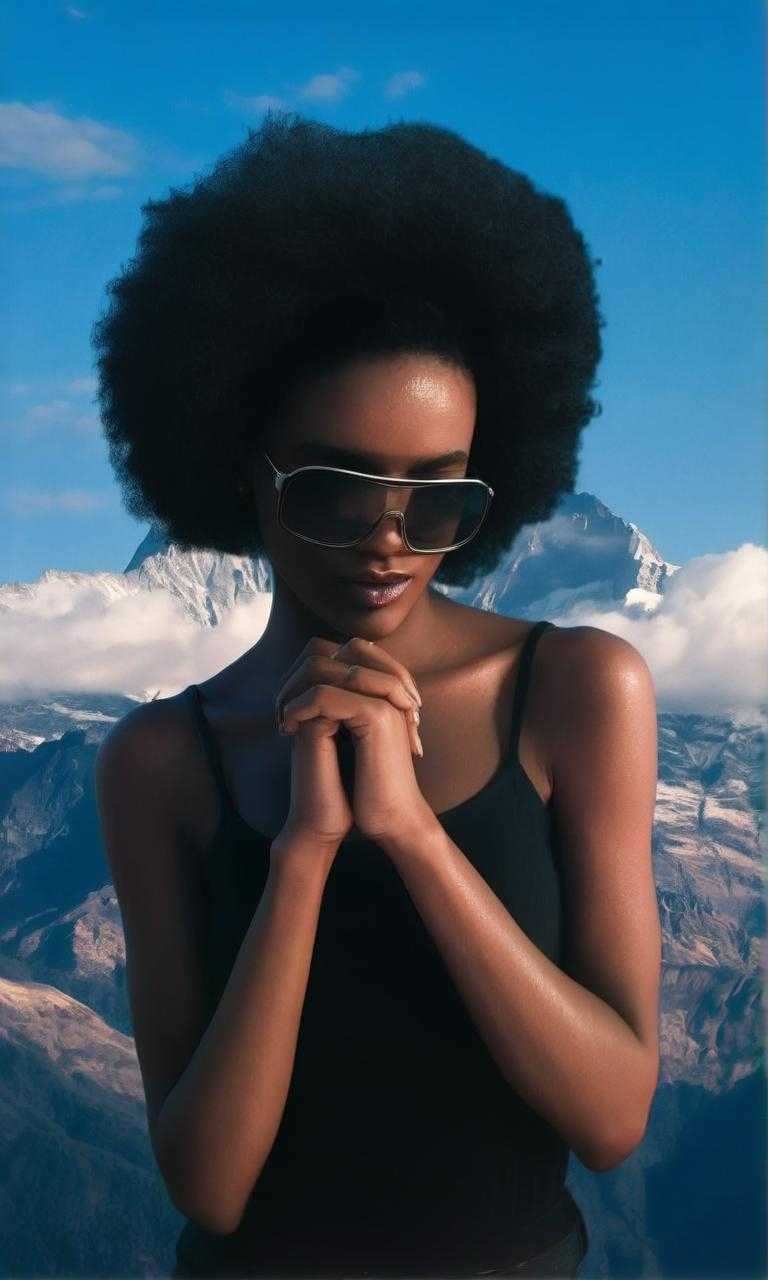}}\\
    \vspace{-0.10em}
    % \subfloat[Composite]{\includegraphics[width=0.14\linewidth]{plots/relighting/composite/78.jpg}}
    % \subfloat[Harmonizer]{\includegraphics[width=0.14\linewidth]{plots/relighting/harmonizer/78.jpg}}
    % \subfloat[PIH]{\includegraphics[width=0.14\linewidth]{plots/relighting/pih/78.jpg}}
    % \subfloat[PCT-Net]{\includegraphics[width=0.14\linewidth]{plots/relighting/pct/78.jpg}}
    % \subfloat[INR]{\includegraphics[width=0.14\linewidth]{plots/relighting/inr/78.jpg}}
    % \subfloat[IC-Light]{\includegraphics[width=0.14\linewidth]{plots/relighting/ic_light/78.jpg}}
    % \subfloat[Ours]{\includegraphics[width=0.14\linewidth]{plots/relighting/rf/78.jpg}}\\
    %\vspace{-0.10em}
    % \vspace{-0.10em}
    % \subfloat[Composite]{\includegraphics[width=0.14\linewidth]{plots/relighting/composite/60.jpg}}
    % \subfloat[Harmonizer]{\includegraphics[width=0.14\linewidth]{plots/relighting/harmonizer/60.jpg}}
    % \subfloat[PIH]{\includegraphics[width=0.14\linewidth]{plots/relighting/pih/60.jpg}}
    % \subfloat[PCT]{\includegraphics[width=0.14\linewidth]{plots/relighting/pct/60.jpg}}
    % \subfloat[INR]{\includegraphics[width=0.14\linewidth]{plots/relighting/inr/60.jpg}}
    % \subfloat[IC-Light]{\includegraphics[width=0.14\linewidth]{plots/relighting/ic_light/60.jpg}}
    % \subfloat[Ours (1 NFE)]{\includegraphics[width=0.14\linewidth]{plots/relighting/rf/60.jpg}}\\
    % \vspace{-0.10em}
    \subfloat{\includegraphics[width=0.124\linewidth]{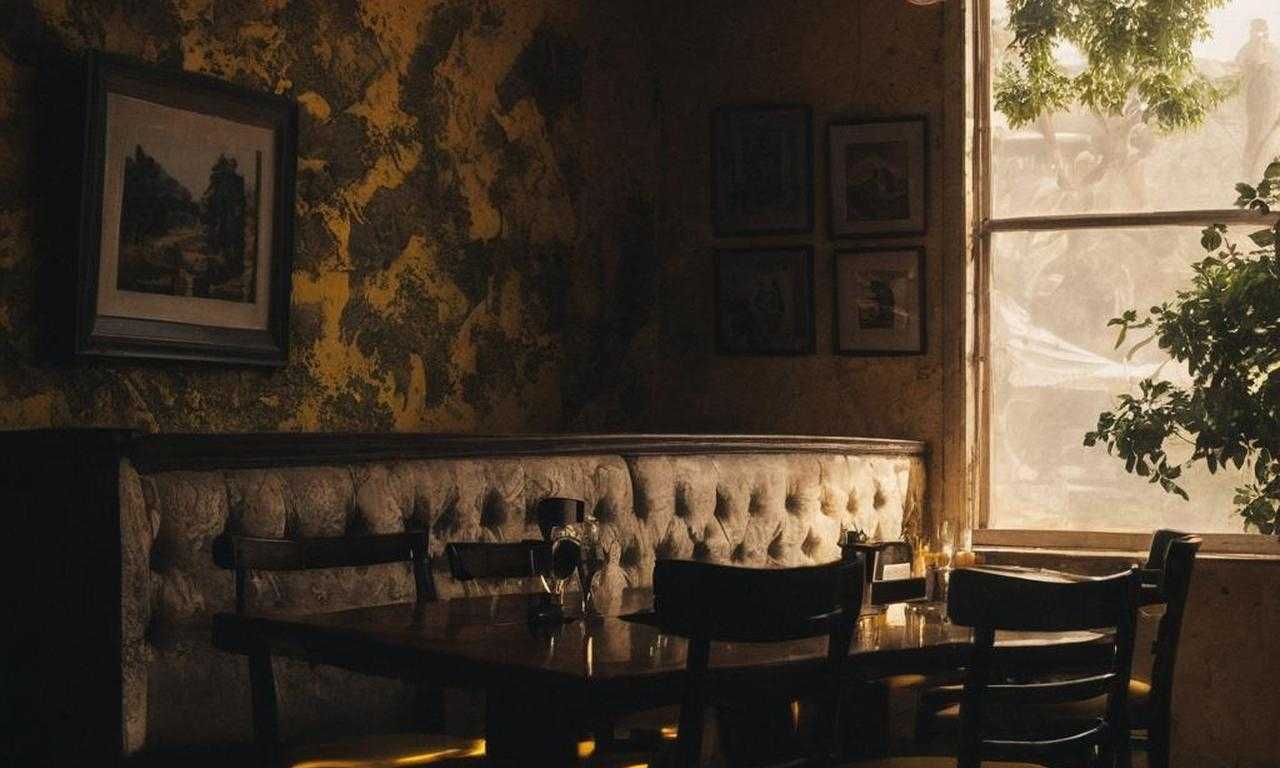}}
    \subfloat{\includegraphics[width=0.124\linewidth]{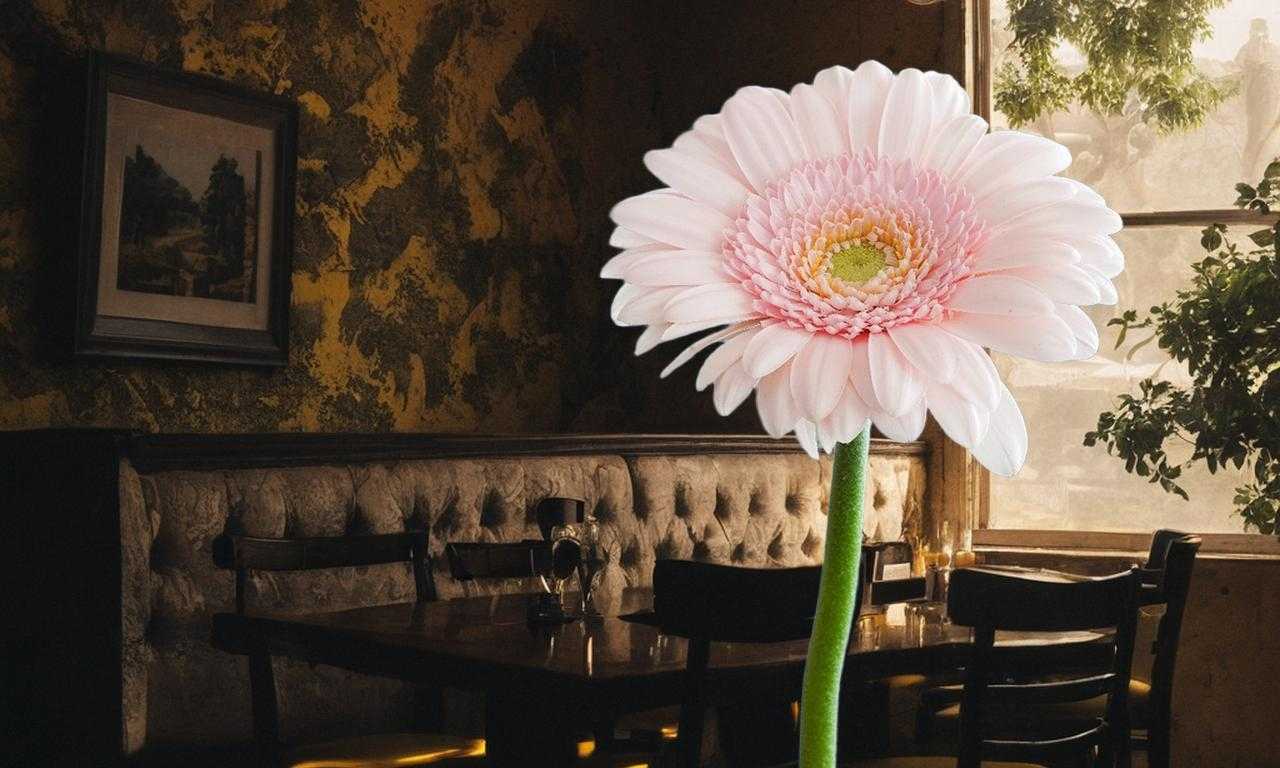}}
    \subfloat{\includegraphics[width=0.124\linewidth]{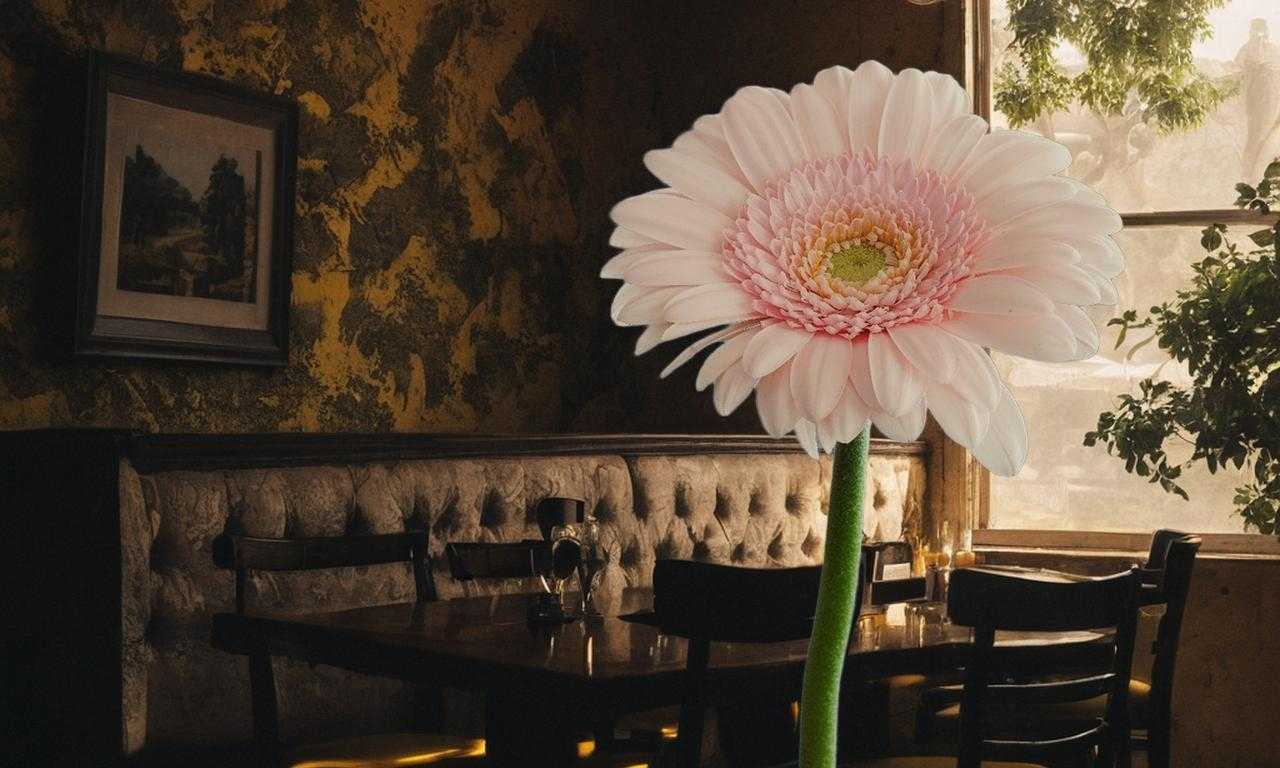}}
    \subfloat{\includegraphics[width=0.124\linewidth]{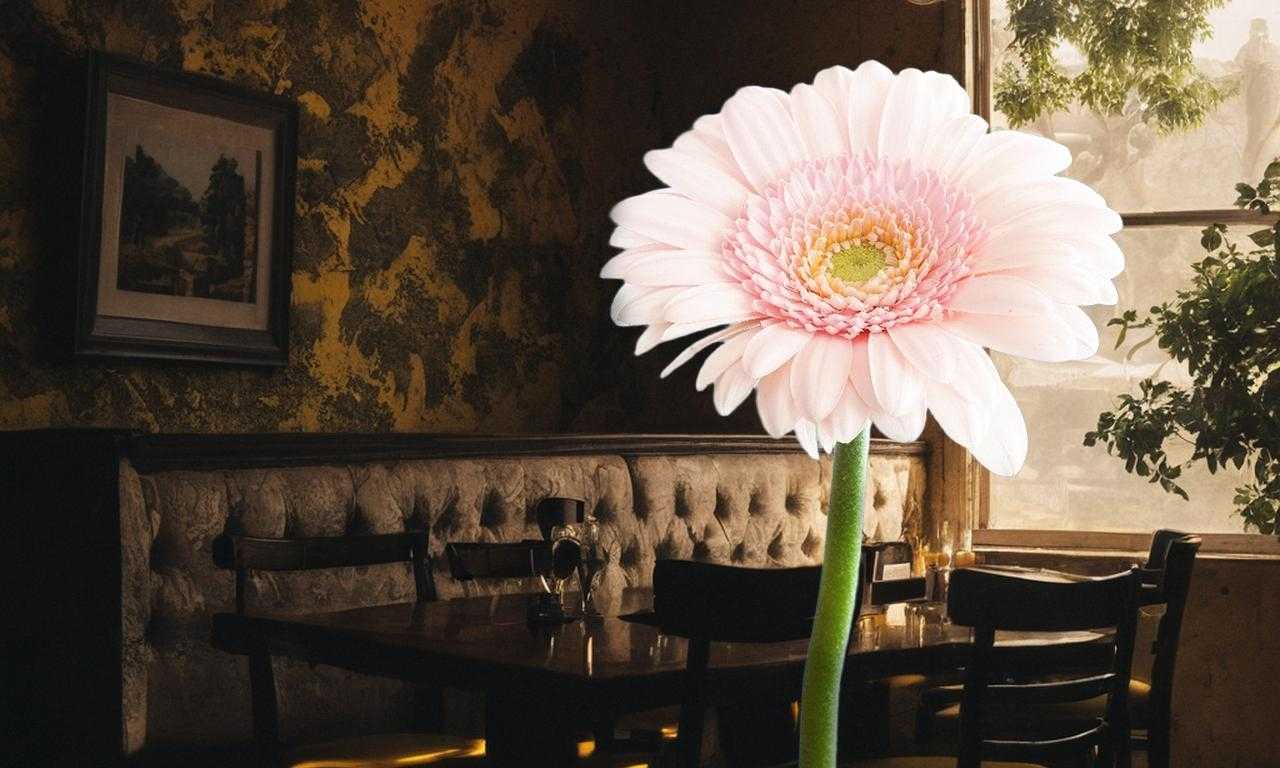}}
    \subfloat{\includegraphics[width=0.124\linewidth]{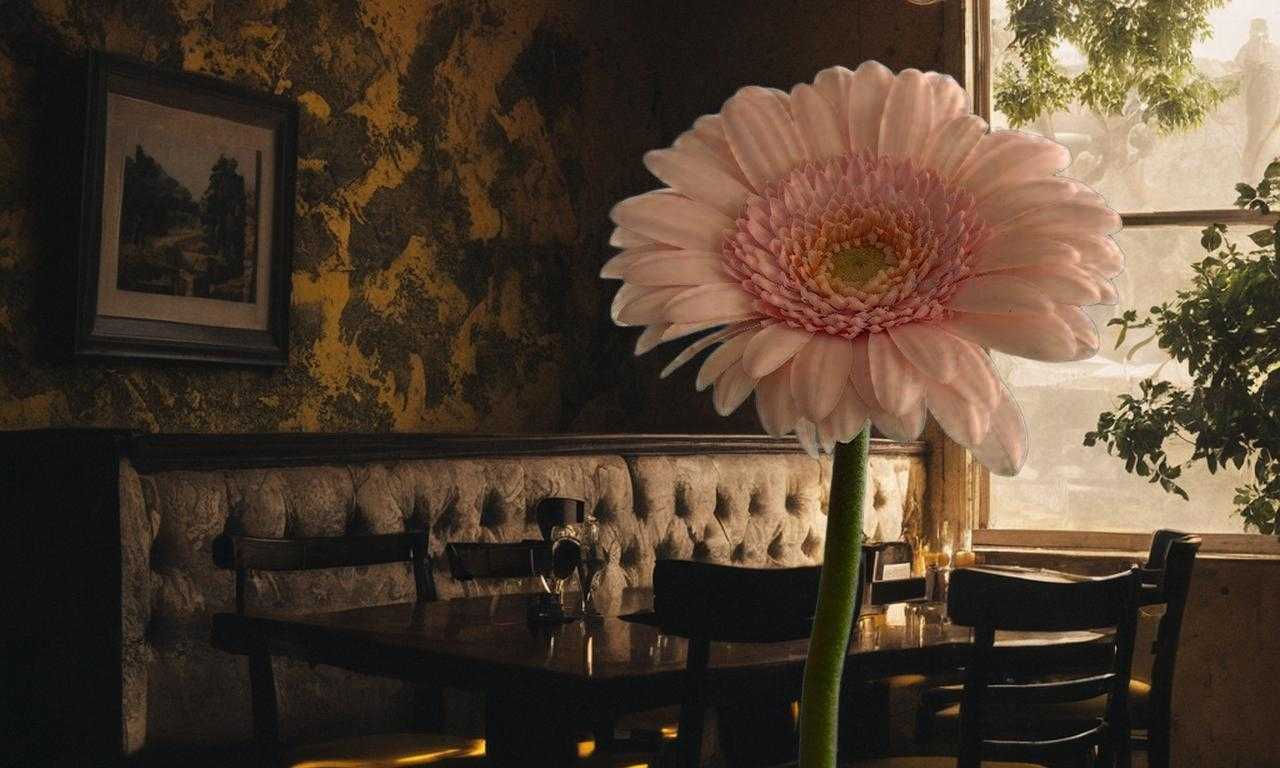}}
    \subfloat{\includegraphics[width=0.124\linewidth]{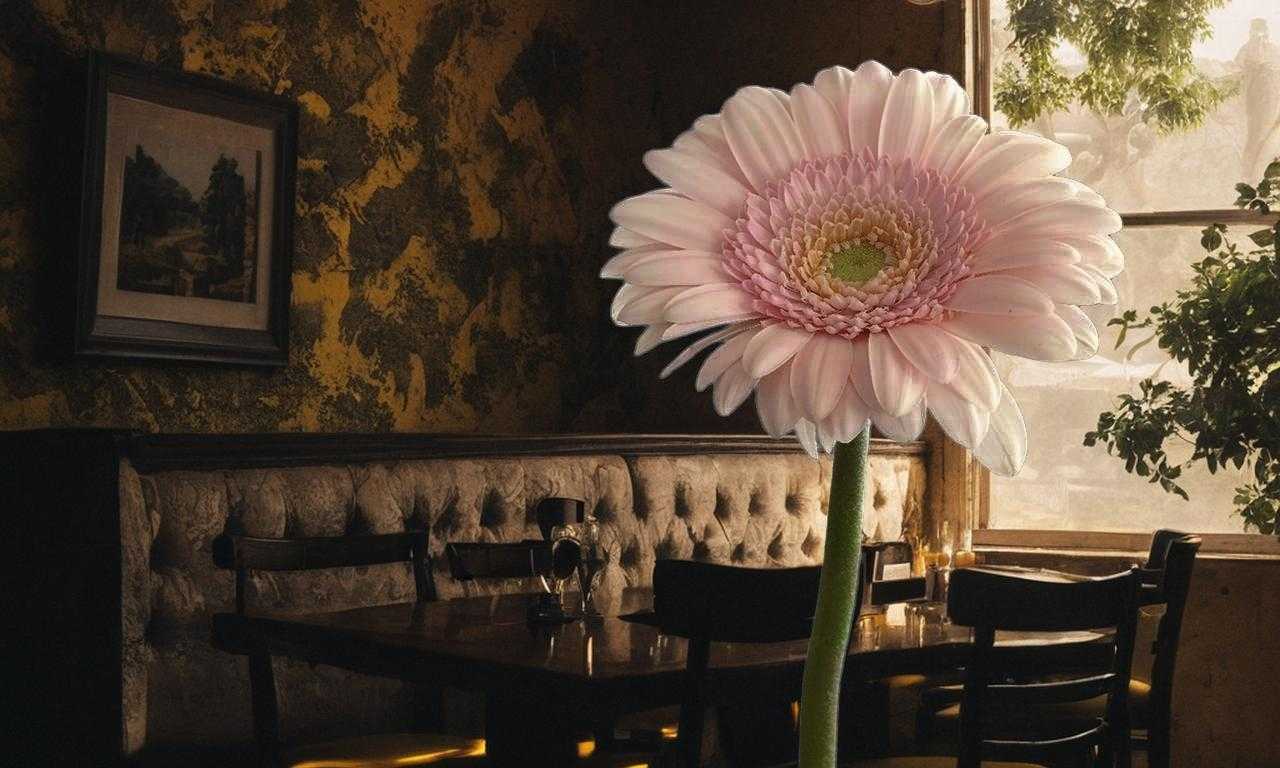}}
    \subfloat{\includegraphics[width=0.124\linewidth]{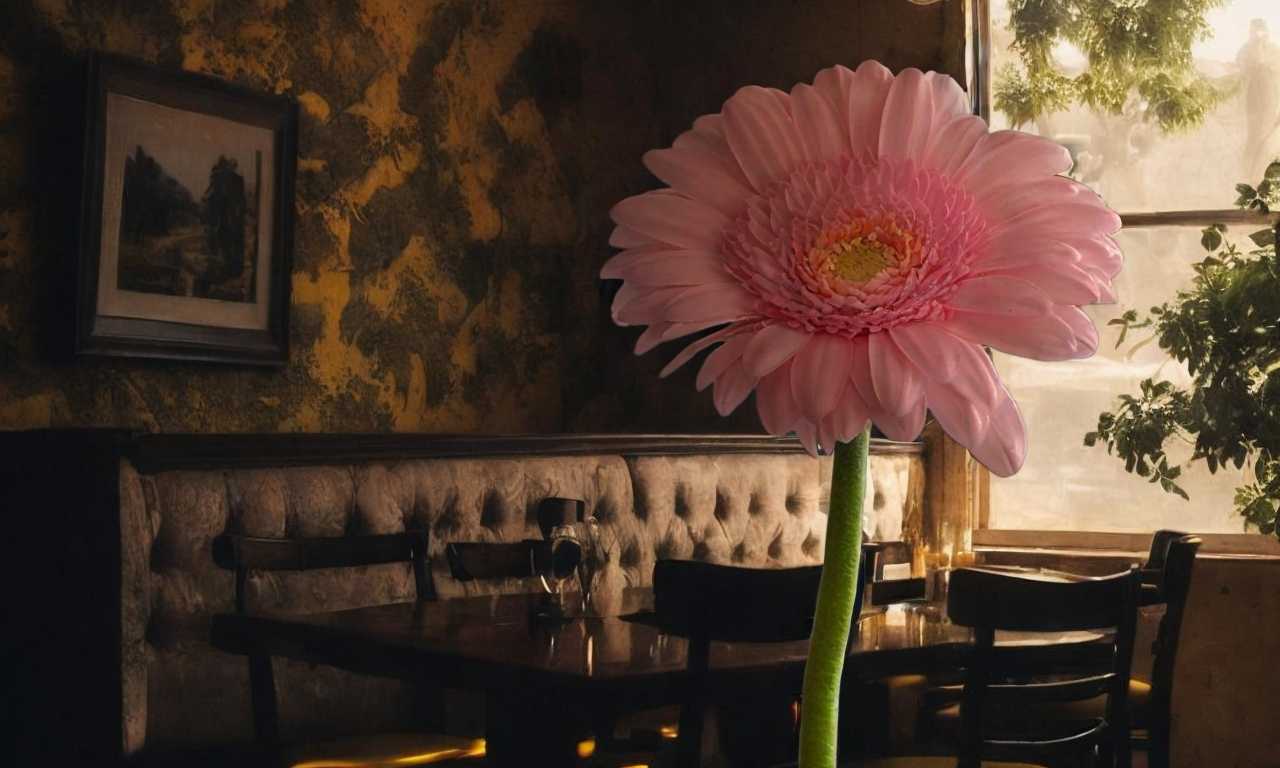}}
    \subfloat{\includegraphics[width=0.124\linewidth]{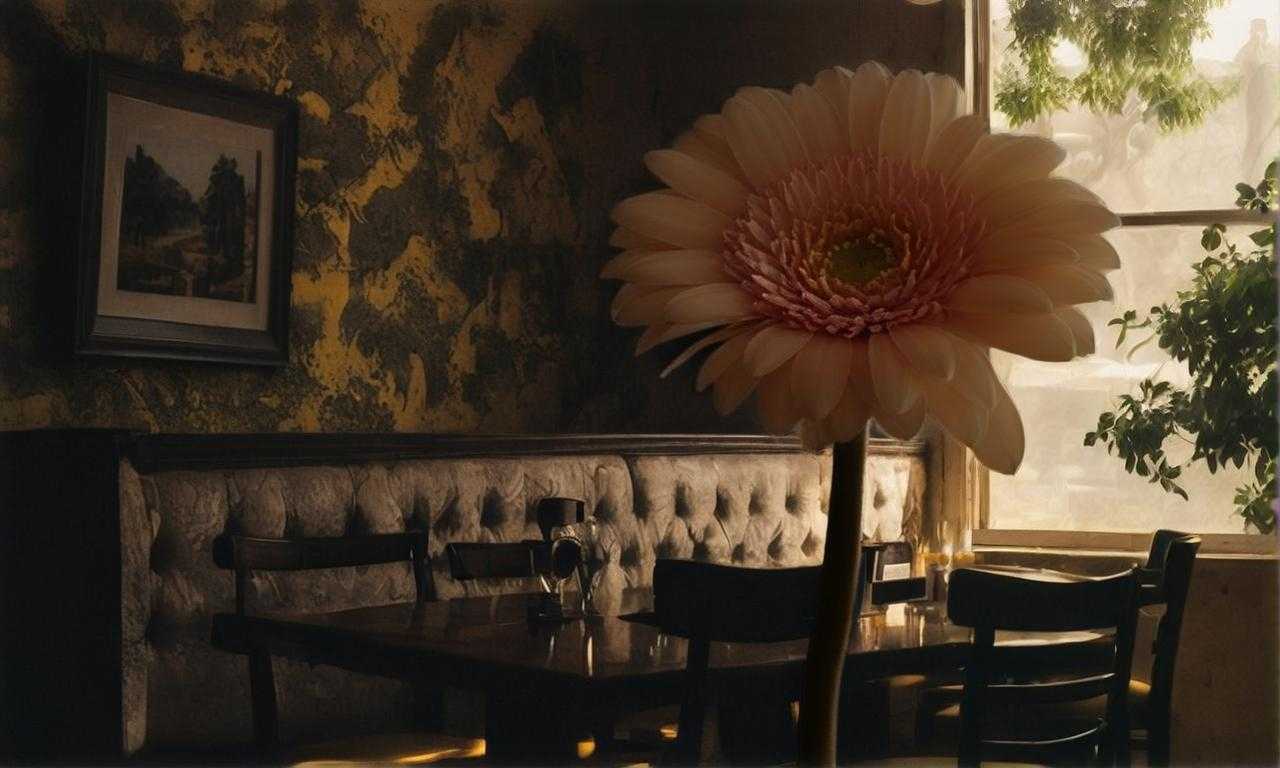}}\\
    \vspace{-0.10em}
    % \subfloat{\includegraphics[width=0.14\linewidth]{plots/relighting/composite/68.jpg}}
    % \subfloat{\includegraphics[width=0.14\linewidth]{plots/relighting/harmonizer/68.jpg}}
    % \subfloat{\includegraphics[width=0.14\linewidth]{plots/relighting/pih/68.jpg}}
    % \subfloat{\includegraphics[width=0.14\linewidth]{plots/relighting/pct/68.jpg}}
    % \subfloat{\includegraphics[width=0.14\linewidth]{plots/relighting/inr/68.jpg}}
    % \subfloat{\includegraphics[width=0.14\linewidth]{plots/relighting/ic_light/68.jpg}}
    % \subfloat{\includegraphics[width=0.14\linewidth]{plots/relighting/rf/68.jpg}}\\
    % \vspace{-0.10em}
    % \subfloat{\includegraphics[width=0.14\linewidth]{plots/relighting/composite/77.jpg}}
    % \subfloat{\includegraphics[width=0.14\linewidth]{plots/relighting/harmonizer/77.jpg}}
    % \subfloat{\includegraphics[width=0.14\linewidth]{plots/relighting/pih/77.jpg}}
    % \subfloat{\includegraphics[width=0.14\linewidth]{plots/relighting/pct/77.jpg}}
    % \subfloat{\includegraphics[width=0.14\linewidth]{plots/relighting/inr/77.jpg}}
    % \subfloat{\includegraphics[width=0.14\linewidth]{plots/relighting/ic_light/77.jpg}}
    % \subfloat{\includegraphics[width=0.14\linewidth]{plots/relighting/rf/77.jpg}}\\
    % \vspace{-0.10em}
    \subfloat{\includegraphics[width=0.124\linewidth]{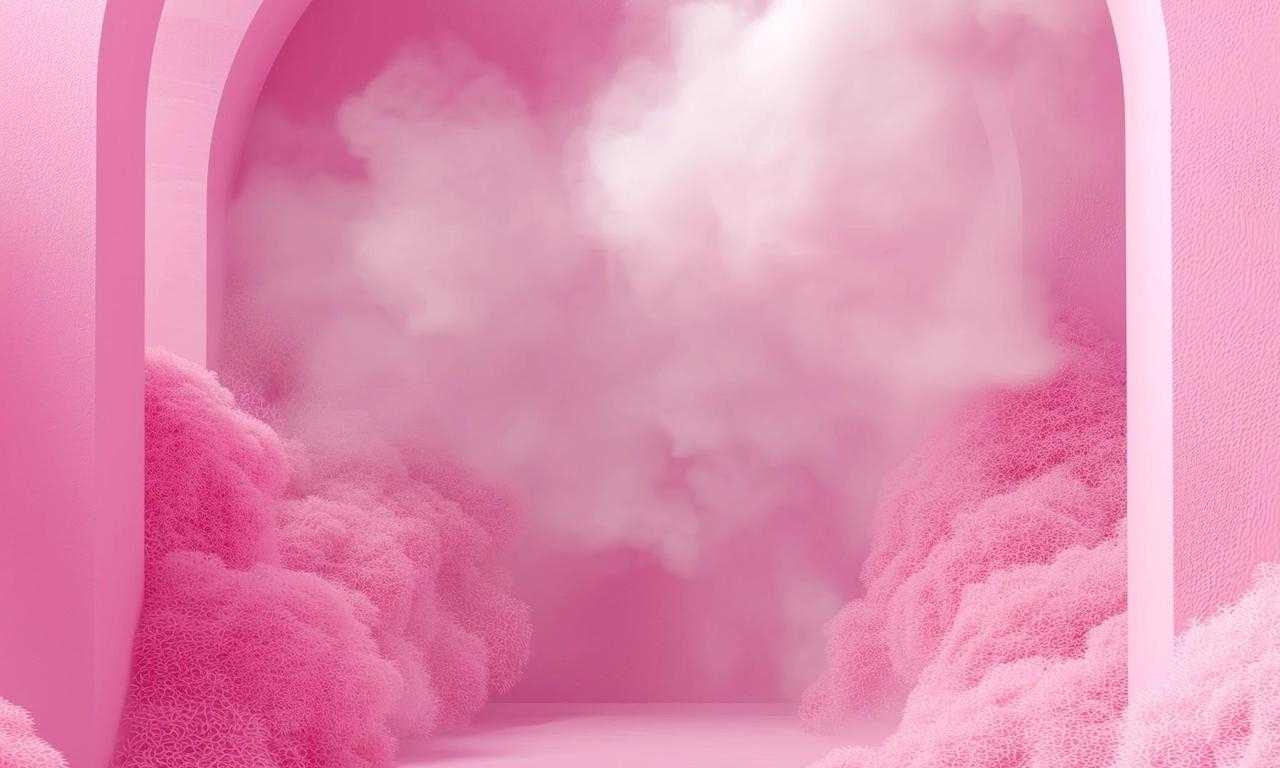}}
    \subfloat{\includegraphics[width=0.124\linewidth]{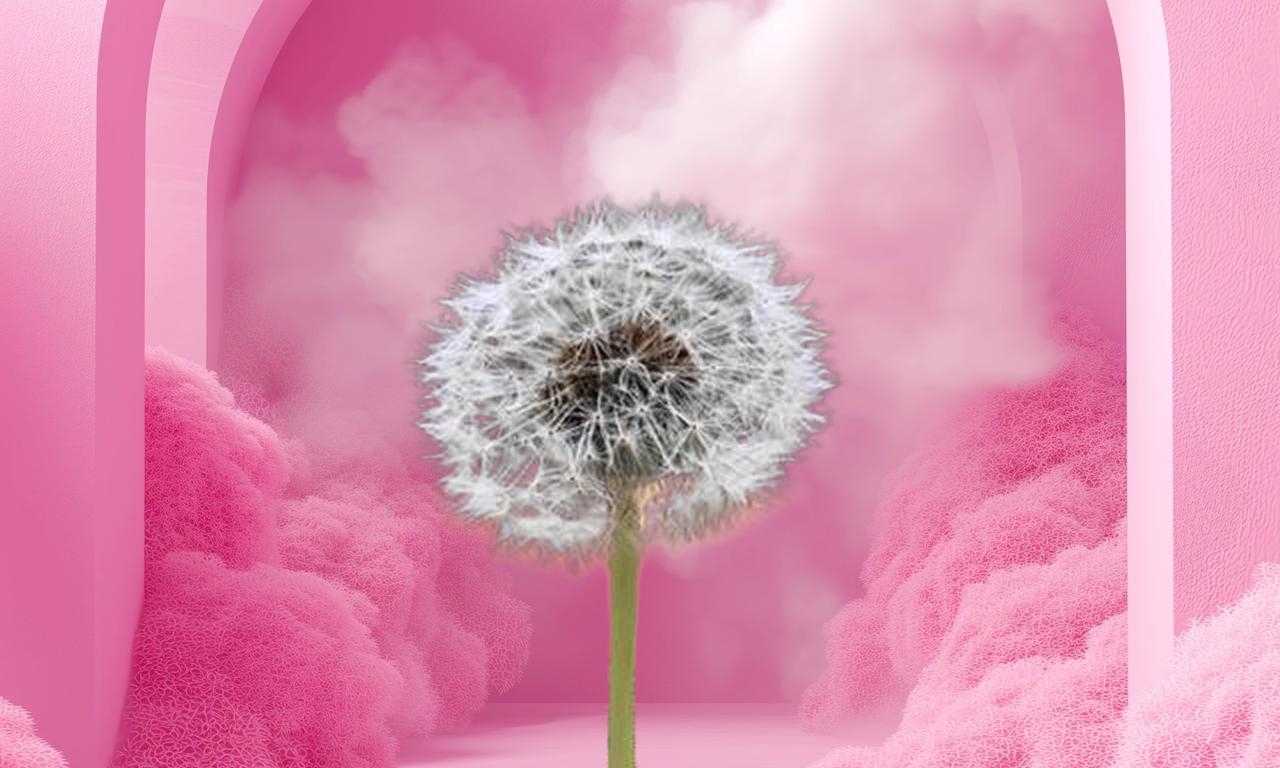}}
    \subfloat{\includegraphics[width=0.124\linewidth]{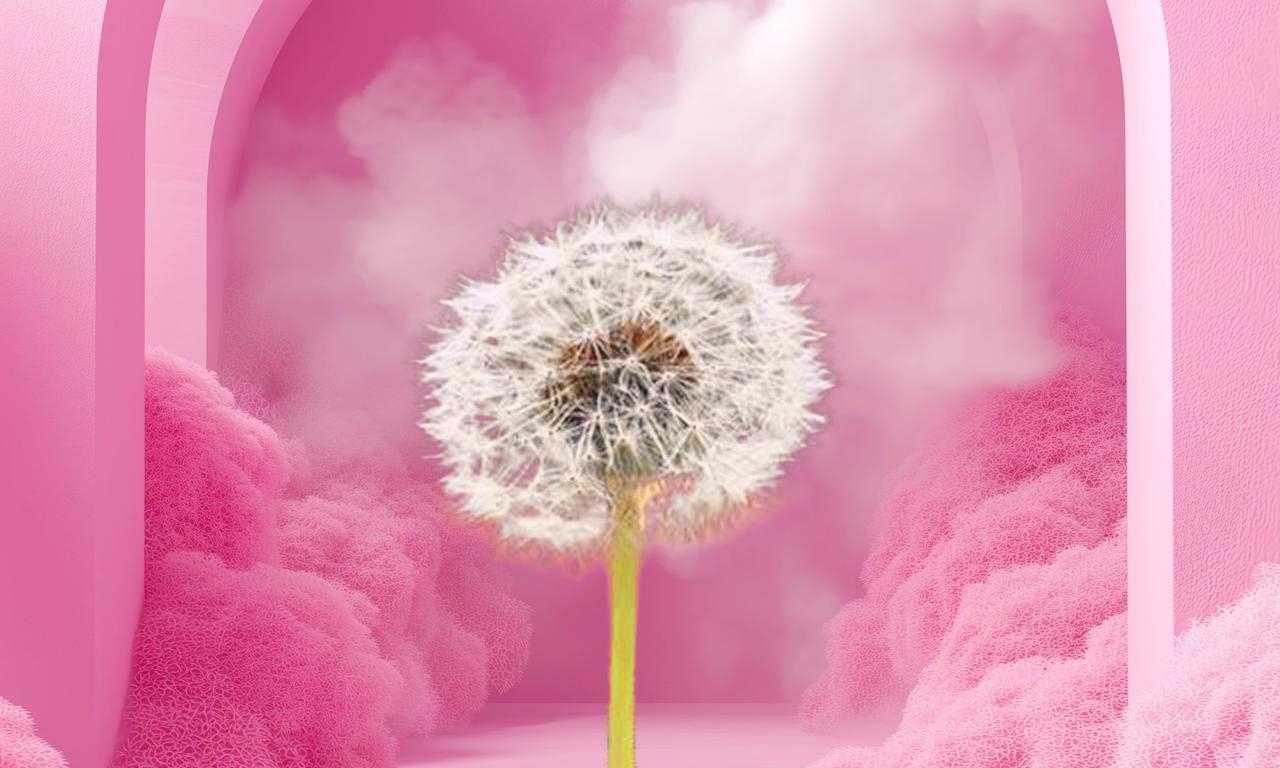}}
    \subfloat{\includegraphics[width=0.124\linewidth]{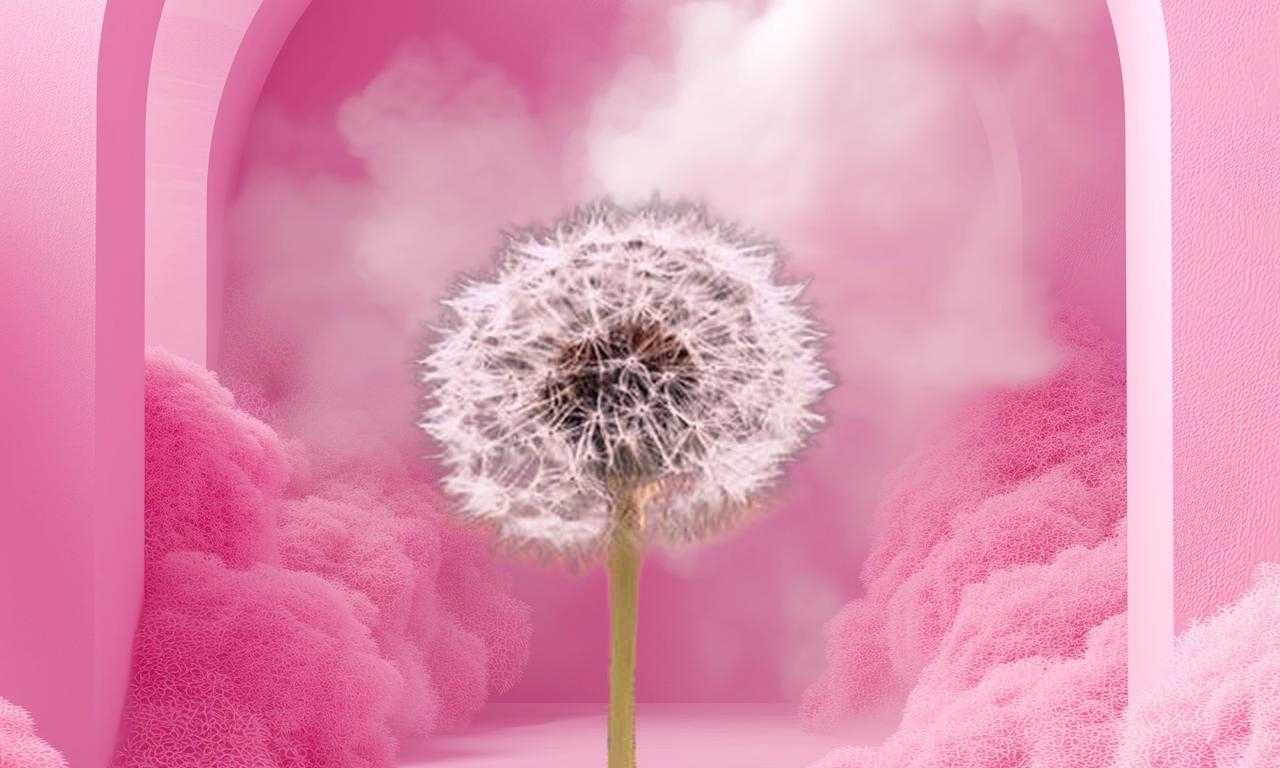}}
    \subfloat{\includegraphics[width=0.124\linewidth]{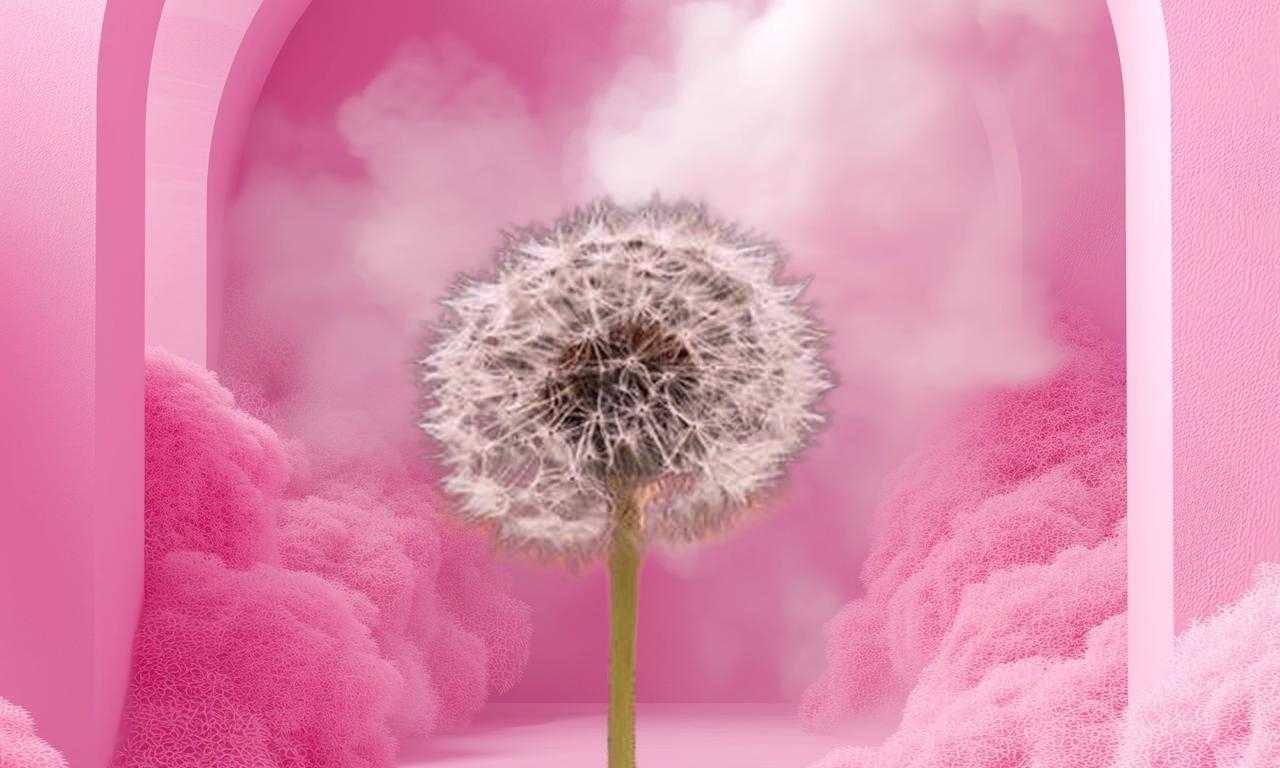}}
    \subfloat{\includegraphics[width=0.124\linewidth]{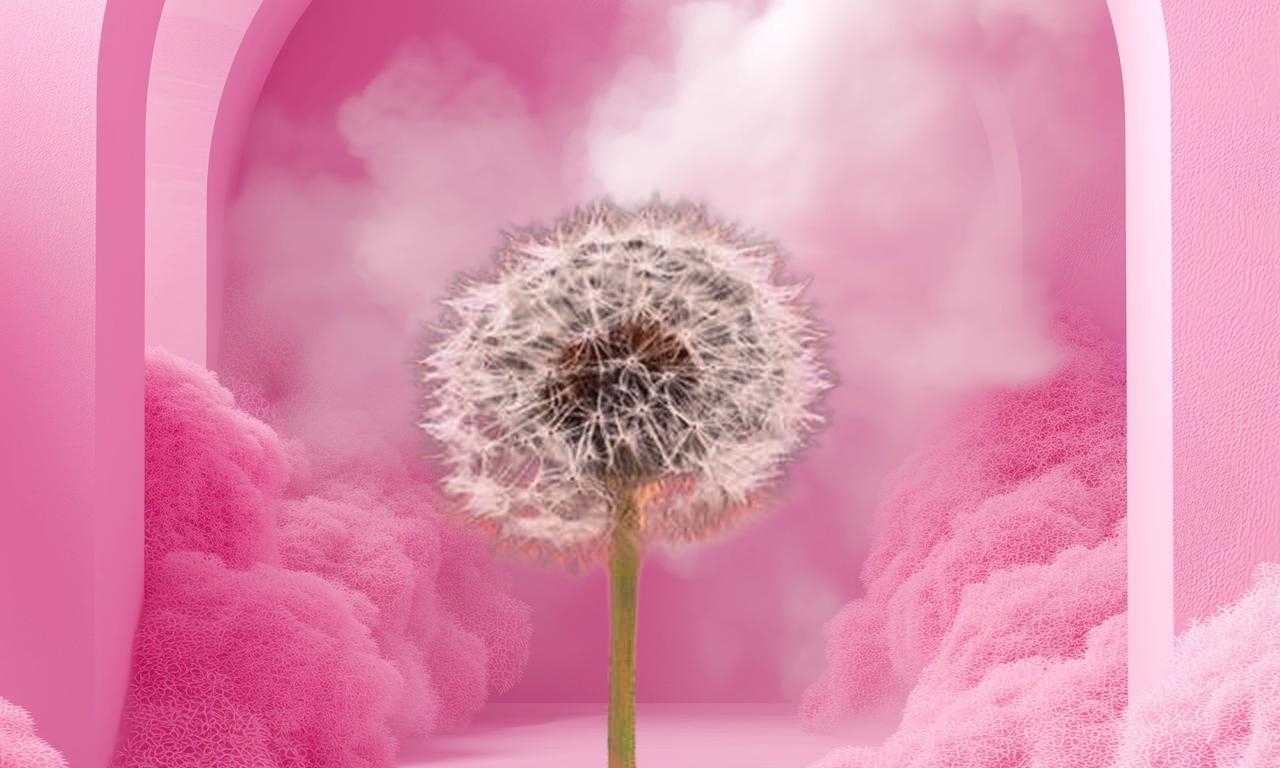}}
    \subfloat{\includegraphics[width=0.124\linewidth]{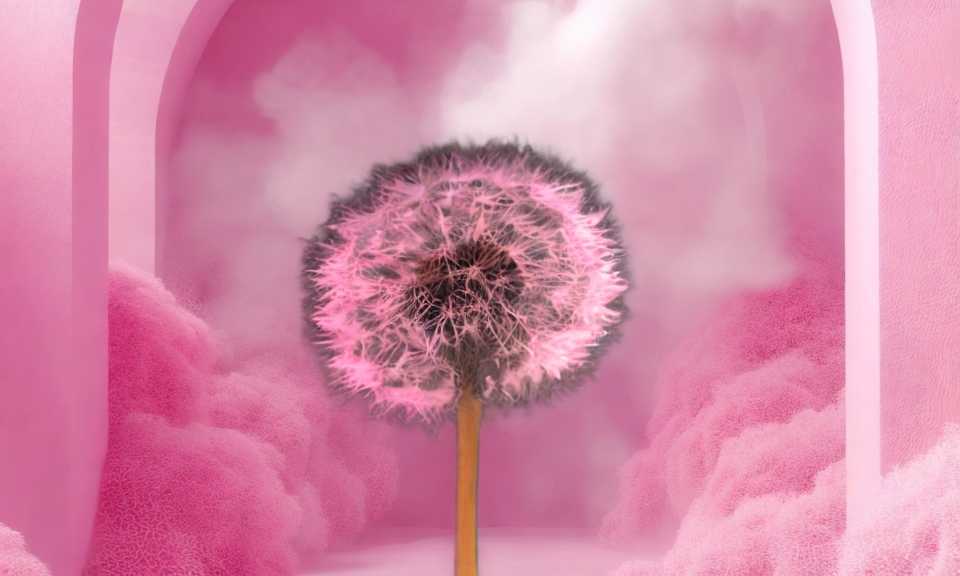}}
    \subfloat{\includegraphics[width=0.124\linewidth]{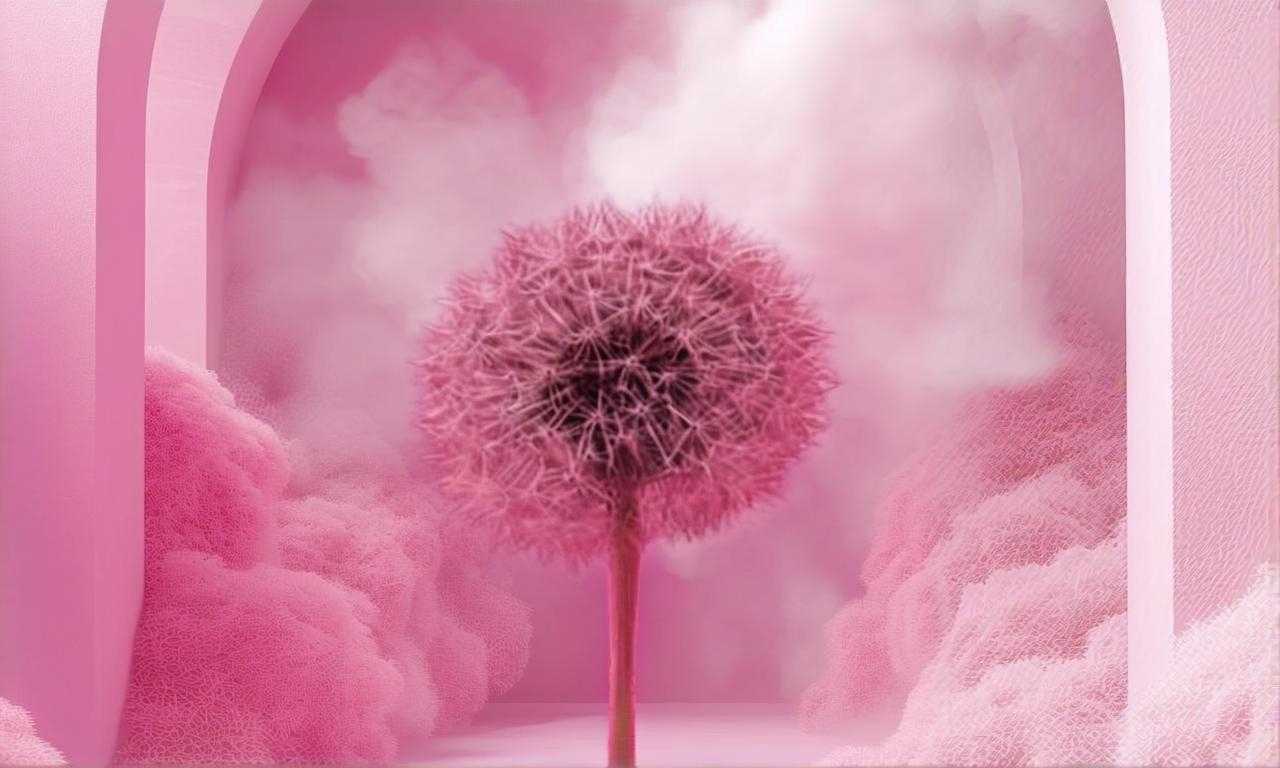}}\\
    % \vspace{-0.10em}
    % \subfloat{\includegraphics[width=0.14\linewidth]{plots/relighting/composite/53.jpg}}
    % \subfloat{\includegraphics[width=0.14\linewidth]{plots/relighting/harmonizer/53.jpg}}
    % \subfloat{\includegraphics[width=0.14\linewidth]{plots/relighting/pih/53.jpg}}
    % \subfloat{\includegraphics[width=0.14\linewidth]{plots/relighting/pct/53.jpg}}
    % \subfloat{\includegraphics[width=0.14\linewidth]{plots/relighting/inr/53.jpg}}
    % \subfloat{\includegraphics[width=0.14\linewidth]{plots/relighting/ic_light/53.jpg}}
    % \subfloat{\includegraphics[width=0.14\linewidth]{plots/relighting/rf/53.jpg}}
    \caption{Qualitative results for object relighting. The model is able to relight the object according to the provided background and also remove existing shadows and reflections. See the appendices for more results.}
    \label{fig:relighting_results}
\end{figure*}

\paragraph{Results} We follow the same approach as described above to create a test set composed of approximately 10k unseen real images and evaluate the performance of the proposed method. For this benchmark, we consider INR \cite{chen2023dense}, Harmonizer \cite{ke2022harmonizer}, PCT-Net \cite{guerreiro2023pct}, PIH \cite{wang2023semi} and IC-light \cite{zhang2025scaling}. We report in \cref{tab:relighting_results} the FID, Local FID, fMSE and PSNR metrics computed on the test set. As highlighted in the table, the method outperforms other competitors for most metrics. Note that for IC-light since it needs a background image as conditioning while other methods take as input directly the composite image, we used the estimated background using our object removal model proposed in \cref{sec:object-removal} which can lead to artifacts in the background. In addition, we also provide qualitative samples for each method in \cref{fig:relighting_results}. The proposed approach appears able to add strong illumination changes to the foreground object while preserving the background. Moreover, it is able to remove existing shadows and reflections so the foreground object appears more realistic. Finally, we also noted that while IC-light model seems to degrade the quality of the foreground object (see the last row of \cref{fig:relighting_results}), the proposed approach allows to keep the foreground object consistent with the input image.

\begin{table}[ht]
    \centering
    \scriptsize
    \begin{tabular}{ccccc}
      \toprule
      Model                   & FID $\downarrow$  & Local FID $\downarrow$ & fMSE $\downarrow$  & PSNR $\uparrow$ \\
      \midrule
      Harmonizer              & 13.91 & \underline{14.21} & 1533.34  & \textbf{23.49} \\
      PCT-NET                 & 13.96 & 14.53 & 1634.24  & \underline{23.25} 
     \\
      PIH                     & 15.17 & 15.45 & 1755.86  & 22.83 \\
      INR Harmonization       & \underline{13.86} & 14.65 & \underline{1480.01}  & \textbf{23.49} \\
      IC-Light$^{*}$           & 20.88 & 22.11 & 1897.19  & 22.39\\
      \midrule
      Ours                    & \textbf{12.79} & \textbf{12.83} & \textbf{1173.02} & 23.24 \\
      \bottomrule
      \multicolumn{5}{l}{\scriptsize$^{*}$ IC-Light uses backgrounds computed using our object removal model} \\
    \end{tabular}
    \caption{Metrics for image relighting task. Our method uses a single NFE. Best results are in bold, second best are underlined.}
    \label{tab:relighting_results}
  \end{table}
  
\begin{figure}[t]
    \centering
    \includegraphics[width=\linewidth]{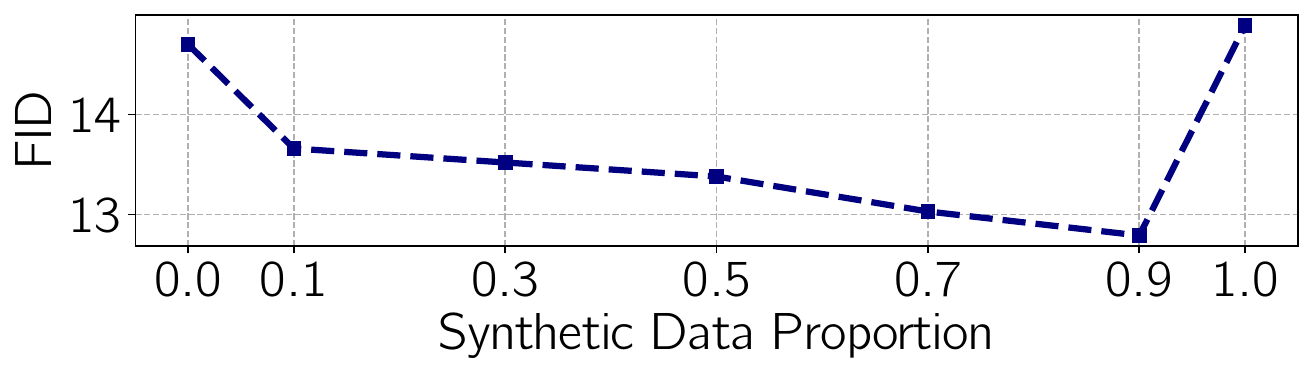}
    \caption{Influence of the synthetic data}
    \label{fig:synthetic_data}
\end{figure}

\paragraph{Influence of synthetic data}
We ablate the influence of the additional synthetic data (created with the rendering engine) on the overall model performance. We noticed that the proportion of synthetic data strongly influences the model performance. In \cref{fig:synthetic_data}, we plot the evolution of the FID score with respect to the proportion of synthetic data in the training set. Interestingly, the more synthetic data, the better the model performance since it clearly helps the model to learn the lighting conditions on simpler and controlled scenes. However, we observe that adding too much synthetic data may eventually lead to a performance drop since the outputs realism is affected.

\subsection{Controllable image relighting and shadow generation}

\begin{figure}[t]
    \centering
    \includegraphics[width=\linewidth]{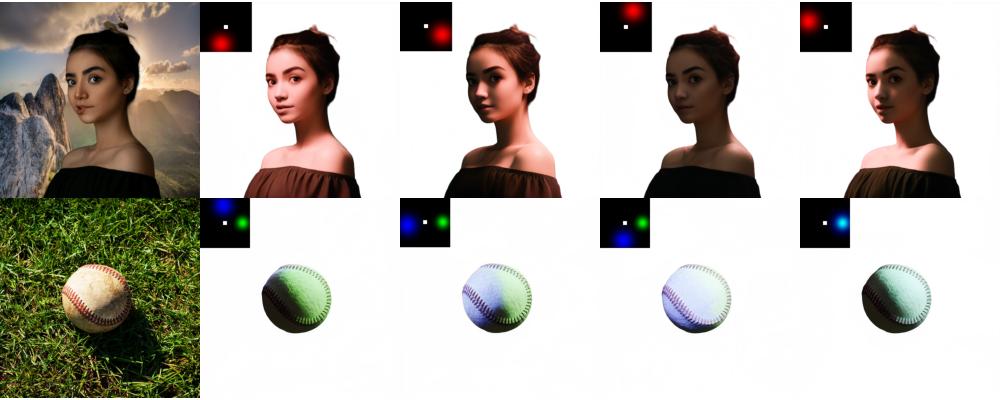}\\
    \vspace{-0.1em}
    \includegraphics[width=\linewidth]{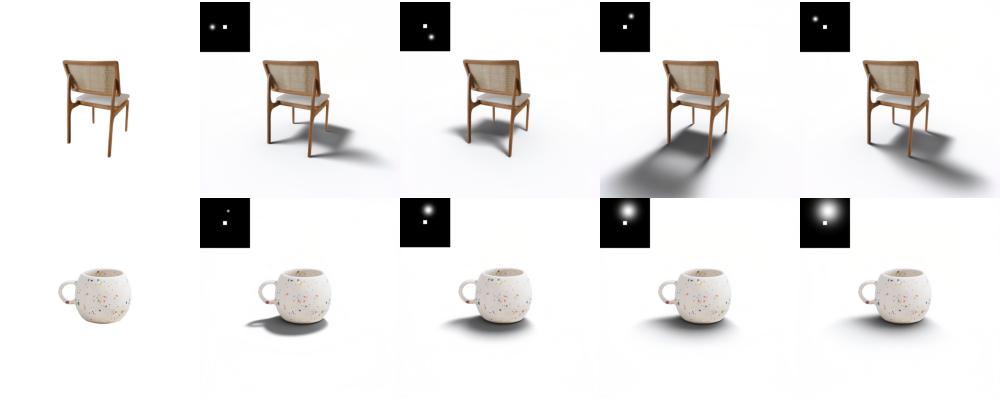}
    \caption{\emph{Left}: Controllable image relighting. \emph{Right}: Controllable shadow generation. For both tasks, the foreground object is extracted from the input image using a matting model \cite{zheng2024birefnet} and pasted on a white background. The control light map is represented at the top left of each image where the white square shows the position of the object. The model is able to relight the object and generate realistic shadows according to the light conditions even when using multiple light sources. Moreover, it can effectively remove existing shadows and reflections present on the original foreground object.}
    \label{fig:control_relighting_results}
\end{figure}

\paragraph{Setting} Finally, we show the effectiveness of our proposed Conditional Latent Bridge Matching model for two tasks. The first one consists of controllable image relighting where the model is additionally conditioned on a light map representing the position, color and intensity of the light sources and must relight the foreground object according to these sources. The second one consists of controllable shadow generation where the model is conditioned on a light map representing the position and sharpness of the light source and must generate a shadow of the foreground object on the ground. For shadow generation, we build a 2D light map inspired by \cite{ssn} and \cite{tasar2024controllable} that represents the light information as a gray-scaled image in which each light source is represented as a mixture of Gaussians the amplitude of which encodes the intensity while the variance represents the softness. For image relighting, we adapt this representation by considering RGB light maps incorporating the color of the light sources.

\paragraph{Dataset creation}
Creating a dataset consisting of real images for controllable relighting or shadow generation typically demands a costly setup, such as a light stage~\cite{pandey2021total}. Therefore, we propose to rely on a synthetic dataset. For controllable relighting, we generate a dataset by placing a randomly selected 3D model at the center of Blender’s coordinate system during each rendering iteration. To illuminate the scene, we position one to three area lights with varying colors and light intensities on the surface of the upper hemisphere of a sphere with a fixed radius. We then render the scene, capturing the desired lighting variations. For shadow generation, we position a sufficiently large plane beneath the 3D model to serve as the shadow receiver. We position a single area light at a random location, emitting white light with a variable area light size. The size determines the sharpness of the shadows.

\paragraph{Results}
In \cref{fig:control_relighting_results}, we present the generated outputs for both tasks under various lighting conditions, including variations in position, intensity, color, and number of light sources. As highlighted in the figure, the model is able to relight the object accordingly even in the context of multiple light sources. Moreover, it respects the position, the intensity and the color of the sources. An interresting property of the approach is also that the model is able to remove existing shadows and reflections present on the original object and add new ones improving realism.

\subsection{Ablation study}\label{sec:ablations}
In this section, we ablate the different components of our method. To do so, we consider the object-removal task and train all the models for 20k iterations on 2 H100 GPUs unless stated otherwise. The ablated parameters are: the timestep distribution $\pi(t)$, the pixel loss weight $\lambda$ in Eq.~\eqref{eq:full_loss}, the magnitude of the noise parameter $\sigma$ in Eq.~\eqref{eq:latent_bridge} and the number of inference steps (NFE).

\begin{figure}[ht]
    \centering
    \subfloat[LBM vs FM]{\includegraphics[width=0.503\linewidth]{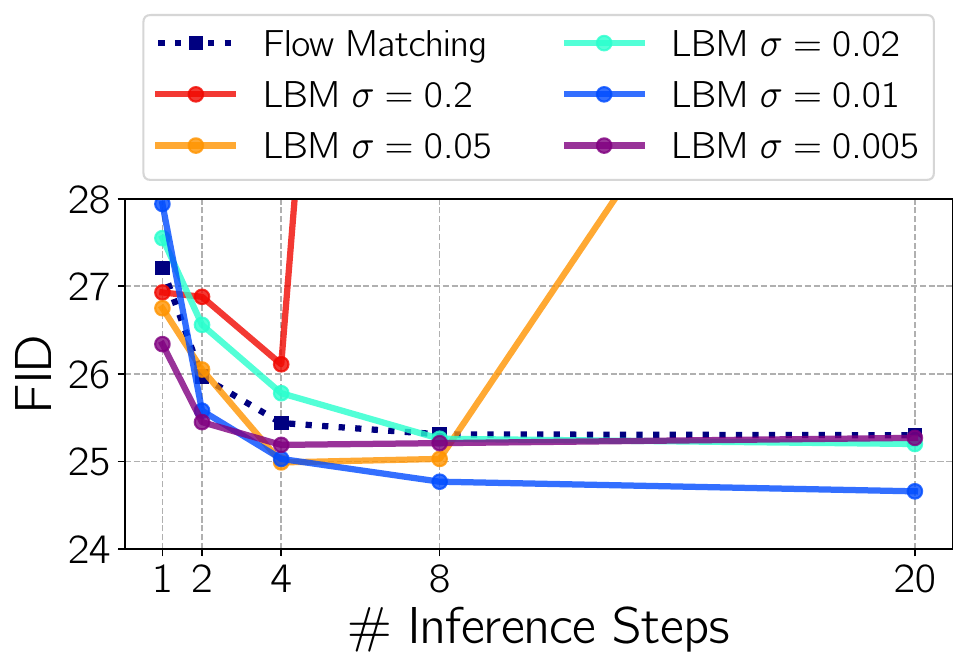}}
    \subfloat[Influence of $\pi(t)$]{\includegraphics[width=0.487\linewidth]{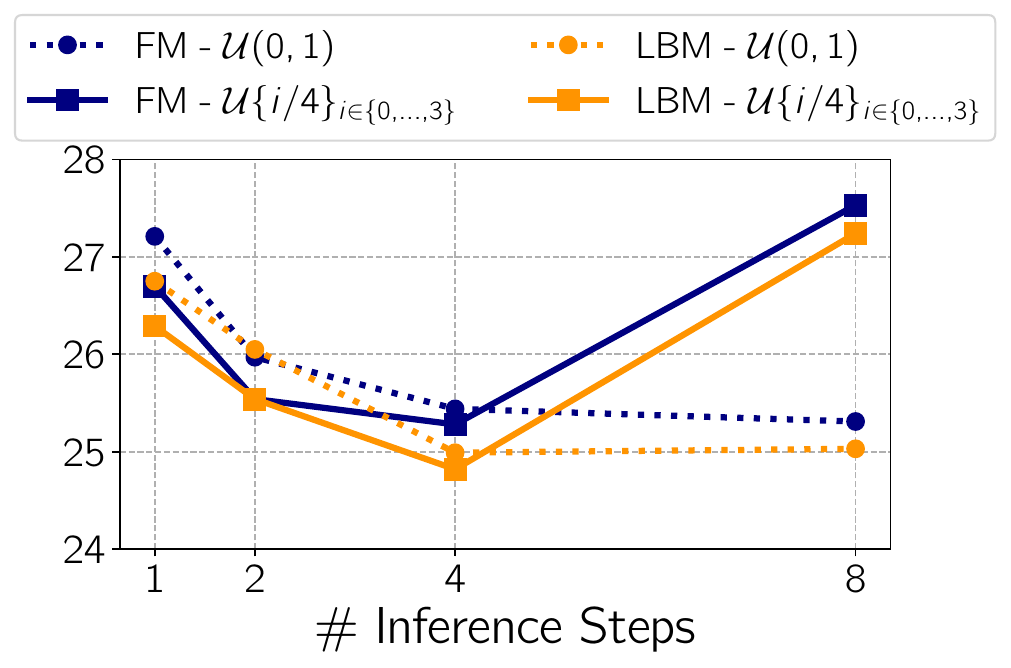}}
    \caption{Influence of the timestep distribution and $\sigma$ as well as the number of inference steps. Notably, the proposed LBM model outperforms flow matching for small enough values of $\sigma$ using either a uniform or discrete timestep distribution.}
    \label{fig:sigma_time}
\end{figure}

\paragraph{Influence of $\sigma$}
First, we ablate the noise parameter $\sigma$ in Eq.~\eqref{eq:latent_bridge}. We consider the same configuration as in \cref{sec:object-removal} and report the FID computed on the RORD validation set for $\sigma$ ranging from 0 to 0.2 where $\sigma=0$ corresponds to flow matching. For the timestep distribution, we consider a uniform distribution for this ablation. In \cref{fig:sigma_time} (left), we plot the evolution of the FID according to the number of inference steps for each considered configuration. The first observation of such a study is that as expected, the method's performance improves as the number of inference steps increases for \emph{small} enough values of $\sigma$. However, if $\sigma$ is too large, the performance drops since too much noise is added when solving Eq.~\eqref{eq:bridge_sde} potentially removing too much information from the source image. Notably, the method \textbf{outperforms flow matching} in terms of FID for small $\sigma$. This can be explained by the fact that adding this noise parameter allows the model to reach a wider diversity of samples. Given a source image, a LBM will indeed solve an SDE and so generate a different sample each time thanks to its intrinsic stochastic nature. On the contrary, in flow matching we solve an ODE the solution of which is unique. This emphasizes the importance of $\sigma$ in the proposed method and provides a hint why bridge models may be better suited than flow models for generative tasks.

\paragraph{Influence of the timestep distribution $\pi(t)$}
We train a model using either a uniform distribution $\pi(t) = \mathcal{U}(0,1)$ or a distribution focusing on 4 discrete timesteps as we propose in \cref{sec:timestep_sampling}. We set $\sigma=0.05$ for the two settings and report the FID computed on the RORD validation set in \cref{fig:sigma_time} (right). Notably, the method outperforms again flow matching 
for $\pi(t)$ set to either a discrete distribution or a uniform distribution. Interestingly, we observe that in most cases, using a sharp distribution that focuses on 4 distinct timesteps leads to better results than using a uniform distribution when inferring with those specific timesteps. This is for instance visible by comparing the FID achieved with flow matching using either a uniform or discrete timestep distribution (blue curves). The same behaviour is observed with LBM as well. However, using a discrete distribution limits the number of inference steps to the number of selected timesteps since we observe a strong performance drop when inferring with more timesteps (solid lines) while the performance with a uniform distribution still improves (dotted lines). In a nutshell, these discrete distributions are valuable because they concentrate the model's knowledge on specific timesteps, thereby improving performance when these timesteps are used during inference. However, this comes at the cost of limiting the number of possible inference steps.

% \begin{table}[t]
%     \centering
%     \scriptsize
%     \begin{tabular}{c|c|c|c|c|c}
%         \toprule
%         Backbone & LPIPS $\downarrow$ & FID $\downarrow$ & Local FID $\downarrow$ & fMSE $\downarrow$ & PSNR $\uparrow$ \\
%         SDXL \\
%         SD15 \\
%         SD2 \\
%         \midrule
%         \bottomrule
%     \end{tabular}
%     \caption{Influence of the Backbone architecture}
%     \label{tab:backbone}\end{table}

\begin{figure}[t]
    \centering
    \captionsetup[subfigure]{position=above, labelformat = empty}
    \subfloat[Input]{\includegraphics[width=0.25\linewidth]{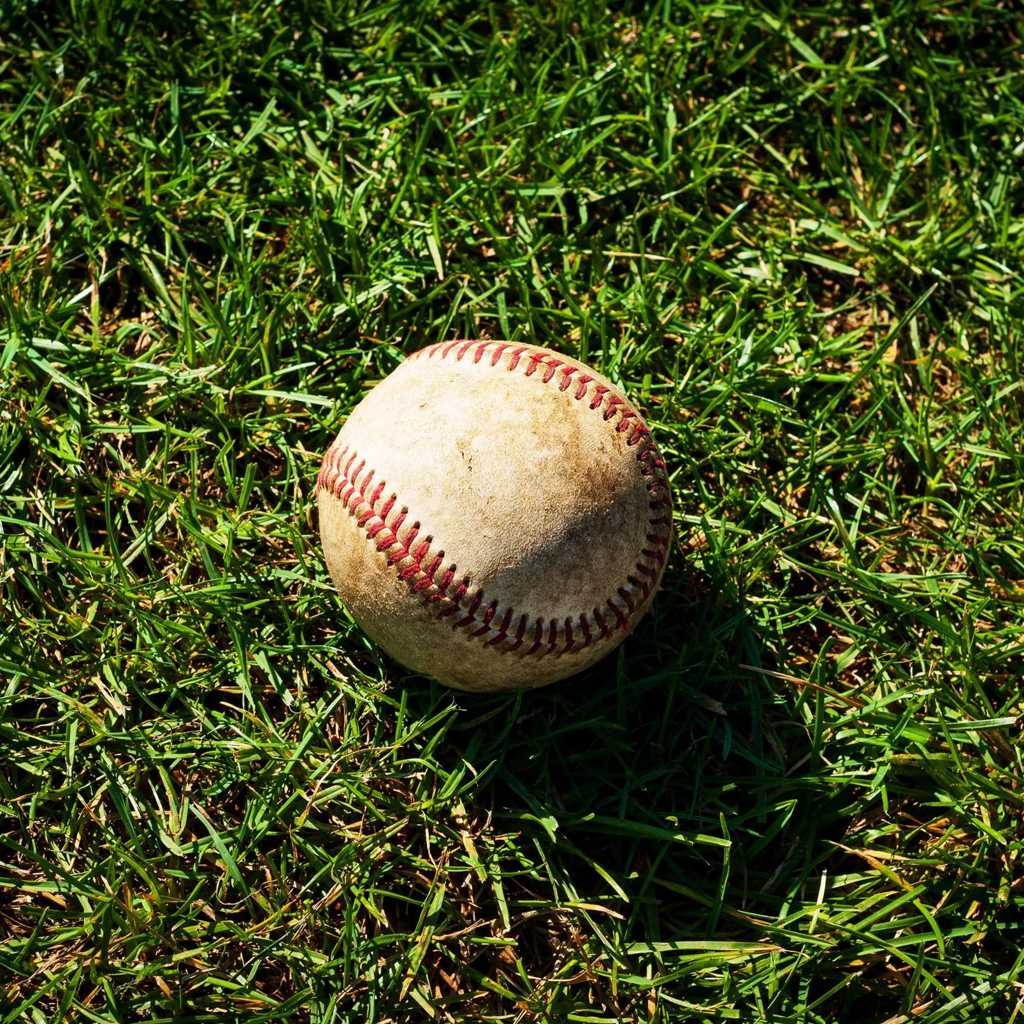}}
    %\subfloat[1 NFE]{\includegraphics[width=0.2\linewidth]{plots/surface/surface_1.jpg}}
    \subfloat[1 NFE]{\includegraphics[width=0.25\linewidth]{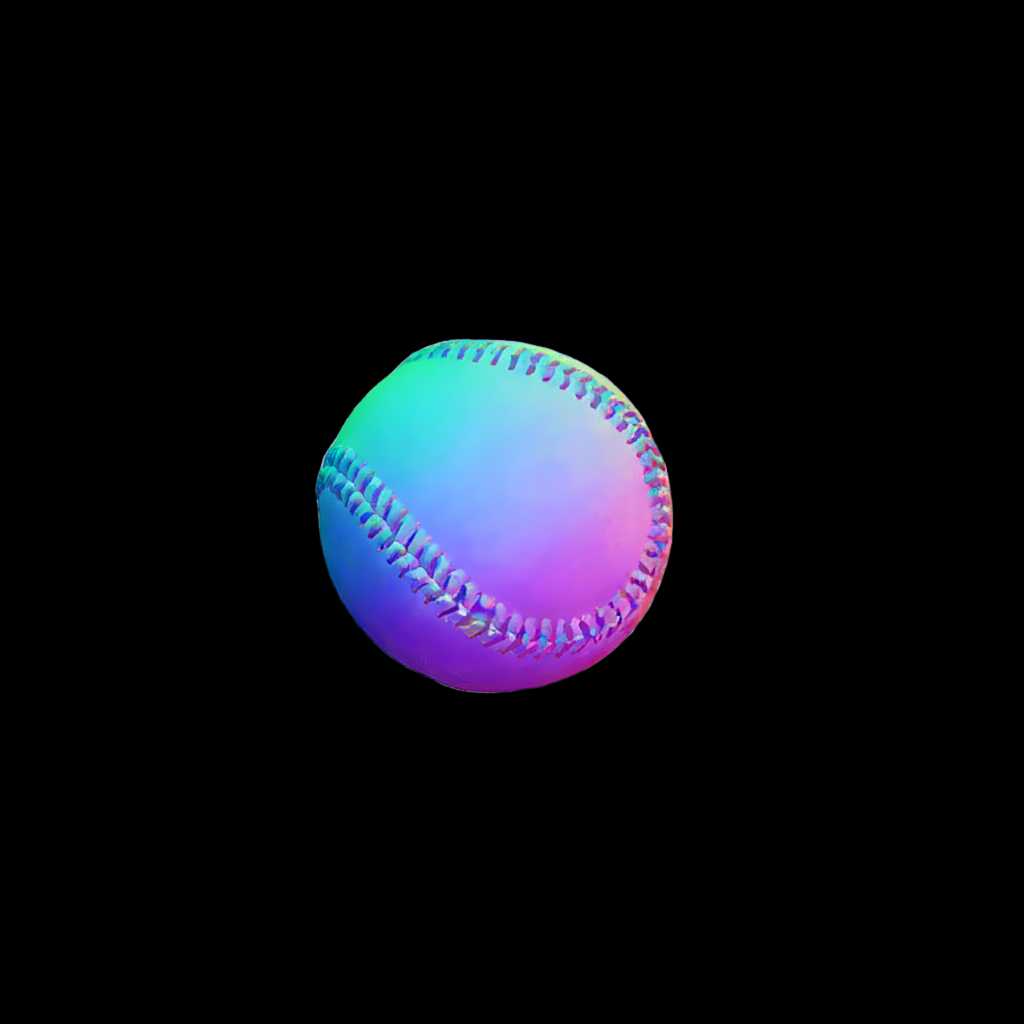}}
    \subfloat[2 NFE]{\includegraphics[width=0.25\linewidth]{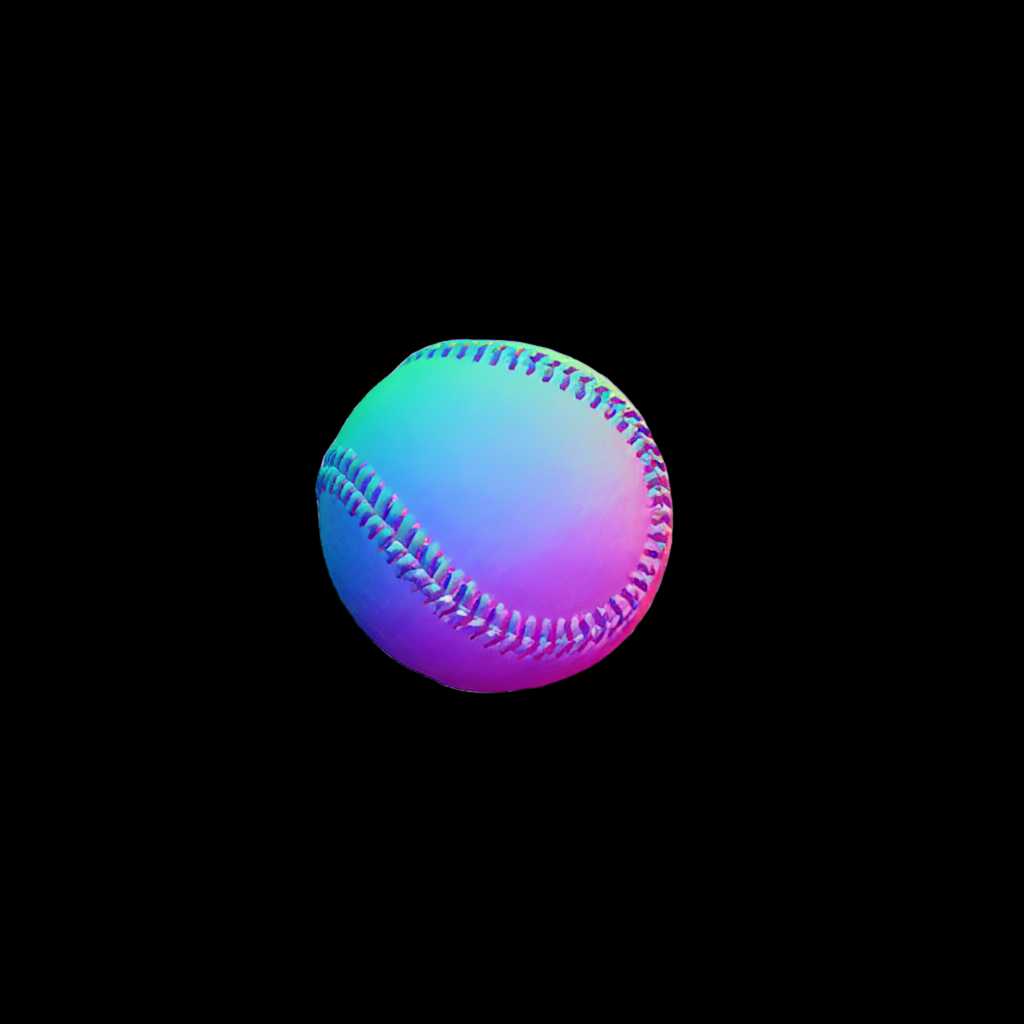}}
    \subfloat[4 NFE]{\includegraphics[width=0.25\linewidth]{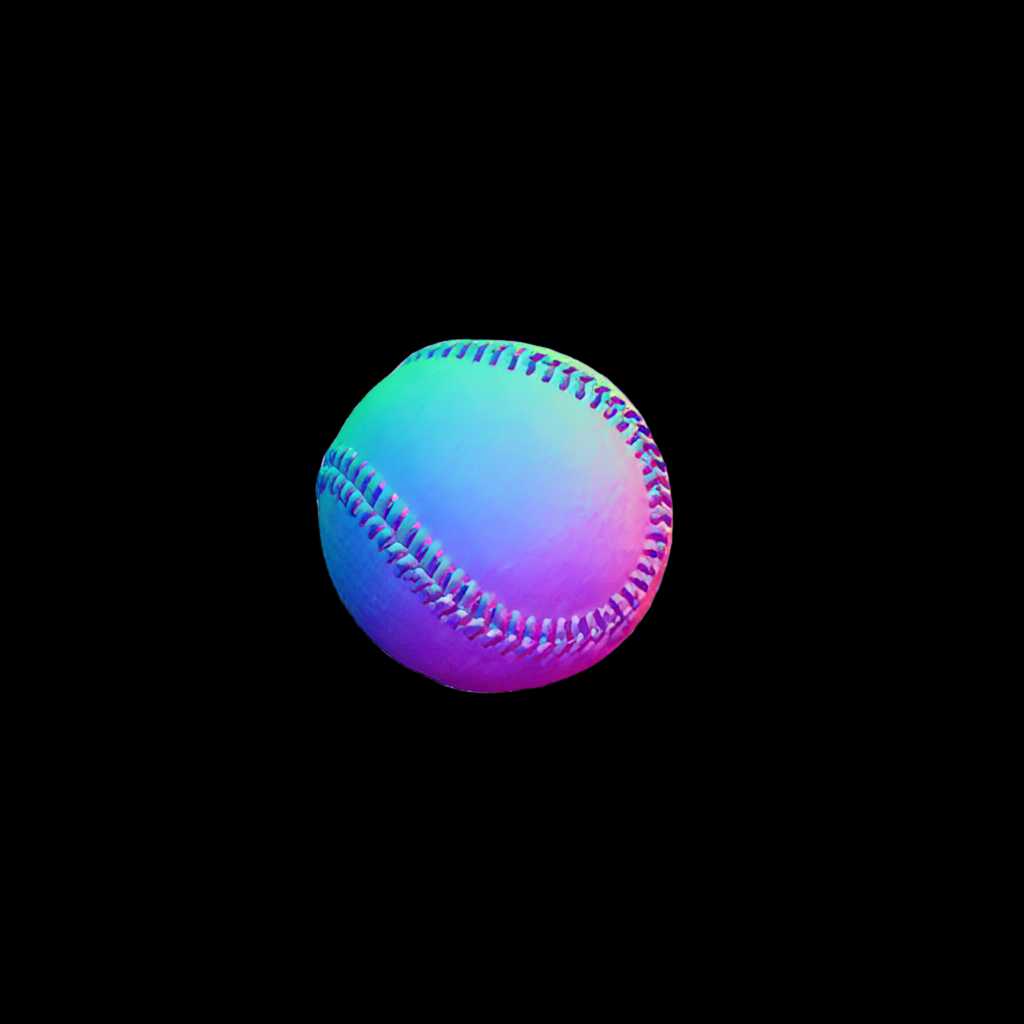}}\\
    \vspace{-0.2em}
    \subfloat{\includegraphics[width=0.25\linewidth]{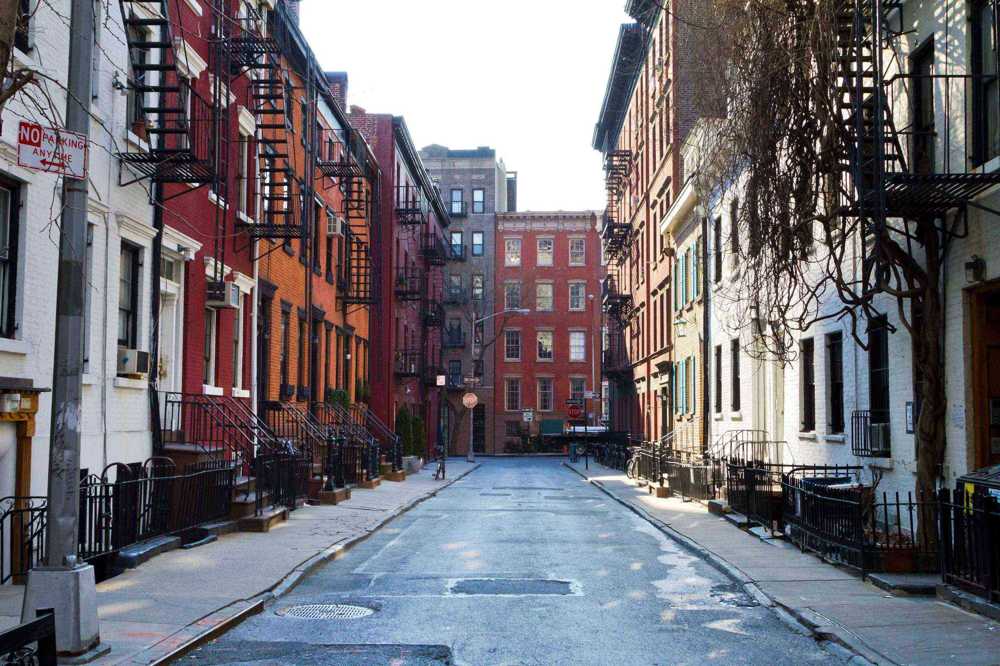}}
    \subfloat{\includegraphics[width=0.25\linewidth]{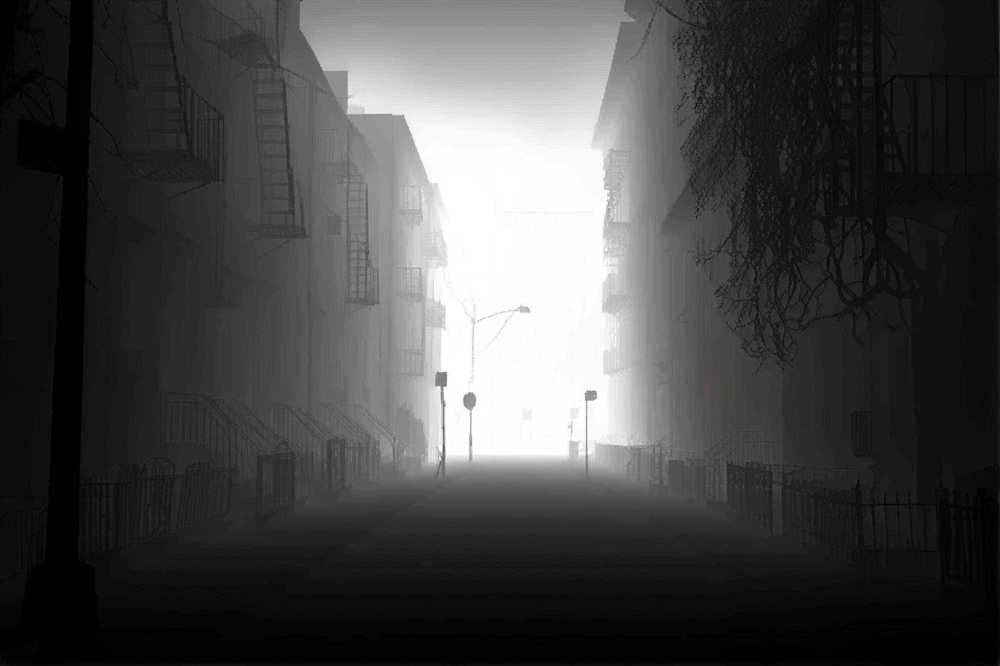}}
    \subfloat{\includegraphics[width=0.25\linewidth]{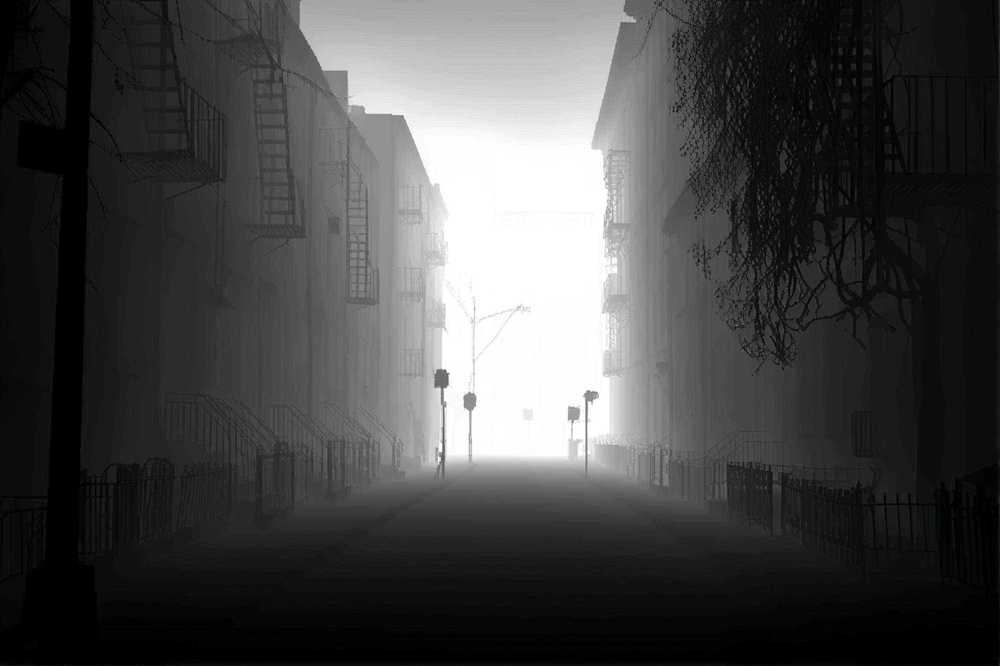}}
    \subfloat{\includegraphics[width=0.25\linewidth]{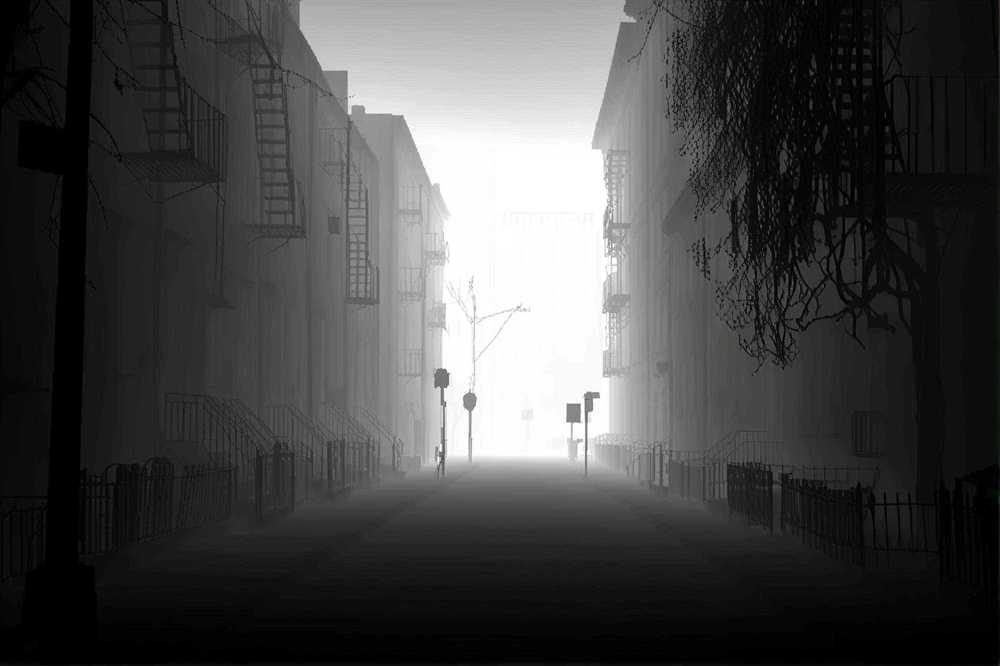}}\\
    \vspace{-0.2em}
    \subfloat{\includegraphics[width=0.25\linewidth]{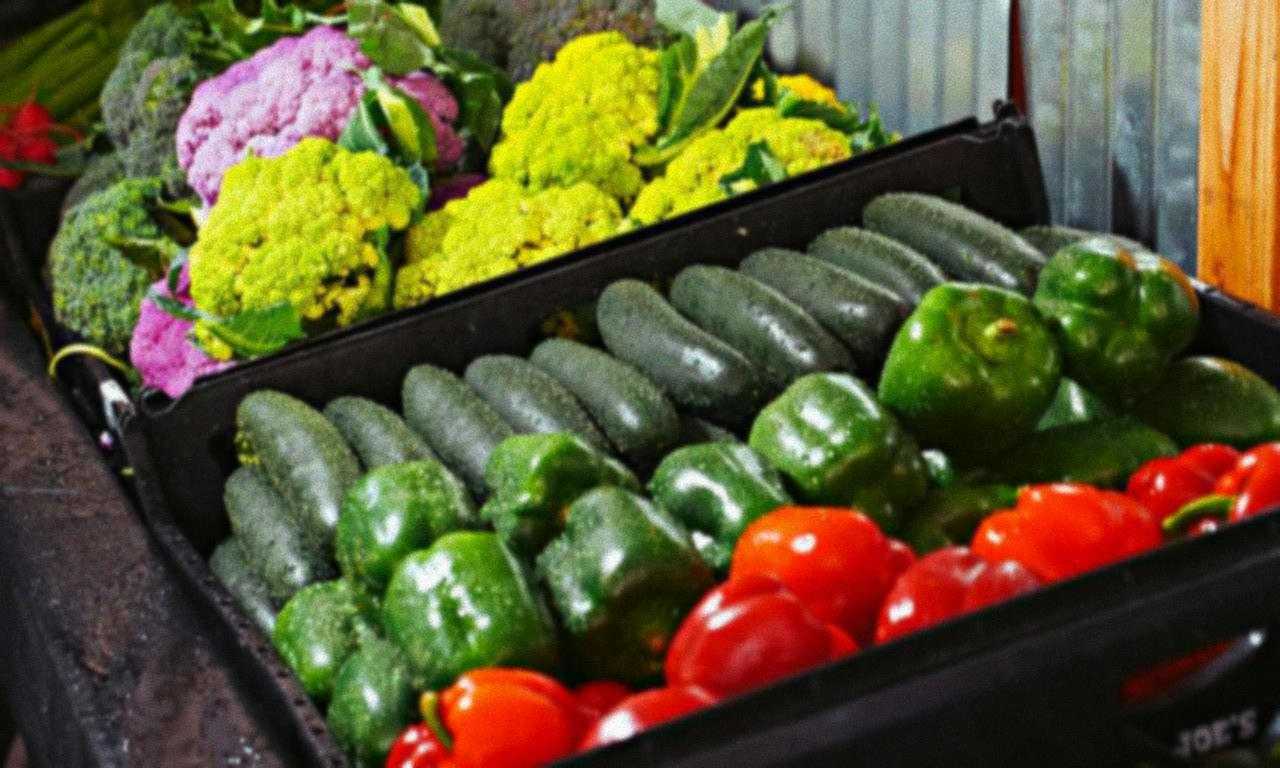}}
    \subfloat{\includegraphics[width=0.25\linewidth]{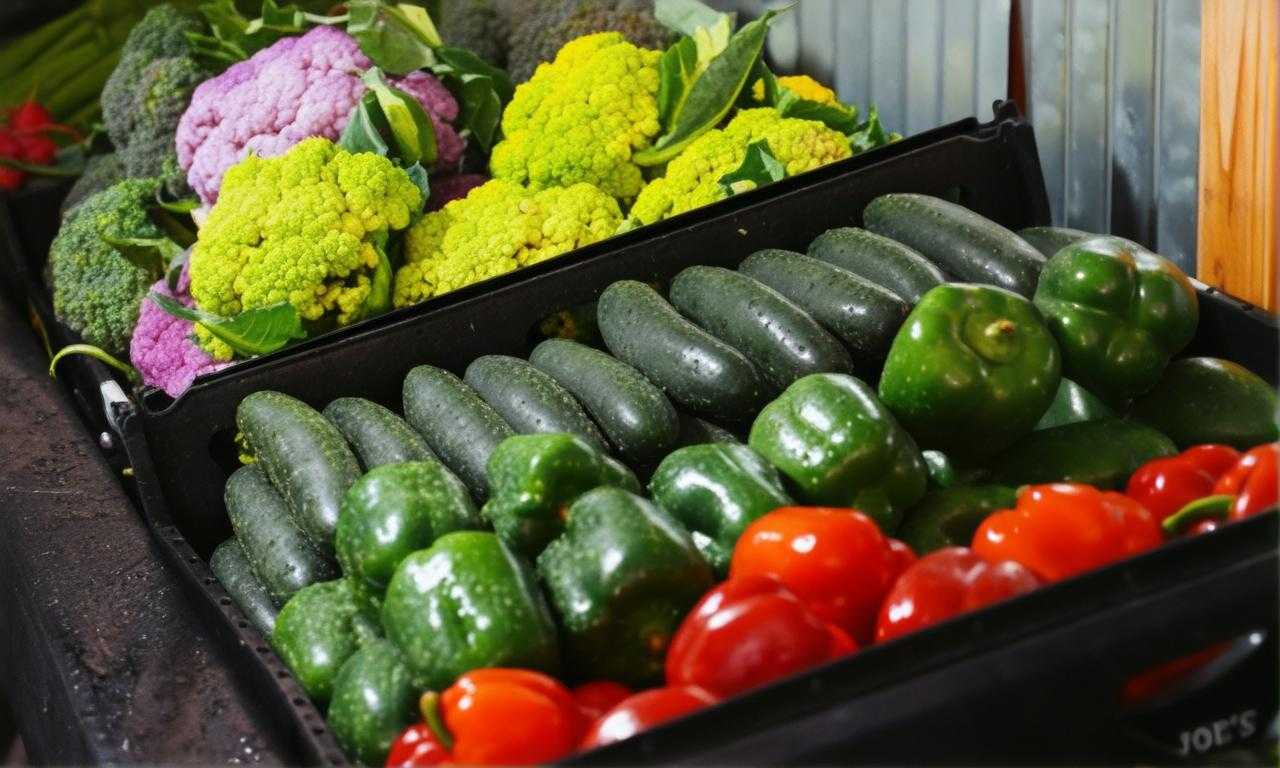}}
    \subfloat{\includegraphics[width=0.25\linewidth]{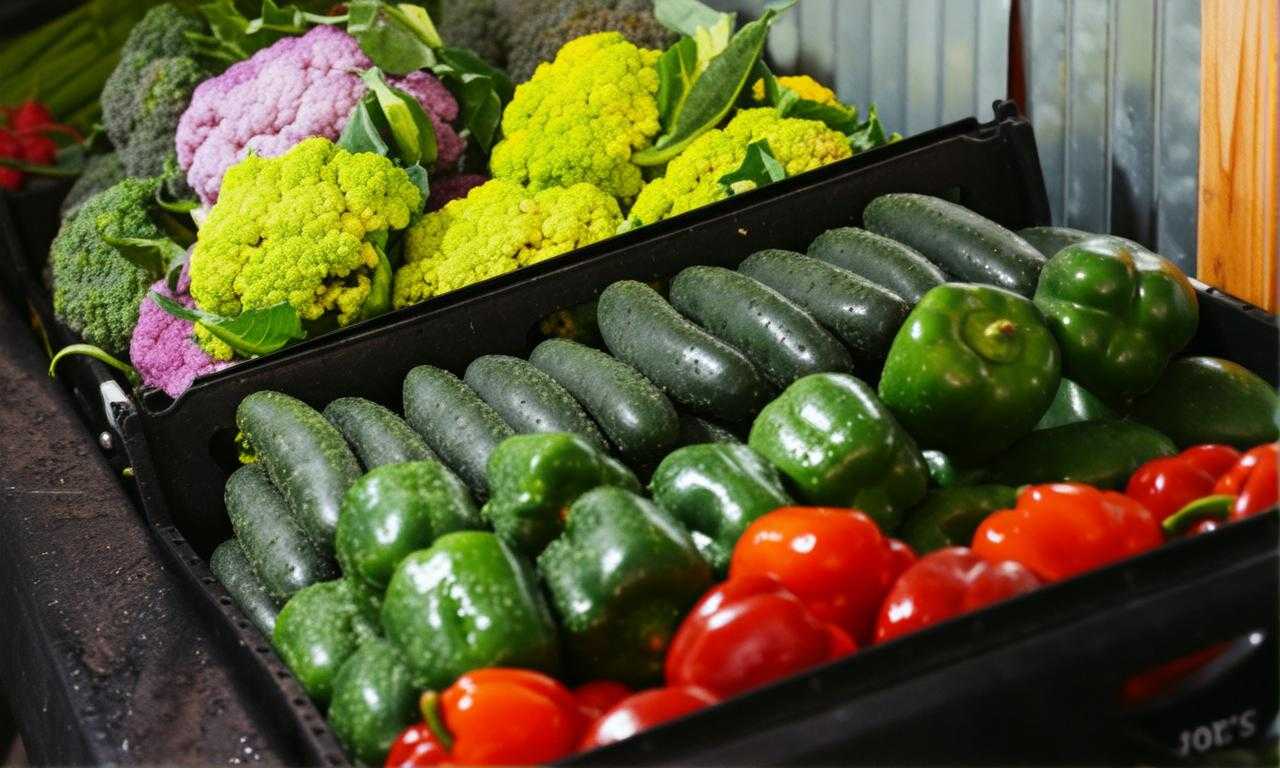}}
    \subfloat{\includegraphics[width=0.25\linewidth]{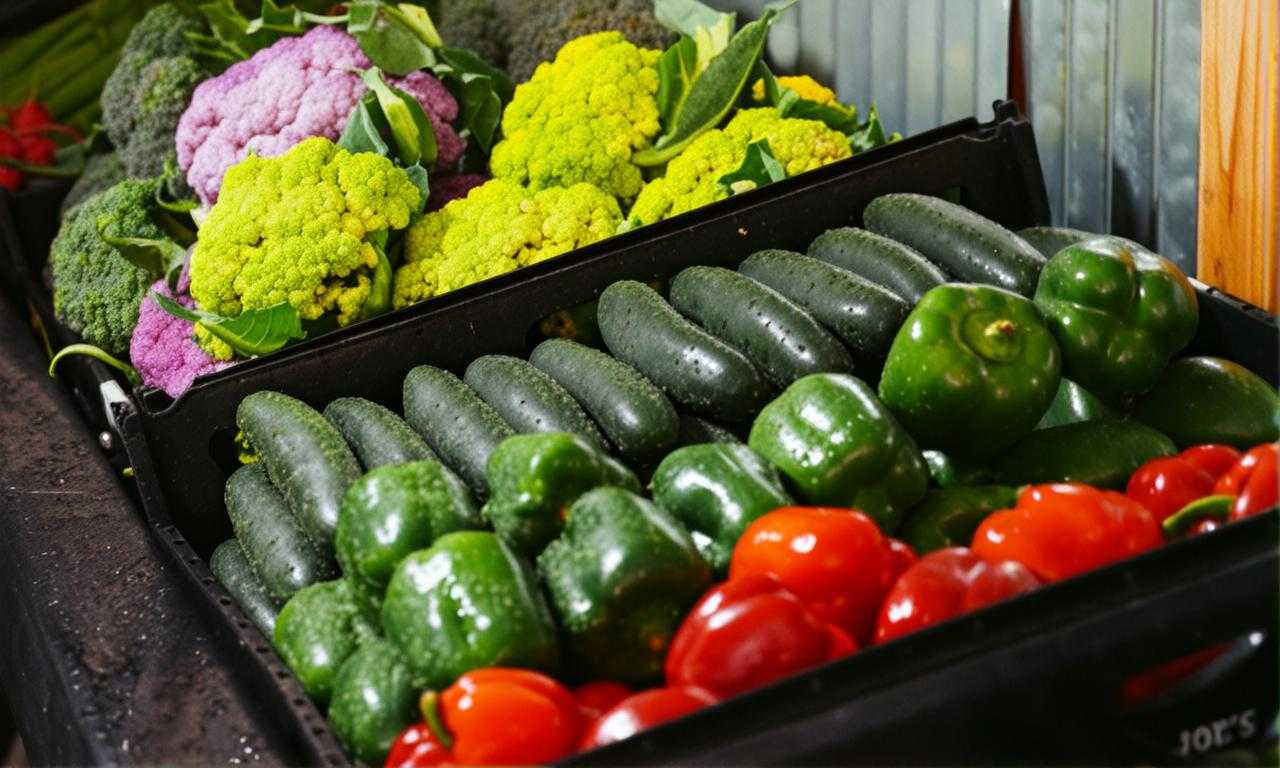}}\\
    \caption{Influence of the number of inference steps for depth and normal estimation as well as image restoration. From left to right: input image, output using a single neural function evaluations (NFE), 2 NFEs or 4 NFEs. Best viewed zoomed in.}
    \label{fig:surface}
\end{figure}

\paragraph{Influence of the Pixel Loss}
For this ablation, we vary the weight $\lambda$ in Eq.~\eqref{eq:full_loss} associated with the pixel loss and report in \cref{tab:pixel_loss} the metrics. As shown in \cref{tab:pixel_loss}, using a pixel loss is clearly beneficial to the model performance. Moreover, it was noted in our experiments that this specific loss also speeds up domain shift and allows to get better quality results.

\begin{table}[t]
    \centering
    \scriptsize
    \begin{tabular}{c|c|c|c|c|c}
        \toprule
        $\lambda$ & LPIPS $\downarrow$ & FID $\downarrow$ & Local FID $\downarrow$ & fMSE $\downarrow$ & PSNR $\uparrow$ \\
        \midrule
        1                 & 24.87 & 30.21 & 34.49 & \textbf{1233.99} & \textbf{22.55} \\
        5                 & 23.57 & 26.84 & 28.94 & 1281.71 & \textbf{22.55} \\
        10                & \textbf{23.29} & \textbf{26.70} & \textbf{28.84} & 1305.68& 22.39 \\  
        20                & 23.35 & 27.09 & 29.27 & 1360.02 & 22.34 \\
        %50                & 23.29 & 26.56 & 28.11 & 1358.09 & 22.32 \\
        \bottomrule
    \end{tabular}
    \caption{Influence of the pixel loss weight $\lambda$ for the object-removal task.}
    \label{tab:pixel_loss}
\end{table}

\section{Conclusion}
In this paper we introduced Latent Bridge Matching, a new method based on bridge matching in a latent space scalable to high-resolution images. We showed that this method is able to demonstrate strong performances for various image translation tasks. We carefully ablated the main elements of the method and underlined the importance of each component. A particularly interesting result of this study is that the stochasticity of the method is beneficial to the model performance. In particular, it outperforms flow matching which corresponds to the \emph{zero-noise} limit of our proposed framework. A noticeable limitation of the method relies on the need to have access to existing couplings of images beforehand to be able to train the model.

\clearpage
{
    \small
    \bibliographystyle{ieeenat_fullname}
    \bibliography{main}
}
\appendix
\clearpage
\section{Training details}
In this section, we provide any relevant parameters used to train our models.
\subsection{Object-removal task}
For the object-removal task, we trained our model for 20k iterations on 2 H100 GPUs. We set $\sigma=0.05$ and used the timestep distribution we propose in the main paper \emph{i.e.} $\pi(t) = \mathcal{U}({i/4})_{i\in\{ {0, 1, 2, 3 }\}}$. We use a bucketing strategy as proposed in \cite{podell2023sdxl} allowing us to handle multiple aspect ratios and resolutions. This strategy consists of defining buckets with pre-defined aspect ratios and pixel budgets and filling them with the data flow. During each training iteration, a target pixel budget is sampled and then the upcoming images are assigned to the bucket with the closest aspect ratio and budget and are resized accordingly. We use the following bucket pixel budgets: $[256^2, 512^2, 768^2, 1024^2]$ sampled with probabilities $[0.1, 0.2, 0.2, 0.5]$. For each budget we consider aspect ratios ranging from $0.25$ to $4$. The batch sizes are respectively set to 32, 16, 8 and 4 for each budget. We trained the model with LPIPS pixel loss with weight $\lambda=10$ and a learning rate of $3e^{-5}$ and we used the AdamW optimizer \cite{loshchilovdecoupled}. For data sources, we randomly sampled data from the RORD train set, our synthetic dataset or our in-the-wild dataset with probabilities $[0.3, 0.3, 0.4]$. For the latter, we used the random masking strategy proposed in \cite{suvorov2022resolution} while for RORD and our synthetic dataset we used the provided semantic masks. The denoiser is initialized using the weights of the pre-trained text-to-image model SDXL \cite{podell2023sdxl}. 

\subsection{Depth estimation}
For depth estimation, we trained our model for 20k iterations on 2 H100 GPUs. We set $\sigma = 0.005$ and set $\lambda=50$ for the pixel loss (LPIPS) scale. We used the following timestep distribution $\pi(t) = 0.025 \cdot \delta_{t=0.75} + 0.05 \cdot \delta_{t=0.5} + 0.025 \cdot \delta_{t=0.25} + 0.9 \cdot \delta_{t=0}$ to favor 1 step inference. We use a batch size of 4 and trained the model with a combination of \emph{hypersim} \cite{roberts2021hypersim} (40\%), \emph{virtual KITTI} \cite{cabon2020virtual} (10\%) and replica \cite{straub2019replica} (50\%) datasets. For \emph{virtual KITTI}, as is common, we set the far plane to 80m. The learning rate is set to $4e^{-5}$ and we used the AdamW optimizer during training.

\subsubsection{Normal estimation}
For surface normal estimation, we trained an LBM model for 25k iterations on 2 H100 GPUs. We set $\sigma = 0.1$ and $\lambda=50$ and used a pixel loss chosen as L1. We used the following timestep distribution $\pi(t) = 0.05 \cdot \delta_{t=0.75} + 0.1 \cdot \delta_{t=0.5} + 0.05 \cdot \delta_{t=0.25} + 0.8 \cdot \delta_{t=0}$ to favour 1 step inference. We used a batch size of 4 and trained the model with a combination of \emph{hypersim} \cite{roberts2021hypersim} (20\%), \emph{virtual KITTI} \cite{cabon2020virtual} (10\%) and replica \cite{straub2019replica} (70\%) datasets. The learning rate is set to $4e^{-5}$ and we used the AdamW optimizer during training.

\subsubsection{Image relighting}
In the case of image relighting, we trained a LBM model for 20k iterations on 2 H100 GPUs. We set $\sigma = 0.01$ and $\lambda=10$ and used a LPIPS pixel loss. We used the same timestep distribution and the same data bucketing strategy as for the object-removal task with the same bucket pixel budgets and probabilities. The training data is composed of synthetic data created using the rendering engine (90\%) and in-the-wild data (10\%). We trained the model with a learning rate of $3e^{-5}$ together with the AdamW optimizer.

\subsubsection{Controllable shadow generation and controllable image relighting}
For these experiments, we trained a conditional LBM for 19k iterations using a pixel loss scale set to $\lambda=2.5$ with LPIPS loss. We used a timestep distribution $\pi(t)$ similar to the one used for the object-removal task. We used a batch size of 4 and trained the model with a learning rate set to $5e^{-5}$ together with AdamW optimizer. The light map conditioning is injected by concatenating it in the latent space along the channels axis. In these cases, we only trained with the synthetic data created using the rendering engine.

\section{Additional object-removal results}
In this appendix, we provide additional results for the object removal task. In this case, instead of considering the \emph{coarse} semantic masks from RORD validation set, we consider the fine semantic masks precisely indicating the object to remove from the source image. We provide in \cref{tab:object-removal-semantic}, the same metrics as in the main paper. Similar to what was observed in the previous experiment, the proposed model is again able to reach the best results.
\begin{table}[ht]
    \centering
    \scriptsize
    \begin{tabular}{cccccc}
        \toprule
        Method (NFE) & FID $\downarrow$ & Local FID $\downarrow$ & fMSE $\downarrow$ & PSNR $\uparrow$ & SSIM $\uparrow$ \\
        \midrule
        LAMA (1) & 30.43 &	36.69	& 2450.60 &	19.74	&56.45\\
        SDXL inp. (50) & 42.55&	45.35		&3976.95&	20.06&	69.49  \\
        PowerPaint (50) & 40.61	& 40.35		&3673.91	&20.71	&66.85 \\
        AE (50)	& \underline{18.43}	& \underline{22.24}		& \underline{1772.99}	& \underline{22.81}	& \underline{70.79}\\
        \midrule
        Ours (1) & \textbf{15.50}	& \textbf{15.62}	& \textbf{1024.67}	& \textbf{24.28}	& \textbf{73.10}\\
        \bottomrule
    \end{tabular}
    \caption{Metrics for object-removal task with models fine-tuned on RORD train set and evaluated on RORD validation set (52k images) using the fine semantic masks. Our method uses a single NFE. Best results are highlighted in bold, second best are underlined.}
    \label{tab:object-removal-semantic}
\end{table}

For the sake of completeness, we also fine-tune LAMA, SDXL-inpaint., PowerPaint and our LBM checkpoint (Attentive Eraser is training-free) only on RORD train set such that all the models see approx. 400k samples, which was enough to reach convergence. For the sake of completeness, we also train a LBM model from scratch only on the RORD train set with the same number of iterations. We share the results in \cref{tab:object-removal-finetune}. As shown in the table, while this fine-tuning step improves competitors’ results, in particular for fine masks, our method still outperforms competitors for most metrics. Also note that our initial model is 047
trained on 2 H100 for $\approx$18h vs. 240h on 8 V100 for LAMA.

\begin{table}[ht]
    \centering
    \tiny
    \begin{tabular}{ccc|cc|cc|cc|cc|cc|cc|c}
      \toprule
      \multirow{2}{*}{Method} &\multicolumn{2}{c|}{\hspace{-0.33cm}FID$\downarrow$} & \multicolumn{2}{c|}{Local FID$\downarrow$} & \multicolumn{2}{c|}{fMSE$\downarrow$} & \multicolumn{2}{c|}{PSNR$\uparrow$} & \multicolumn{2}{c|}{SSIM$\uparrow$}  & Inf.\\
      & \hspace{-0.33cm}Coa. & \hspace{-0.33cm} Fin. & Coa. & \hspace{-0.33cm} Fin. & Coa. & \hspace{-0.33cm} Fin. & Coa. & \hspace{-0.33cm} Fin. & Coa. & \hspace{-0.33cm} Fin. & time (s) \\
      \midrule
      LAMA &\hspace{-0.33cm}30.3 &\hspace{-0.33cm}21.4 &	38.0 &\hspace{-0.33cm}28.2&	\underline{1592.2} &\hspace{-0.33cm}\underline{1350.3}&	19.7 &\hspace{-0.33cm}20.6&	55.9 &\hspace{-0.33cm}57.1&	\textbf{0.1}\\
      SDXL-inp. &\hspace{-0.33cm}\underline{27.2}	&\hspace{-0.33cm}18.5 &\textbf{27.3} &\hspace{-0.33cm}\underline{18.0}	&2297.3	&\hspace{-0.33cm}2213.1 &19.8	&\hspace{-0.33cm}21.4 &64.9	&\hspace{-0.33cm}69.0&7.2\\
      PowerPaint &\hspace{-0.33cm}29.9	&\hspace{-0.33cm}27.0 &\underline{30.0}	&\hspace{-0.33cm}23.7 &2871.2	&\hspace{-0.33cm}2679.7&18.5	&\hspace{-0.33cm}19.9 &58.3	&\hspace{-0.33cm}63.4&4.2\\
      AE & \hspace{-0.33cm}29.7 &\hspace{-0.33cm}\underline{18.4}& 33.2 &\hspace{-0.33cm}22.2& 2029.0 &\hspace{-0.33cm}1773.0& \underline{20.9} &\hspace{-0.33cm}\underline{22.8}& \underline{65.7}&\hspace{-0.33cm}\underline{70.8}& 8.0\\
      \midrule
      Ours &\hspace{-0.33cm}\textbf{26.9} &\hspace{-0.33cm}\textbf{15.7}&	30.5 &\hspace{-0.33cm}\textbf{15.6}&	\textbf{1306.6} &\hspace{-0.33cm}\textbf{997.4}&	\textbf{22.5} &\hspace{-0.33cm}\textbf{24.5}&	\textbf{69.2} &\hspace{-0.33cm}\textbf{73.2}&	\underline{0.3} \\
      Ours (scratch) &\hspace{-0.33cm}27.9 &\hspace{-0.33cm}16.7 &30.7 &\hspace{-0.33cm}16.9 &	1329.5	&\hspace{-0.33cm}1032.2&22.4 &\hspace{-0.33cm}24.4&	69.0	&\hspace{-0.33cm}72.9& 0.3\\
      \bottomrule
  \end{tabular}
    \caption{Metrics for object-removal task computed on RORD validation set using the coarse (Coa.) and fine (Fin.) masks. Our method and LAMA use a single neural function evaluation (NFE), others use 50 NFEs. Inference time is averaged over 50 images and computed on a single H100 GPU.}
    \label{tab:object-removal-finetune}
  \end{table}

% \subsection{Influence of the denoiser backbone}
% In this section, we evaluate the influence of the denoiser backbone on the performance of the proposed method. We considered 3 different U-Net architectures extracted fron the pretrained text-to-image diffusion models SDXL \cite{podell2023sdxl}, SD2 and SD1.5 \cite{rombach2022high}. We train the models for 20k iterations following the same approach as for the main paper. The value of $\sigma$ is set to $0.01$ and the pixel loss is chosen as LPIPS with weight $\lambda=10$. We provide in \cref{tab:backbone} the same metrics as in the main paper. As we can see, the proposed method is able to reach the best results.

% \begin{table}[t]
%     \centering
%     \scriptsize
%     \begin{tabular}{c|c|c|c|c|c}
%         \toprule
%         $\lambda$ & LPIPS $\downarrow$ & FID $\downarrow$ & Local FID $\downarrow$ & fMSE $\downarrow$ & PSNR $\uparrow$ \\
%         \midrule
%         SD1.5               & 24.51 & 29.16 & 30.79 & 1426.23 & 21.85 \\
%         SD2                 & 24.69 & 30.31 & 31.38 & 1409.58 & 21.96 \\
%         SDXL                &  \\  
%         %50                & 23.29 & 26.56 & 28.11 & 1358.09 & 22.32 \\
%         \bottomrule
%     \end{tabular}
%     \caption{Influence of the denoiser backbone for the object-removal task.}
%     \label{tab:backbone}
% \end{table}

\section{Results for depth estimation}
As mentioned in the main paper, we also consider the monocular depth estimation task which consists of estimating a depth map from a two dimensional image. We provide in \cref{tab:depth} the zero-shot results of our method compared to the state-of-the-art methods on commonly used evaluation datasets such as NYUv2 \cite{silberman2012indoor}, KITTI \cite{geiger2013vision}, ETH3D \cite{schops2017multi}, Scannet \cite{dai2017scannet} and DIODE \cite{vasiljevic2019diode}. As shown in the table , the proposed method is able to outperform or be competitive with the state-of-the-art methods and achieves the best average ranking across all metrics and datasets.

\begin{table*}[t]
    \centering
    \tiny
    \begin{tabular}{cccc|ccc|ccc|ccc|ccc|c}
        \toprule
        \multirow{2}{*}{Method} & \multicolumn{3}{c|}{NYUv2} & \multicolumn{3}{c|}{KITTI} & \multicolumn{3}{c|}{ETH3D} & \multicolumn{3}{c}{ScanNet} & \multicolumn{3}{c}{DIODE}  & Avg\\
        & AbsRel$\downarrow$ &  $\delta 1 \uparrow$ & $\delta 2 \uparrow$ & AbsRel$\downarrow$ &  $\delta 1 \uparrow$ & $\delta 2 \uparrow$ & AbsRel$\downarrow$ &  $\delta 1 \uparrow$ & $\delta 2 \uparrow$ & AbsRel$\downarrow$ &  $\delta 1 \uparrow$ & $\delta 2 \uparrow$ & AbsRel$\downarrow$ &  $\delta 1 \uparrow$ & $\delta 2 \uparrow$ & Rank\\
        \midrule
        DiverseDepth & 11.7 &  87.5 & - & 19.0 & 70.4 & - & 22.8 & 69.4 & - & 10.9 & 88.2 & - & 37.6 & 63.1 & - & 16.6\\
        MiDaS & 11.1 & 88.5 & - & 23.6 & 63.0 & - & 18.4 & 75.2 & - & 12.1 & 84.6 & - & 33.2 & 71.5 & - & 16.1\\
        LeRes & 9.0 & 91.6 & - & 14.9 & 78.4 & - & 17.1 & 77.7 & - & 9.1 & 91.7 & - & 27.1 & 76.6 & - & 13.2\\
        Omnidata & 7.4 & 94.5 & - & 14.9 & 83.5 & - & 16.6 & 77.8 & - & 7.5 & 93.6 & - & 33.9 & 74.2 & - & 13.2\\
        DPT & 9.8 & 90.3 & - & 10.0 & 90.1 & - & 7.8 & 94.6 & - & 8.2 & 93.4 & - & \textbf{18.2} & 75.8 & - & 10.8\\
        HDN & 6.9 & 94.8 & - & 11.5 & 86.7 & - & 12.1 & 83.3 & - & 8.0 & 93.9 & - & 24.6 & \underline{78.0} & - & 10.2\\
        DepthFM & 6.0 & 95.5 & -& 9.1 &90.2 & - &6.5 & 95.4 & - & 6.6 &  94.9 & -&\underline{22.4}& \textbf{78.5}& - &7.2\\
        GenPercept & 5.6 & 96.0 & 99.2 & 13.0 & 84.2 & 97.2 & 7.0 & 95.6 & 98.8 & 6.2 & 96.1 & 99.1 & 35.7 & 75.6 & 86.6 & 8.3\\
        Diff.-E2E-FT & 5.4 & 96.5 & 99.1 & 9.6 & 92.1 & 98.0 & 6.4 & 95.9 & 98.7 & 5.8 & 96.5 & 98.8 & 30.3 & 77.6 & 87.9 & 5.6 \\
        DepthAnything V2 & \underline{4.5} & \underline{97.9} & \underline{99.3} & \textbf{7.4} & \underline{94.6} & 98.6 & 13.1 & 86.5 & 97.5 & \textbf{4.2} & \underline{97.8} & \underline{99.3} & 26.5 & 73.4 & 87.1 & 5.4\\
        DepthAnything & \textbf{4.3} & \textbf{98.1} & \textbf{99.6} & \underline{7.6} & \textbf{94.7} & \textbf{99.2} & 12.7 & 88.2 & 98.3 & \underline{4.3} & \textbf{98.1} & \textbf{99.6} & 26.0 & 75.9 & 87.5 & 4.1 \\
        GeoWizard & 5.6 & 96.3 & 99.1 & 14.4 & 82.0 & 96.6 & 6.6 & 95.8 & 98.4 & 6.4 & 95.0 & 98.4 & 33.5 & 72.3 & 86.5 & 9.6 \\
        Marigold (LCM) & 6.1 & 95.8 & 99.0 & 9.8 & 91.8 & 98.7 & 6.8 & 95.6 & 99.0 & 6.9 & 94.6 & 98.6 & 30.7 & 77.5 & \textbf{89.3} & 7.7\\
        Marigold & 5.5 & 96.4 & 99.1 & 9.9 & 91.6 & 98.7 & 6.5 & 95.9 & 99.0 & 6.4 & 95.2 & 98.8 & 30.8 & 77.3 & \underline{88.7} & 6.4\\
        Lotus-D & 5.1 & 97.2 & 99.2 & 8.1 & 93.1 & 98.7 & \underline{6.1} & \textbf{97.0} & 99.1 & 5.5 & 96.5 & 99.0 & 22.8 & 73.8 & 86.2 & \underline{4.0}\\
        Lotus-G & 5.4 & 96.8 & 99.2 & 8.5 & 92.2 & 98.4 & \textbf{5.9} & \textbf{97.0} & \underline{99.2} & 5.9 & 95.7 & 98.8 & 22.9 & 72.9 & 86.0  & 5.3\\
        \midrule
        Ours & 5.6 & 97.2 & 99.2 & 9.4 & 93.0 & \underline{98.9} & 6.3 & \underline{96.5} & \textbf{99.3} & 5.7 & 97.0 & 99.2 & 30.3 & 77.5 & \textbf{89.3}& \textbf{3.7} \\
        \bottomrule
    \end{tabular}
    \caption{Metrics for depth estimation task. Our method uses a single NFE. Competitors results are taken from \cite{he2024lotus}. Best results are highlighted in bold, second best are underlined.}
    \label{tab:depth}
\end{table*}

\section{Failure cases}
In this section, we present some identified failure cases of our model for the different tasks considered.
\subsection{Object-removal}
\begin{figure}[t]
    \centering
    \captionsetup[subfigure]{position=above, labelformat = empty}
    \subfloat[Input]{\includegraphics[width=0.33\linewidth]{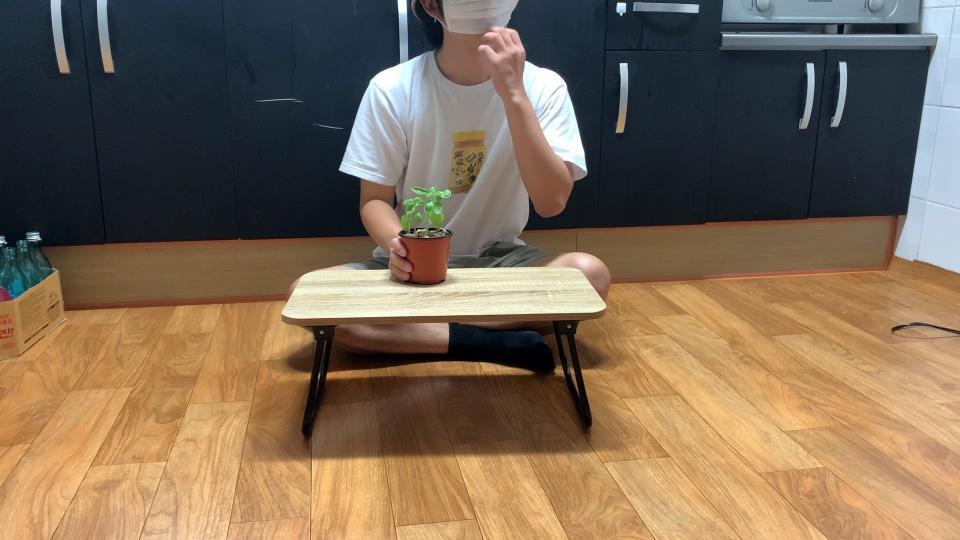}}
    \subfloat[Mask]{\includegraphics[width=0.33\linewidth]{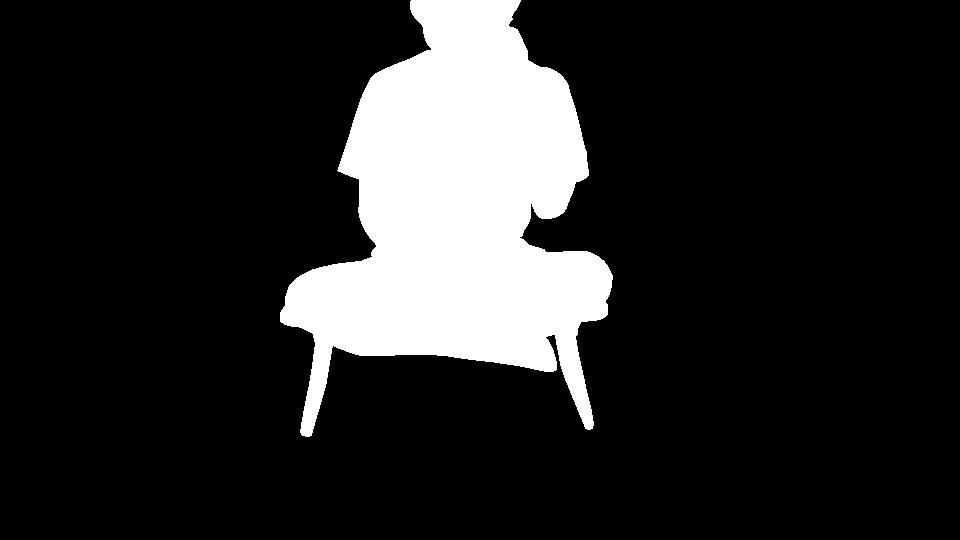}}
    \subfloat[Ours]{\includegraphics[width=0.33\linewidth]{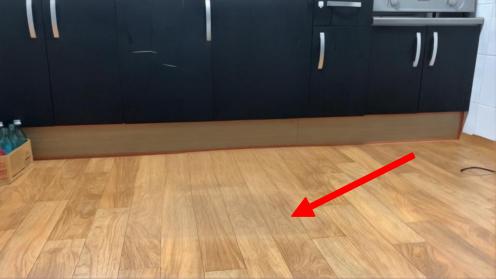}}\\
    \vspace{-0.1em}
    \subfloat{\includegraphics[width=0.33\linewidth]{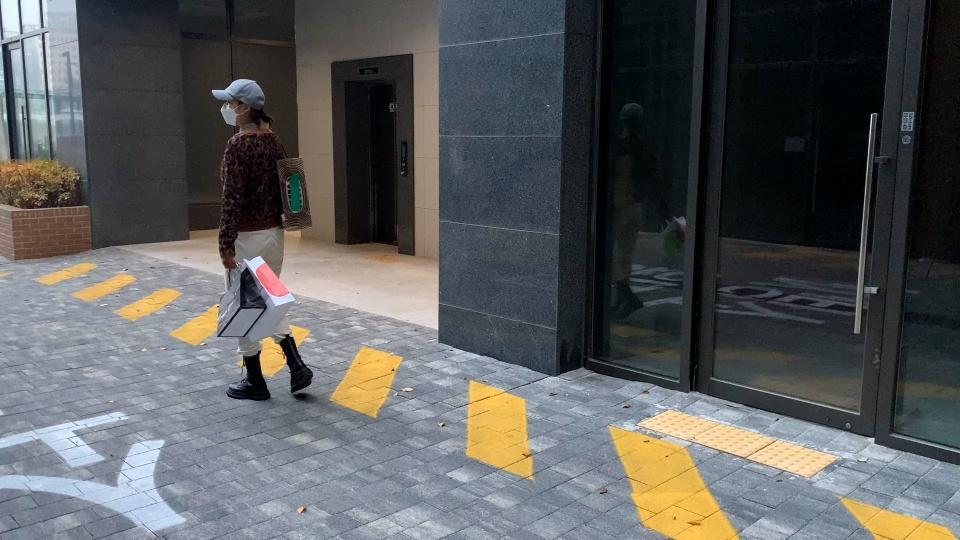}}
    \subfloat{\includegraphics[width=0.33\linewidth]{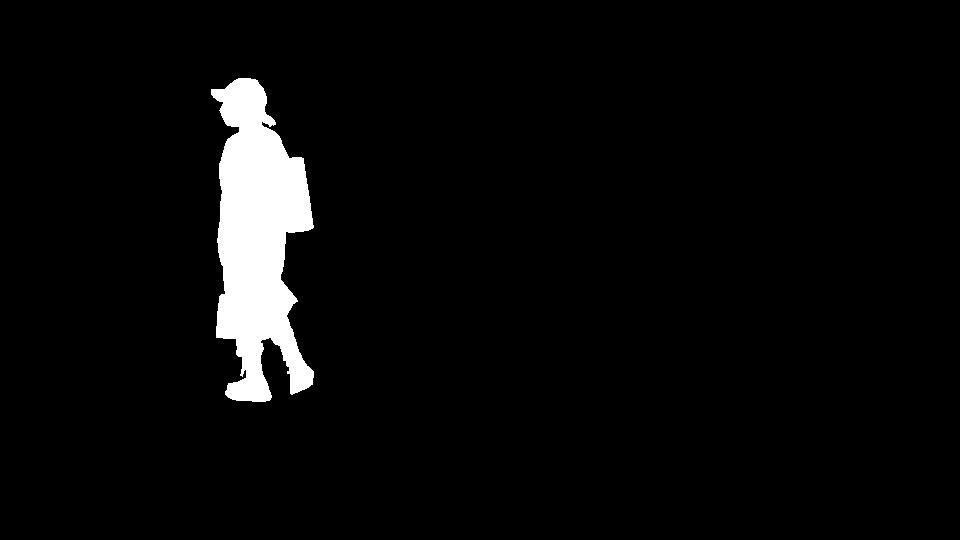}}
    \subfloat{\includegraphics[width=0.33\linewidth]{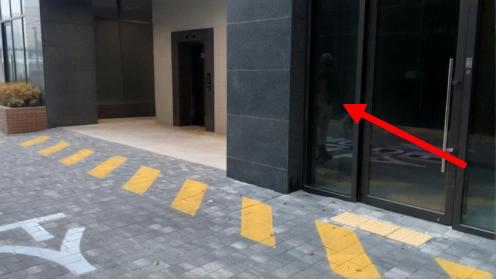}}
    \caption{Failure cases for object-removal. In the first row the model is not able to remove completely the shadow underneath the object. In the second row the model is not able to remove the reflection on the glass. }
    \label{fig:object-removal-failure-case-1}
\end{figure}

For object-removal, we noticed that our method can remove shadows more efficiently than all the existing methods as shown in the main paper, but there still exists some cases where it is not able to remove the shadow perfectly. Moreover, sometimes the model is not able to remove complex reflections of the object in the environment. These two failure cases are illustrated in \cref{fig:object-removal-failure-case-1}. On the top row, the shadow underneath the object to remove is still visible in the output image. On the bottom row, the model successfully removed the person and associated shadow but failed to remove the reflection on the glass door.

\subsection{Image relighting}
While the proposed method is able to handle most cases, we noticed that it can sometimes fail to remove existing reflections on the foreground image, induce a color shift or add a \emph{plastic} effect to the output image due to the use of synthetic data for training. We believe that these three failure cases can be addressed with a more careful training data curation and through more realistic renderings of the synthetic data.

\begin{figure}[t]
    \centering
    \captionsetup[subfigure]{position=above, labelformat = empty}
    \subfloat[Composite]{\includegraphics[width=0.205\linewidth]{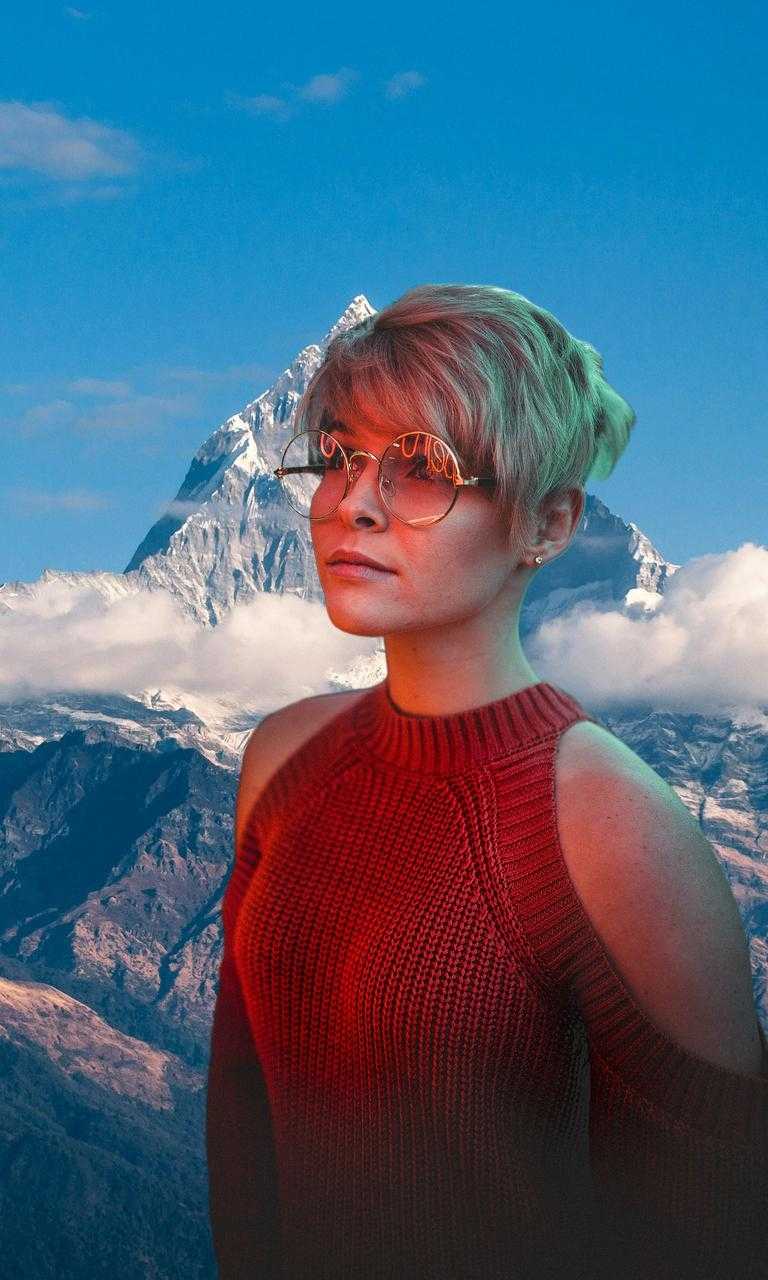}}
    \subfloat[Ours]{\includegraphics[width=0.205\linewidth]{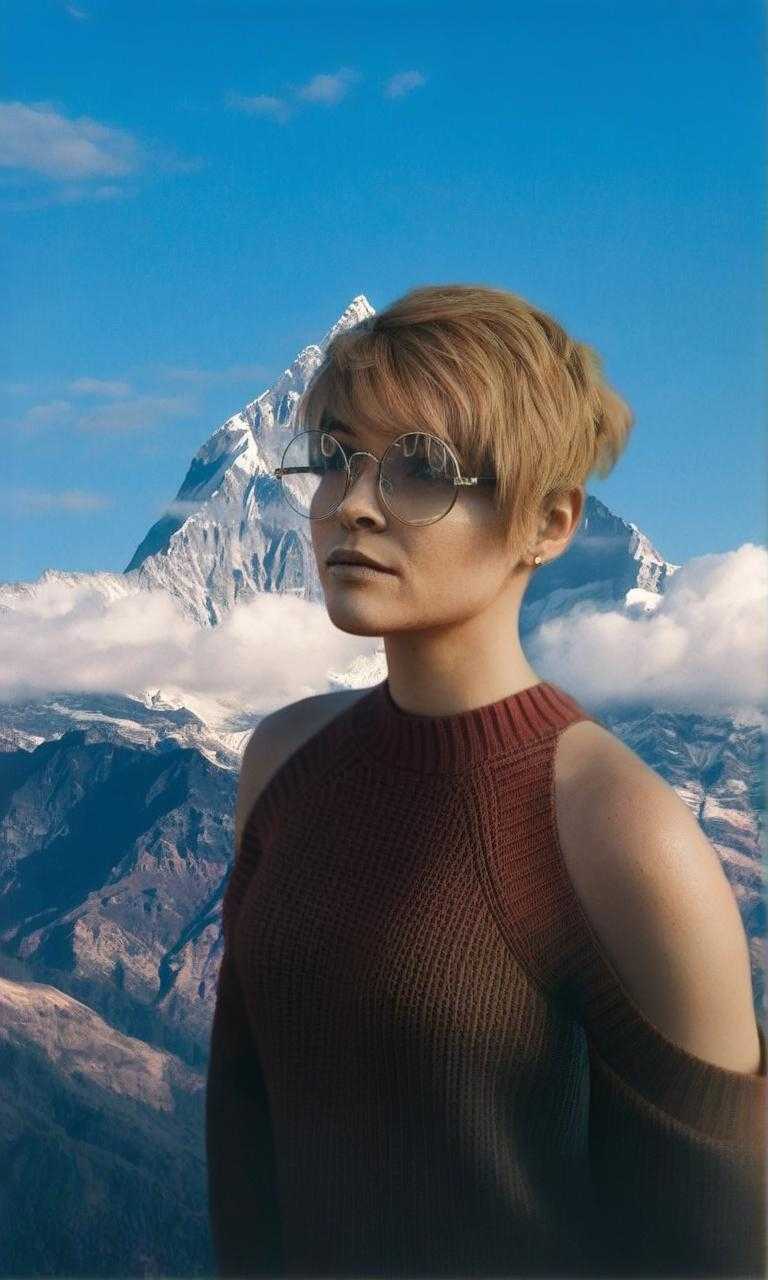}}
    \hspace{1em}
    \subfloat[Composite]{\includegraphics[width=0.266\linewidth]{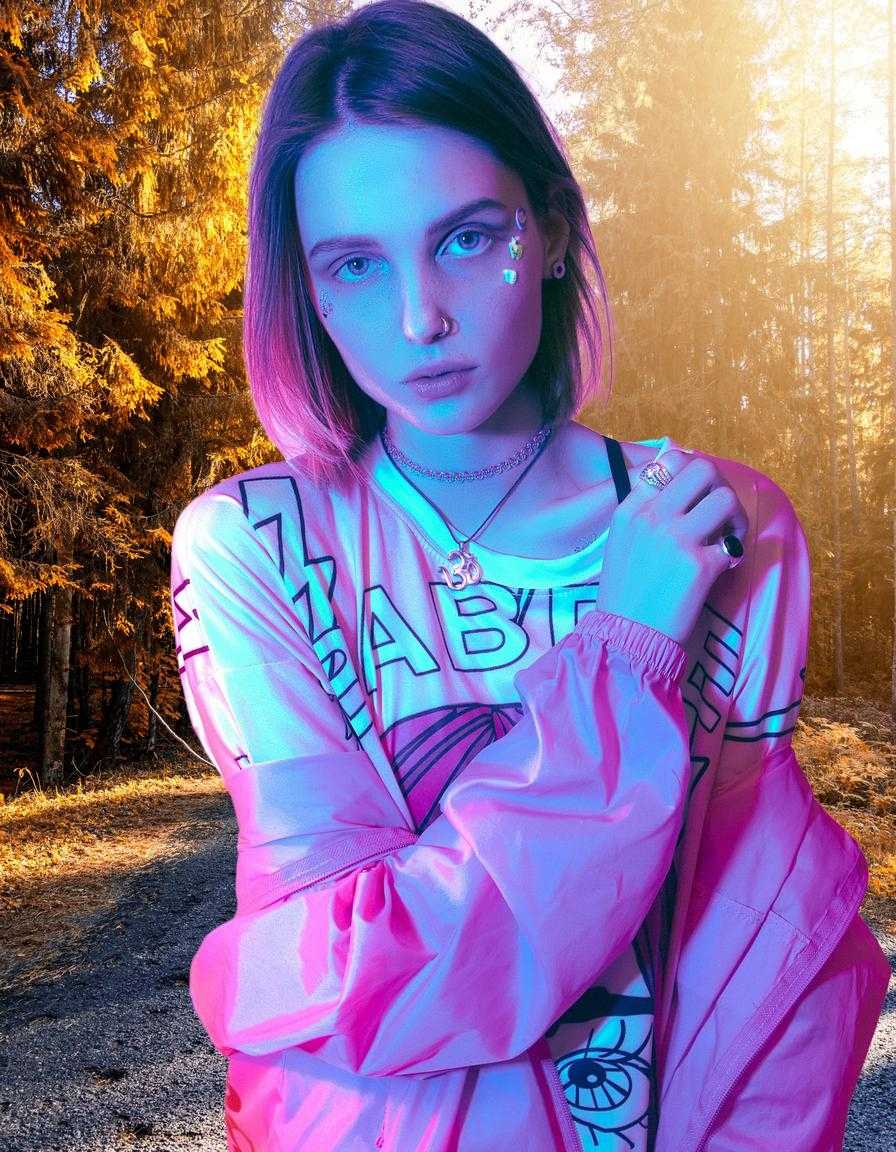}}
    \subfloat[Ours]{\includegraphics[width=0.266\linewidth]{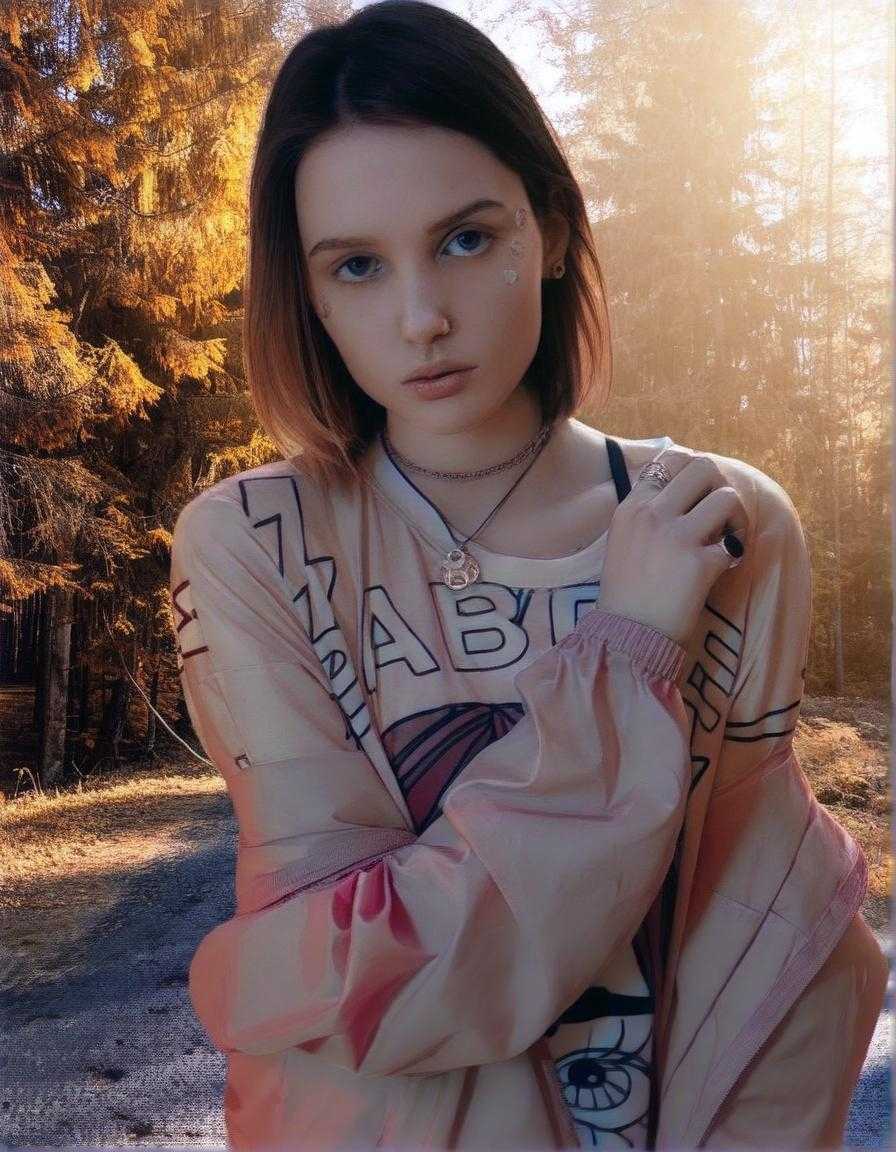}}
    \caption{Failure cases for image relighting. On the left, the model is not able to remove the reflection in the subject glasses. On the right, the model changes the color of the person's jacket and create a \emph{plastic} effect on the face.}
    \label{fig:object-relighting-failure-case-1}
\end{figure}

\section{Memory footprint and inference time}
Our intuition to use a latent model is motivated by the key observations made in \cite{rombach2022high} where the authors scale image generation from diffusion models. Nevertheless, we quantitatively report in \cref{tab:memory-latency} the memory/latency comparison between a pixel model and a latent model both for training and inference. Note that the VAE compresses the source image by a factor of 8 and is frozen during training drastically reducing the memory footprint of the model as shown in the table.

\begin{table}[ht]
    \tiny
    \begin{tabular}{ccccc}
        \toprule
        Mode (Resolution) & Metric & Pixel Model & Latent Model & Gain\\
        \midrule
        \multirow{2}{*}{Inference (256 / 1024)}& Latency (s) & 0.19 / 20.11 &	0.14 / 0.27 & 26.3\% / 98.7\% \\
        & Peak Memory (Gb) & 5.63 / 15.71 &	 5.29 / 7.70 & 6.0\% / 51.0\% \\
        \midrule
        \multirow{2}{*}{Training (256 / 1024)}& Latency (s) & 0.71 / - &	0.43 / 0.58 & 39.4\% / - \\
        & Peak Memory (Gb) & 43.19 / OoM	& 24.85/ 25.35 & 41.3\% / -\\
      \bottomrule
    \end{tabular}
    \caption{Training and inference memory usage and per-iteration latency for a \emph{pixel} and a \emph{latent} bridge model. The metrics are averaged over 10 images using a batch size of 1 with AdamW for training and 1 NFE for inference on a single H100 80Gb GPU.}
    \label{tab:memory-latency}
  \end{table}
\section{Additional samples}
\begin{figure*}[t]
    \captionsetup[subfigure]{position=above, labelformat = empty}
    \centering
    \subfloat{\includegraphics[width=\linewidth]{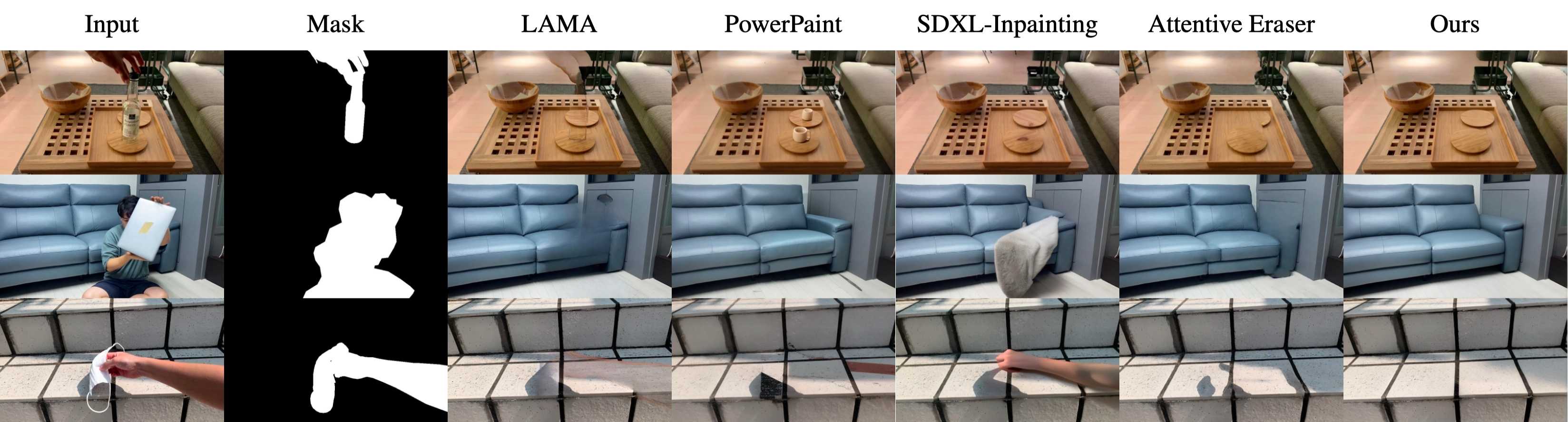}}\\
    \vspace{-0.1em}
    \subfloat{\includegraphics[width=\linewidth]{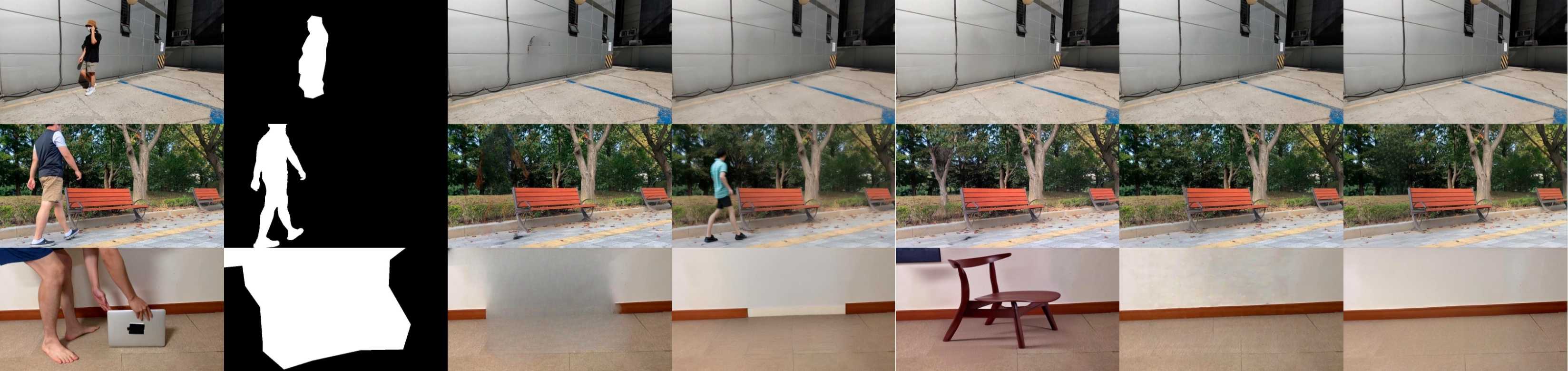}}\\
    \vspace{-0.1em}
    \subfloat{\includegraphics[width=\linewidth]{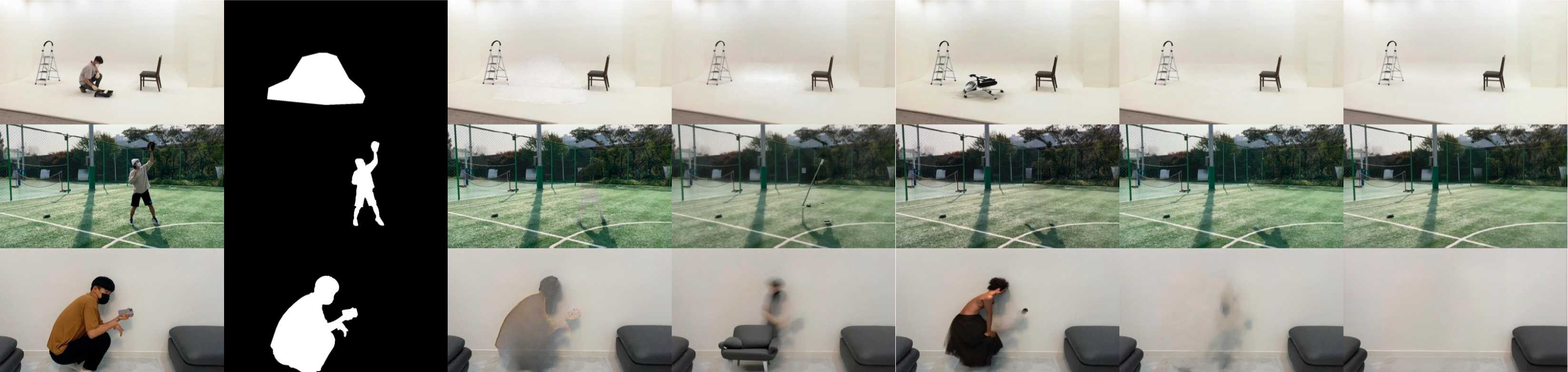}}
    \vspace{-0.1em}
    \subfloat{\includegraphics[width=\linewidth]{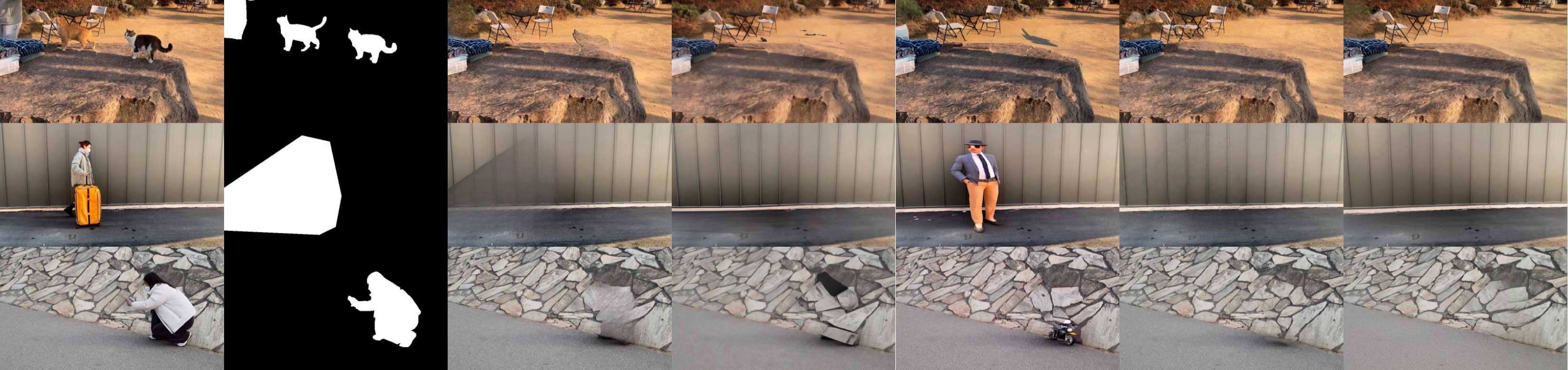}}

    \caption{Qualitative results for object-removal on RORD validation dataset \cite{sagong2022rord}. Best viewed zoomed in. Our model uses a single NFE and is able to successfully remove not only the object but also its shadow.}
    \label{fig:appendix-object-removal}
\end{figure*}

\begin{figure*}[ht]
    \captionsetup[subfigure]{position=above, labelformat = empty}
    \centering
    \subfloat{\includegraphics[width=\linewidth]{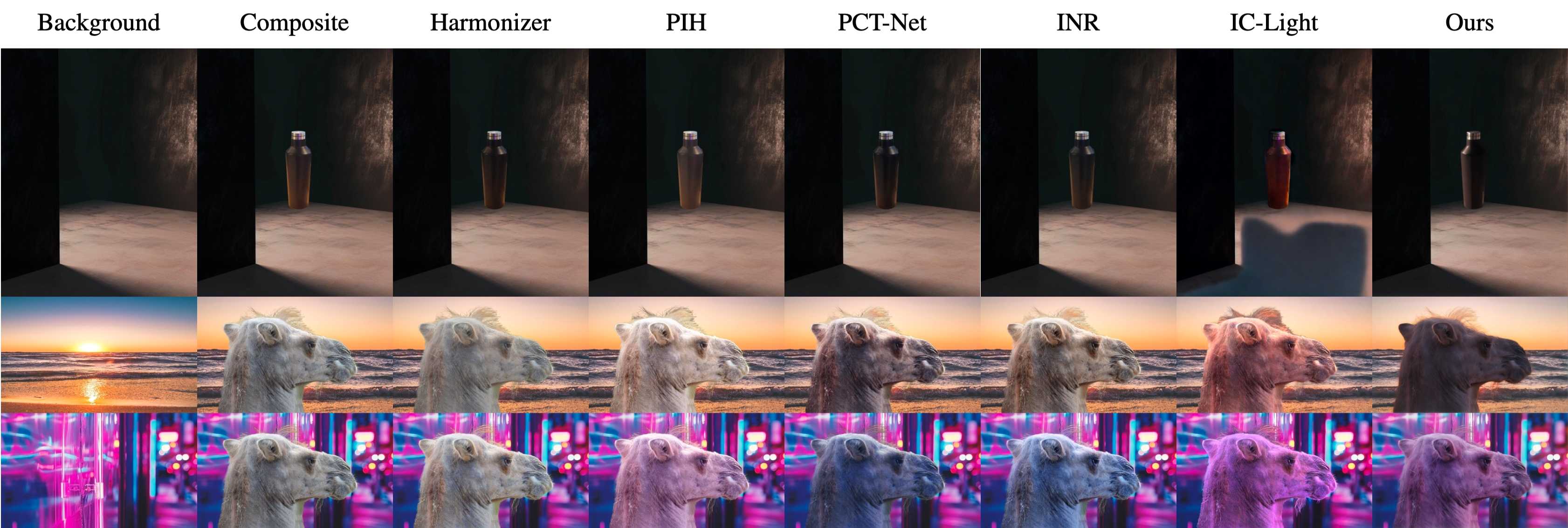}}\\
    \vspace{-0.1em}
    \subfloat{\includegraphics[width=\linewidth]{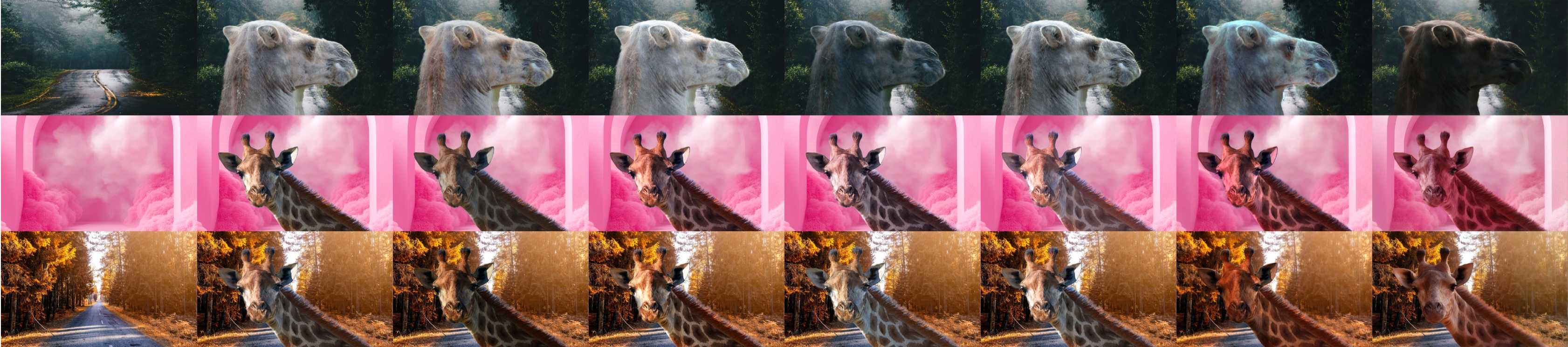}}\\
    \vspace{-0.1em}
    \subfloat{\includegraphics[width=\linewidth]{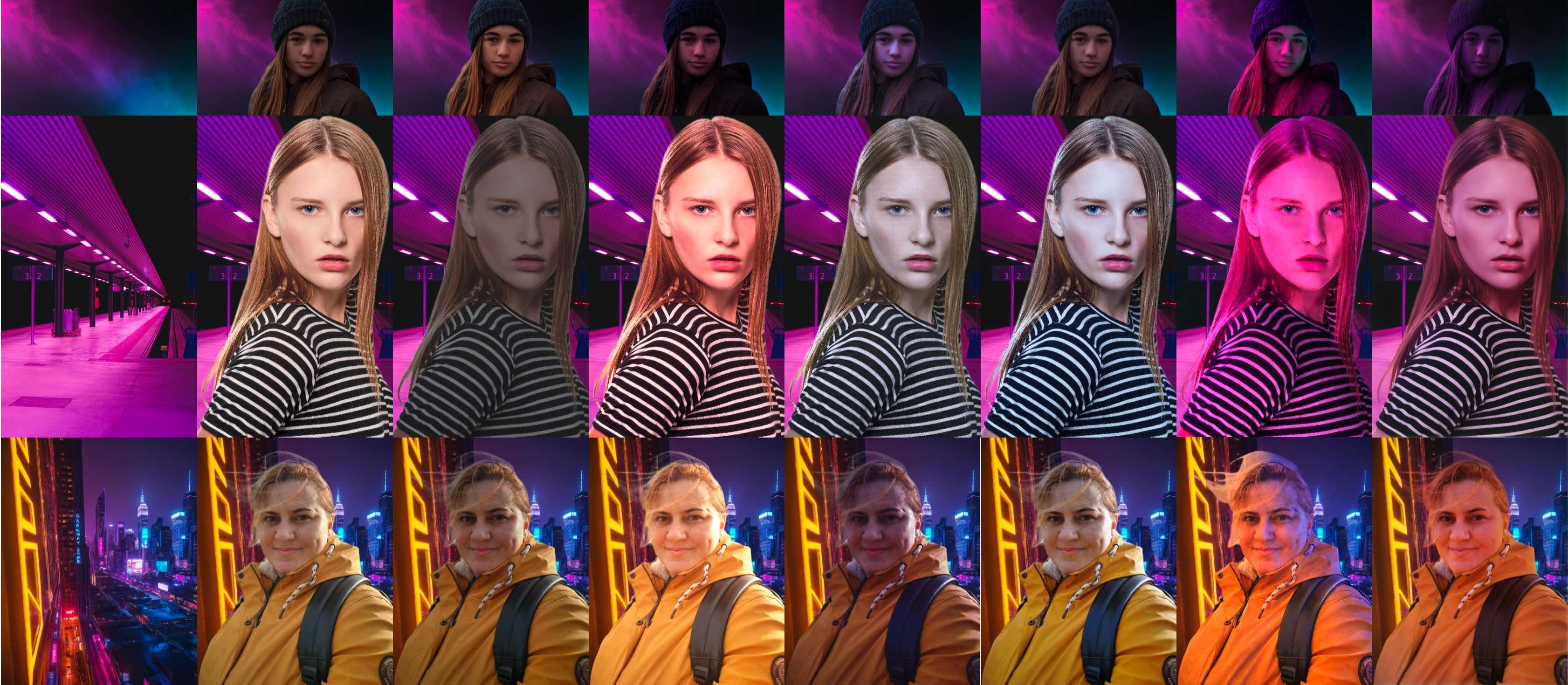}}
    \vspace{-0.1em}
    \subfloat{\includegraphics[width=\linewidth]{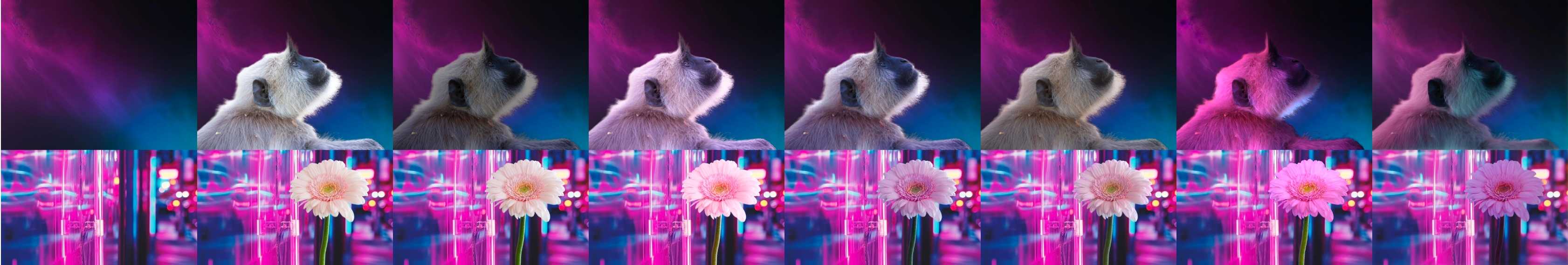}}

    \caption{Qualitative results for object relighting. The model is able to relight the object according to the provided background and also remove existing shadows and reflections.}
    \label{fig:app_relighting_results}
\end{figure*}

\begin{figure*}[ht]
    \captionsetup[subfigure]{position=above, labelformat = empty}
    \centering
    \subfloat[Original]{\includegraphics[width=0.24\linewidth]{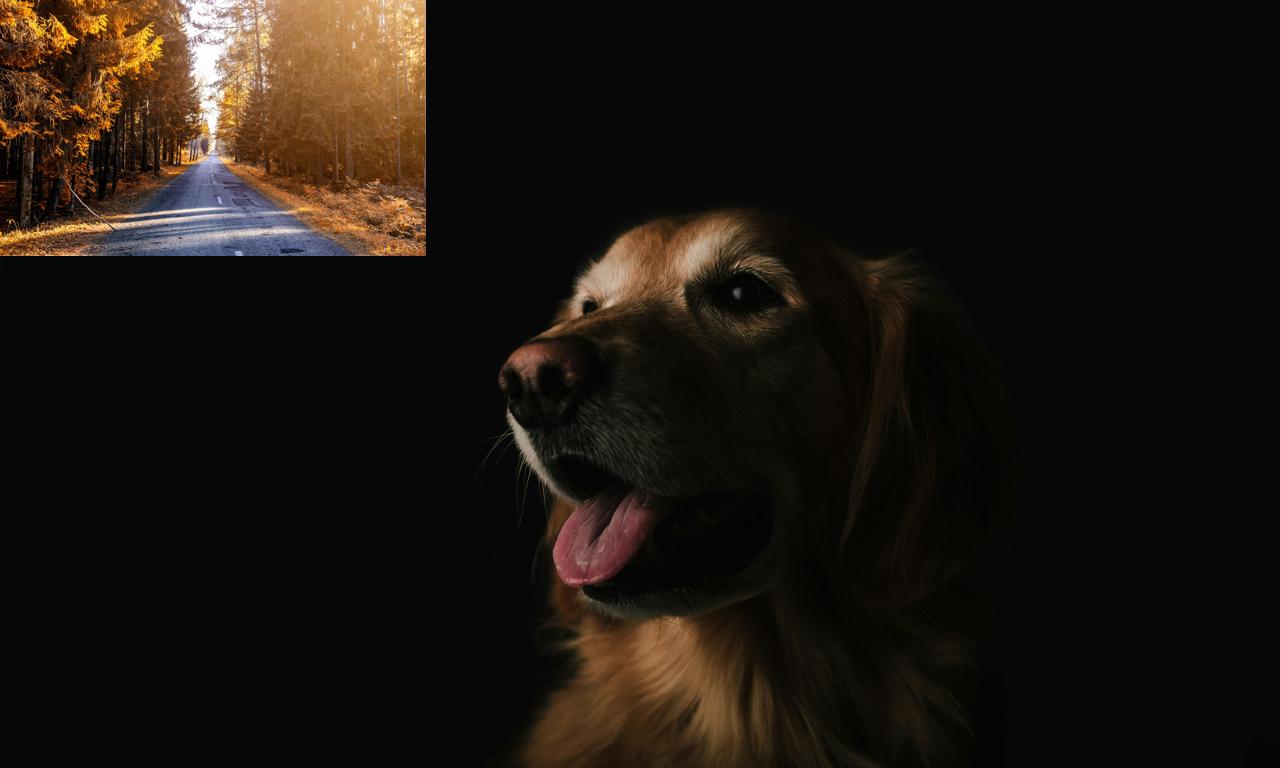}}
    \subfloat[Ours (1 NFE)]{\includegraphics[width=0.24\linewidth]{plots/relighting/additional_paper_plots_rf/output_image_0.jpg}}
    \hspace{0.1em}
    \subfloat[Original]{\includegraphics[width=0.24\linewidth]{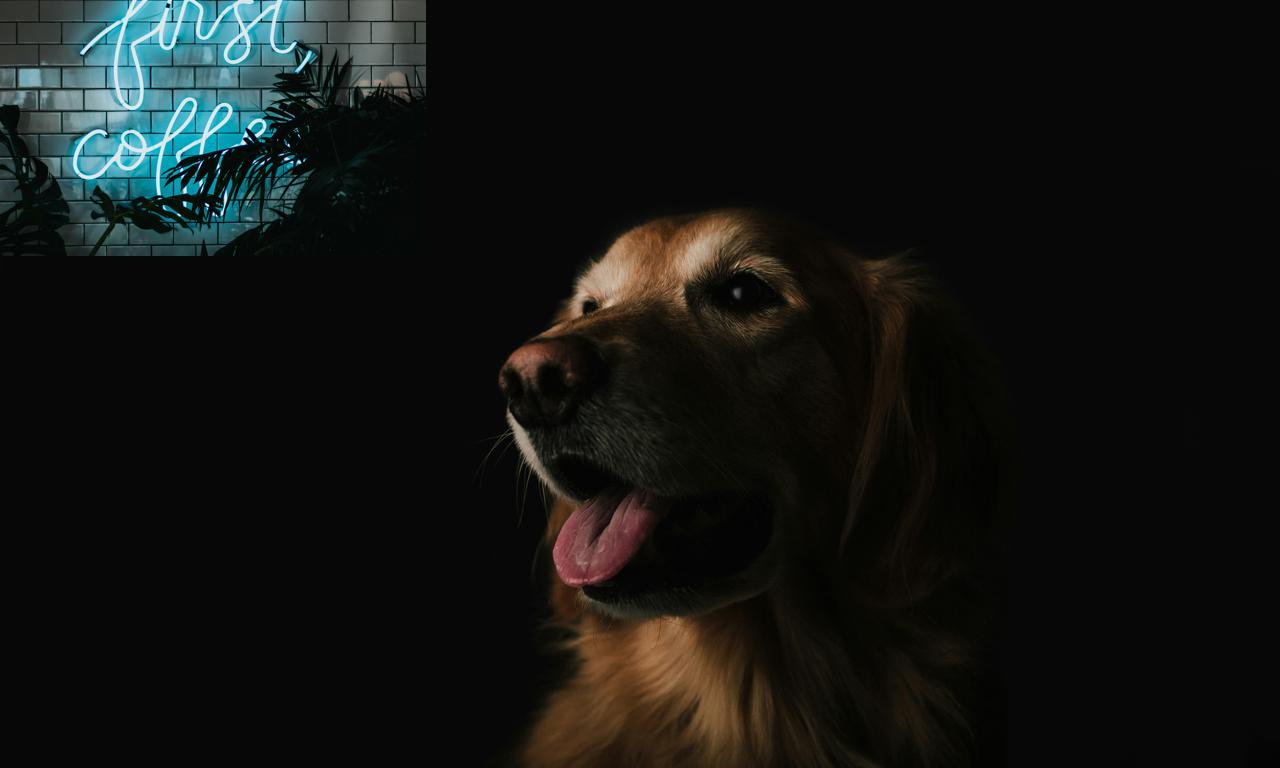}}
    \subfloat[Ours (1 NFE)]{\includegraphics[width=0.24\linewidth]{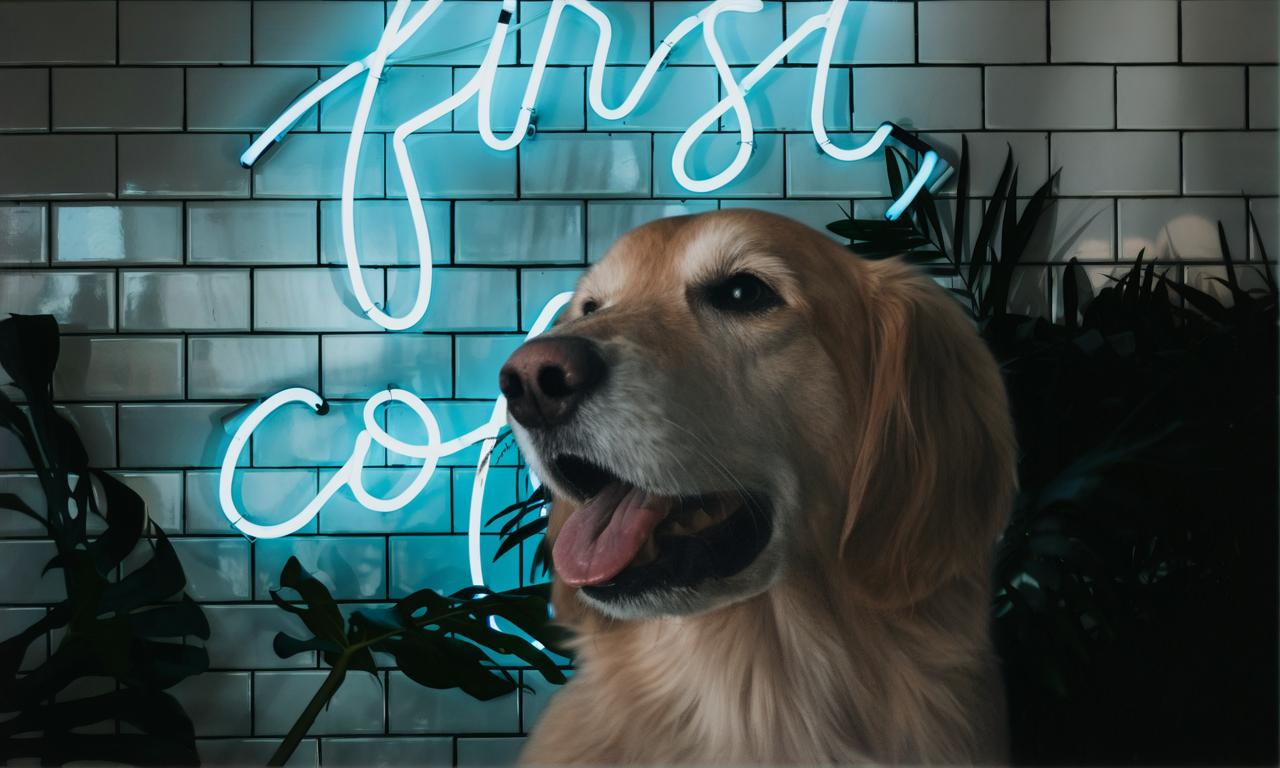}}\\
    \vspace{-0.1em}
    \subfloat{\includegraphics[width=0.24\linewidth]{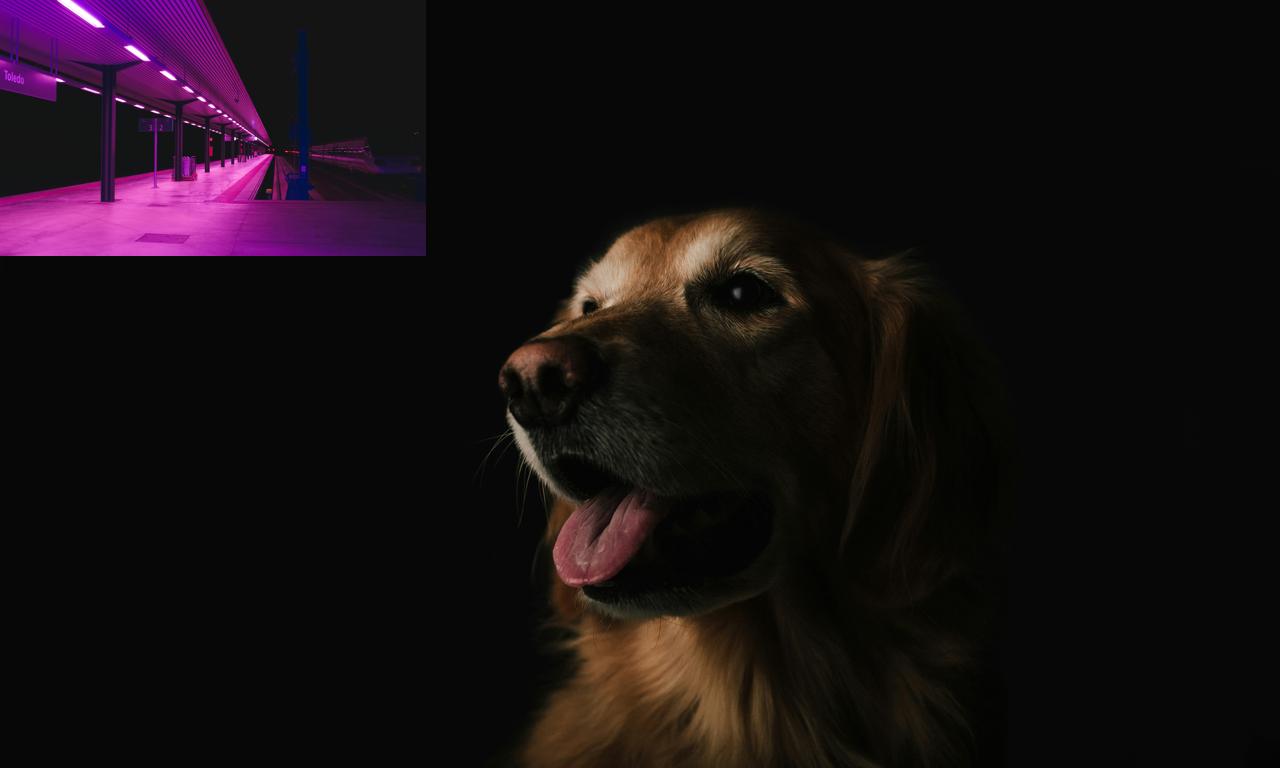}}
    \subfloat{\includegraphics[width=0.24\linewidth]{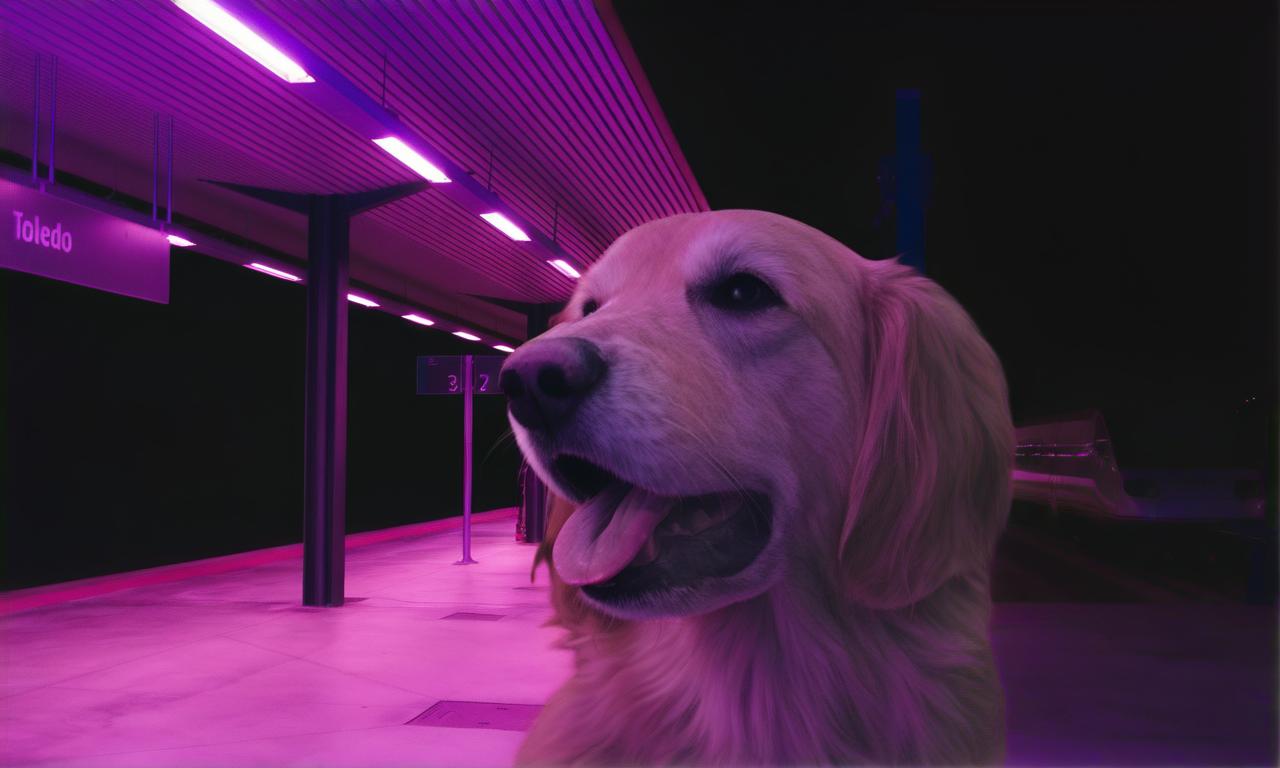}}
    \hspace{0.1em}
    \subfloat{\includegraphics[width=0.24\linewidth]{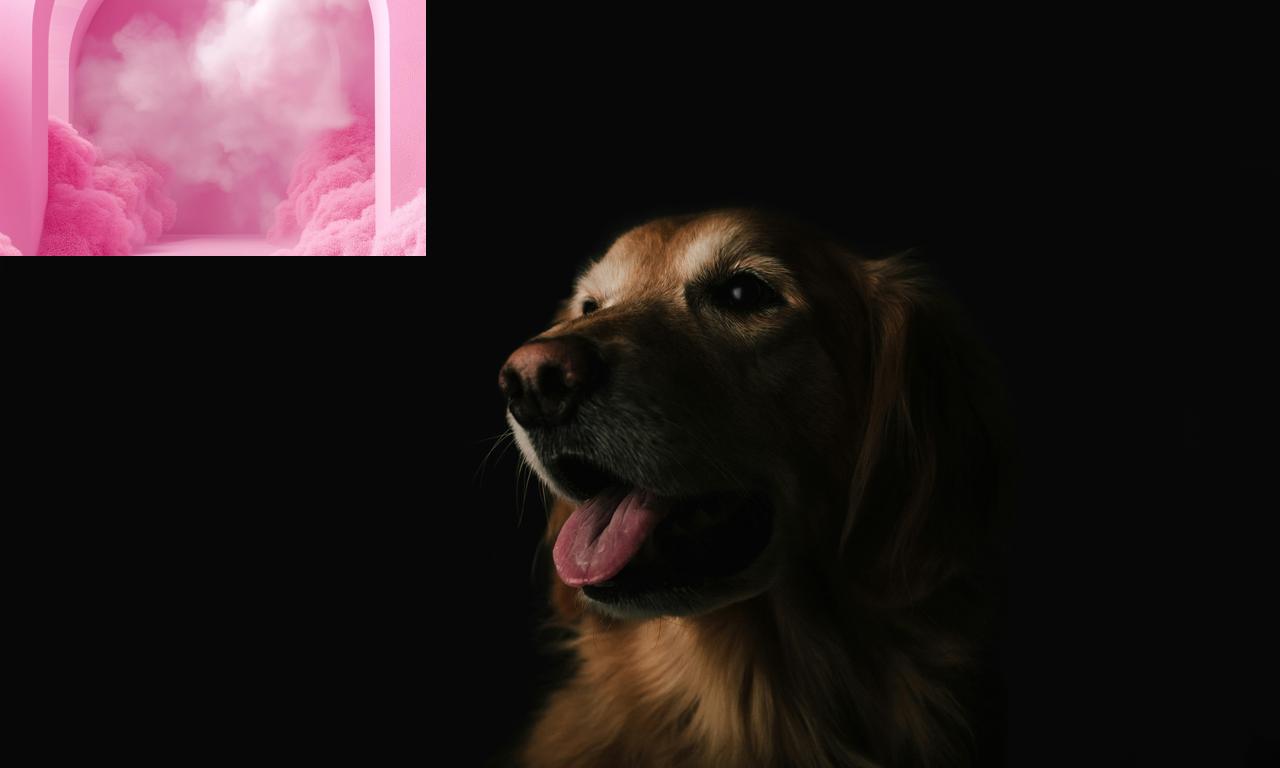}}
    \subfloat{\includegraphics[width=0.24\linewidth]{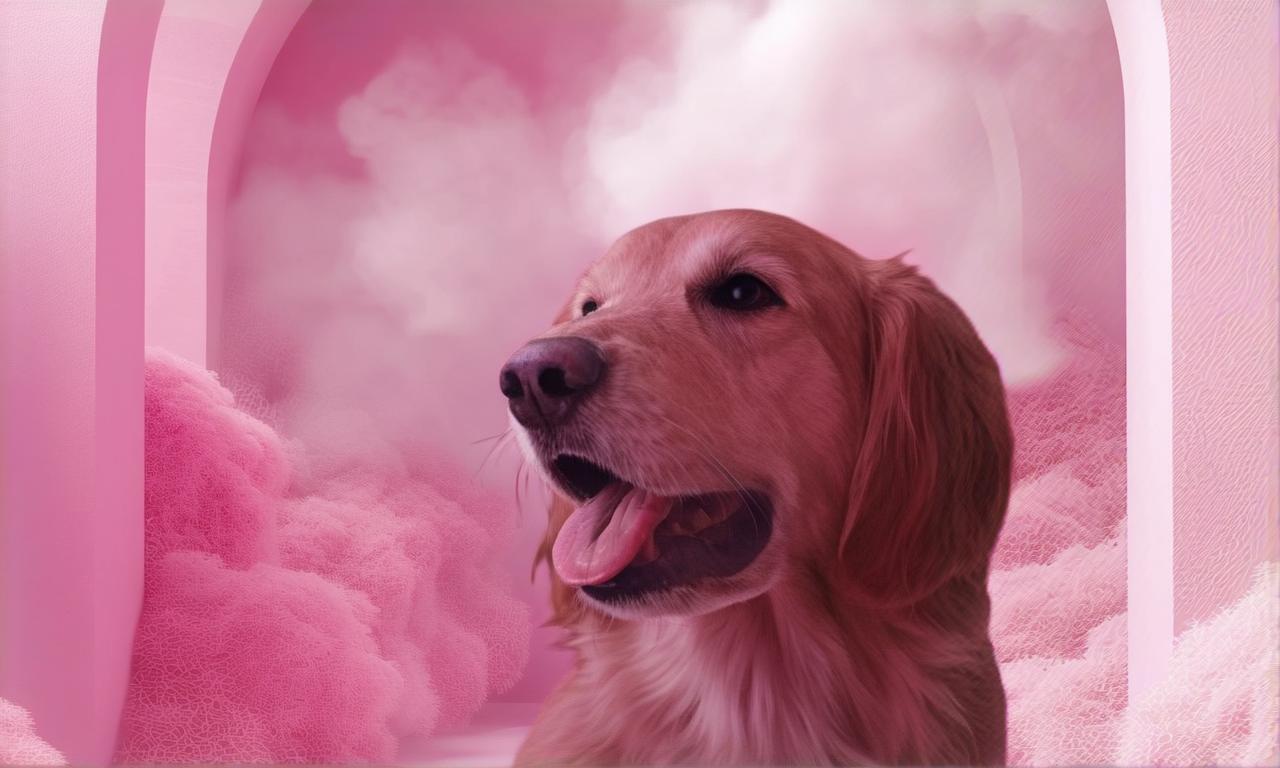}}\\
    \vspace{-0.1em}
    \subfloat{\includegraphics[width=0.24\linewidth]{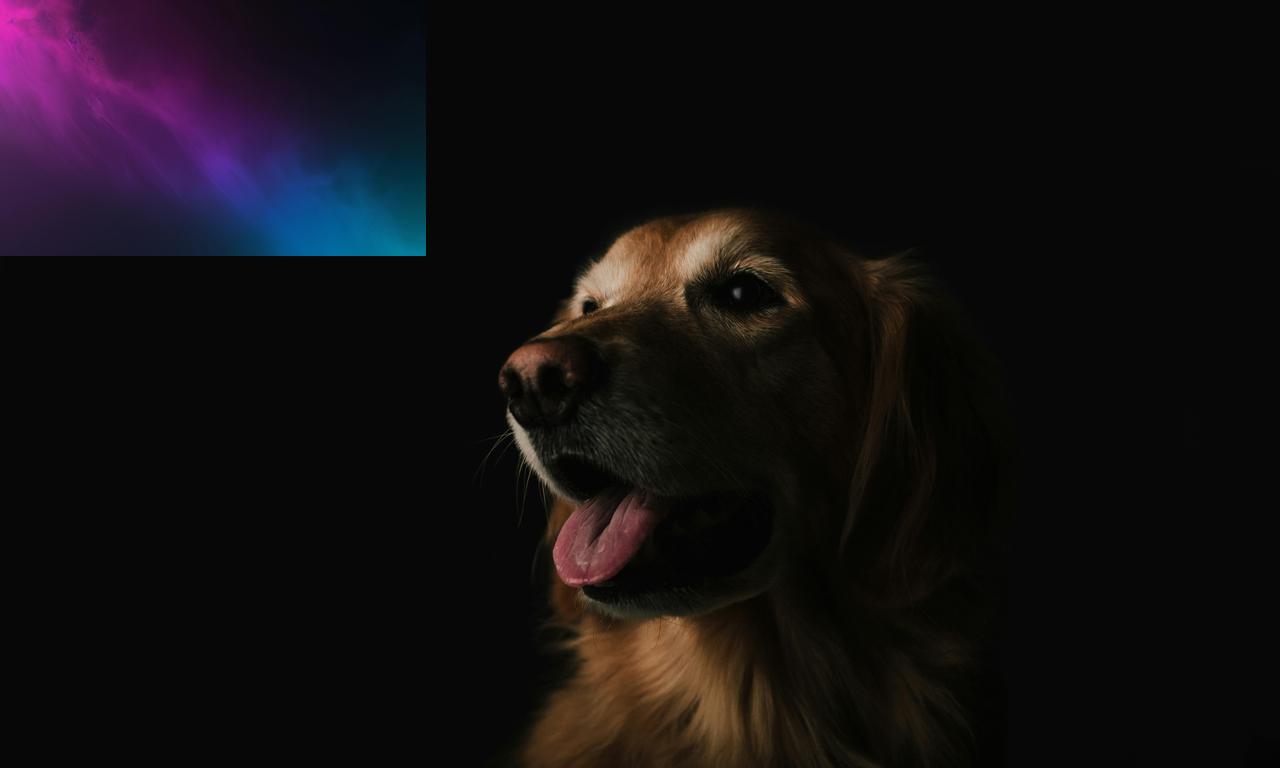}}
    \subfloat{\includegraphics[width=0.24\linewidth]{plots/relighting/additional_paper_plots_rf/output_image_4.jpg}}
    \hspace{0.1em}
    \subfloat{\includegraphics[width=0.24\linewidth]{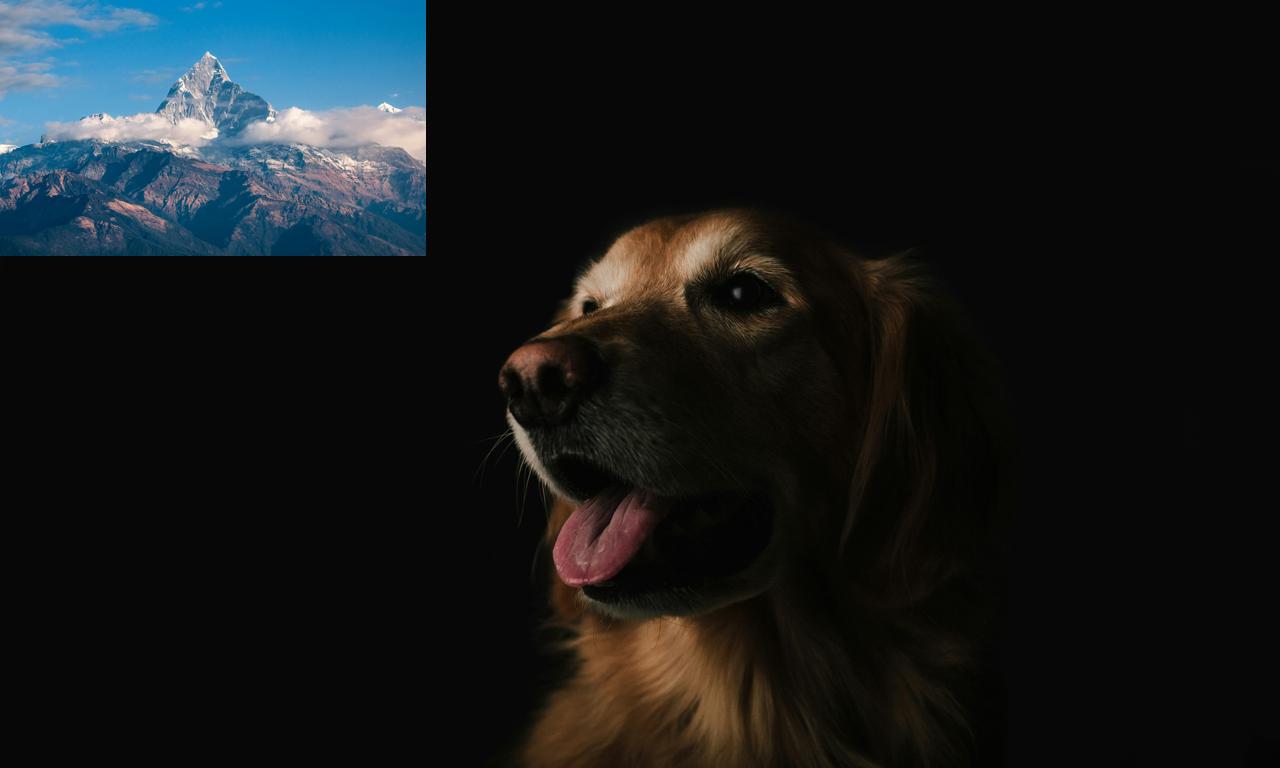}}
    \subfloat{\includegraphics[width=0.24\linewidth]{plots/relighting/additional_paper_plots_rf/output_image_5.jpg}}\\
    \vspace{-0.1em}
    \subfloat{\includegraphics[width=0.24\linewidth]{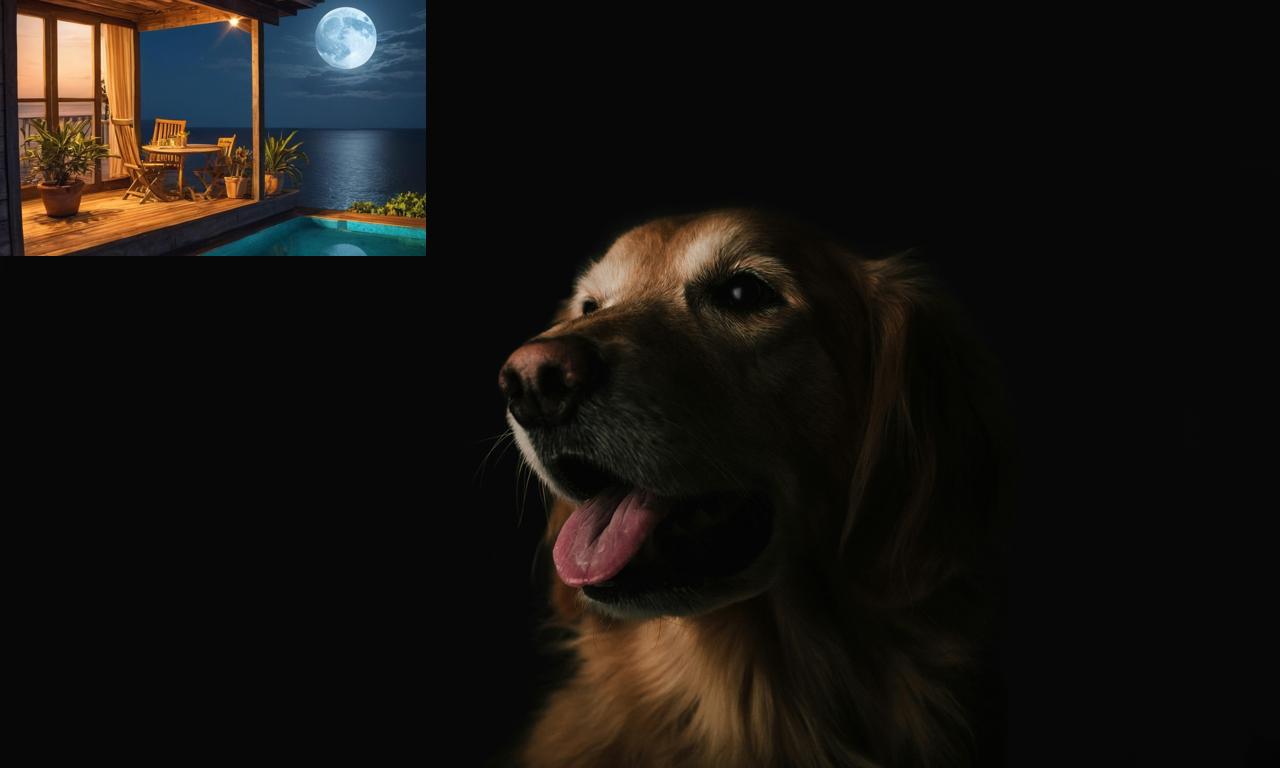}}
    \subfloat{\includegraphics[width=0.24\linewidth]{plots/relighting/additional_paper_plots_rf/output_image_6.jpg}}
    \hspace{0.1em}
    \subfloat{\includegraphics[width=0.24\linewidth]{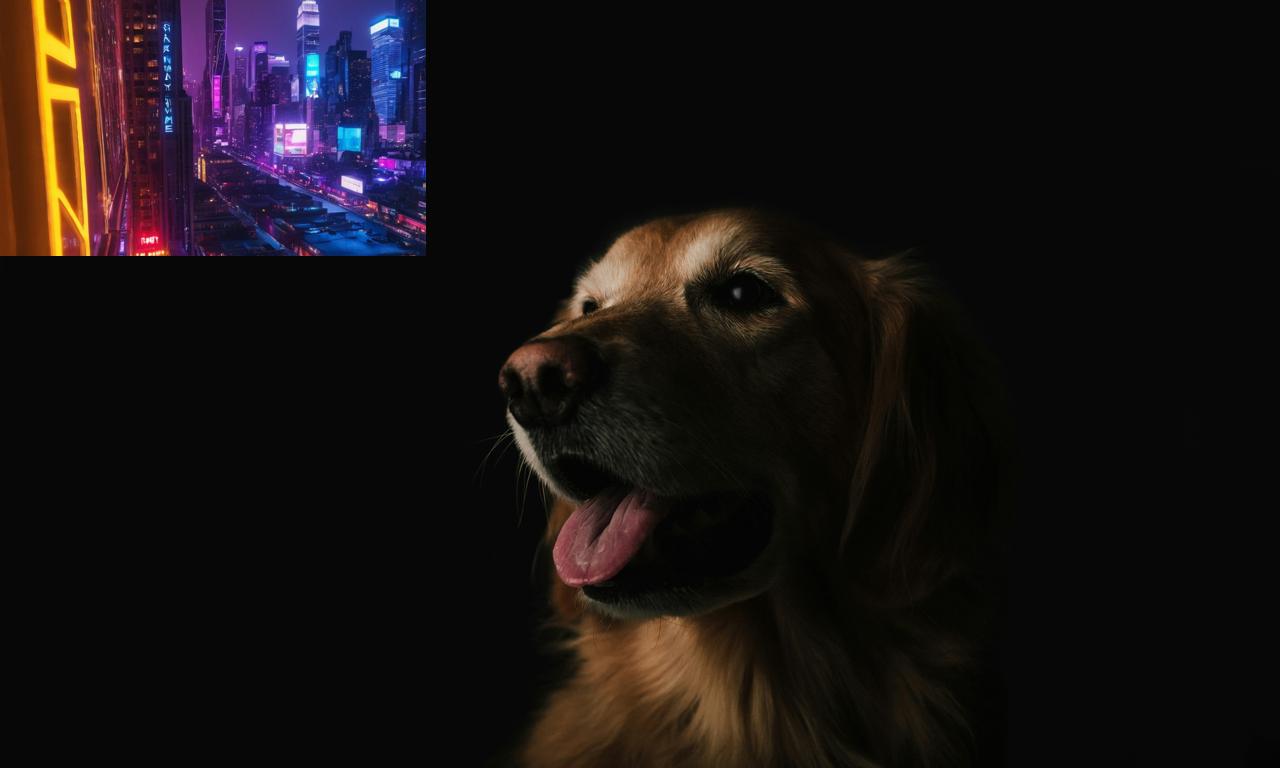}}
    \subfloat{\includegraphics[width=0.24\linewidth]{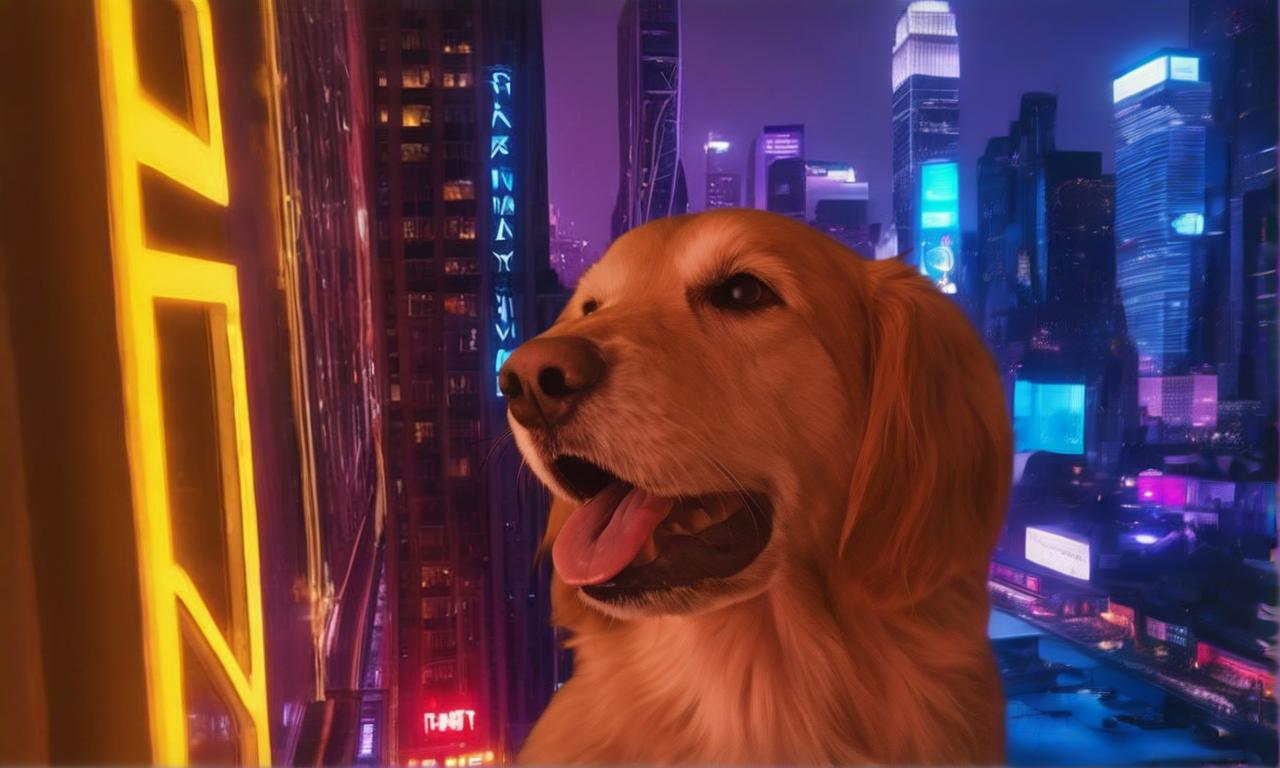}}\\
    \vspace{-0.1em}
    \subfloat{\includegraphics[width=0.24\linewidth]{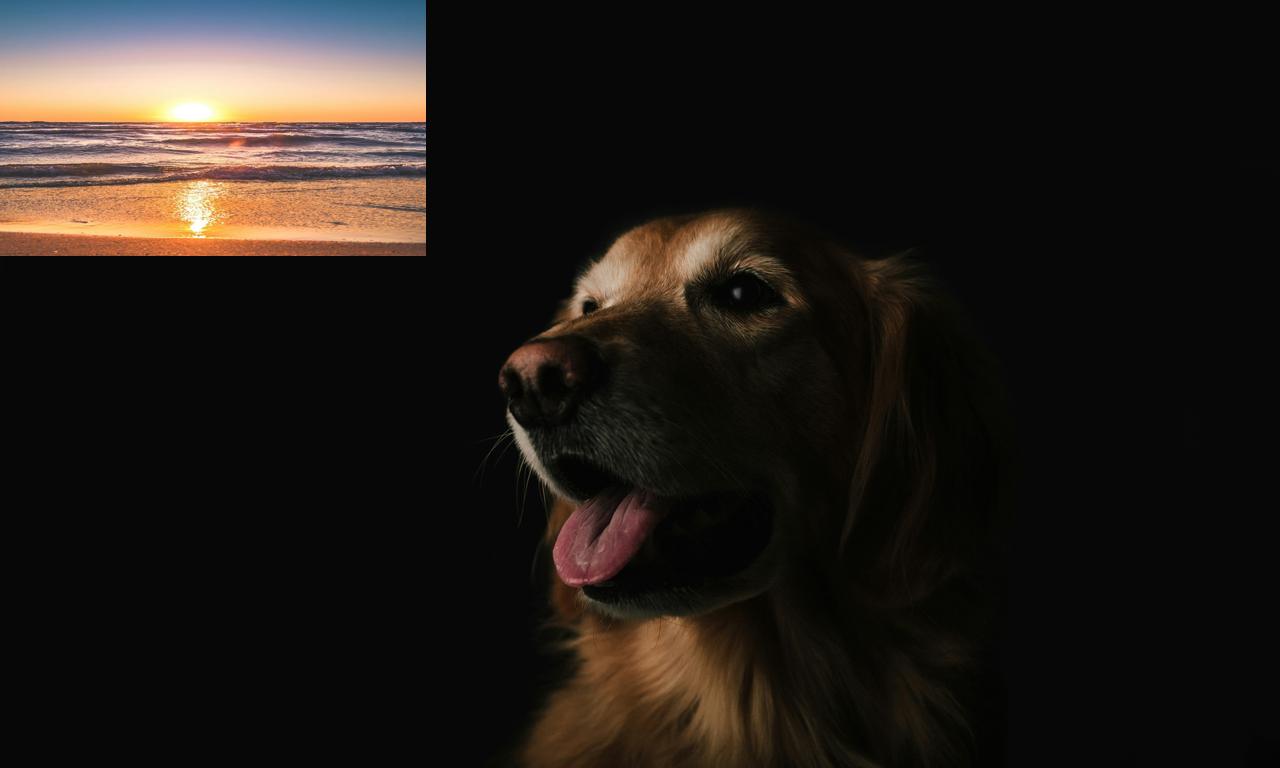}}
    \subfloat{\includegraphics[width=0.24\linewidth]{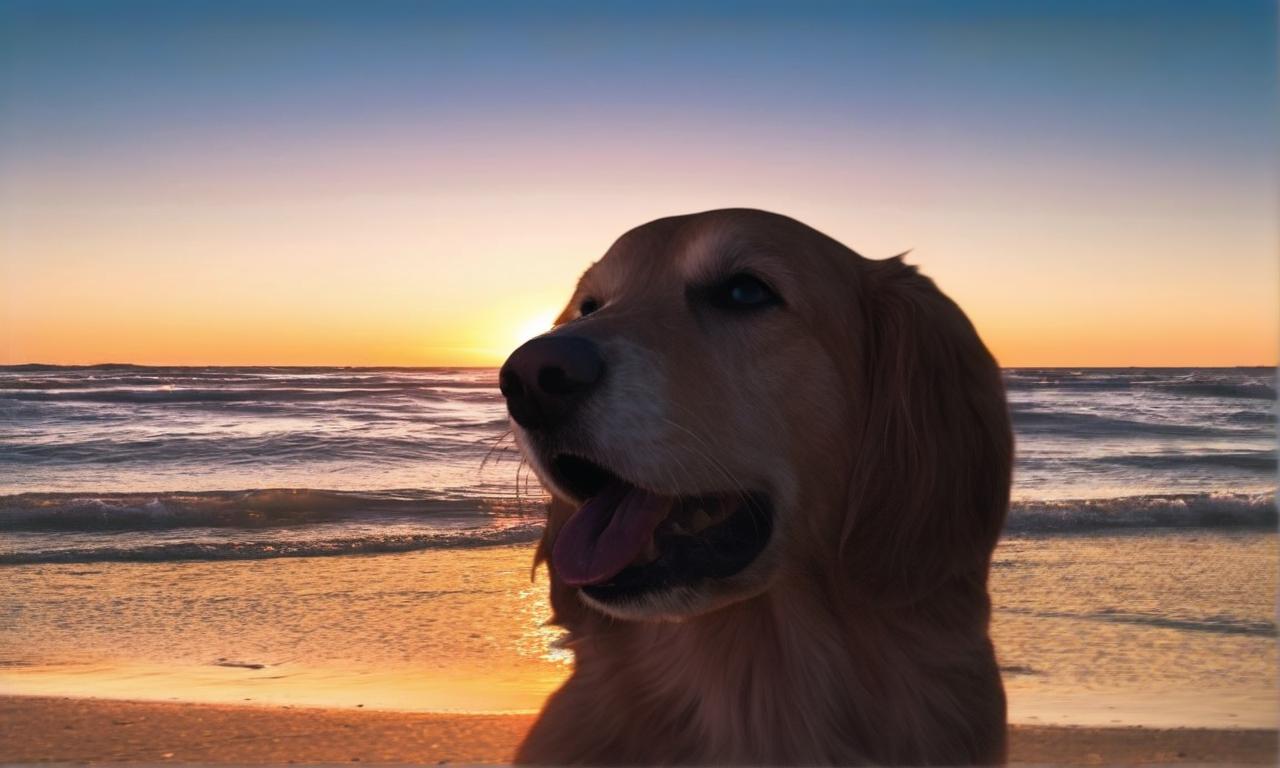}}
    \hspace{0.1em}
    \subfloat{\includegraphics[width=0.24\linewidth]{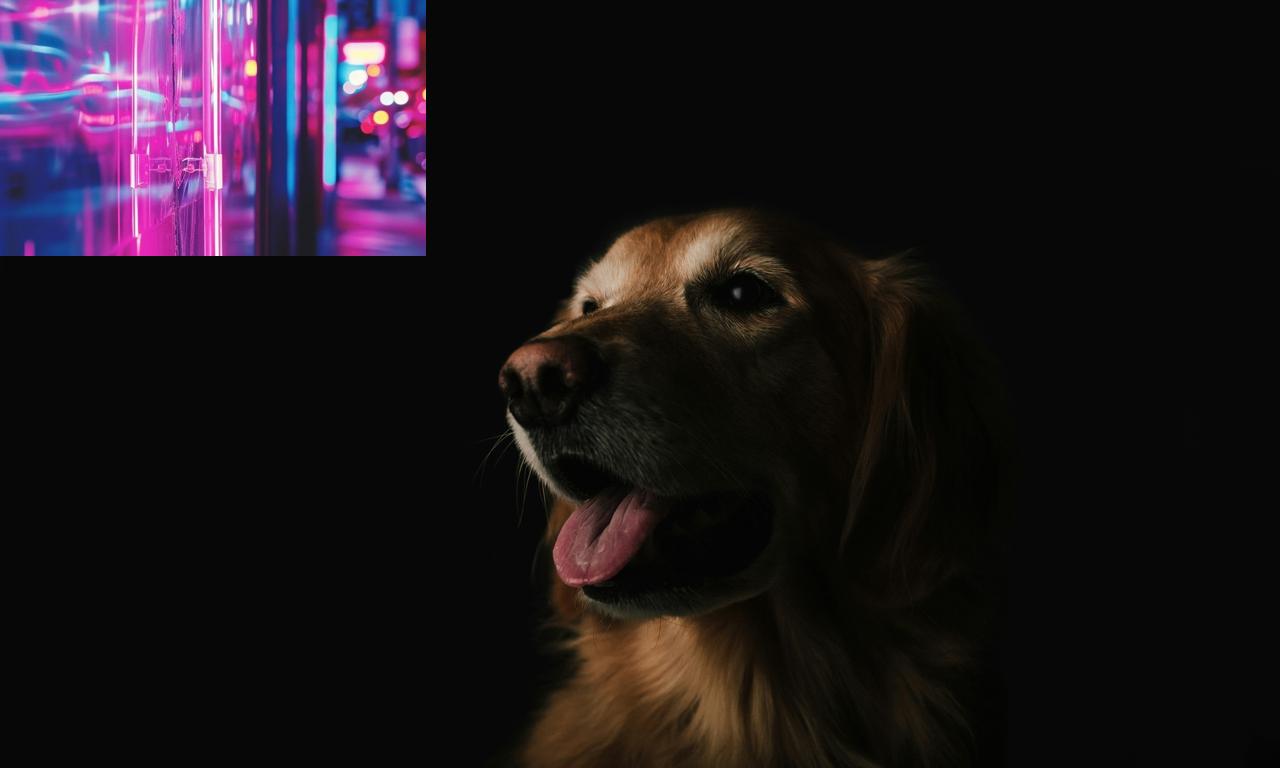}}
    \subfloat{\includegraphics[width=0.24\linewidth]{plots/relighting/additional_paper_plots_rf/output_image_9.jpg}}

    \caption{Qualitative results for object relighting. The model is able to relight the object according to the provided background and also remove existing shadows and reflections.}
    \label{fig:app_relighting_results_bis}
\end{figure*}

\begin{figure*}[ht]
    \captionsetup[subfigure]{position=above, labelformat = empty}
    \centering
    \subfloat[Composite]{\includegraphics[width=0.2\linewidth]{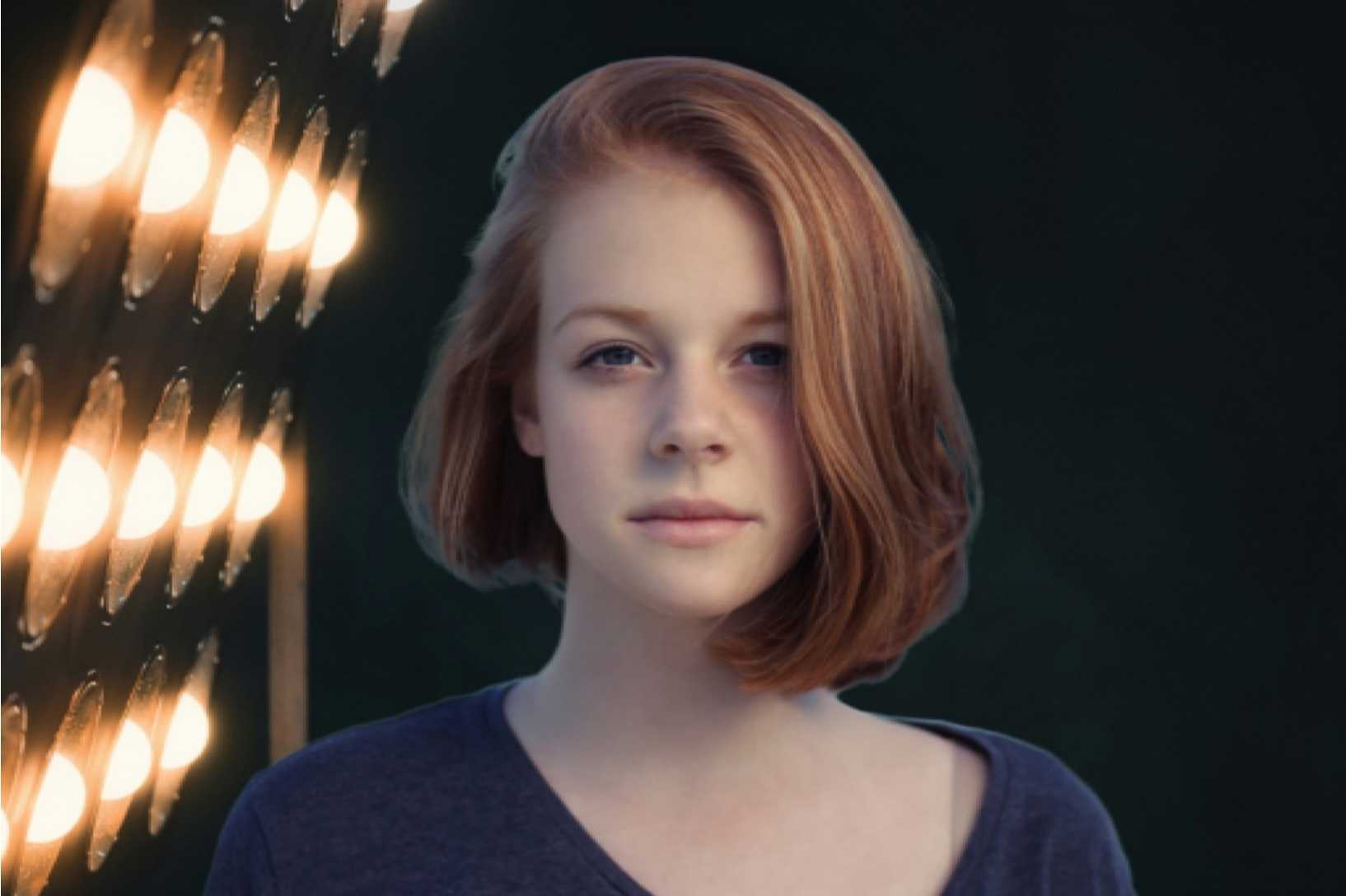}}
    \subfloat[Ours (1 NFE)]{\includegraphics[width=0.2\linewidth]{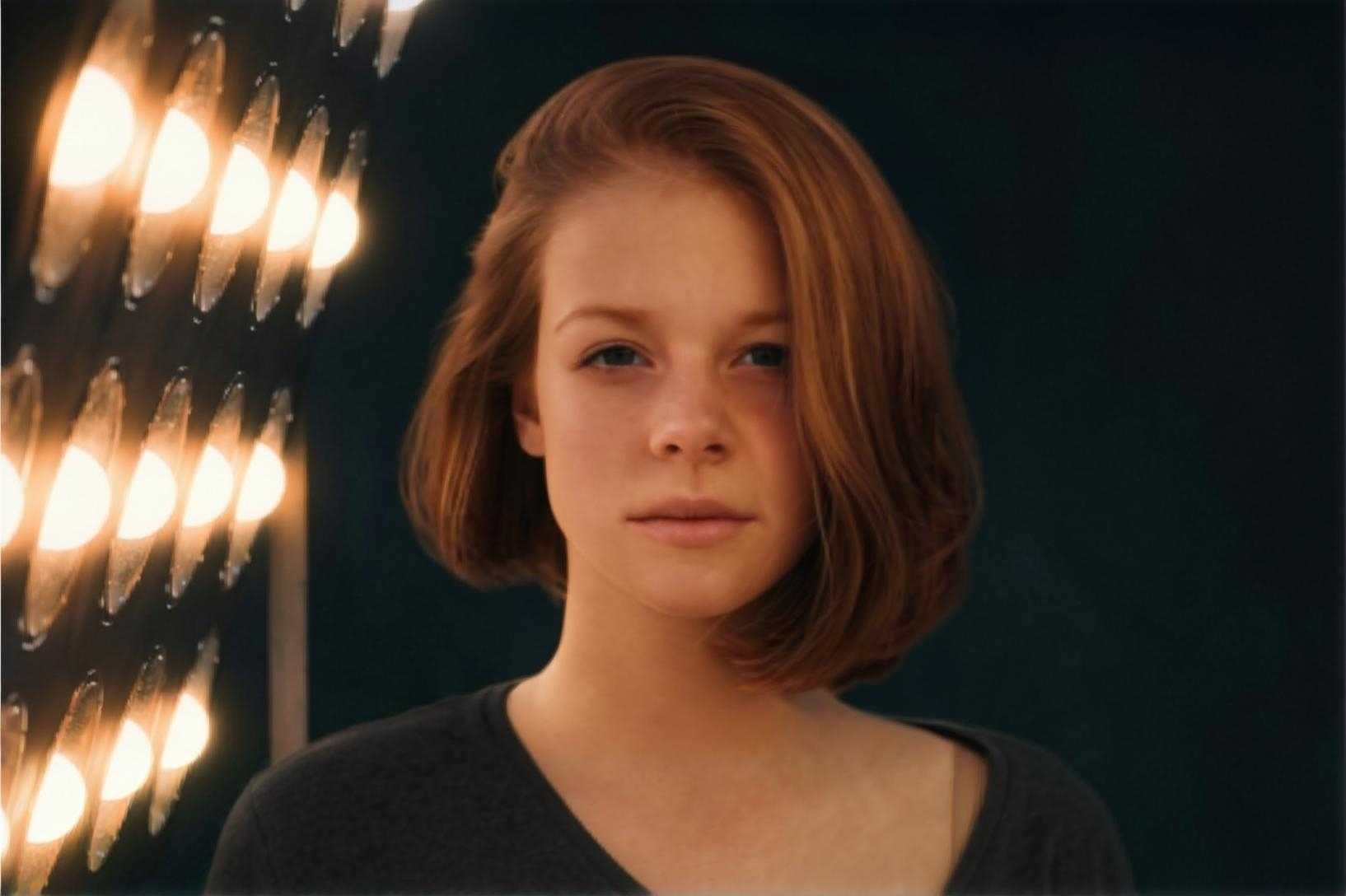}} \hspace{0.1em}
    \subfloat[Composite]{\includegraphics[width=0.2\linewidth]{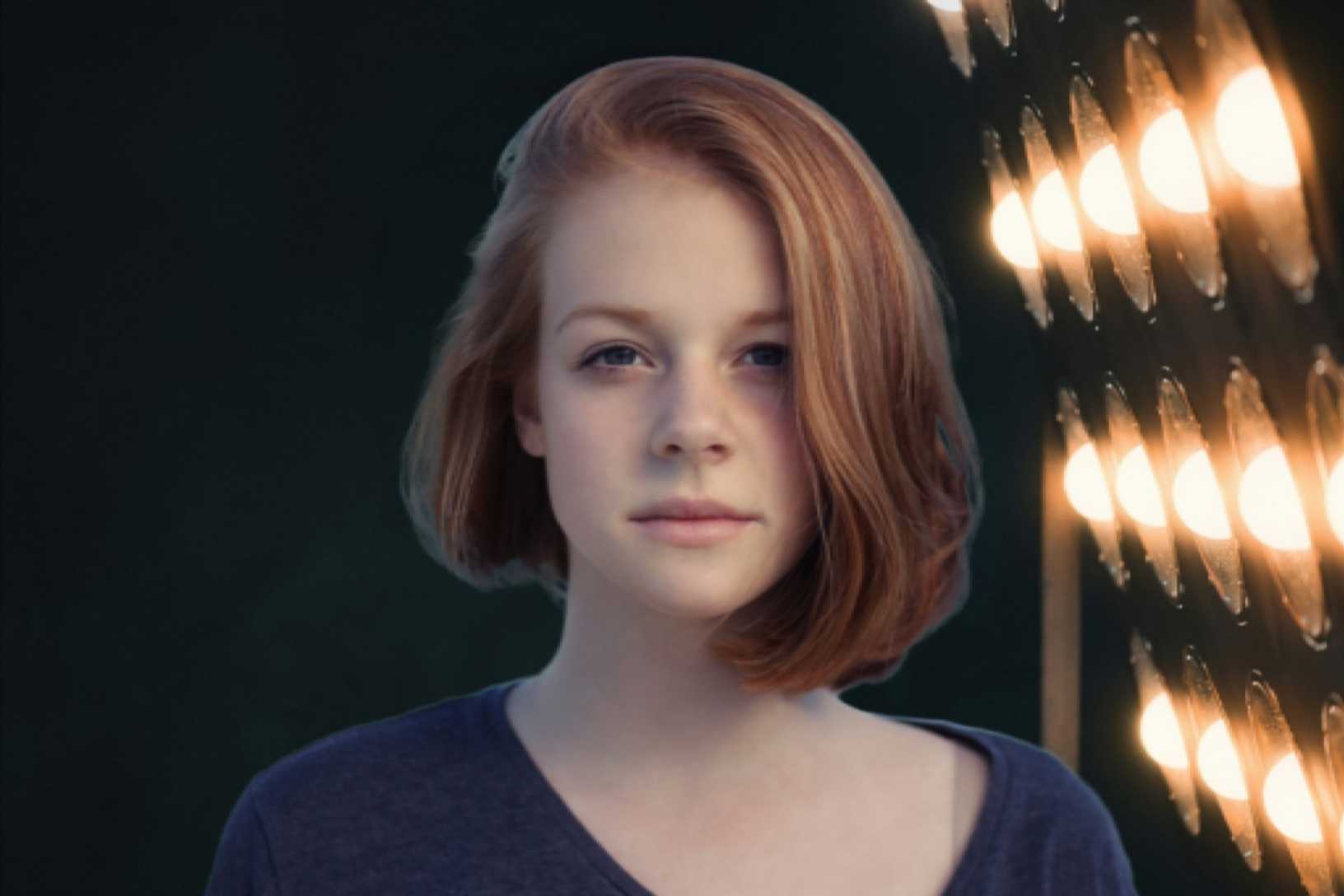}}
    \subfloat[Ours (1 NFE)]{\includegraphics[width=0.2\linewidth]{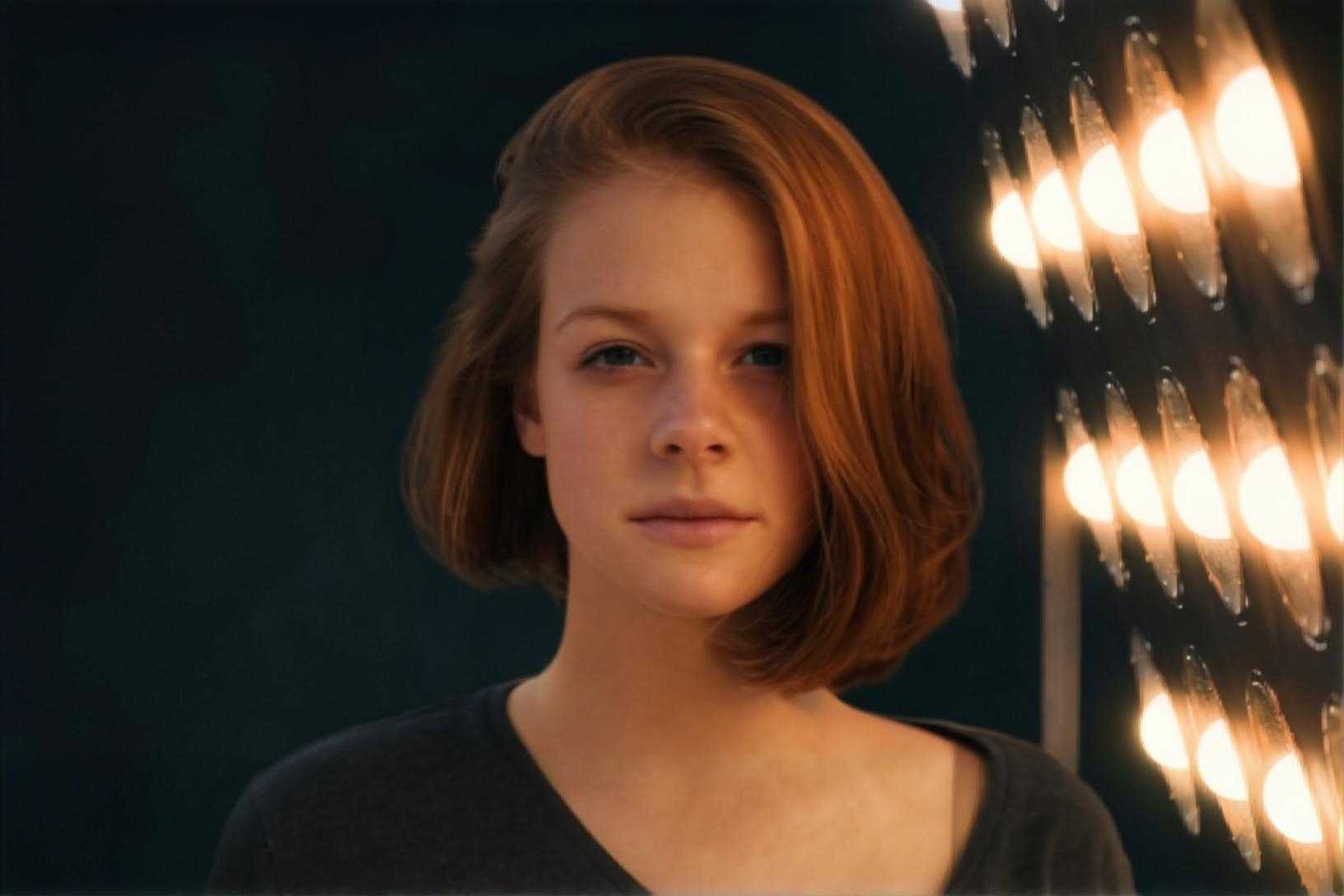}}\\
    \vspace{-0.1em}
    \subfloat{\includegraphics[width=0.2\linewidth]{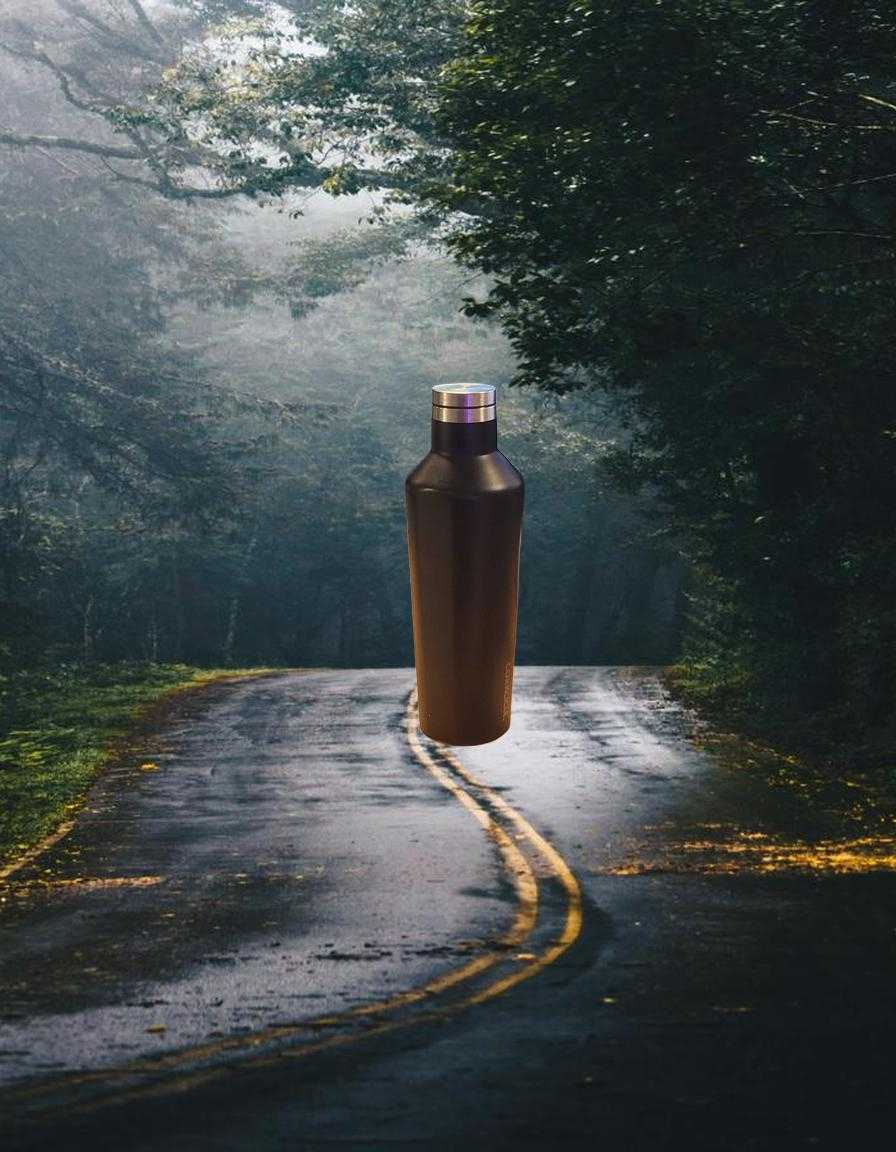}}
    \subfloat{\includegraphics[width=0.2\linewidth]{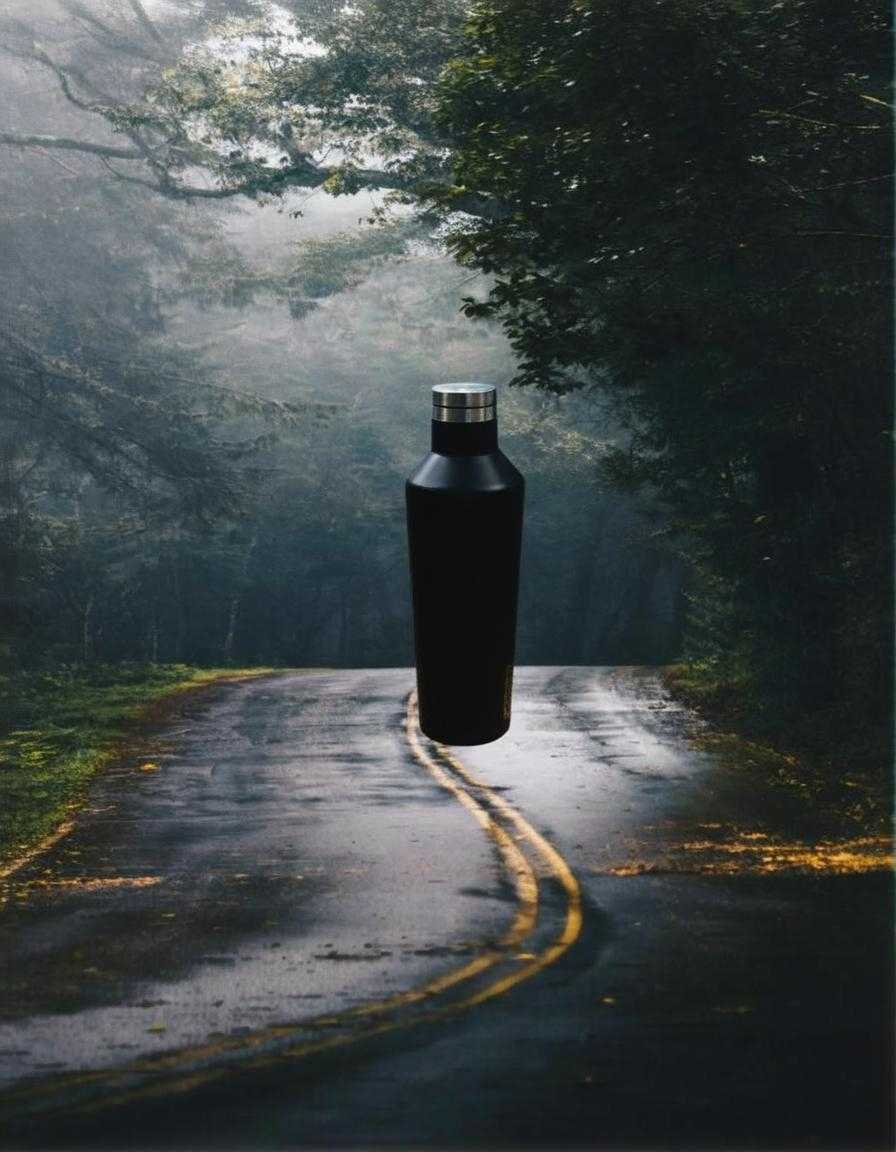}} \hspace{0.1em}
    \subfloat{\includegraphics[width=0.2\linewidth]{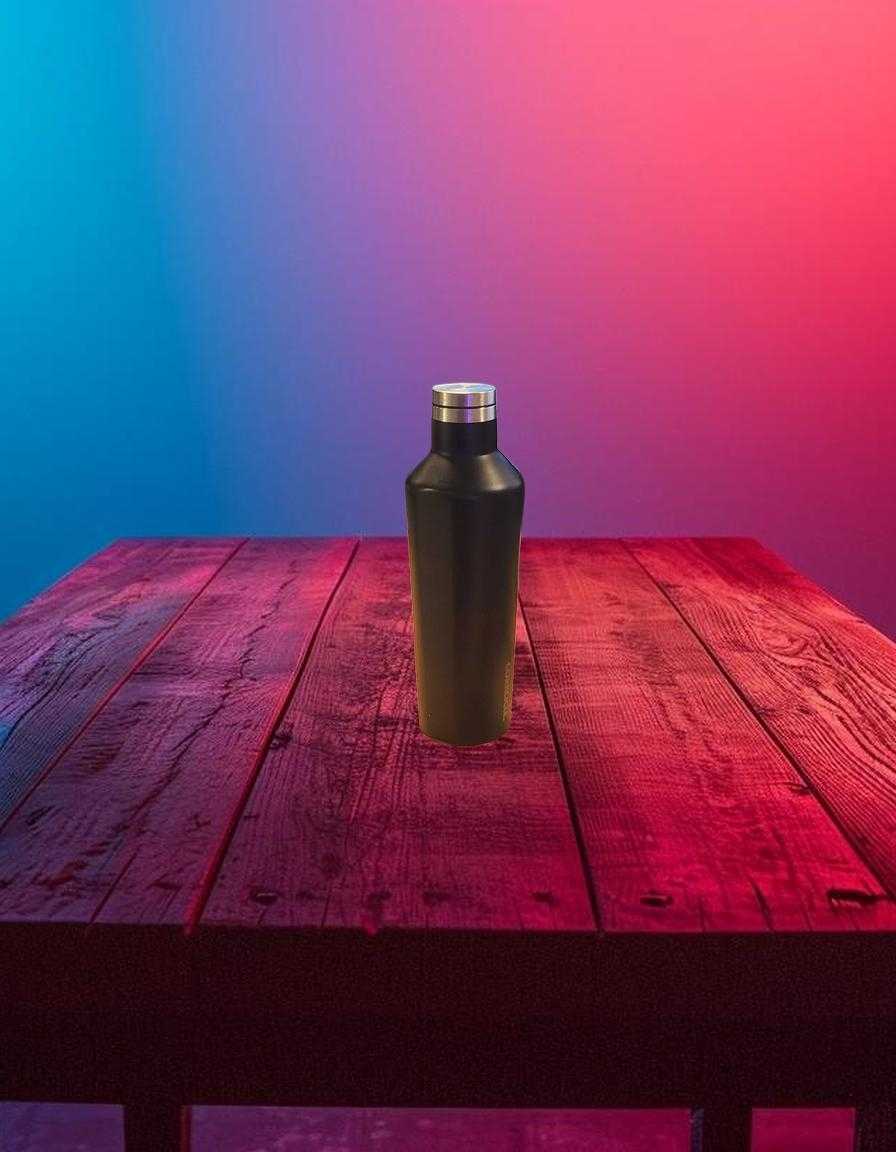}}
    \subfloat{\includegraphics[width=0.2\linewidth]{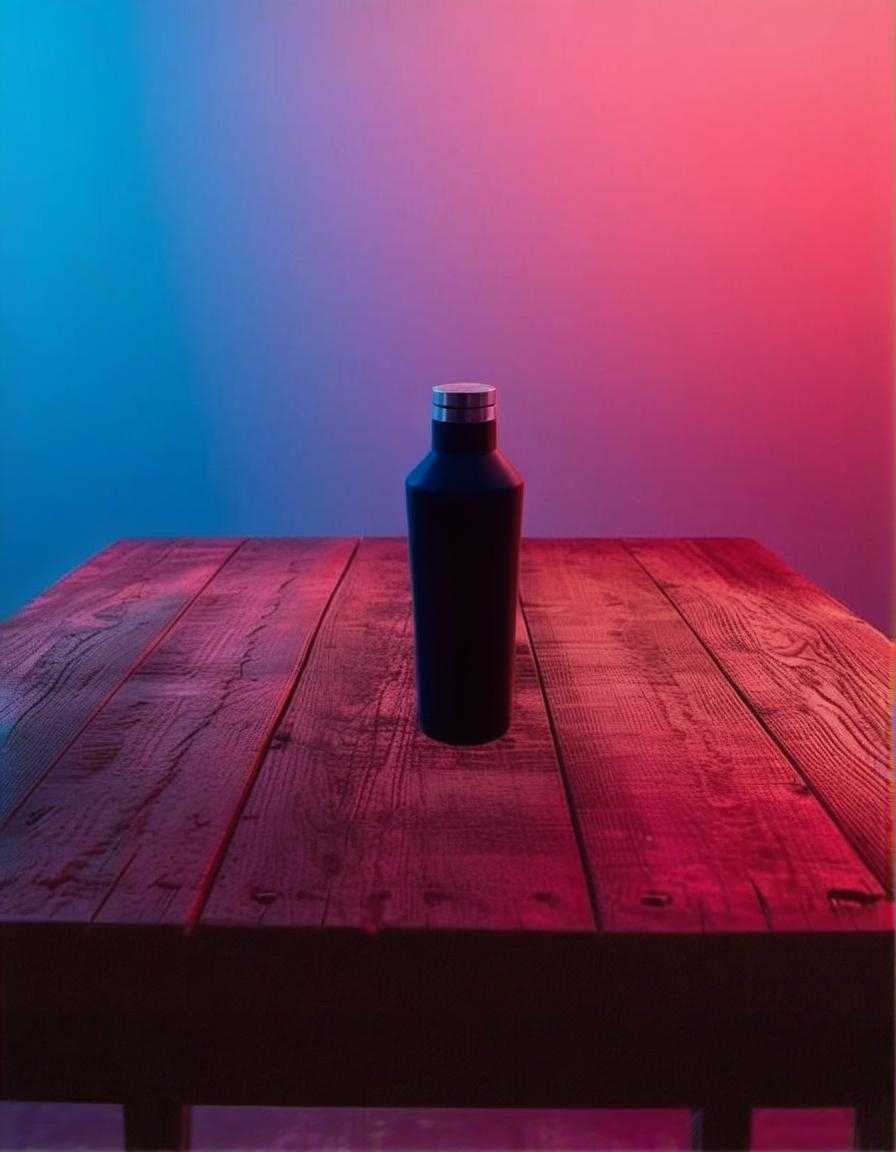}}\\
    \vspace{-0.1em}
    \subfloat{\includegraphics[width=0.2\linewidth]{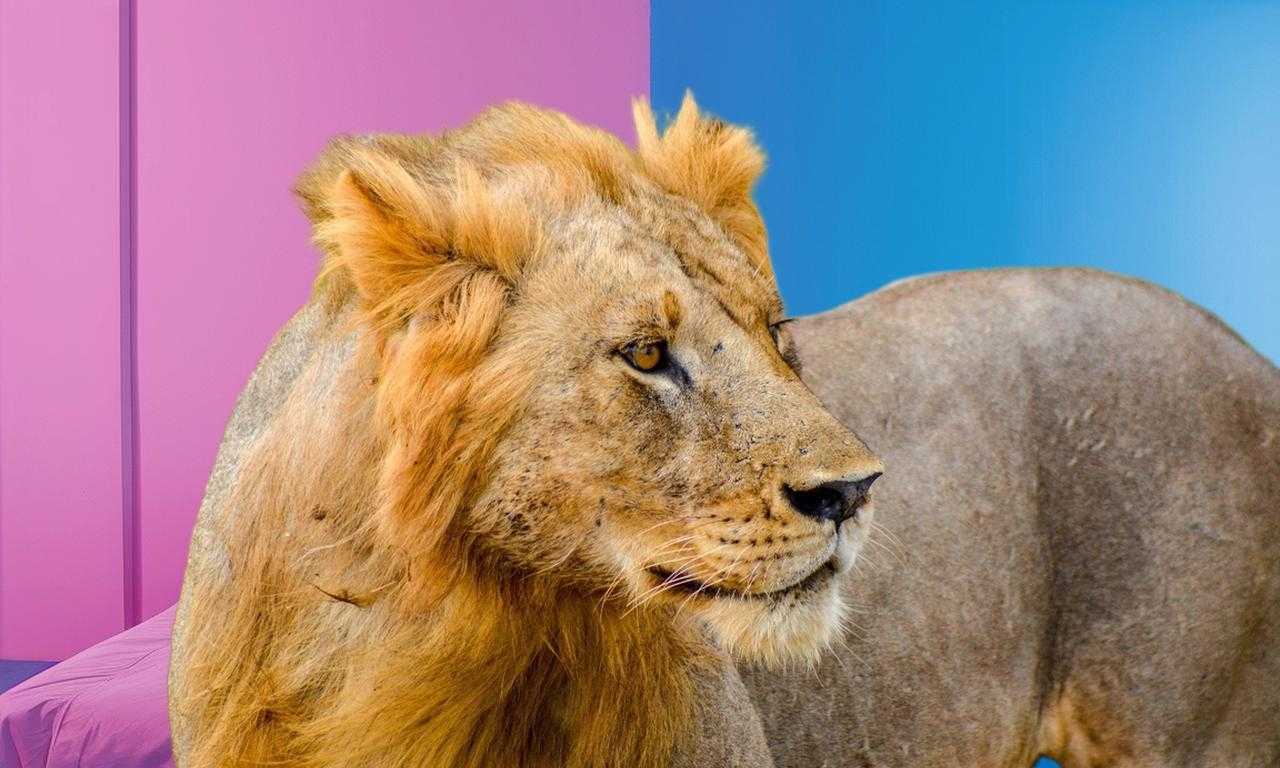}}    \subfloat{\includegraphics[width=0.2\linewidth]{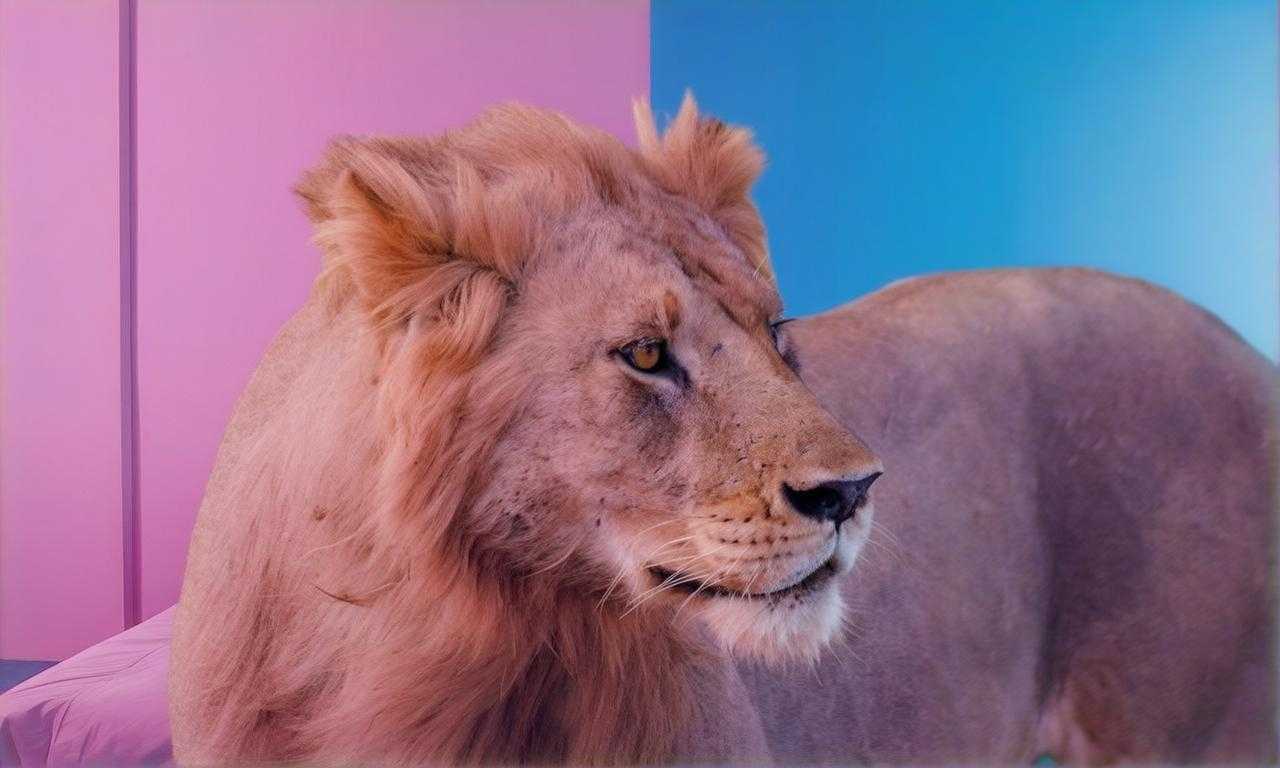}}
    \hspace{0.1em}
    \subfloat{\includegraphics[width=0.2\linewidth]{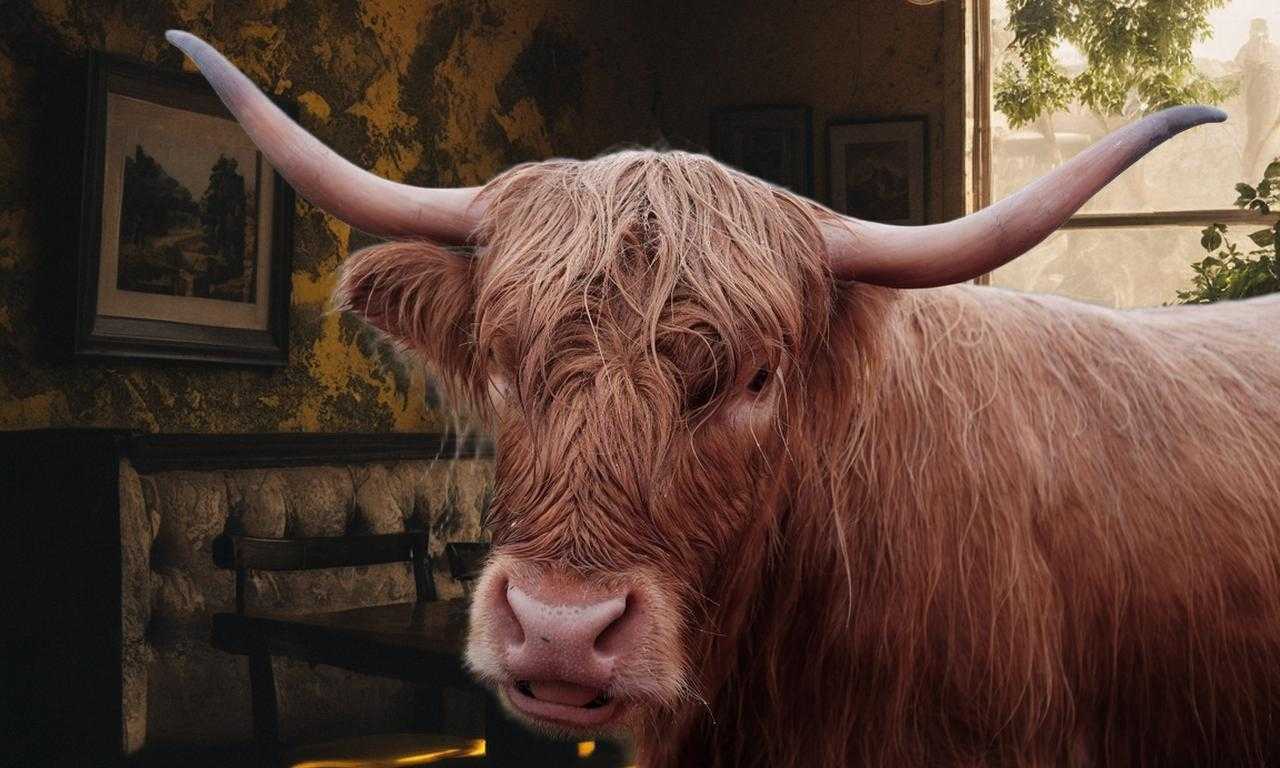}}
    \subfloat{\includegraphics[width=0.2\linewidth]{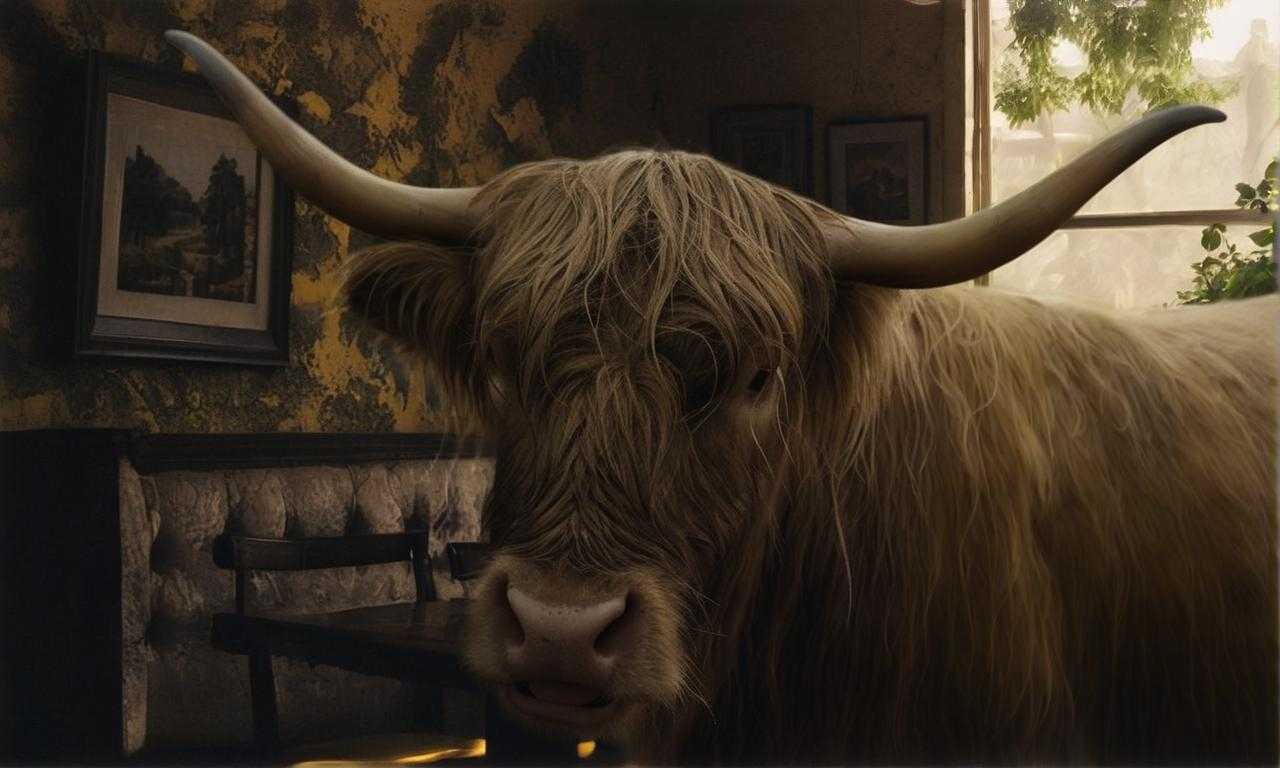}}\\
    \vspace{-0.1em}
    \subfloat{\includegraphics[width=0.2\linewidth]{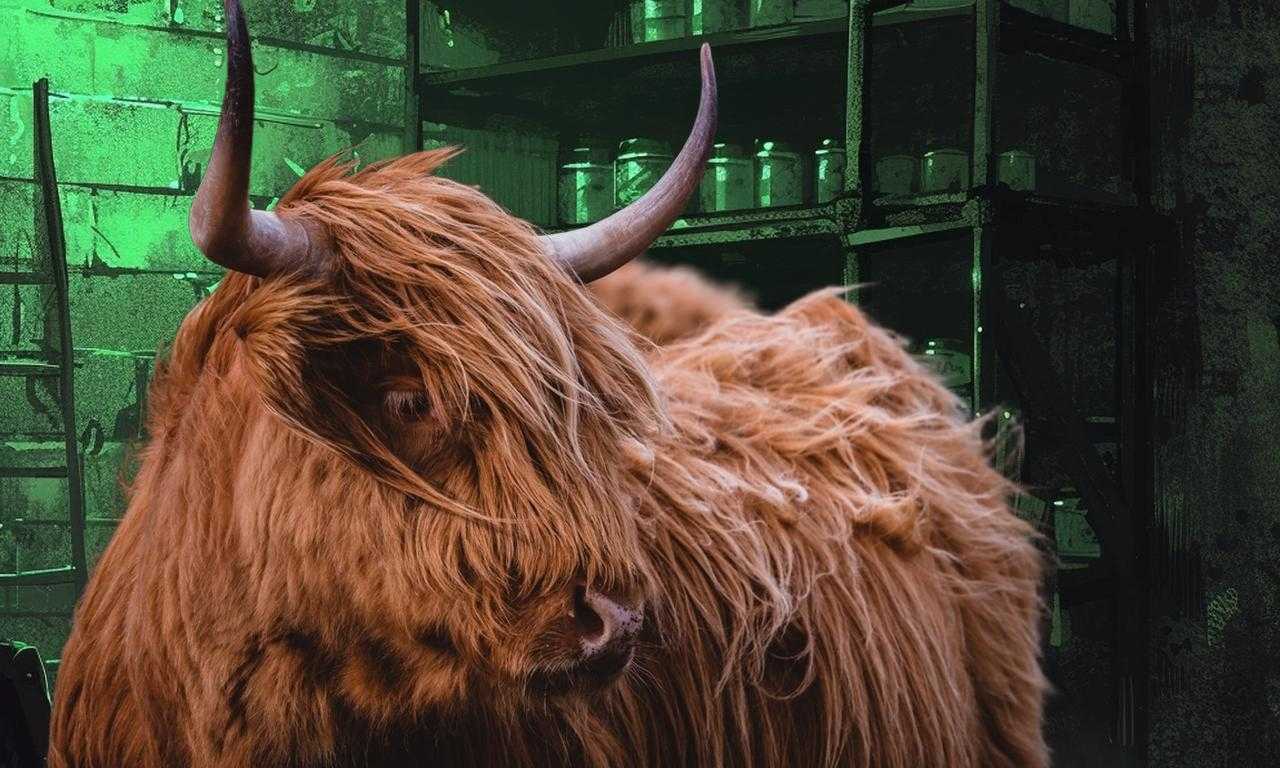}}
    \subfloat{\includegraphics[width=0.2\linewidth]{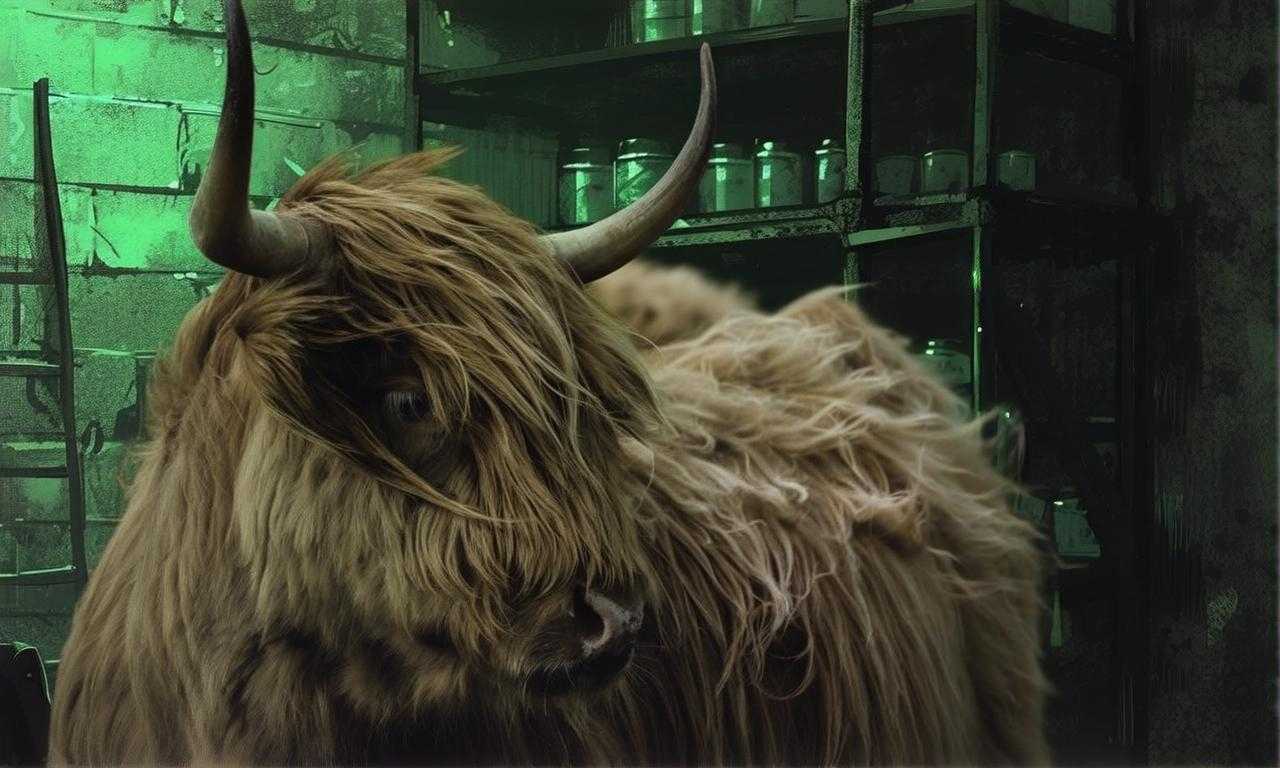}} \hspace{0.1em}
    \subfloat{\includegraphics[width=0.2\linewidth]{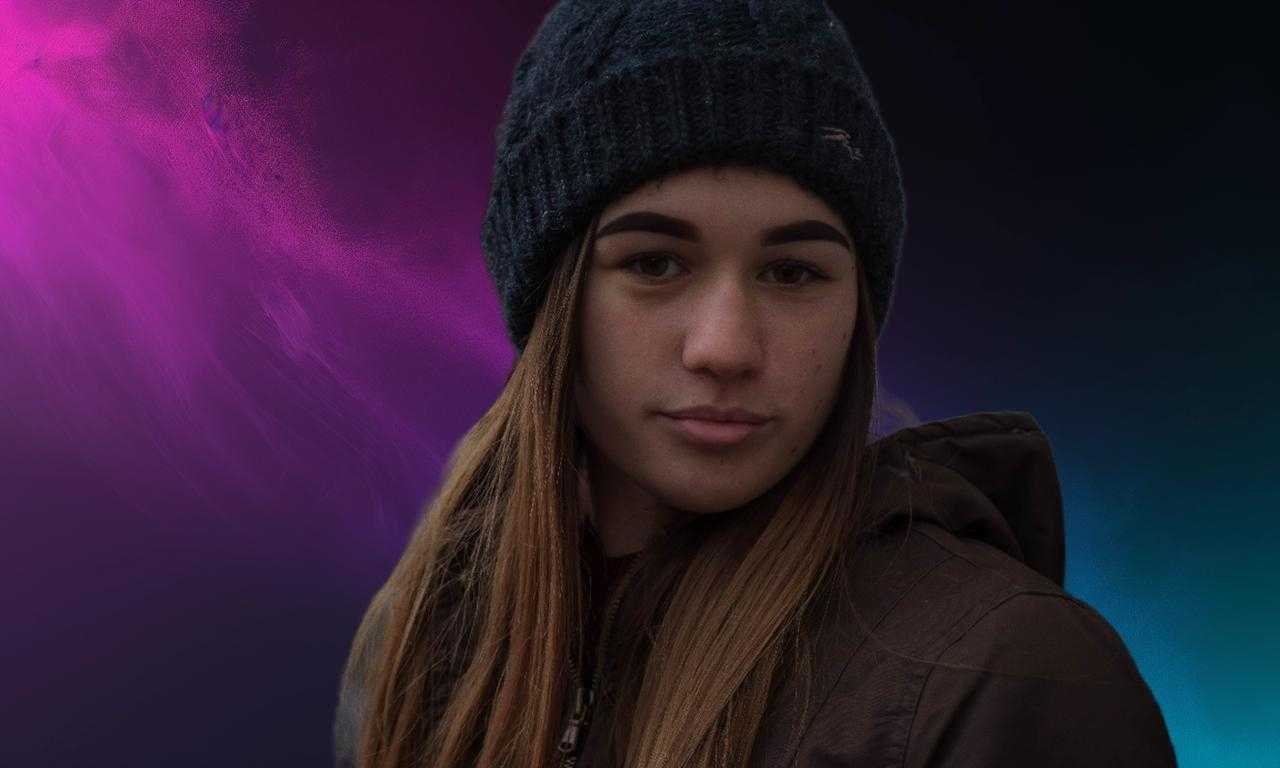}}
    \subfloat{\includegraphics[width=0.2\linewidth]{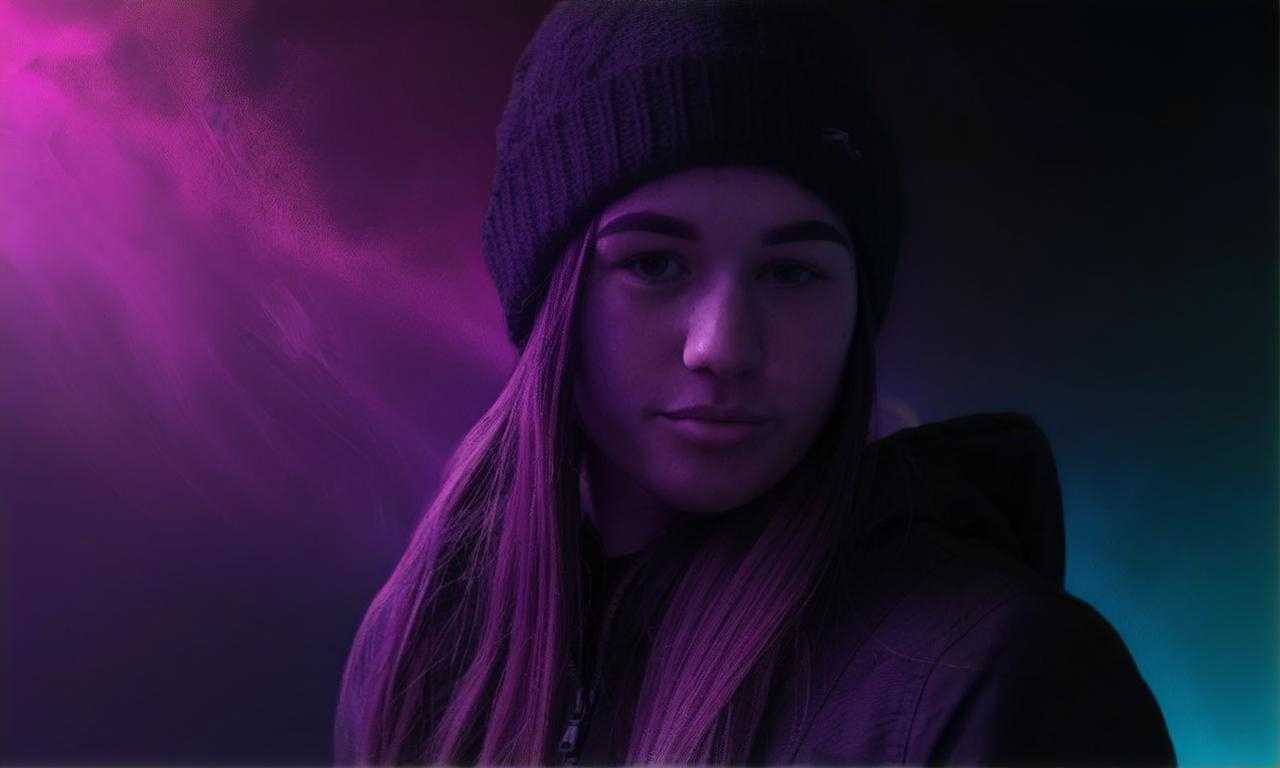}}\\
    \vspace{-0.1em}
    \subfloat{\includegraphics[width=0.2\linewidth]{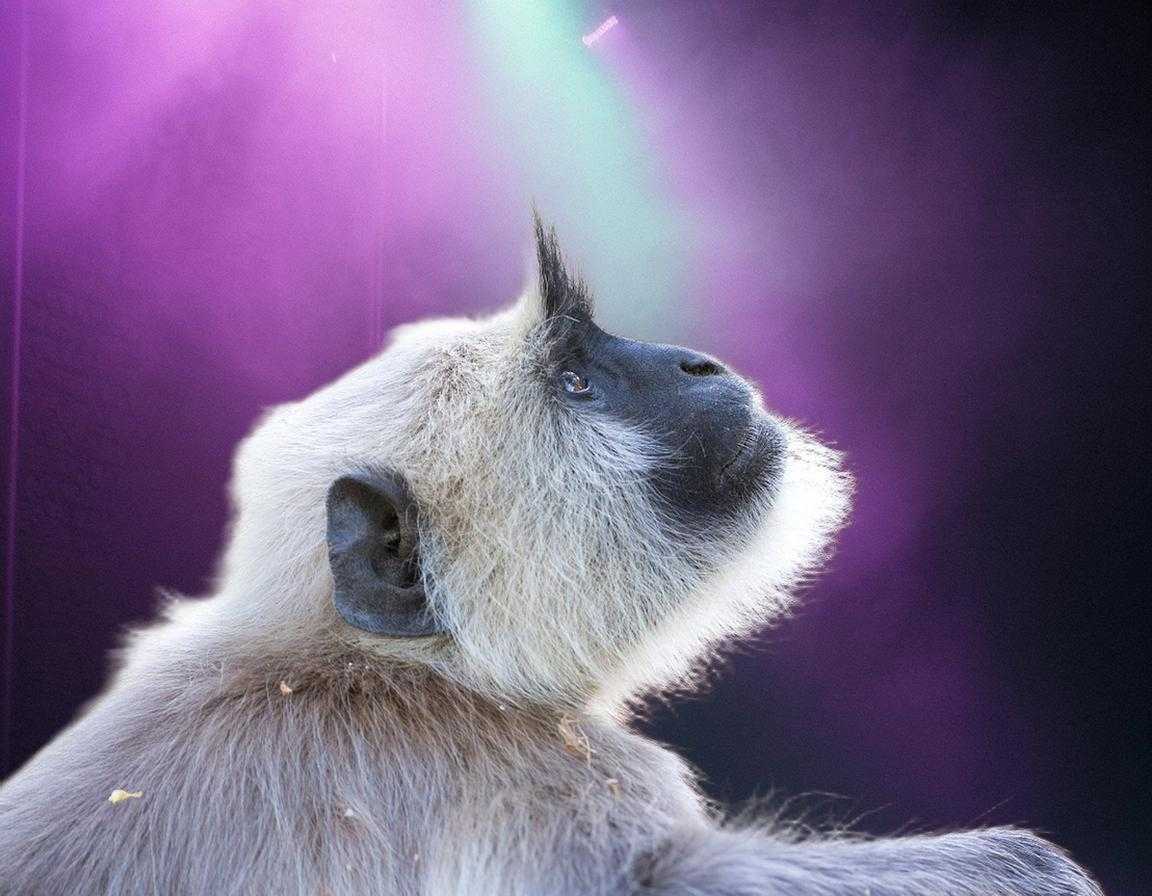}}
    \subfloat{\includegraphics[width=0.2\linewidth]{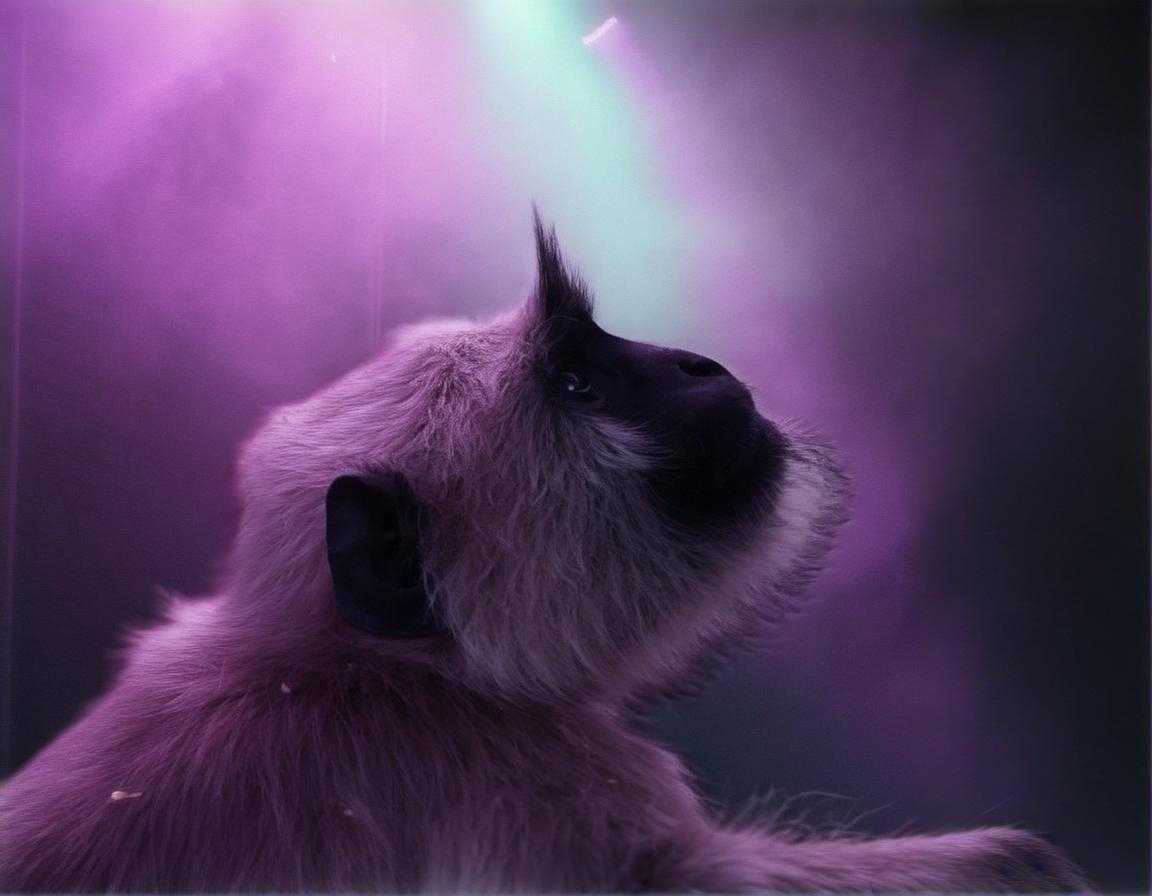}} \hspace{0.1em}
    \subfloat{\includegraphics[width=0.2\linewidth]{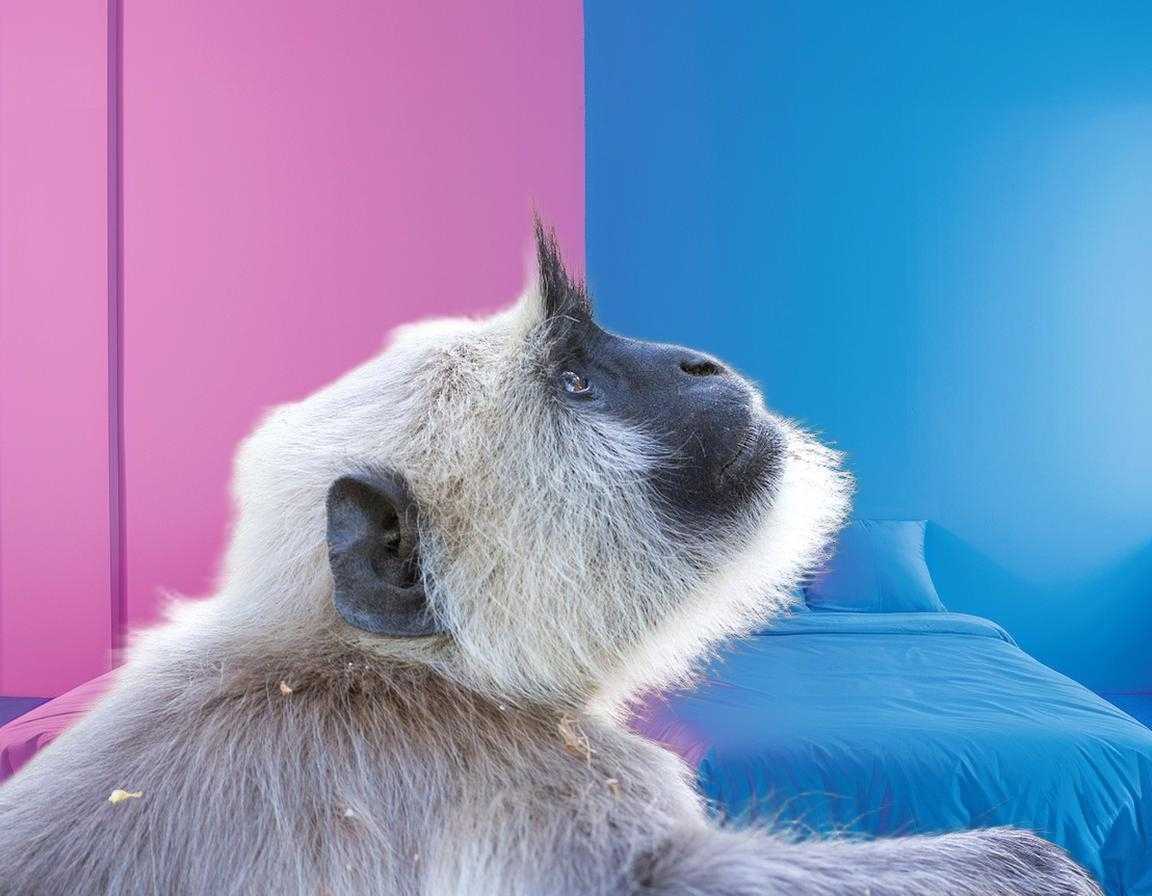}}
    \subfloat{\includegraphics[width=0.2\linewidth]{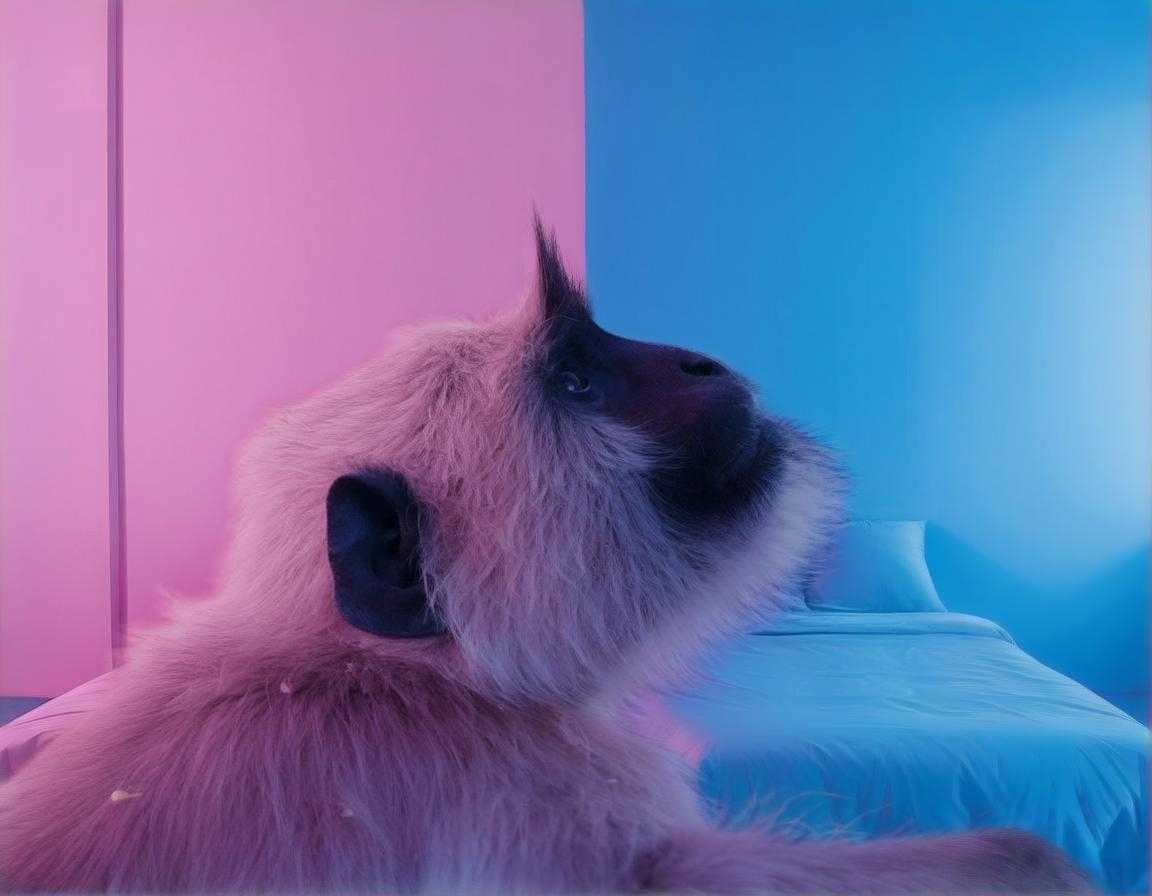}}\\
    \vspace{-0.1em}
    \subfloat{\includegraphics[width=0.2\linewidth]{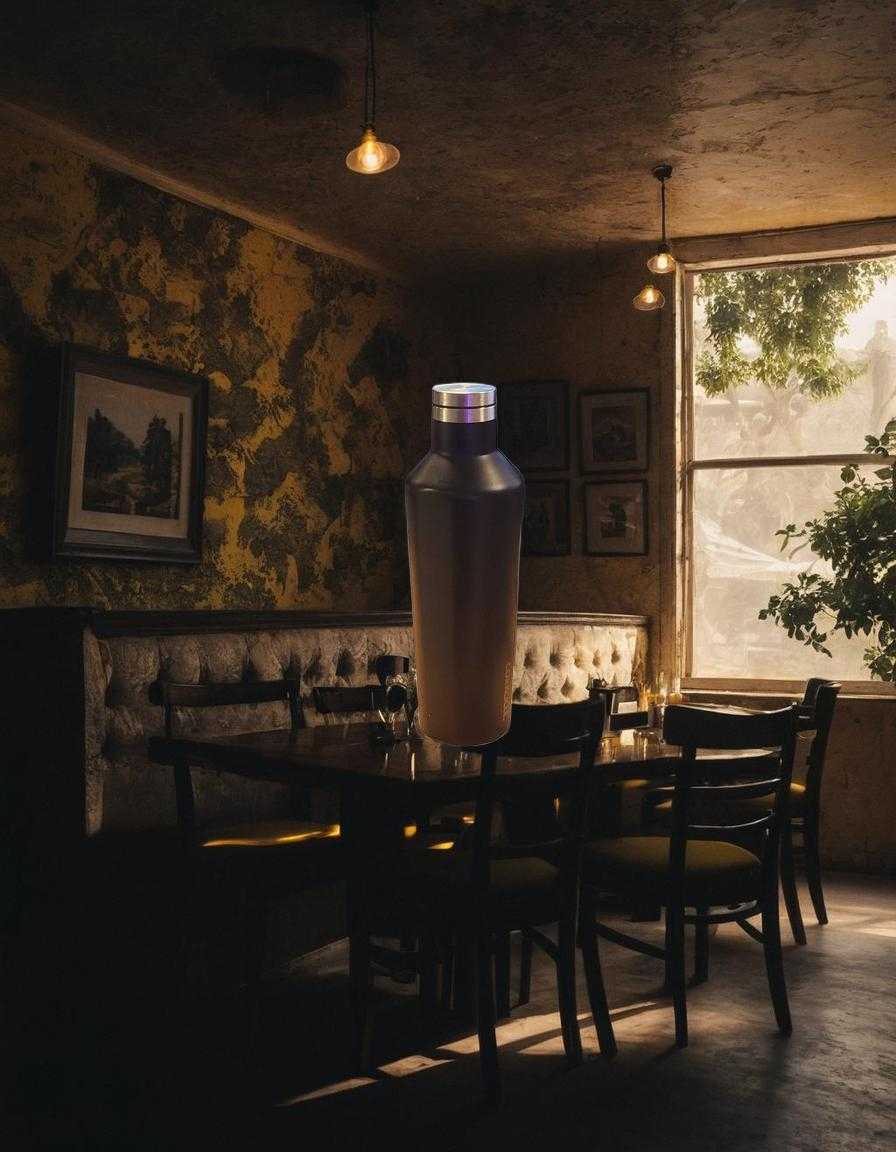}}
    \subfloat{\includegraphics[width=0.2\linewidth]{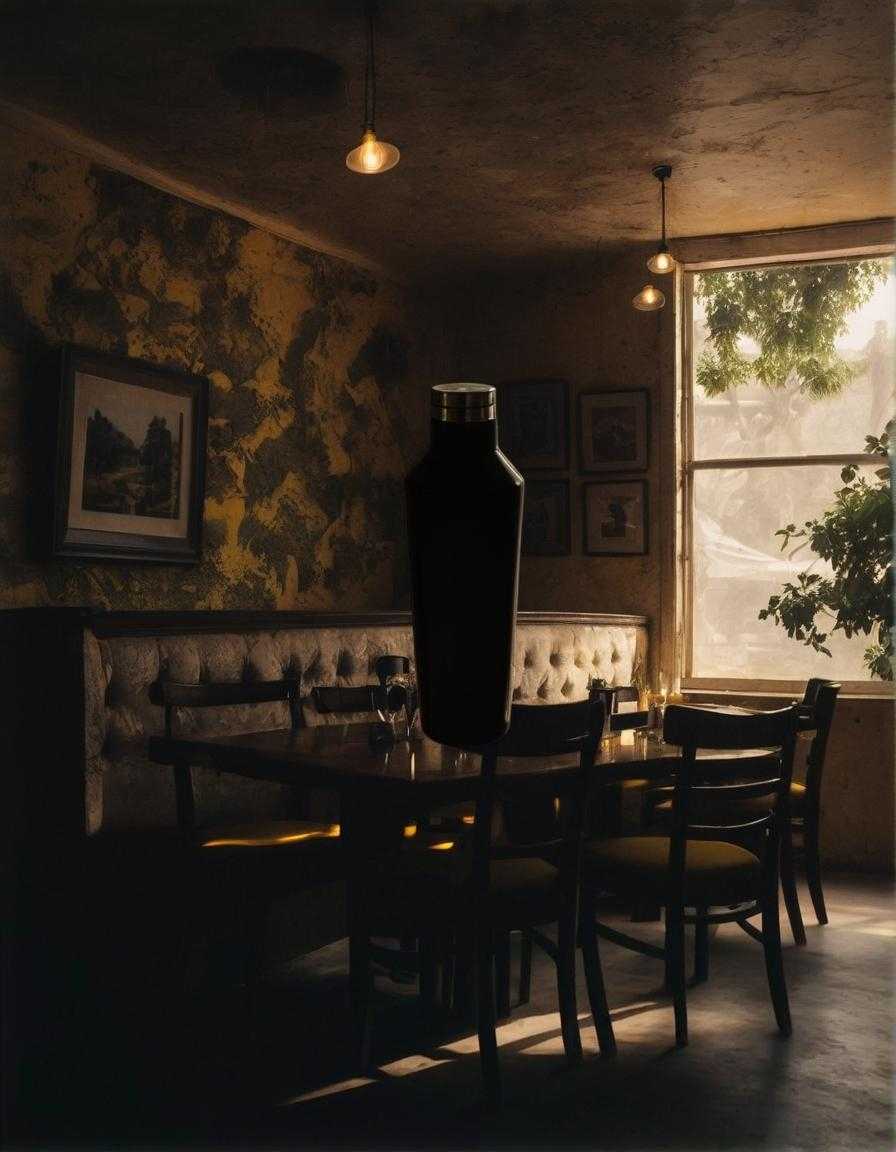}} \hspace{0.1em}
    \subfloat{\includegraphics[width=0.2\linewidth]{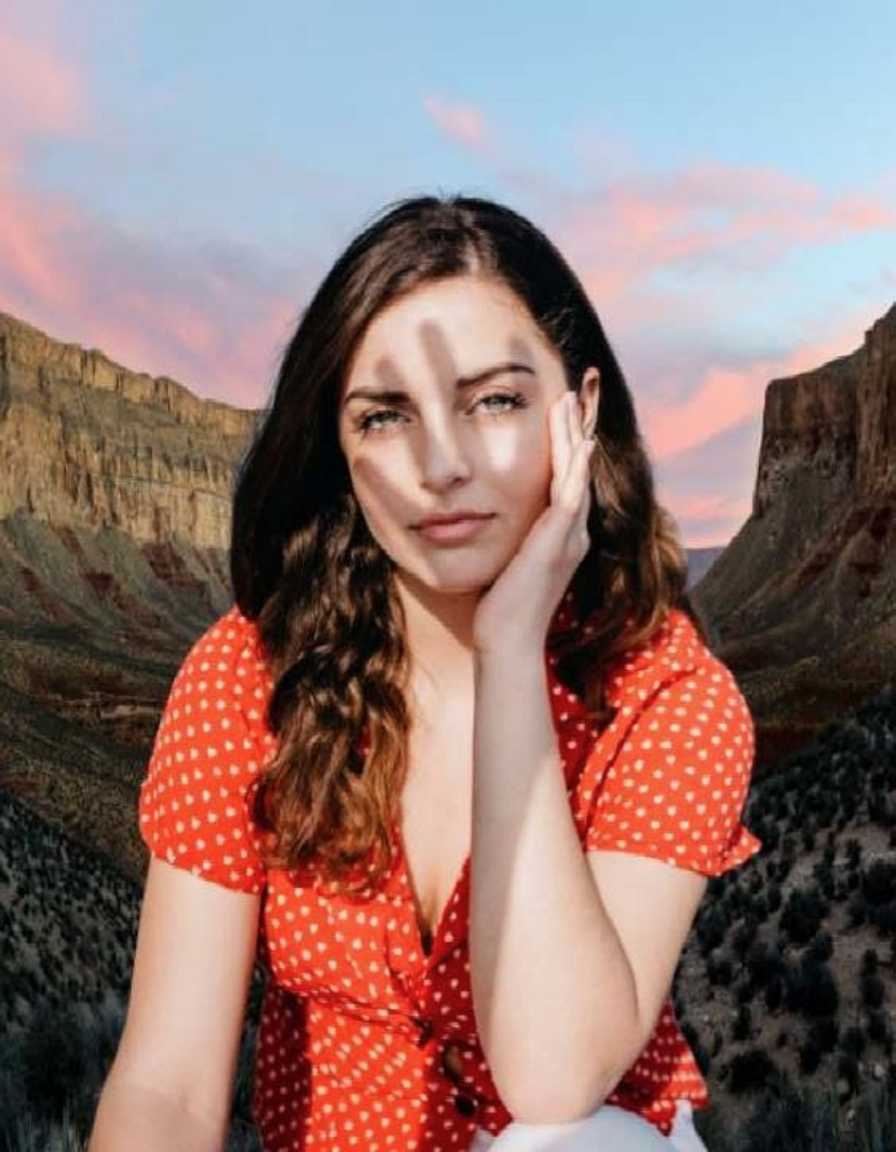}}
    \subfloat{\includegraphics[width=0.2\linewidth]{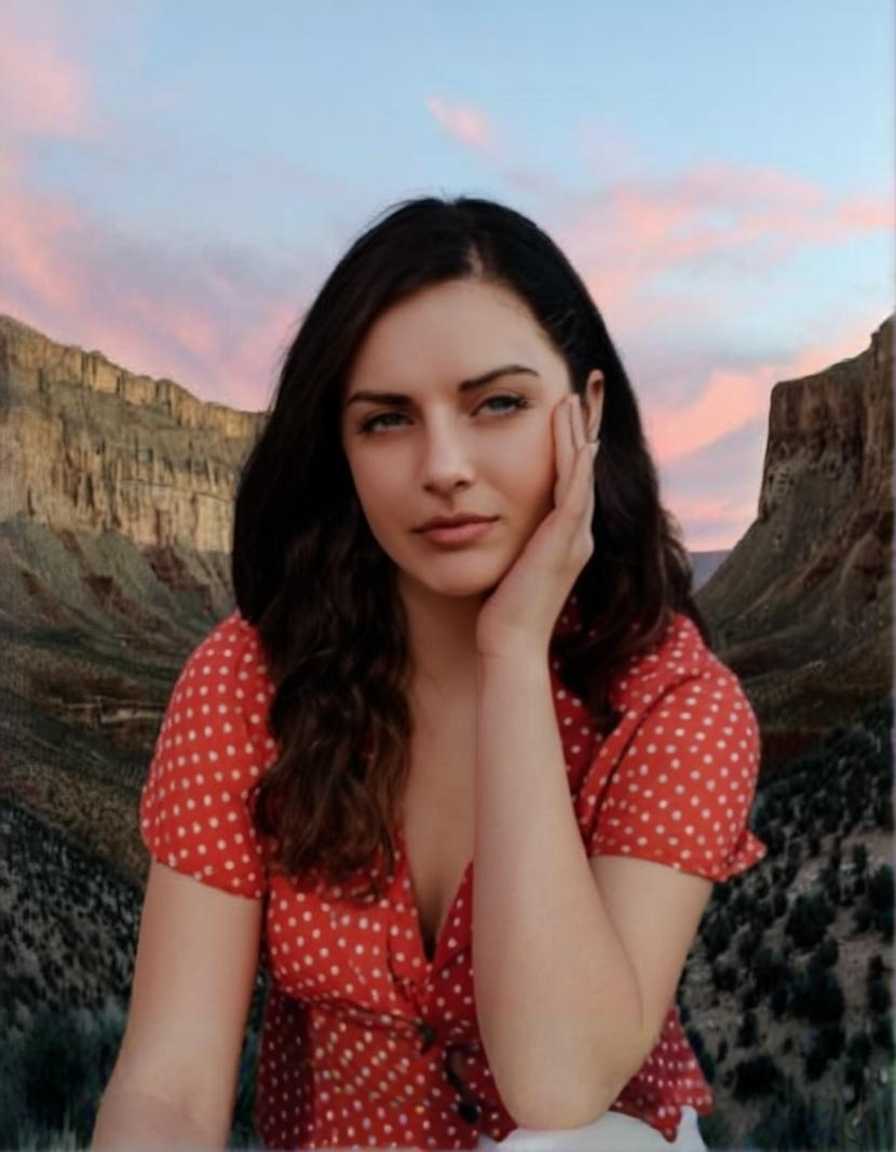}}\\

    \caption{Qualitative results for object relighting. The model is able to relight the object according to the provided background and also remove existing shadows and reflections.}
    \label{fig:app_relighting_results_2}
\end{figure*}

\begin{figure*}[h]
    \captionsetup[subfigure]{position=above, labelformat = empty}
    \centering
    \subfloat[Background]{\includegraphics[width=0.165\linewidth]{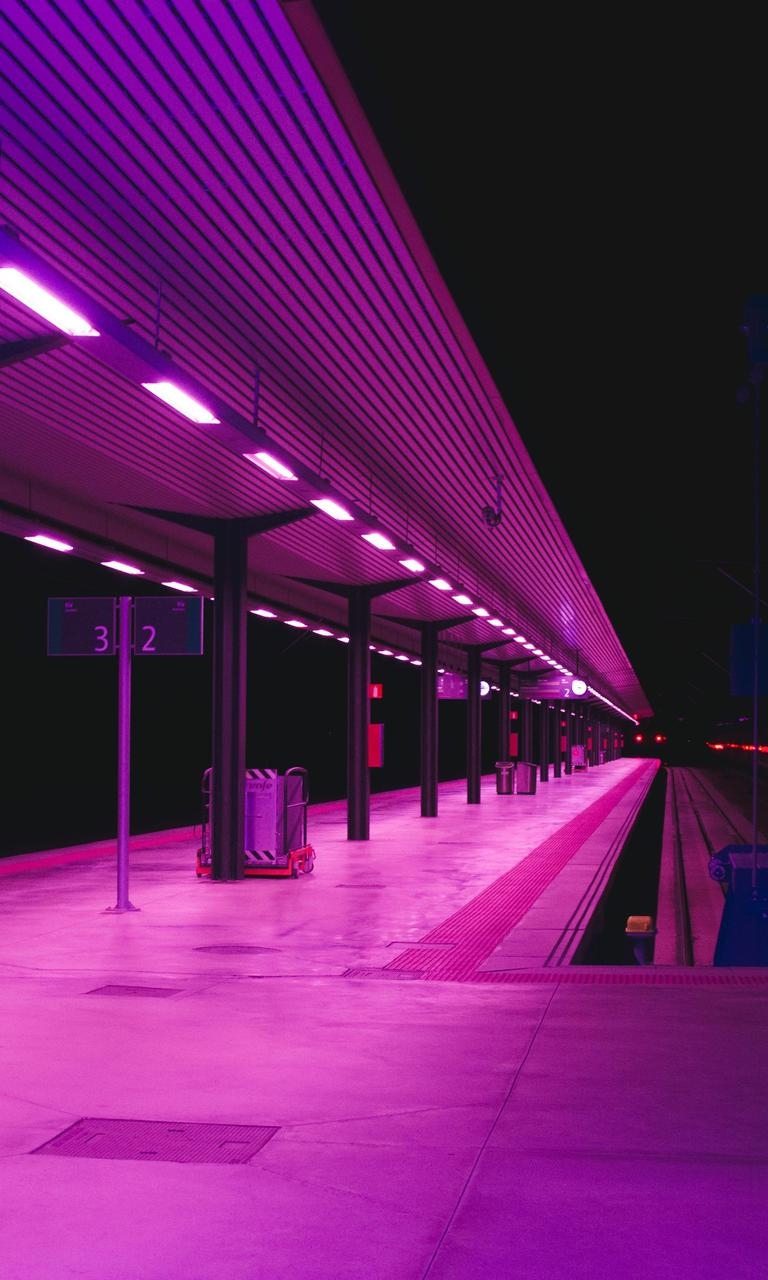}}
    \subfloat[Composite]{\includegraphics[width=0.165\linewidth]{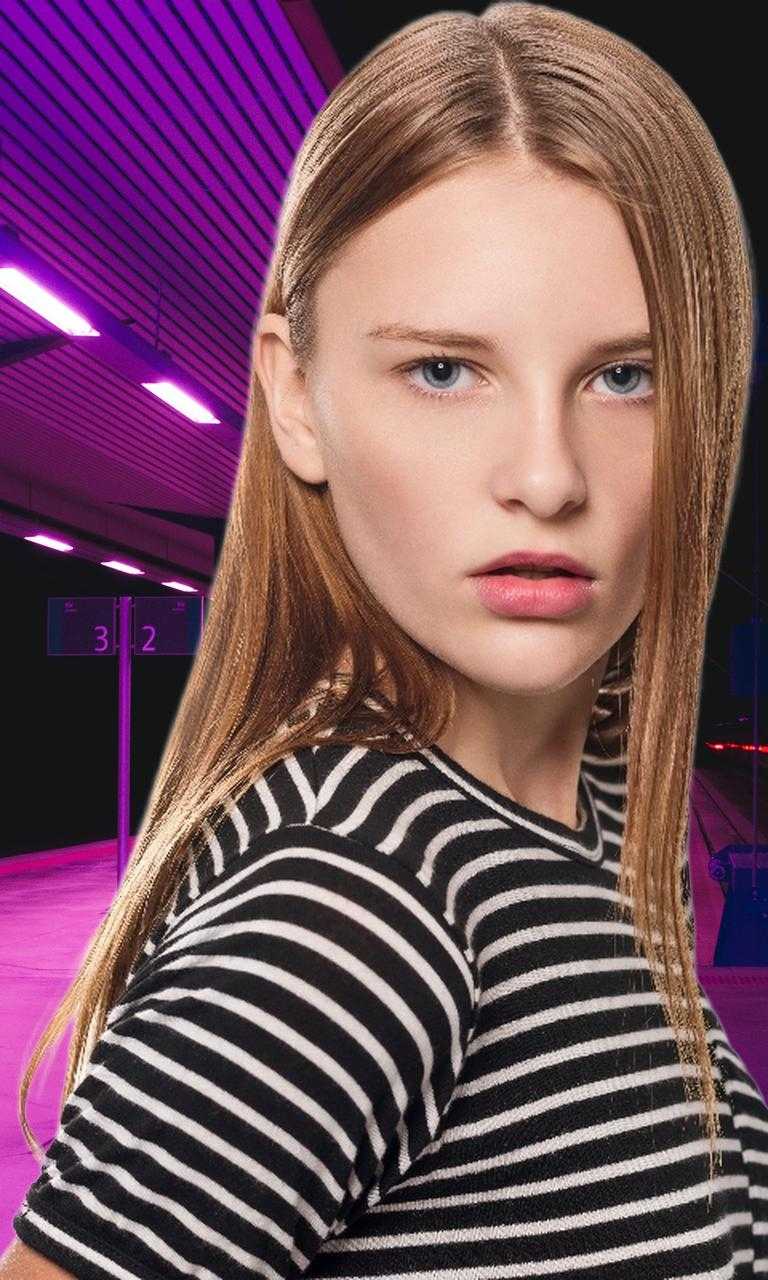}}
    \subfloat[Ours (1 NFE)]{\includegraphics[width=0.165\linewidth]{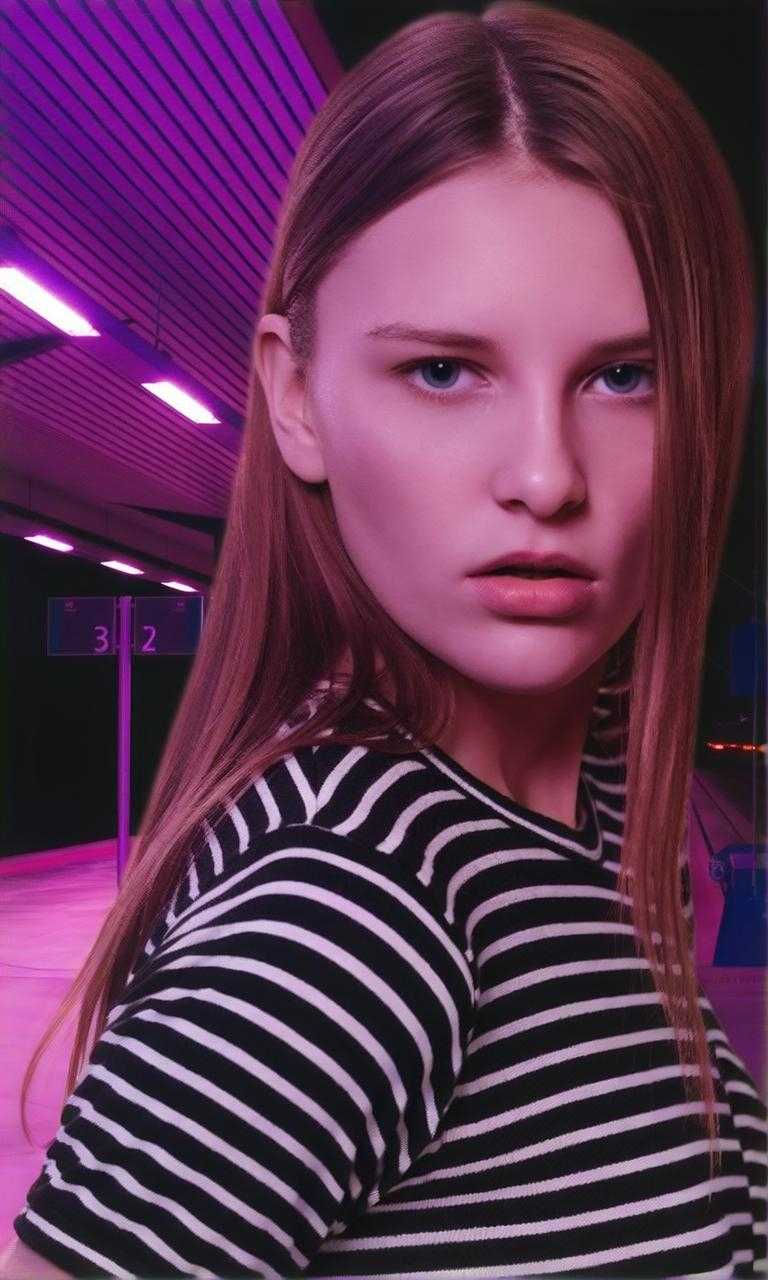}}
    \hspace{0.1em}
    \subfloat[Background]{\includegraphics[width=0.165\linewidth]{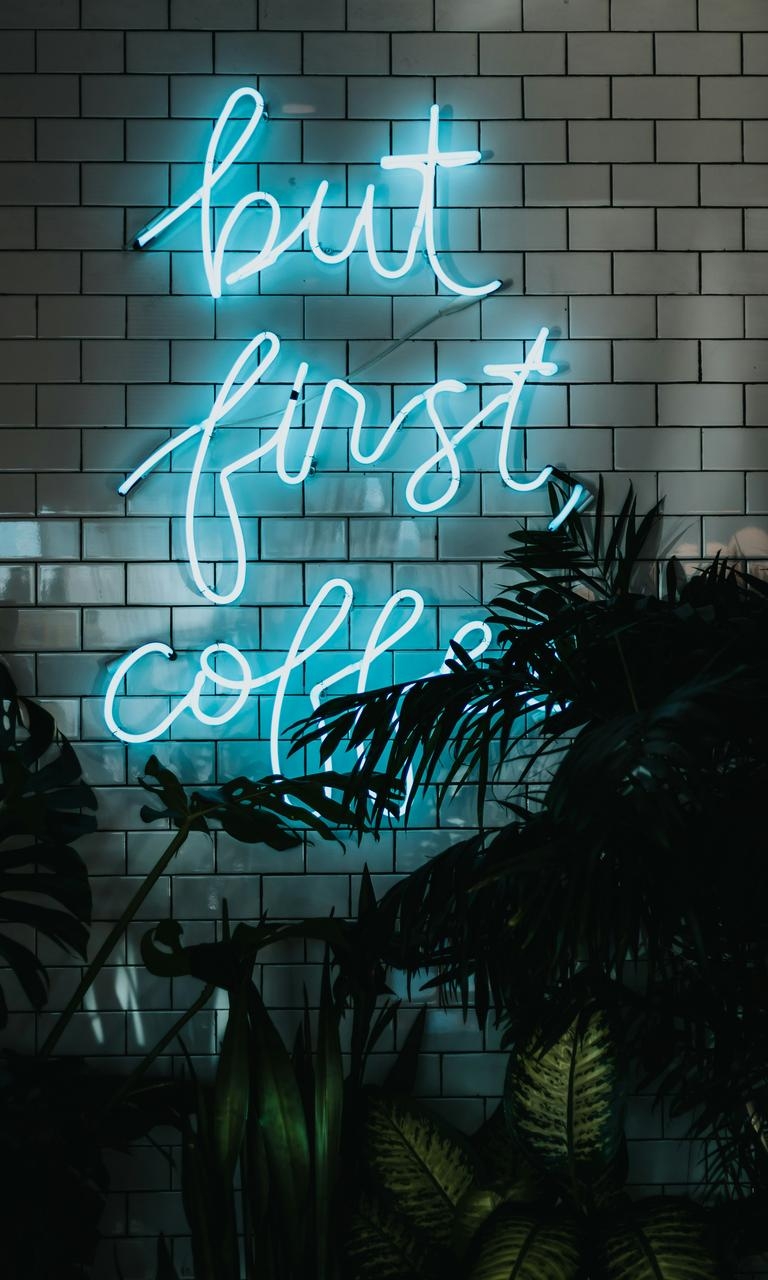}}
    \subfloat[Composite]{\includegraphics[width=0.165\linewidth]{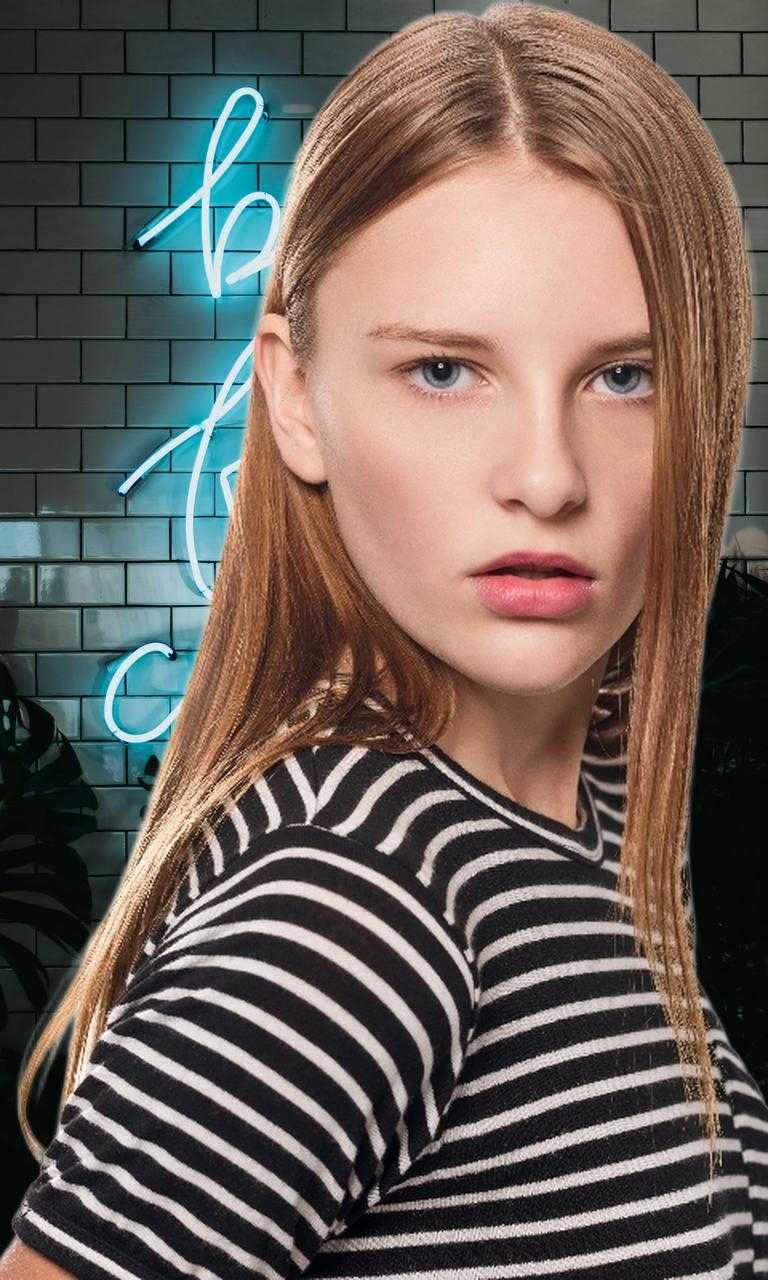}}
    \subfloat[Ours (1 NFE)]{\includegraphics[width=0.165\linewidth]{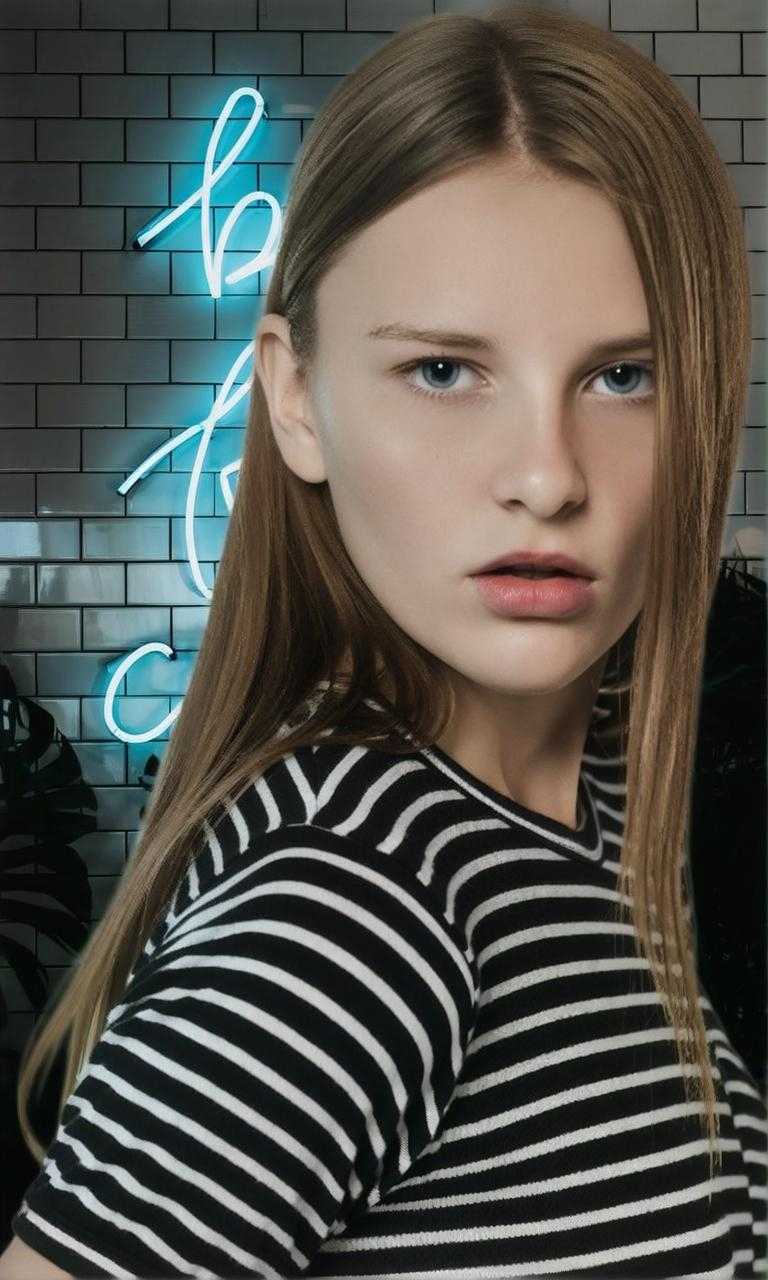}}\\
    \vspace{-0.1em}
    \subfloat{\includegraphics[width=0.165\linewidth]{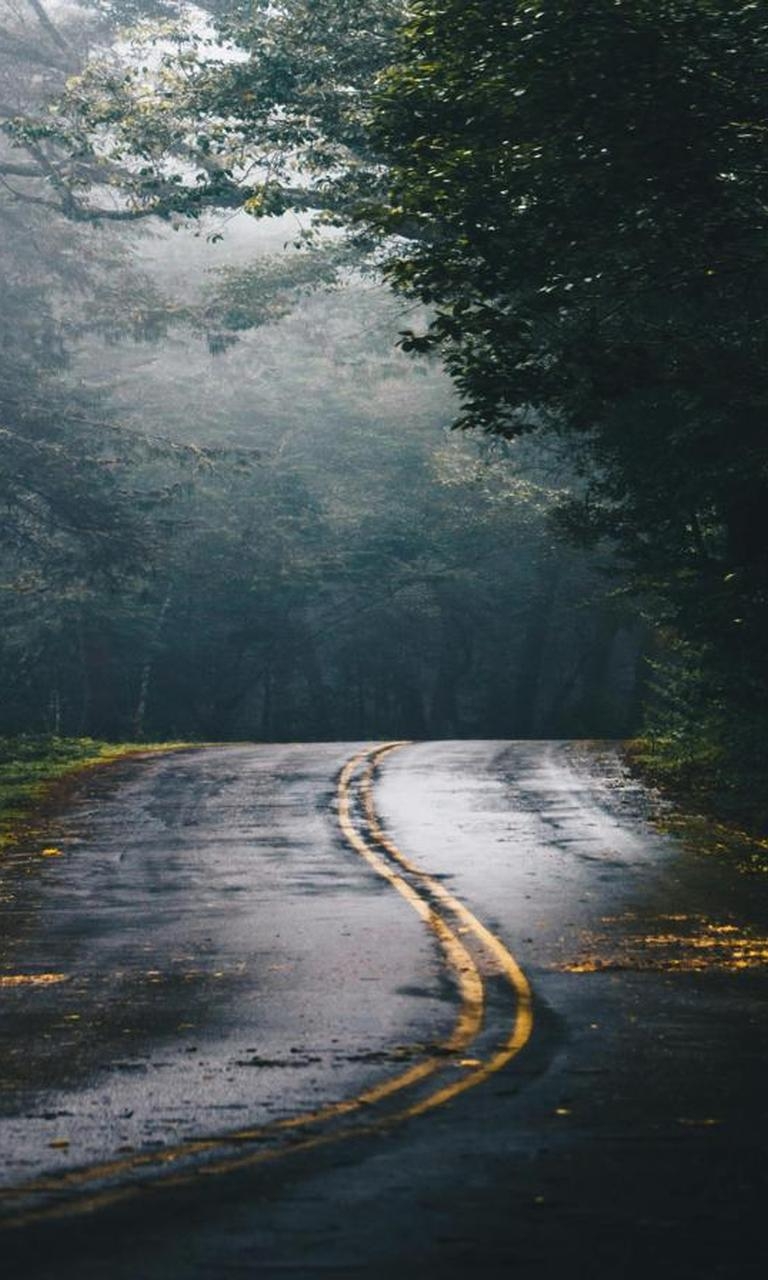}}
    \subfloat{\includegraphics[width=0.165\linewidth]{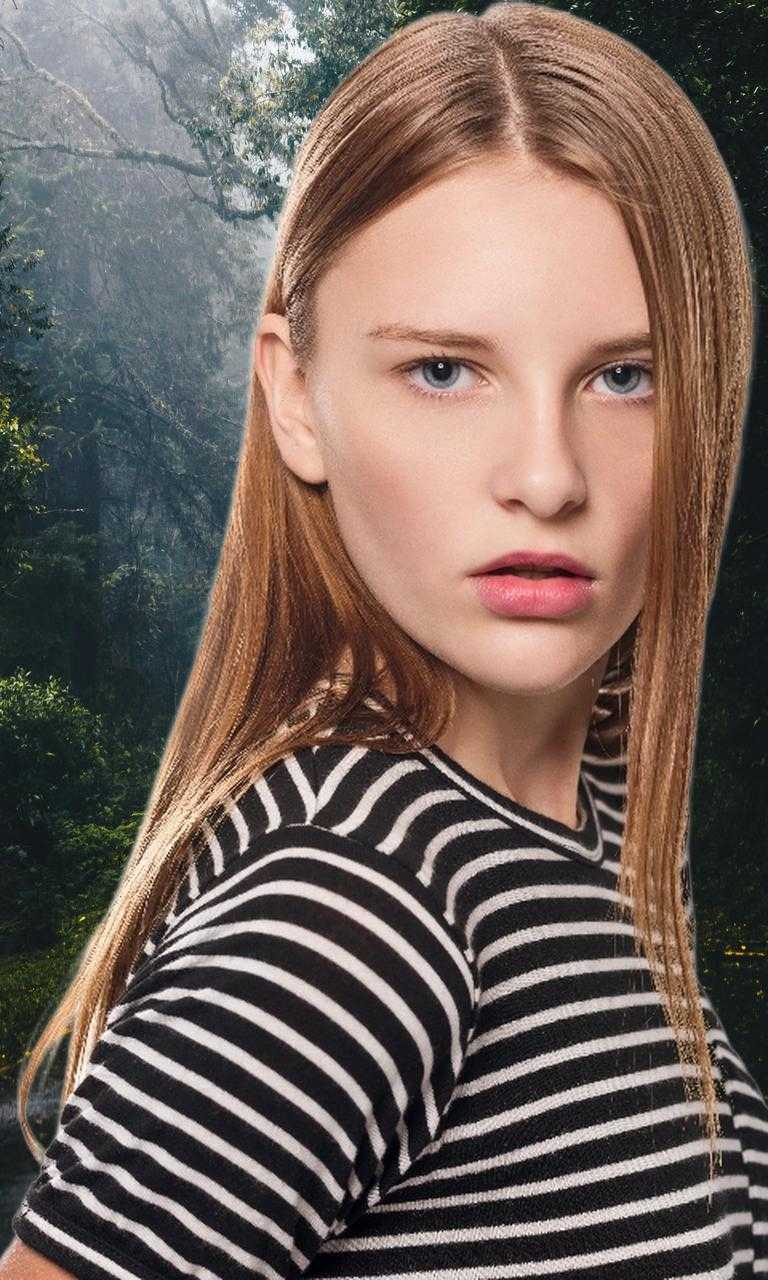}}
    \subfloat{\includegraphics[width=0.165\linewidth]{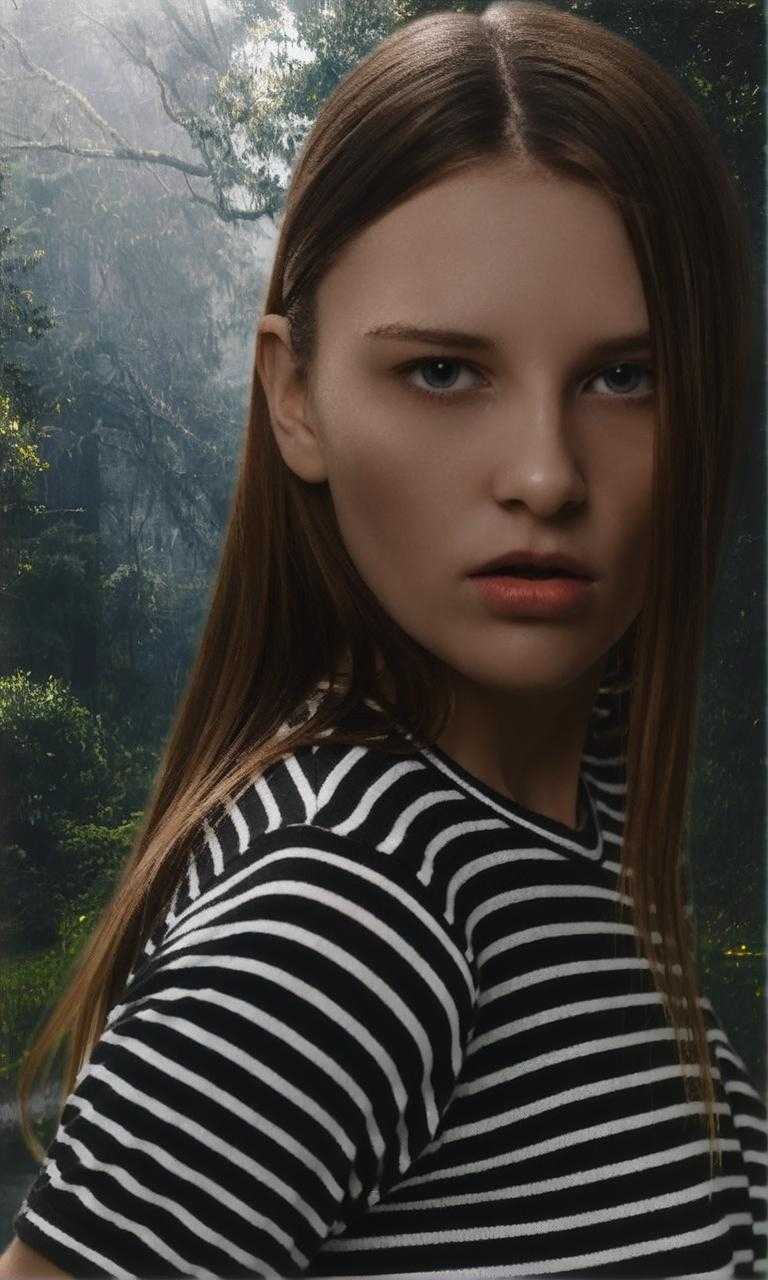}}
    \hspace{0.1em}
    \subfloat{\includegraphics[width=0.165\linewidth]{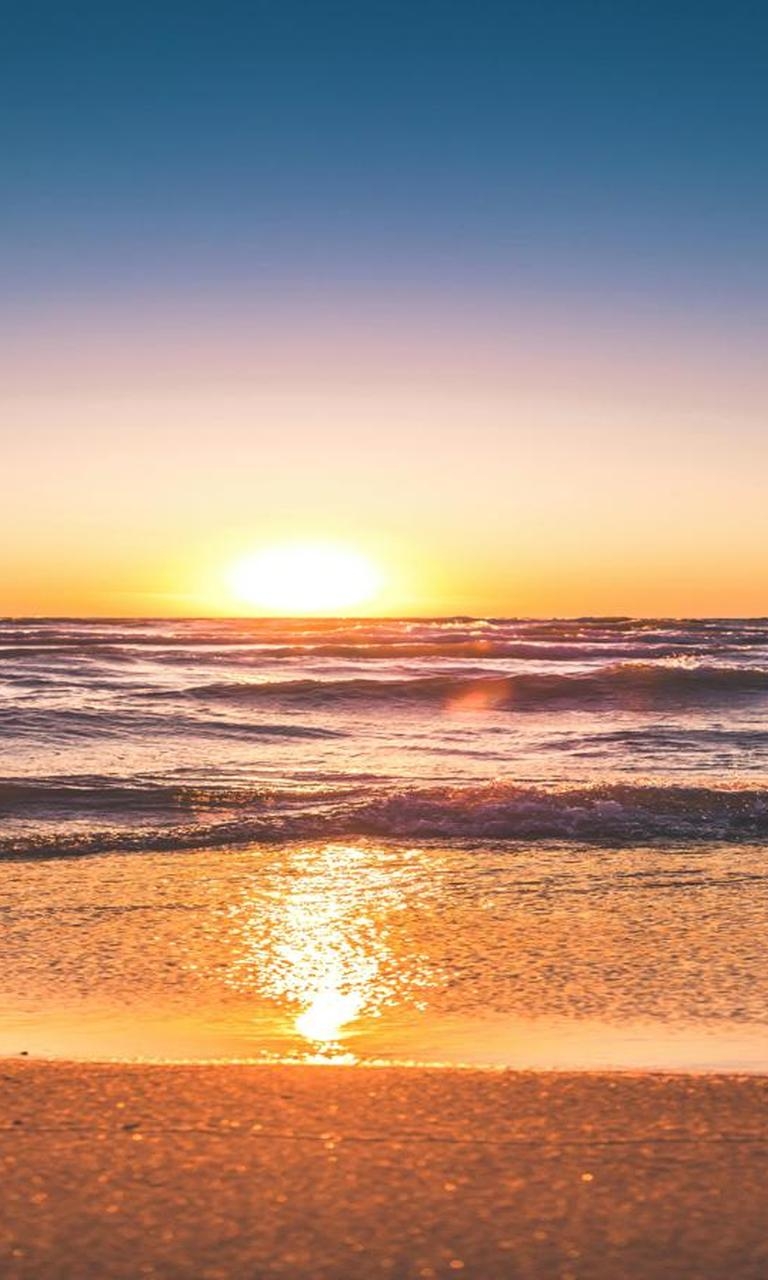}}
    \subfloat{\includegraphics[width=0.165\linewidth]{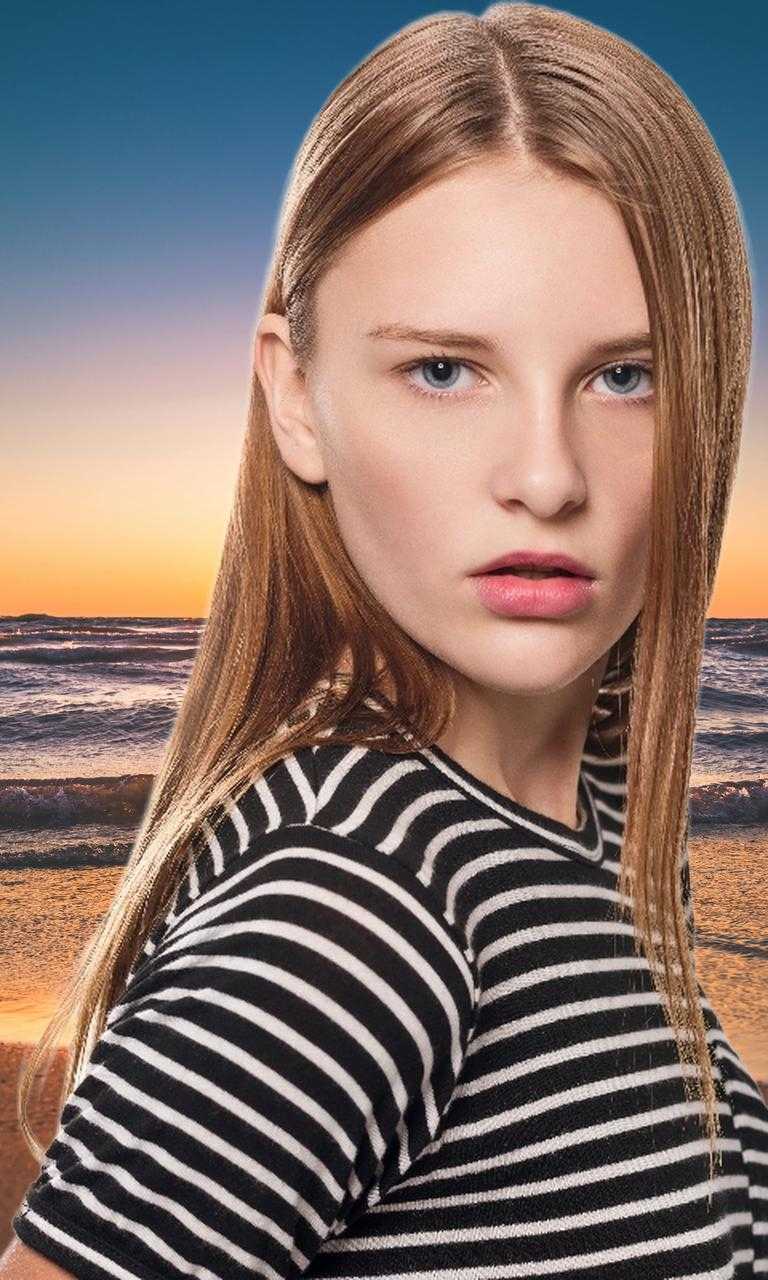}}
    \subfloat{\includegraphics[width=0.165\linewidth]{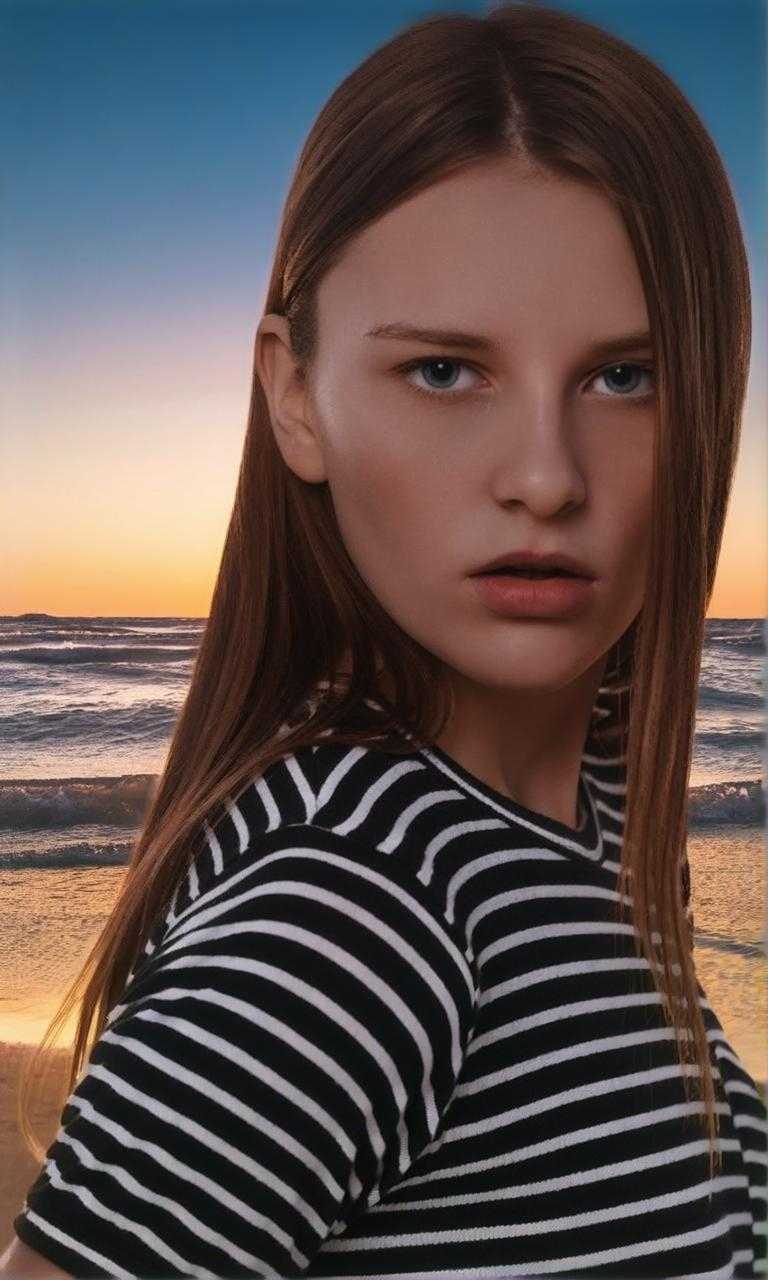}}\\
    
    \caption{Qualitative results for object relighting. The model is able to relight the object according to the provided background and also remove existing shadows and reflections.}
    \label{fig:app_relighting_results_3}
\end{figure*}

\begin{figure*}[ht]
    \captionsetup[subfigure]{position=above, labelformat = empty}
    \centering
    \subfloat{\includegraphics[width=\linewidth]{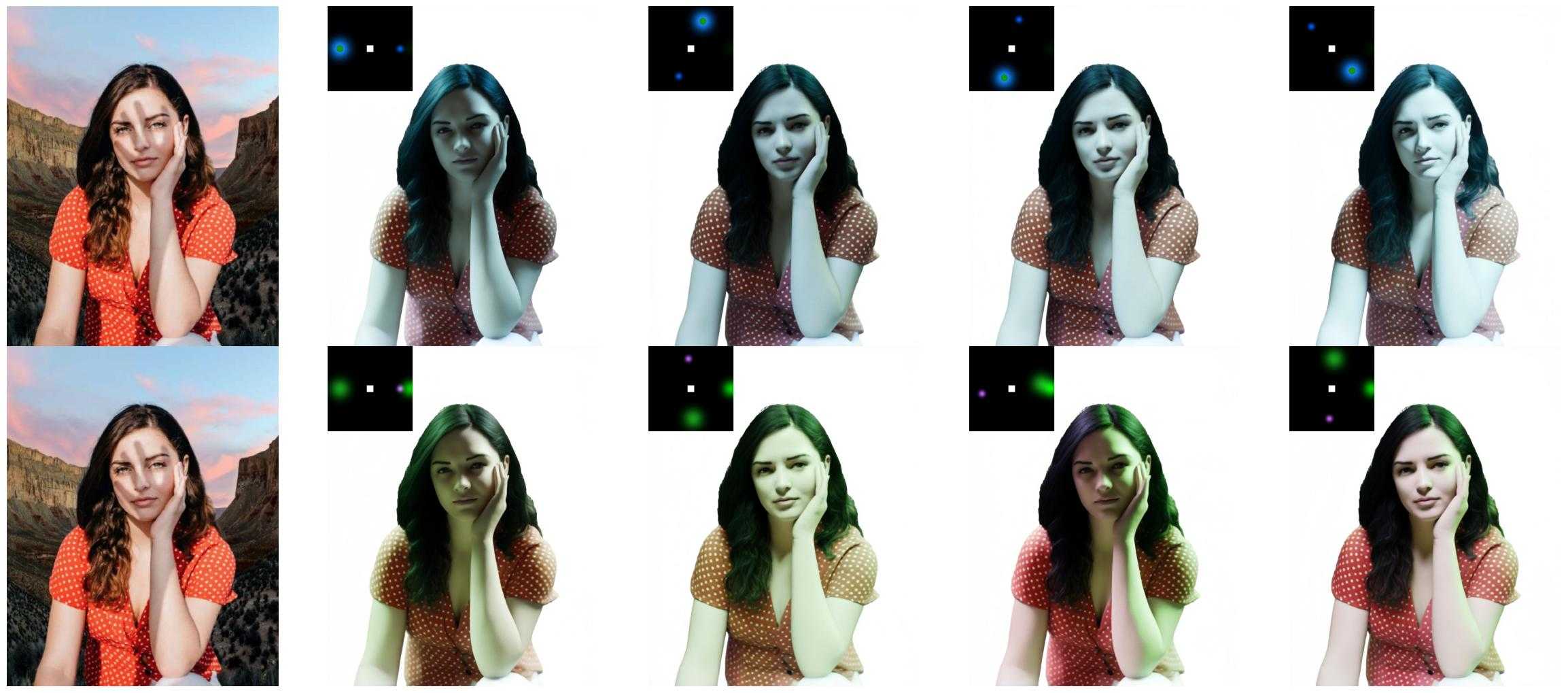}}\\    \subfloat{\includegraphics[width=\linewidth]{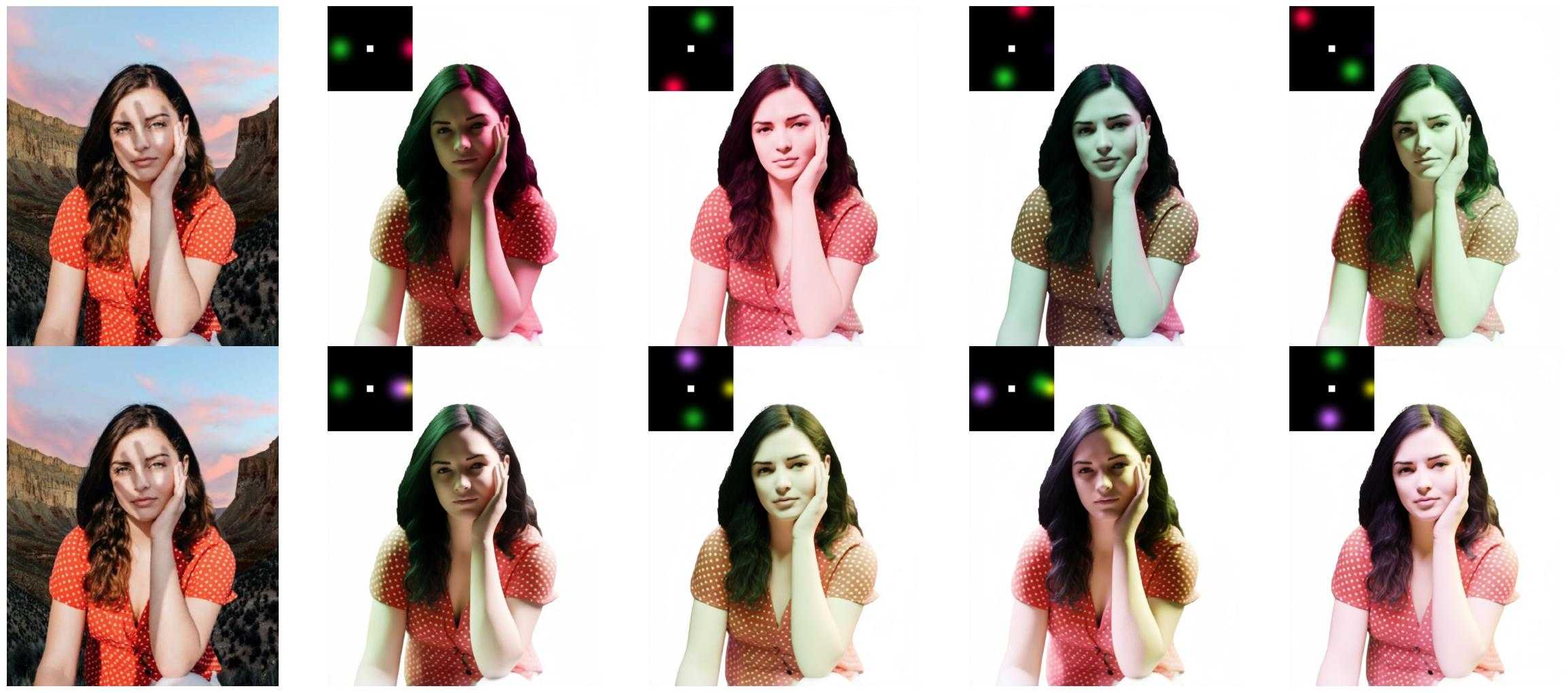}}
    \caption{Qualitative results for controllable image relighting.}
    \label{fig:app_relighting_results_4}
\end{figure*}

\begin{figure*}[h]
    \captionsetup[subfigure]{position=above, labelformat = empty}
    \centering
    \subfloat{\includegraphics[width=\linewidth]{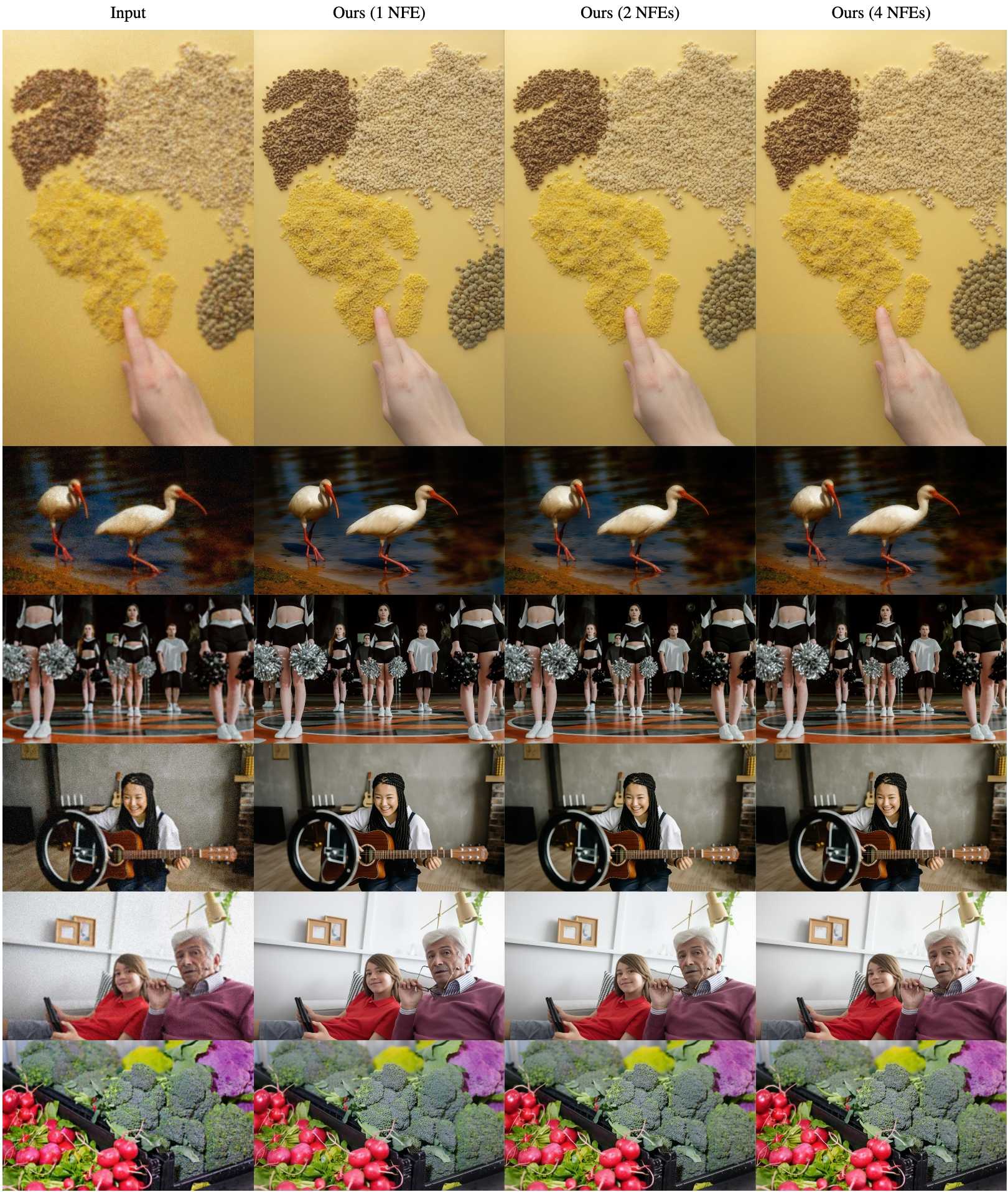}}
    \caption{Qualitative results for object image restoration.}
    \label{fig:app_restoration_results}
\end{figure*}

\begin{figure*}[h]
    \captionsetup[subfigure]{position=above, labelformat = empty}
    \centering
    \subfloat{\includegraphics[width=\linewidth]{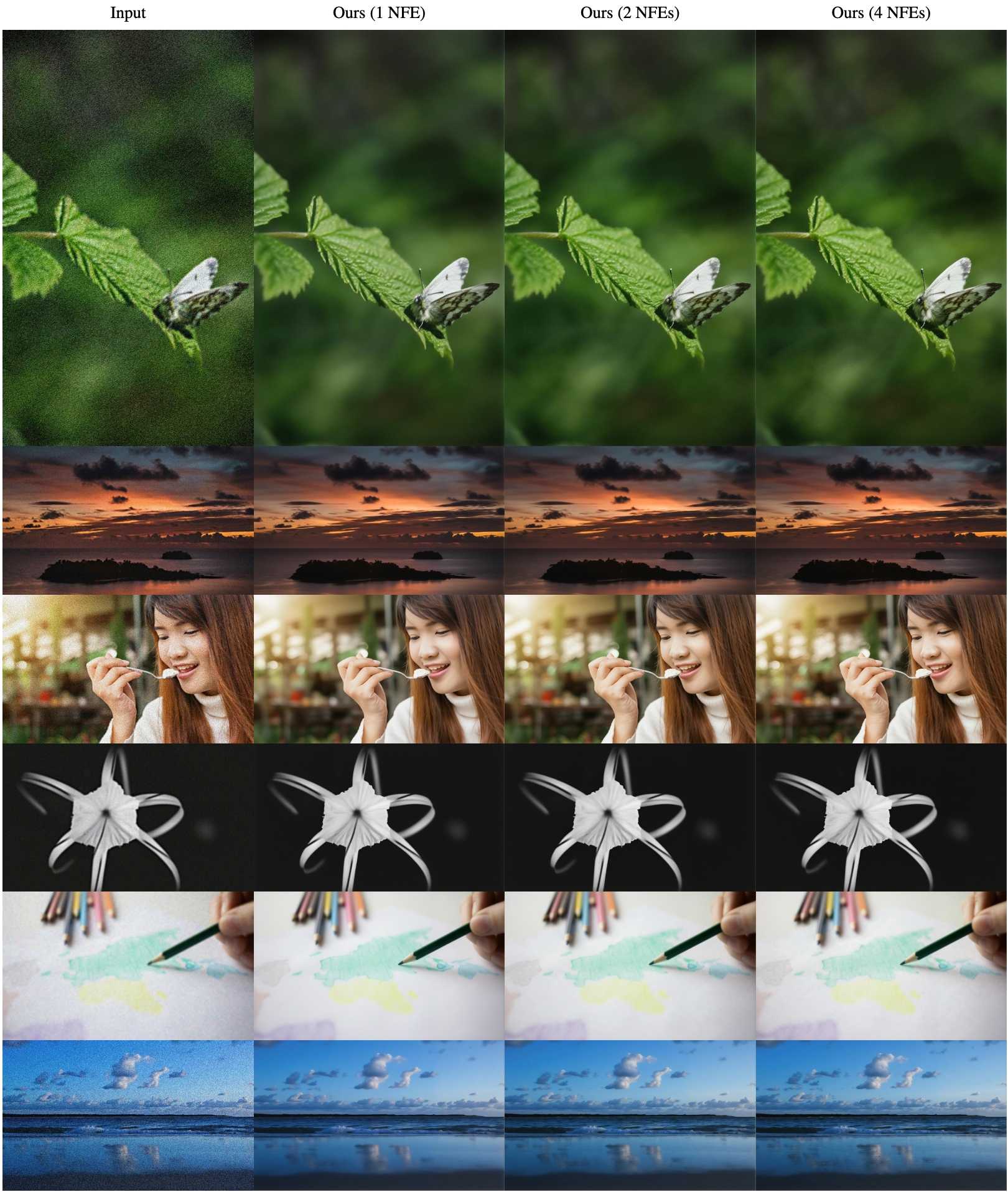}}
    \caption{Qualitative results for object image restoration.}
    \label{fig:app_restoration_results_2}
\end{figure*}

Finally, we provide additional samples for object-removal in \cref{fig:appendix-object-removal} and for image relighting in \cref{fig:app_relighting_results,fig:app_relighting_results_bis,fig:app_relighting_results_2,fig:app_relighting_results_3,fig:app_relighting_results_4}. For object-removal, our model remains the only one capable of removing the target object as well as the associated shadows. For image relighting, the proposed approach can create strong illumination effects on the foreground object and can handle complex lighting conditions. To further stress the method's versatility, we also consider an image restoration task and provide qualitative samples in \cref{fig:app_restoration_results,fig:app_restoration_results_2}. For this task, $\pi_0$ corresponds to the distribution of the latents of the degraded images while $\pi_1$ is the distribution of the latents of the clean images. We artificially create degraded images using the method proposed in \cite{wang2021real}.
In line with the performance observed for the tasks considered in the paper, the proposed method is able to create realistic outputs from degraded images.

\end{document}